\documentclass{article}

\usepackage[dvipsnames]{xcolor}
\usepackage[numbers]{natbib}
\usepackage[preprint]{neurips_2024}

\usepackage{hyperref}
\usepackage{url}
\usepackage{graphicx}
\usepackage{subcaption}
\usepackage{amsmath, nccmath}
\usepackage{bbm}
\usepackage{multirow}
\usepackage{soul}
\usepackage{makecell}
\usepackage{empheq}
\usepackage{listings}
\usepackage{epigraph}
\usepackage{pdfpages}
\usepackage{wrapfig}
\hypersetup{hidelinks}
\usepackage{booktabs}
\usepackage{dashrule}
\usepackage{tabularx,ragged2e}
\usepackage{diagbox}
\usepackage{caption}
\usepackage{multicol}
\usepackage{ulem}
\newcommand{\msout}[1]{\text{\sout{\ensuremath{#1}}}}

\usepackage{longtable}

\usepackage[skins,theorems]{tcolorbox}
\tcbset{highlight math style={enhanced,
  colframe=red,colback=white,arc=0pt,boxrule=1pt}}

\newcommand{\mb}{\mathbf}
\newcommand{\bs}{\boldsymbol}
\newcommand{\mc}{\mathcal}

\newcommand{\ie}{\textit{i.e.,}}
\newcommand{\eg}{\textit{e.g.,}}

\newtheorem{definition}{\textsc{Definition}}

\newtheorem{example}{\textsc{Example}}

\newcommand{\old}{\textsc{RPN}}
\newcommand{\our}{\textsc{RPN} 2}
\newcommand{\new}{\textsc{RPN} 2}
\newcommand{\oldtoolkit}{\textsc{tinyBIG}}
\newcommand{\toolkit}{\href{https://www.tinybig.org}{\textcolor{Plum}{\textbf{\textsc{tinyBIG v0.2.0}}}}}

\newcommand{\insertquote}[2]{%
    \begin{quote}
        \textit{#1}
        \par
        \hfill --- #2
    \end{quote}%
}

\usepackage{array}
\usepackage{makecell}

\newcommand{\ind}{\perp\!\!\!\perp} 
\newcommand{\nind}{\not\!\ind}

\usepackage{diagbox, xcolor}

\usepackage{longtable}

\usepackage[edges]{forest}
\tikzset{%
    parent/.style =          {align=center,yshift=-0cm,text width=1.2cm,rounded corners=3pt},
    child/.style =           {align=center,text width=1.4cm,rounded corners=3pt},
    grandchild/.style =      {align=left,text width=2.7cm,rounded corners=3pt},
    greatgrandchild/.style = {align=left,text width=5.0cm,rounded corners=3pt},
    referenceblock/.style =  {align=left,text width=1.5cm,rounded corners=3pt}
}

\title{{\our}: On Interdependence Function Learning\\ [1ex]
\large{
Towards Unifying and Advancing CNN, RNN, GNN, and Transformer
}}
\author{%
  Jiawei Zhang\\
  IFM Lab\thanks{\textcopyright\ 2024 IFM Lab. All rights reserved. The {\our} project and {\toolkit} toolkit are developed and maintained by IFM Lab.}\\
  Department of Computer Science\\
  University of California, Davis\\
  \texttt{jiawei@ifmlab.org} \\
  {Github: \color{magenta}{https://github.com/jwzhanggy/tinyBIG}}\\
  {Project Official Website: \color{magenta}{https://www.tinybig.org}}
 }
  
\begin{document}
\maketitle

\begin{figure*}[h]
    \begin{minipage}{\textwidth}
    \centering
    	\vspace{-10pt}
	\href{https://www.tinybig.org}{
    	\includegraphics[width=0.6\linewidth]{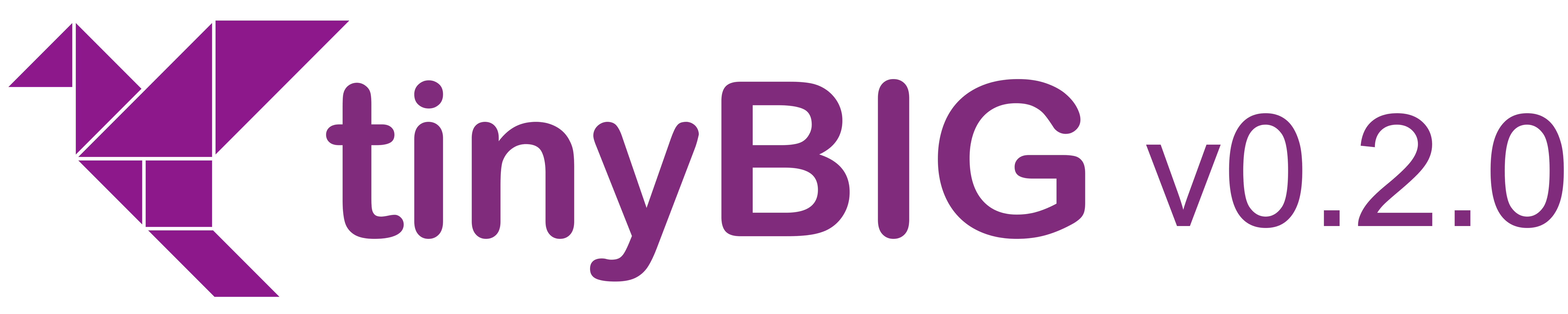}
	}
	\vspace{10pt}
    \end{minipage}%
\end{figure*}

\insertquote{``Nothing in the world stands by itself. Every object is a link in an endless chain and is thus connected with all the other links."\\}{\textsc{Dialectical Materialism} (\textit{Alexander Spirkin})}


\begin{abstract}

This paper builds upon our previous work on the Reconciled Polynomial Network ({\old}) \cite{zhang2024rpnreconciledpolynomialnetwork}. In our prior research, we introduced {\old} as a general model architecture comprising three component functions: data expansion function, parameter reconciliation function, and remainder function. By strategically combining these components functions, we demonstrated {\old}'s versatility in constructing models for addressing a wide range of function learning tasks on multi-modal data. Furthermore, {\old} also unified diverse base models, including PGMs, kernel SVM, MLP, and KAN, into its canonical representation.

The original {\old} model was designed under the assumption of input data independence, presuming the independence among both individual instances within data batches and attributes in each data instance. However, this assumption often proves invalid for function learning tasks involving complex, interdependent data such as language, images, time series, and graphs. Ignoring such data interdependence may inevitably lead to significant performance degradation.

To overcome these limitations, we introduce the new \textbf{Reconciled Polynomial Network (version 2)}, namely \textbf{\our}, in this paper. By incorporating data and structural interdependence functions, {\our} explicitly models data interdependence via new component functions in its architecture.

This enhancement not only significantly improves {\our}'s learning performance but also substantially expands its unifying potential, enabling it to encompass a broader range of contemporary dominant backbone models within its canonical representation. These backbones include, but are not limited to, convolutional neural networks (CNNs), recurrent neural networks (RNNs), graph neural networks (GNNs), and Transformers. Our analysis reveals that the fundamental distinctions among these backbone models primarily stem from their diverse approaches to defining the interdependence functions. Furthermore, this unified representation opens up new opportunities for designing innovative architectures with the potential to surpass the performance of these dominant backbones.

To evaluate the effectiveness of {\our}, we conducted extensive empirical experiments on a diverse set of benchmark datasets spanning multiple modalities, including images, language, time series, and graphs. The results demonstrate that {\our}, with its enhanced interdependence functions, significantly outperforms the previous {\old} model across all evaluated benchmarks. Furthermore, by strategically selecting component functions, we developed novel {\our}-based models that enhance existing backbones in various function learning tasks, such as image classification, language classification, time series forecasting, and graph learning. These findings underscore the versatility and potential of {\our} in advancing backbones across diverse domains for addressing different function learning tasks.

The new Reconciled Polynomial Network (version 2), {\ie} {\our}, presents significant improvements over its predecessor.  The data interdependence component functions substantially enhance {\our}'s learning effectiveness in complex function learning tasks involving interdependent data. Building on {\old}'s unification capabilities, {\our} integrates a broader range of backbone models into its canonical framework. Importantly, {\our} reveals that the primary distinction between existing backbone architectures lies in their definitions of data interdependence functions. This key insight opens avenues for designing new superior backbones, positioning {\our} as a powerful framework for advancing function learning development.

To reflect these advancements, we have also upgraded the {\oldtoolkit} toolkit into the new {\toolkit}. This updated version seamlessly incorporates interdependence functions into {\our} model design, substantially enhancing its learning capabilities. Additionally, {\toolkit} also introduces a new family of data compression and fusion functions and expands the existing repertoire of data expansion and parameter reconciliation functions. Detailed information about the updated {\toolkit} toolkit is available at the project's GitHub repository and dedicated website, with their URLs provided above.

\end{abstract}



{
\newpage
\vspace{3em}
\hrule
\vspace{1em}
\setcounter{tocdepth}{2}
\tableofcontents
\vspace{3em}
\hrule
\vspace{1em}
}

\newpage
\section{Introduction}\label{sec:introduction}

This paper is a follow-up work of our previous study on the Reconciled Polynomial Network ({\old}) \cite{zhang2024rpnreconciledpolynomialnetwork}. In our prior work, we introduced {\old} as a general model architecture comprising three component functions: data expansion function, parameter reconciliation function, and remainder function. Inspired by Taylor's Theorem, {\old} proposes to disentangle the underlying function to be inferred as the inner product of a data expansion function with a parameter reconciliation function. Together with the remainder function, {\old} can accurately approximate the underlying functions that govern data distributions. Our previous research demonstrated {\old}'s versatility in constructing models with varying complexities, capacities, and levels of completeness, which can also serve as a framework for unifying diverse base models, including PGMs, kernel SVM, MLP, and KAN.

\begin{figure*}[h]
    \begin{minipage}{\textwidth}
    \centering
    	\includegraphics[width=1.0\linewidth]{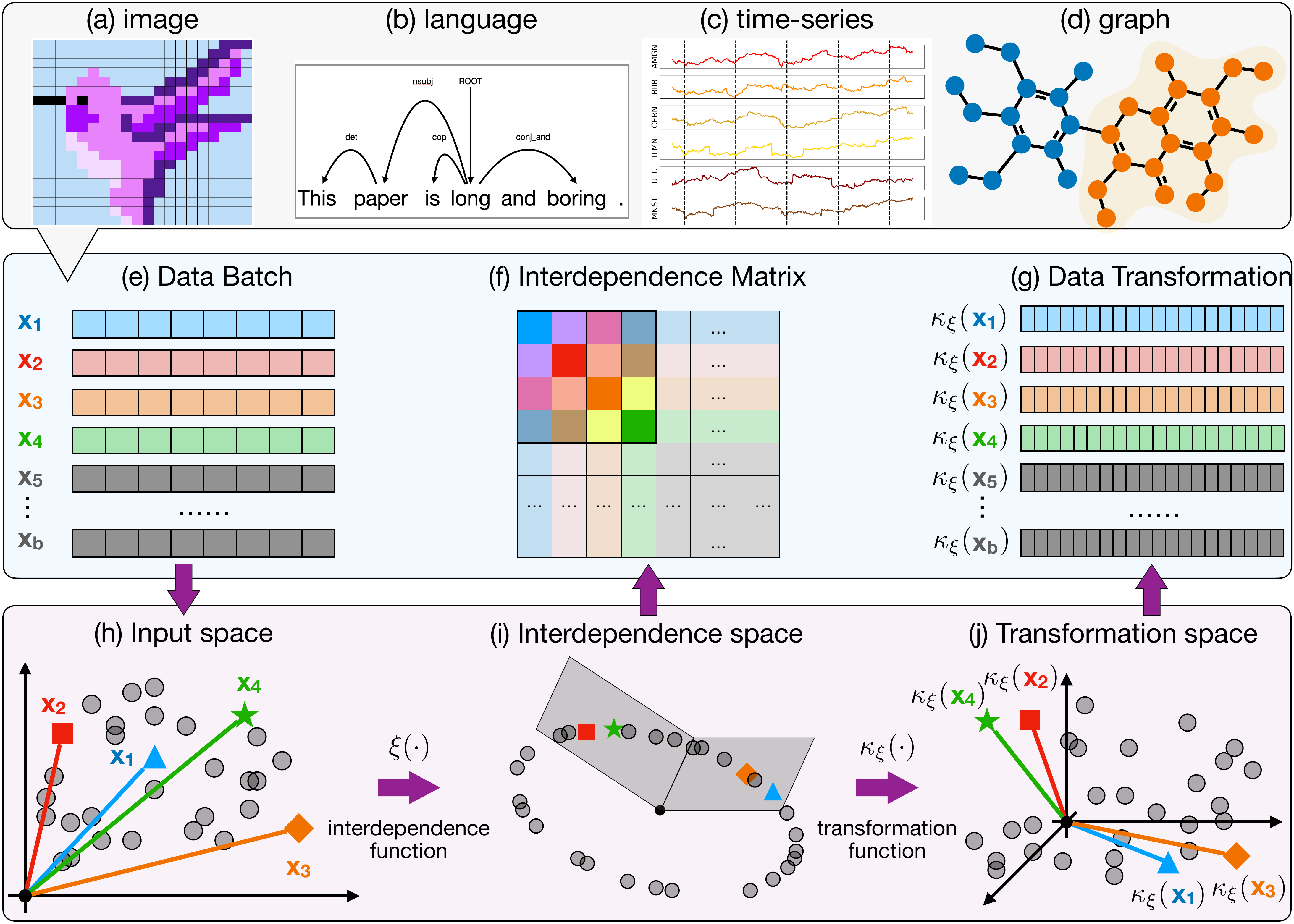}
    	\caption{An illustration of data interdependence modeling in {\our}. Plots (a)-(d) show some examples of interdependent data examples: (a) a colored image of a hummingbird, (b) a sentence and its parsing structure, (c) time-series price data of six stocks, and (d) the Myricetin molecular graph. These data instances in different modalities can all be fed as inputs to the {\our} model. Plots (e)-(g) provide the matrices representations of the input data in {\our}: (e) input data batch, (f) calculated (instance) interdependence matrix, and (g) output data batch after transformation. Plots (h)-(j) indicate the learning space of {\our}: (h) input data space, (i) interdependence space, and (j) data transformation space used for defining the interdependence function and data transformation function. }
    	\label{fig:interdependency_example}
    \end{minipage}%
\end{figure*}

Meanwhile, the previous {\old} model was built on the assumption that data instances in the training batches are independent and identically distributed. Moreover, within each data instance, {\old} also presumed the involved attributes to be independent as well, treating them separately in the expansion functions. However, as illustrated in Figure~\ref{fig:interdependency_example} (a)-(d), these assumptions often prove invalid for function learning tasks on complex, and interdependent data such as images, language, time series, and graphs. In such data, strong interdependence relationships typically exist among both instances and attributes. Ignoring these data interdependencies, as the previous {\old} model does, will significantly degrade learning performance.

In this paper, we propose a redesign of the {\old} architecture, introducing the new \textbf{{\new}} ({\ie} Reconciled Polynomial Network version 2.0) model. As illustrate by Figure~\ref{fig:interdependency_example}, {\new} incorporates a novel component, the \textbf{interdependence functions}, to explicitly model diverse relationships among both data instances and attributes. While we refer to this component as ``\textit{interdependence}'', this function actually captures a wide range of relationships within the input data, including structural interdependence, logical causality, statistical correlation, and numerical similarity or dissimilarity.

Technically, as shown in Plots (e)-(j) of Figure~\ref{fig:interdependency_example}, {\our} employs interdependence functions to generate matrices that capture relationships among data instances and attributes. These functions take data batches as input, and some also incorporate additional structural information, such as underlying topological connections and geometric shapes, to compute comprehensive interdependence matrices. The resulting matrices are typically sparse and are applied efficiently to input data batches (both before and after expansion) using sparse matrix multiplication, optimizing computational resources in terms of both space and time.

The introduction of these interdependence functions significantly enhances {\our}'s ability to model complex function learning tasks involving interdependent data. Moreover, this advancement greatly broadens {\our}'s unifying capacity, enabling it to encompass a wider range of prevalent backbone architectures within its canonical representation, including but not limited to, convolutional neural networks (CNNs), recurrent neural networks (RNNs), graph neural networks (GNNs), and Transformers. Notably, their unified representations reveal that existing backbone architectures primarily differ in their definitions of data interdependence functions. This key insight not only opens new avenues for designing more advanced models but also positions {\our} as a powerful framework for further innovation in function learning architecture design.

To evaluate the effectiveness of the new {\our} model for deep function learning tasks on interdependent data, this paper presents extensive empirical experiments across a diverse set of benchmark datasets, including image, language, time-series, and graph datasets. Leveraging grid-based geometric structural interdependence functions, {\our} effectively captures local interdependence among image patches on benchmarks like MNIST and CIFAR-10. These grid-based functions allow image patches to adopt various shapes, such as cuboids and cylinders, each offering distinct modeling advantages. For language and time-series data, {\our} utilizes chain-structured topological interdependence functions, excelling in tasks such as language classification and time-series forecasting. In the case of graph-structured data, {\our} demonstrates superior performance by effectively modeling structural relationships within graphs using its interdependence functions. With these diverse interdependence functions, {\our} achieves performance comparable to leading models like CNNs, RNNs, and GCNs across these multimodal benchmarks.

What's more, to facilitate the adoption, implementation, and experimentation of the new {\our}, we have updated the {\oldtoolkit} toolkit introduced in our previous paper \cite{zhang2024rpnreconciledpolynomialnetwork} to the new {\toolkit}. This updated version incorporates interdependence modeling capabilities in the {\our} model design and learning, updating the head and layer modules, and the {\our} model architecture. Additionally, {\toolkit} introduces a new family of data compression and multi-input function functions for embedding data vectors into lower-dimensional spaces. {\our} has also updated the existing repertoire of data expansion and parameter reconciliation functions to facilitate the implementation of {\our}-based models. This updated toolkit enables researchers to rapidly design, customize, and deploy new {\our} models across a wide spectrum of function learning tasks on various interdependent datasets.

We summarize the contributions of this paper as follows:

\begin{itemize}

\item \textbf{{\our} for Data Interdependence Modeling}: We redesign the reconciled polynomial network model, introducing {\our} with data interdependence modeling capabilities. Equipped with interdependence functions, {\our} can learn interdependence relationships among both instances and attributes of the data batch, significantly improving its learning performance for function learning tasks on complex and interdependent data.

\item \textbf{Data Interdependence Functions}: We present a suite of interdependence functions capable of learning various categories of interdependence relationships among instances and attributes of the input data batch. These functions utilize the input data batches along with their necessary geometric and topological structure information to calculate interdependence matrices comprising scores between pairwise instances and attributes.

\item \textbf{Backbone Model Unification and Advancing}: The data interdependence functions significantly extend {\our}'s unifying potential, encompassing a broader range of frequently utilized backbone architectures (including CNN, RNN, GNN, and Transformer) within its canonical representation. This unified representation reveals that existing backbone architectures primarily differ in their data interdependence function definitions, opening new avenues for designing superior models and positioning {\our} as a powerful framework for advancing future backbone model design.

\item \textbf{Experimental Investigation}: To evaluate {\our}'s learning performance, we present a series of empirical experiments on numerous real-world benchmark datasets across various function learning tasks, including image classification, language classification, time-series forecasting, and graph classification. The results demonstrate the effectiveness and superior performance of new backbone models designed based on {\our} compared to existing dominant backbone architectures.

\item \textbf{Toolkit Updating}: We update the {\oldtoolkit} toolkit to the new {\toolkit}, incorporating implementations of all data interdependence functions introduced in this paper. Additionally, we introduce a new family of data compression and data fusion functions that can compress and fuse input data into lower-dimensional spaces. The existing repertoire of data expansion and parameter reconciliation functions has also been expanded to include several new component function implementations in the {\toolkit} toolkit.

\end{itemize}

\noindent \textbf{Paper Organization}: The remaining parts of this paper are organized as follows. Section~\ref{sec:formulation} covers notations, function learning task formulations, and essential background knowledge of the reconciled polynomial network model introduced in our previous paper \cite{zhang2024rpnreconciledpolynomialnetwork}. Section~\ref{sec:data_interdependence} introduces the new data interdependence concept, several data interdependence examples and modeling approaches. Section~\ref{sec:method} provides detailed descriptions of the new {\our} model's architecture and design mechanisms. Our library of new data interdependence functions, data compression functions, and other newly added component functions will be presented in Sections~\ref{sec:function}, \ref{sec:function_2} and \ref{sec:function_3}, respectively. Section~\ref{sec:backbone_unification} demonstrates how {\our} unifies and represents existing backbone architectures. Experimental evaluation of {\our}'s performance on numerous benchmark datasets is provided in Section~\ref{sec:experiments}. Interpretations of {\our} with the interdependence functions are provided in Section~\ref{sec:interpretation} from the theoretic machine learning and biological neuroscience perspectives. Section~\ref{sec:discussion} discusses {\our}'s intellectual merits, limitations, and potential future opportunities. Finally, we introduce related works in Section~\ref{sec:related_work} and conclude the paper in Section~\ref{sec:conclusion}.


\section{Notation System and Background Knowledge}\label{sec:formulation}

To ensure the self-containment of this paper, we preface the technical descriptions of the novel {\our} model with a concise overview of both the function learning task and the original Reconciled Polynomial Network ({\old}) introduced in the previous paper \cite{zhang2024rpnreconciledpolynomialnetwork} in this section. Before introducing the background knowledge, we will also briefly describe the notations that will be used in this paper.

\subsection{Notation System}

In the sequel of this paper, unless otherwise specified, we adopt the following notational conventions: lower-case letters ({\eg} $x$) represent scalars, upper-case letters ({\eg} $X$) represent variables, lower-case bold letters ({\eg} $\mathbf{x}$) denote column vectors, boldface upper-case letters ({\eg} $\mathbf{X}$) denote matrices and high-order tensors, and upper-case calligraphic letters ({\eg} $\mathcal{X}$) denote sets. 

For a vector $\mathbf{x}$, we denote its $i$-th element as $\mathbf{x}(i)$ or $\mb{x}_i$, which will be used interchangeably. We use $\mathbf{x}^\top$ to represent the transpose of vector $\mathbf{x}$. For vector $\mathbf{x}$, its $L_p$-norm is defined as $\left\| \mathbf{x} \right\|_p = \left( \sum_i |\mathbf{x}(i)|^p \right)^{\frac{1}{p}}$. The elementwise product of vectors $\mathbf{x}$ and $\mathbf{y}$ of the same dimension is denoted by $\mathbf{x} \odot \mathbf{y}$, their inner product by $\left\langle \mathbf{x}, \mathbf{y} \right\rangle$, and their Kronecker product by $\mathbf{x} \otimes \mathbf{y}$.

For a matrix $\mathbf{X}$, we represent its $i$-th row and $j$-th column as $\mathbf{X}(i,:)$ and $\mathbf{X}(:,j)$, respectively. The $(i,j)$-th entry of matrix $\mathbf{X}$ is denoted as $\mathbf{X}(i,j)$, and its transpose is represented as $\mathbf{X}^{\top}$. The elementwise and Kronecker product operations extend to matrices $\mathbf{X}$ and $\mathbf{Y}$ as $\mathbf{X} \odot \mathbf{Y}$ and $\mathbf{X} \otimes \mathbf{Y}$, respectively. The Frobenius-norm of matrix $\mathbf{X}$ is represented as $\left\| \mathbf{X} \right\|_F = \left( \sum_{i,j} |\mathbf{X}(i,j)|^2 \right)^{\frac{1}{2}}$, and its infinity-norm is defined as its maximum absolute row sums, {\ie} $\left\| \mathbf{X} \right\|_{\infty} = \max_{i} \left( \sum_j | \mb{X}(i,j) | \right)$. The two-to-infinity subordinate vector norm of matrix $\mb{X}$ is defined as $\left\| \mb{X} \right\|_{2 \to \infty} = \sup_{\left\| \mb{z} \right\|_2 = 1} \left\| \mb{X} \mb{z} \right\|_{\infty}$.

For two variables $X$ and $Y$, we denote their independence as $X \ind Y$, and their conditional independence given a condition $C$ as $X \ind Y | C$. Conversely, their interdependence and conditional interdependence are represented as $X \nind Y$ and $X \nind Y | C$, respectively. If $X$ exhibits direct dependence on $Y$, we express this as $X \gets Y$ or $Y \to X$.


\subsection{Function Learning Task}

As outlined in the previous {\old} paper \cite{zhang2024rpnreconciledpolynomialnetwork}, \textbf{function learning}, as the most fundamental task in machine learning, aims to construct a general model comprising a sequence of component functions that infer relationships between inputs and outputs. The term ``function'' in this context refers to not only the {mathematical function} components constituting the {\old} model but also the {cognitive function} of {\old} as an intelligent system generating the desired output responses from input signals.

In function learning, without prior assumptions about data modalities, the corresponding input and output data can manifest in various forms, including but not limited to continuous numerical values ({\eg} continuous functions and time series), discrete categorical features ({\eg} images, point clouds and language), probabilistic variables (defining dependency relationships between inputs and outputs), interconnected structures ({\eg} grids, graphs and chains) and other forms.

\begin{definition}
(\textbf{Function Learning}): Formally, given input and output spaces $\mathbbm{R}^m$ and $\mathbbm{R}^n$, respectively, the underlying mapping governing the data projection between these spaces can be denoted as:

\begin{equation}
f: \mathbbm{R}^m \to \mathbbm{R}^n.
\end{equation}

Function learning aims to construct a model $g$ as a composition of mathematical function sequences $g_1, g_2, \cdots, g_K$ to project data across different vector spaces, which can be represented as:

\begin{equation}
g: \mathbbm{R}^m \to \mathbbm{R}^n, \text{ and } g = g_1 \circ g_2 \circ \cdots \circ g_K,
\end{equation}

where the $\circ$ notation denotes component function integration and composition operators. The component functions $g_i$ can be defined on either the input data or the model parameters.

For input vector $\mathbf{x} \in \mathbbm{R}^m$, if the output generated by the model approximates the desired output, {\ie} 

\begin{equation}
g(\mb{x} | \mb{w}, \bs{\theta}) \approx f(\mb{x}),
\end{equation} 

then model $g$ can serve as an approximated of the underlying mapping $f$ for the provided input $\mb{x}$.

Notations $\mathbf{w} \in \mathbbm{R}^{l}$ and $\boldsymbol{\theta} \in \mathbbm{R}^{l'}$ denote the learnable parameters and hyper-parameters of the function learning model, respectively. By convention, the hyper-parameter vector $\boldsymbol{\theta}$ may be omitted from the model representation, which simplifies the model notation to be $g(\cdot | \mb{w})$.

\end{definition}

\subsection{Reconciled Polynomial Network ({\old}) Model}

To address the function learning tasks, our previous paper \cite{zhang2024rpnreconciledpolynomialnetwork} introduced the \textbf{Reconciled Polynomial Network} (\textbf{{\old}}) model as a general framework with versatile architectures. The {\old} model comprises three component functions, including data expansion function, parameter reconciliation function, and remainder function. This architecture disentangles input data from model parameters and approximates the target functions as the inner product of the {data expansion function} with the {parameter reconciliation function}, subsequently summed with the {remainder function}.

Formally, given the underlying data distribution mapping $f: \mathbbm{R}^m \to \mathbbm{R}^n$, we represent the {\old} model proposed to approximate function $f$ as follows:

\begin{equation}\label{equ:rpn_layer}
{
g(\mb{x} | \mb{w}) = \left\langle \kappa(\mb{x}), \psi(\mb{w}) \right\rangle + \pi(\mb{x}),
}
\end{equation}
where
\begin{itemize}
\item $\kappa: \mathbbm{R}^m \to \mathbbm{R}^{D}$ is named as the \textbf{data expansion function} and $D$ is the target expansion space dimension.
\item $\psi: \mathbbm{R}^l \to \mathbbm{R}^{n \times D}$ is named as the \textbf{parameter reconciliation function}, which is defined only on the parameters without any input data.
\item $\pi: \mathbbm{R}^m \to \mathbbm{R}^n$ is named as the \textbf{remainder function}. 
\end{itemize}

This tripartite set of compositional functions, {\ie} data expansion, parameter reconciliation, and remainder functions, serves as the foundation for the {\old} model. By strategically selecting and combining these component functions, we will be able to construct a {\old} model to address a wide spectrum of learning challenges across diverse function learning tasks. To enhance {\old}'s modeling capabilities, in the previous {\old} paper \cite{zhang2024rpnreconciledpolynomialnetwork}, we introduced both a wide architecture featuring multi-heads and multi-channels (within each layer), and a deep architecture comprising multiple layers. Furthermore, we equipped {\old} with a more adaptable and lightweight mechanism for constructing models with comparable capabilities through nested and extended data expansion functions.

Furthermore, as outlined earlier, the {\old} model was based on the assumption that data instances in training batches are independent and identically distributed (\textit{i.i.d.}). It also assumed that the attributes within each instance were independent, treating them separately within its expansion functions. These restrictive assumptions significantly limit {\old}'s effectiveness in real-world learning tasks involving complex, interdependent data, such as language, images, time series, and graphs. In the following section, we will delve into the concept of data interdependence, which is explicitly modeled in the newly redesigned {\our} model, building upon the foundations of {\old}.


\section{Data Interdependence}\label{sec:data_interdependence}

This section explores the concept of \textbf{data interdependence} in function learning tasks. In practice, interdependence relationships within a data batch can be classified into various granularities, such as \textit{attribute interdependence} and \textit{instance interdependence}, among others. By drawing on real-world data, we will illustrate concrete examples of these relationships. Furthermore, we will explore various metrics and methods for quantifying and modeling interdependence, which will serve as the foundation for developing the robust and effective {\our} model to address function learning tasks in complex and interdependent datasets.

\subsection{What is Data Interdependence?}

Conceptually, the ``\textit{Principle of Universal Connection and Development}'' discussed in \textsc{Dialectical Materialism} asserts that ``\textit{Nothing in the world stands by itself. Every object is a link in an endless chain and is thus connected with all the other links.}'' Data collected from the real-world should inherently reflect such universal and extensive connections. Technically, understanding the data interdependence is also critical for conducting accurate analyses, developing robust models, and making correct decisions in the construction of intelligent function learning systems. 

\begin{definition}
(\textbf{Data Interdependence}): Formally, data interdependence refers to the relationships and interactions between different individual attributes or the entire data instances within a system. In this context, the state or behavior of one element (either an attribute or an instance) can influence or be influenced by others, creating a network of interdependent relationships.
\end{definition}

\noindent \textbf{Instance Interdependence vs. Attribute Interdependence}: Data interdependence may manifest in input data at various granularities, such as attribute interdependence among the attributes within each data instance and instance interdependence among the instances within the data batch. From the perspective of function learning tasks, the distinction between attribute and instance interdependence often becomes blurred, as certain data elements can be regarded as either attributes or instances—or even both. For example, image frames may be considered as individual instances in image classification tasks but can serve as attributes in video content understanding tasks. Similar ambiguities arise in various modalities, such as point clouds, languages, graphs, and time series data.

From a technical implementation perspective, the differences between attribute and instance interdependence are primarily a matter of measuring interdependence across different dimensions of the input data batch ({\eg} rows for instance interdependence and columns for attribute interdependence). By leveraging data batch transposition and reshaping techniques, a unified implementation can effectively calculate interdependence across various dimensions of the data batch, enabling consistent modeling of both attribute and instance interdependence.

To maintain generality, we will illustrate interdependence using examples of general vectors sampled from a vector space below and explore methods to measure these dependencies. In specific applications, these vector variables can represent either attributes or instances within their respective vector spaces, offering a versatile approach to interdependence analysis.


\subsection{Data Interdependence Quantitative Measurements}\label{subsec:quantitative_metrics}

Data interdependence among vector variables representing attribute or instance vectors ({\ie} the columns and rows of the input data batch) can be quantified using various methods, including statistical and numerical approaches. Statistical interdependence measurements, grounded in probability theory and statistics, explicitly model and quantify uncertainty in the interdependence calculation. Conversely, numerical interdependence measurements, based on linear algebra and computational theory, often assume the data interdependence relationships to be deterministic instead. In addition to statistical and numerical approaches, many other quantitative methods also exist, such as topological and geometric measurements, which assess data interdependence relationships from the perspective of topological and geometric structures. These distinct approaches give rise to different categories of quantitative metrics for defining data interdependence functions, all of which will be introduced in detail in the following Section~\ref{sec:function}.


\subsubsection{Statistical Data Interdependence Metrics}\label{subsubsec:statistical_interdependence_matrix}

Formally, given variables $X_1, X_2, \cdots, X_k$ representing vectors $\mb{x}_1, \mb{x}_2, \cdots, \mb{x}_k \in \mathbbm{R}^d$ from a $d$-dimensional vector space (each representing either an attribute or an instance), statistical interdependence measurements assume each follows certain distributions, such as the multivariate Gaussian distribution:

\begin{equation}
X_i \sim \mc{N}(\bs{\mu}_i, \bs{\Sigma}_i), \forall i \in \{1, 2, \cdots, k\},
\end{equation}

where the notation $\mc{N}(\bs{\mu}_i, \bs{\Sigma}_i)$ denotes a multivariate Gaussian distribution with mean vector $\bs{\mu}_i \in \mathbbm{R}^d$ and variance matrix $\bs{\Sigma}_i \in \mathbbm{R}^{d \times d}$. 

The joint distribution of these $k$ variables also follows the multivariate Gaussian distribution, {\ie}

\begin{equation}\label{equ:joint_gaussian_interdependence}
\begin{bmatrix}
X_1\\
X_2\\
\cdots\\
X_k
\end{bmatrix}
\sim \mc{N} \left(
\begin{bmatrix}
\bs{\mu}_1\\
\bs{\mu}_2\\
\cdots\\
\bs{\mu}_k
\end{bmatrix},
\begin{bmatrix}
\bs{\Sigma}_1 & \bs{\Sigma}_{1,2} & \cdots & \bs{\Sigma}_{1,k} \\
\bs{\Sigma}_{2, 1} & \bs{\Sigma}_{2} & \cdots & \bs{\Sigma}_{2,k} \\
\vdots & \vdots & \ddots & \vdots \\
\bs{\Sigma}_{k, 1} & \bs{\Sigma}_{k, 2} & \cdots & \bs{\Sigma}_{k} \\
\end{bmatrix}
\right),
\end{equation}

where $\bs{\Sigma}_{i,j} = Cov(X_i, X_j)$ denotes the covariance matrix of variables $X_i$ and $X_j$ (for $i, j \in \{1, 2, \cdots, k\}$ and $i \neq j$), and $\bs{\Sigma}_{j,i} = Cov(X_j, X_i) = \bs{\Sigma}_{i,j}^\top$. The diagonal variance matrix $\bs{\Sigma}_i$ of variable $X_i$ (for $\forall i \in \{1, 2, \cdots, k\}$) can also be calculated in a similar way as $\bs{\Sigma}_i = Cov(X_i, X_i)$, which is symmetric by default.

Statistically, two variables $X_i$ and $X_j$ are independent if and only if their corresponding covariance matrix is zero, i.e., $\boldsymbol{\Sigma}_{i,j} = \mathbf{0}$ (or equivalently $\boldsymbol{\Sigma}_{j,i} = \mathbf{0}$). If variables $X_1, X_2, \ldots, X_k$ are statistically jointly independent, all the off-diagonal block matrices of the joint covariance matrix in Equation~(\ref{equ:joint_gaussian_interdependence}) will be zeros, rendering the joint covariance matrix to be diagonal.

Based on these above descriptions, various statistical metrics can measure interdependence among the vector variables, such as correlation coefficients and mutual information metrics.

\noindent \textbf{RV Coefficient based Interdependence Metric}: In statistics, the RV coefficient, a multivariate generalization of the squared Pearson correlation coefficient, measures the \textit{linear} dependence between two variables. It ranges from 0 to 1, with 1 indicating perfect linear dependence (or similarity) and 0 indicating no linear dependence. The RV coefficient for pairs of variables can be directly calculated based on their covariance matrices, as introduced above.

Given variables $X_i$ and $X_j$, as well as their variance and covariance matrices $\bs{\Sigma}_i$, $\bs{\Sigma}_j$, $\bs{\Sigma}_{i,j}$ and $\bs{\Sigma}_{j,i}$, their RV-coefficient is defined as:

\begin{equation}
RV(X_i, X_j) = \frac{tr \left( \bs{\Sigma}_{i,j} \bs{\Sigma}_{j,i} \right)}{ \sqrt{ tr \left( \bs{\Sigma}_i^2 \right) tr \left( \bs{\Sigma}_j^2 \right) }} \in \mathbbm{R},
\end{equation}

where the notation $tr(\cdot)$ denotes the \textit{trace} of the input matrix.

\noindent \textbf{Mutual Information (Gaussian) based Interdependence Metric}: In addition to the RV coefficient, mutual information measures the amount of information one random variable contains about another, defining the \textit{non-linear} interdependence relationships of variables.

For random variables $X_i$ and $X_j$ following a multivariate Gaussian distribution, based on their variance and covariance matrices, their mutual information metric is calculated as:

\begin{equation}
MI(X_i, X_j) = \frac{1}{2} \log \left( \frac{ det \left( \bs{\Sigma}_i \right) det \left( \bs{\Sigma}_j \right)}{det \left( \bs{\Sigma} \right)} \right) \in \mathbbm{R},
\end{equation}

where $\bs{\Sigma} = \begin{bmatrix} \bs{\Sigma}_i & \bs{\Sigma}_{i,j} \\ \bs{\Sigma}_{j,i} & \bs{\Sigma}_j \end{bmatrix}$ is the covariance matrix of the joint variables $\begin{bmatrix} X_i\\ X_j \end{bmatrix}$ and $det(\cdot)$ denotes the \textit{determinant} of the input matrix.


\subsubsection{Numerical Data Interdependence Metrics}\label{subsubsec:numerical_interdependence_matrix}

Unlike statistical metrics, numerical metrics measure deterministic interdependence relationships among variables without prior assumptions about their distributions. For variables $X_1, X_2, \ldots, X_k$ representing vectors sampled from a vector space (either attribute or instance vector spaces), numerical methods can quantify interdependence using vector similarity or distance metrics. The simplest form of numerical interdependence in vector space is linear interdependence.

Assuming these vector variables take values $\mathbf{x}_1, \mathbf{x}_2, \ldots, \mathbf{x}_k \in \mathbbm{R}^d$, respectively, they are linearly interdependent if there exist scalar coefficients $\alpha_1, \alpha_2, \ldots, \alpha_k \in \mathbbm{R}$, not all zero, such that:

\begin{equation}\label{equ:numerical_linear_interdependence}
\alpha_1 \mb{x}_1 + \alpha_2 \mb{x}_2 + \cdots + \alpha_k \mb{x}_k = \mb{0},
\end{equation}

where $\mb{0}$ denotes the zero vector of the corresponding vector space. These coefficients $\alpha_1, \alpha_2, \ldots, \alpha_k$ can be calculated using the Gaussian elimination method, which illustrates the linear interdependence relationships among the vectors. Gaussian elimination, also known as \textit{row reduction}, is a method for solving systems of linear equations. It involves a sequence of row-wise reduction operations performed on the corresponding matrix of coefficients. 

To demonstrate how Gaussian elimination reveals interdependence among vectors, consider an example with $\mathbf{x}_1 = [2, -1, -1]^\top$, $\mathbf{x}_2 = [ 3, -4, -2 ]^\top$ and $\mathbf{x}_3= [ 5, -10, -8 ]^\top$. Equation~(\ref{equ:numerical_linear_interdependence}) defined on these three vectors can be rewritten as:

\begin{equation}
\begin{bmatrix}
0\\
0\\
0
\end{bmatrix}
= 
\alpha_1 \begin{bmatrix} 2 \\ -1 \\ -1 \end{bmatrix}
+ \alpha_2 \begin{bmatrix} 3 \\ -4 \\ -2 \end{bmatrix}
+ \alpha_3 \begin{bmatrix} 5 \\ -10 \\ -8 \end{bmatrix}
= 
\begin{bmatrix} 
2 & 3 & 5 \\ 
-1 & -4 & -10 \\ 
-1 & -2 & -8
\end{bmatrix}
\begin{bmatrix} \alpha_1 \\ \alpha_2 \\ \alpha_3 \end{bmatrix}.
\end{equation}

This homogeneous system of linear equations can be transformed into its row-reduced form using Gaussian elimination as follows:

\begin{equation}
\left[
\begin{array}{ccc|c}
2 & 3 & 5 & 0 \\
-1 & -4 & -10 & 0 \\
-1 & -2 & -8 & 0
\end{array}
\right]
\overset{\text{row-reduction}}{\longrightarrow}
\left[
\begin{array}{ccc|c}
2 & 3 & 5 & 0 \\
0 & 1 & 3 & 0 \\
0 & 0 & 0 & 0
\end{array}
\right].
\end{equation}

From this reduction, we observe that: (1) the first and second columns have pivots, indicating that vectors $\mathbf{x}_1$ and $\mathbf{x}_2$ are linearly independent, and (2) the third column has no pivot, indicating that vector $\mathbf{x}_3$ can be linearly represented by $\mathbf{x}_1$ and $\mathbf{x}_2$. Furthermore, we can deduce that $2\alpha_1 + 3\alpha_2 + 5\alpha_3 = 0$ and $\alpha_2 + 3\alpha_3 = 0$, implying $\alpha_2 = -3\alpha_3$ and $\alpha_1 = 2\alpha_3$. Thus:

\begin{equation}
\mb{0} = \alpha_1 \mb{x}_1 + \alpha_2 \mb{x}_2 + \alpha_3 \mb{x}_3 = 2 \alpha_3 \mb{x}_1 - 3 \alpha_3 \mb{x}_2 + \alpha_3 \mb{x}_3,
\end{equation}
or
\begin{equation}
\mb{x}_3 = -2 \mb{x}_1 + 3 \mb{x}_2.
\end{equation}

However, the row-reduced matrix indicates relationships between coefficients and cannot directly define interdependence relationships of data vectors. The row-reduction operation typically has a time complexity of $\mathcal{O}(d^3)$, which can be computationally expensive for high-dimensional data vectors. Moreover, such linear interdependence among instances or attributes is rare in real-world data batches. Instead of directly using linear relationship analysis, we propose to compute interdependence metrics based on other operators, such as \textit{inner product} and \textit{bilinear form}.

\noindent \textbf{Inner Product based Interdependence Metric}: The inner product, a generalization of the dot product, multiplies vectors to produce a scalar. It quantifies the relationship between two vectors (or variables) by calculating the angle between them in the vector space, capturing both vector magnitudes and relative orientation. Formally, for two variables $X_i = \mathbf{x}_i$ and $X_j = \mathbf{x}_j$ taking vectors of the same dimension, their inner product is calculated as:

\begin{equation}
I(X_i, X_j) = \mb{x}_i \mb{x}_j^\top \in \mathbbm{R}.
\end{equation}


\noindent \textbf{Bilinear Form based Interdependence Metric}: A bilinear form is a function linear in both arguments that maps two vectors to a scalar. It measures interdependence by quantifying vector interactions or correlations, with properties like symmetry and orthogonality providing additional insights. The inner product metric can also be viewed as a special case of bilinear forms.

Formally, for variables $X_i$ and $X_j$ taking value vectors $\mathbf{x}_i$ and $\mathbf{x}_j$, their bilinear form-based interdependence metric can be calculated as follows:

\begin{equation}
B(X_i, X_j) = \mb{x}_i \mb{W} \mb{x}_j^\top \in \mathbbm{R}.
\end{equation}

The square matrix $\mathbf{W} \in \mathbbm{R}^{d \times d}$ is normally a constant, with elements defined by $\mathbf{W}(p, q) = B(\mathbf{e}_p, \mathbf{e}_q)$, is called the \textit{matrix of the bilinear form} on the $d$ basis vectors ${\mathbf{e}_1, \mathbf{e}_2, \ldots, \mathbf{e}_d} \in \mathbbm{R}^d$ of the vector space. Additionally, there exist other ways of defining the matrix. Specifically, when matrix $\mathbf{W}$ is the identity matrix, this bilinear form reduces to the inner product. In practice, matrix $\mathbf{W}$ can also be defined as a learnable parameter, and its low-rank representation helps define the interdependence matrix for the Transformer model, which will be introduced later in Section~\ref{subsubsec:parameterized_bilinear_interdependence_function}.


\subsection{Data Interdependence Examples}

In this section, we present examples of real-world data to illustrate the interdependence present in data collected from various sources. These examples span different modalities, including images, language, graphs, and time series. As previously discussed, interdependence relationships can exist at both instance and attribute granularities, with the categorization largely dependent on the data representation format and specific function learning tasks. It is important to note that for the examples discussed below, we present just one potential approach to defining data interdependence relationships. Depending on the specific problem and learning settings, other valid methods for defining interdependence relationships within the data batch may exist. Readers are encouraged to select the most appropriate approach for modeling such dependence relationships in their particular contexts.


\subsubsection{Image Data Interdependence}

\begin{wrapfigure}{r}{0.4\textwidth}
    \vspace{-20pt}
    \centering
    \includegraphics[width=0.4\textwidth]{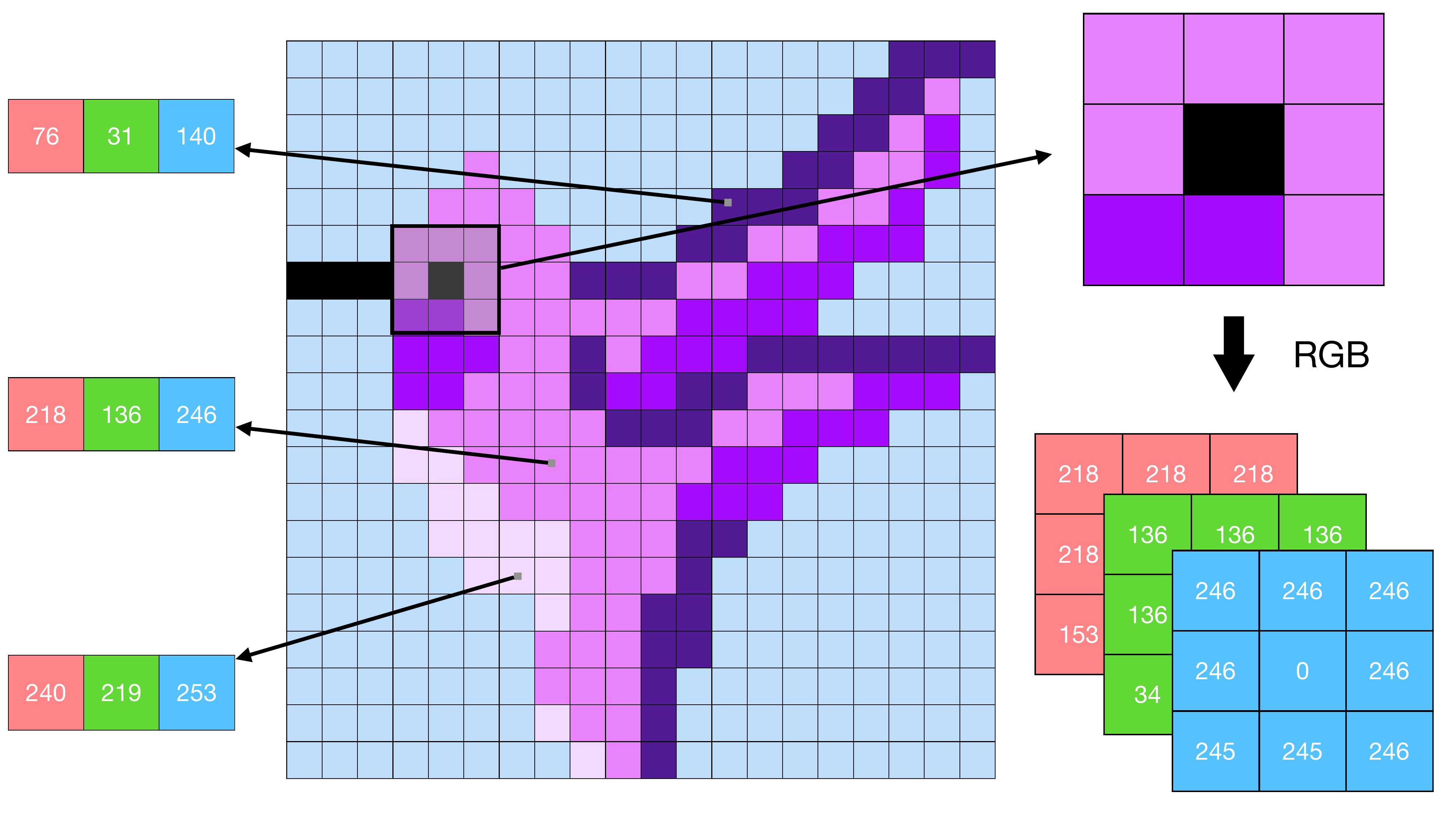}
    \caption{Interdependence in Images.}
    \label{fig:image_spatial_interdependency}
    \vspace{-10pt}
\end{wrapfigure}

Images can be viewed as a sequence of pixels organized into a square or rectangular shape with specific height and width. In the right plot, we illustrate a colored image of a hummingbird in a square shape with $20 \times 20$ pixels, {\ie} both the image height and width are $20$. For colored images, depending on the encoding method used (such as RGB, YCbCr, and CMYK), each pixel is represented by multiple integers. We show the RGB codes of three randomly picked pixels on the left-hand side of the plot. As demonstrated, it is difficult to interpret the physical meanings of these individual pixel RGB values or their potential contribution to identifying objects in the image.

The significance of individual image pixels in addressing the target learning task is profoundly influenced by their surrounding context, which are normally their nearby pixels. The collective variation patterns of these adjacent pixels provide crucial information about the objects present within the images. On the right-hand side of the illustration, we present a $3 \times 3$ pixel segment along with its corresponding RGB color codes. This small-scale representation, when juxtaposed with individual pixels shown on the left, offers a more nuanced perspective. The significant fluctuations in the central pixel’s values compared to its surrounding pixels within this segment indicate its position at the boundary, conveying substantially more information than isolated pixels.


\subsubsection{Language Data Interdependence}

\begin{wrapfigure}{r}{0.45\textwidth}
    \vspace{-10pt}
    \centering
    \includegraphics[width=0.4\textwidth]{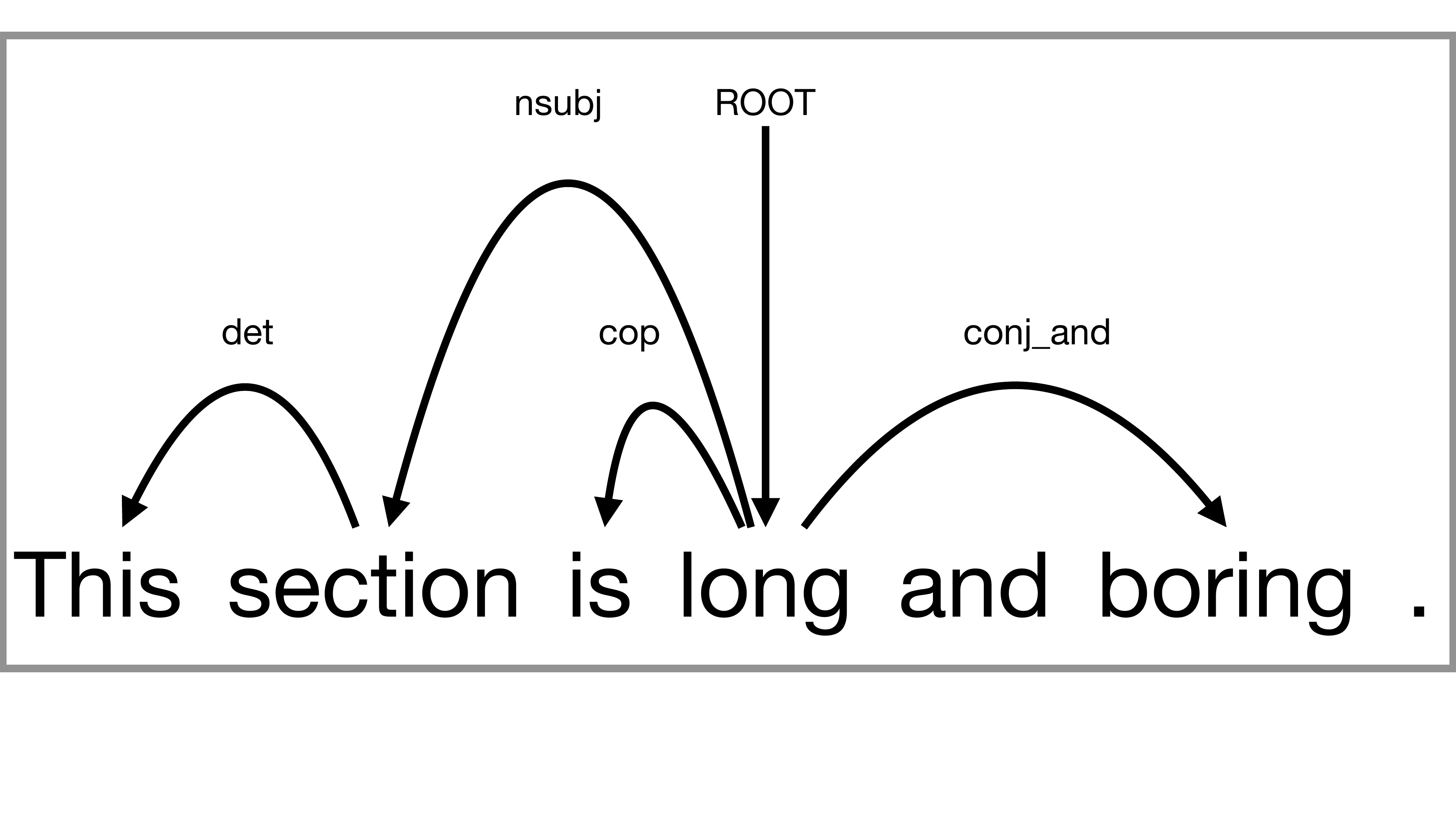}
    \vspace{-15pt}
    \caption{Interdependence in Language.}
    \label{fig:language_temporal_interdependency}
    \vspace{-10pt}
\end{wrapfigure}

Language data, as an important carrier of information, typically appears in an ordered sequence structure, which may include natural language, programming language, and mathematical language. People read and write language data sequentially, which can convey rich semantic information. In the right plot, we illustrate an example sentence ``This section is long and boring.'', and provide its dependency parsing tree. In natural language processing, sentences can typically be decomposed into a sequential list of tokens via a tokenizer based on a pre-defined vocabulary set. For this discussion, we will consider tokens as the smallest units to introduce the interdependence of language data.

In language data, extensive dependence relationships exist among tokens within the same sentence or paragraph (and even across documents). For instance, ``this section" in the above example determines the use of ``is" rather than ``are", and the word ``is" constrains the following words to be adjectives, {\eg} ``long". Furthermore, the conjunction ``and" indicates that the two adjectives should have close semantic meanings, {\ie} ``long" and ``boring". The semantic meaning of each word depends on its sentence context, which is a crucial factor that should be incorporated into model design and learning.


\subsubsection{Time-Series Data Interdependence}

\begin{wrapfigure}{r}{0.45\textwidth}
    \vspace{-30pt}
    \centering
    \includegraphics[width=0.45\textwidth]{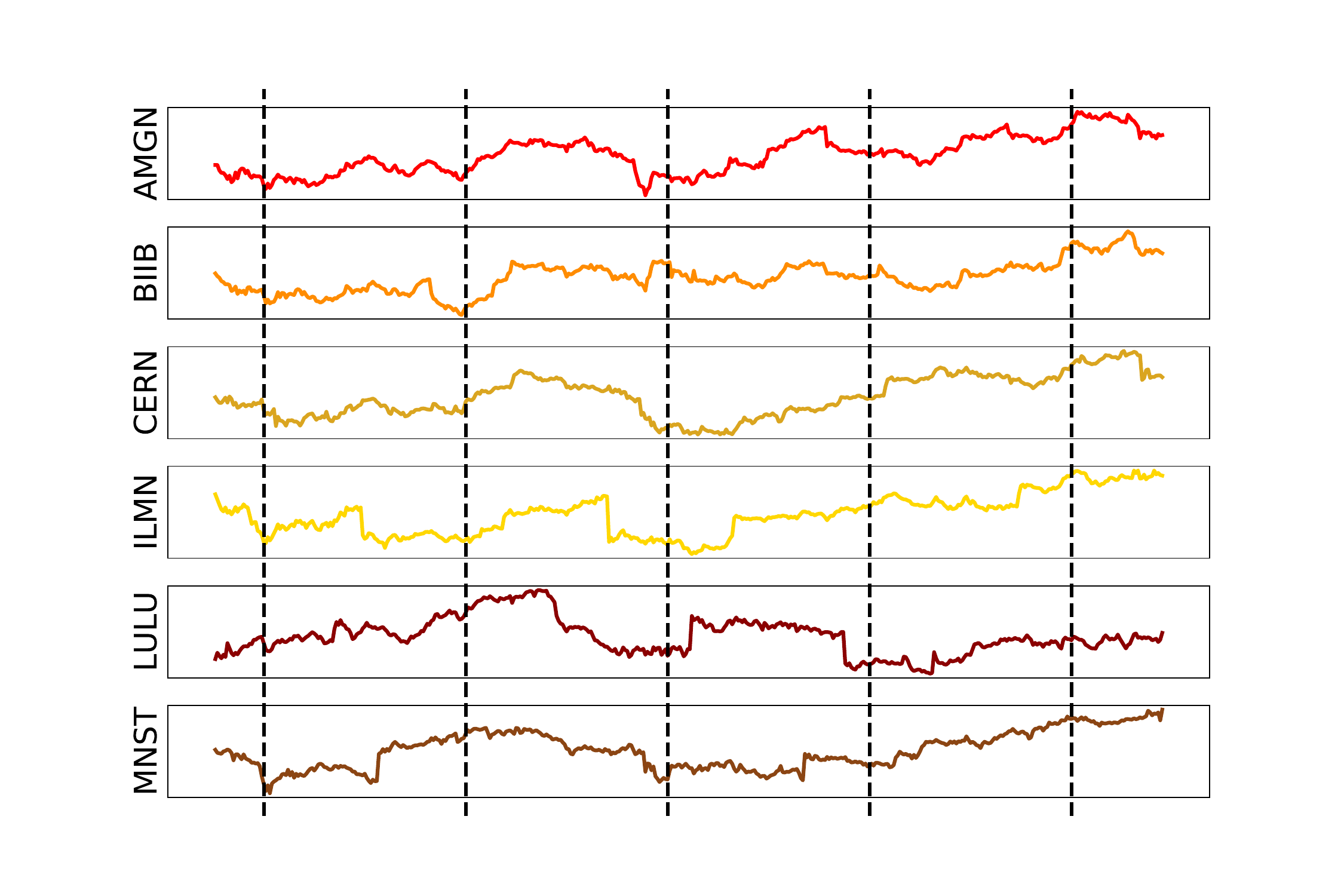}
    \vspace{-20pt}
    \caption{Interdependency in Time-Series.}
    \label{fig:timeseries_temporal_interdependency}
    \vspace{-10pt}
\end{wrapfigure}

Time series data, such as daily temperature readings, stock market prices, and annual GDP growth, provide another representative example of interdependence in data. In the right plot, we show the stock price curves of six biotechnology companies - AMGN, BIIB, CERN, ILMN, LULU, and MNST - over a four-year period. In time series data, data points at later timestamps typically depend on those at previous timestamps. Moreover, for some time series data exhibiting long-term periodic patterns, this interdependence may span a much longer time period, {\eg} a month or even several years.

For the stock price data illustrated in the right plot, these selected stocks belong to the same sector, and the price of one stock may also depend on other correlated stocks. In such stock price time series data, individual data points alone can hardly reveal any information about the underlying price changing patterns. To extract useful features and signals, it may be necessary to include other data points spanning both temporal and sector dimensions.


\subsubsection{Graph Data Interdependence}

\begin{wrapfigure}{r}{0.40\textwidth}
    \vspace{-20pt}
    \centering
    \includegraphics[width=0.40\textwidth]{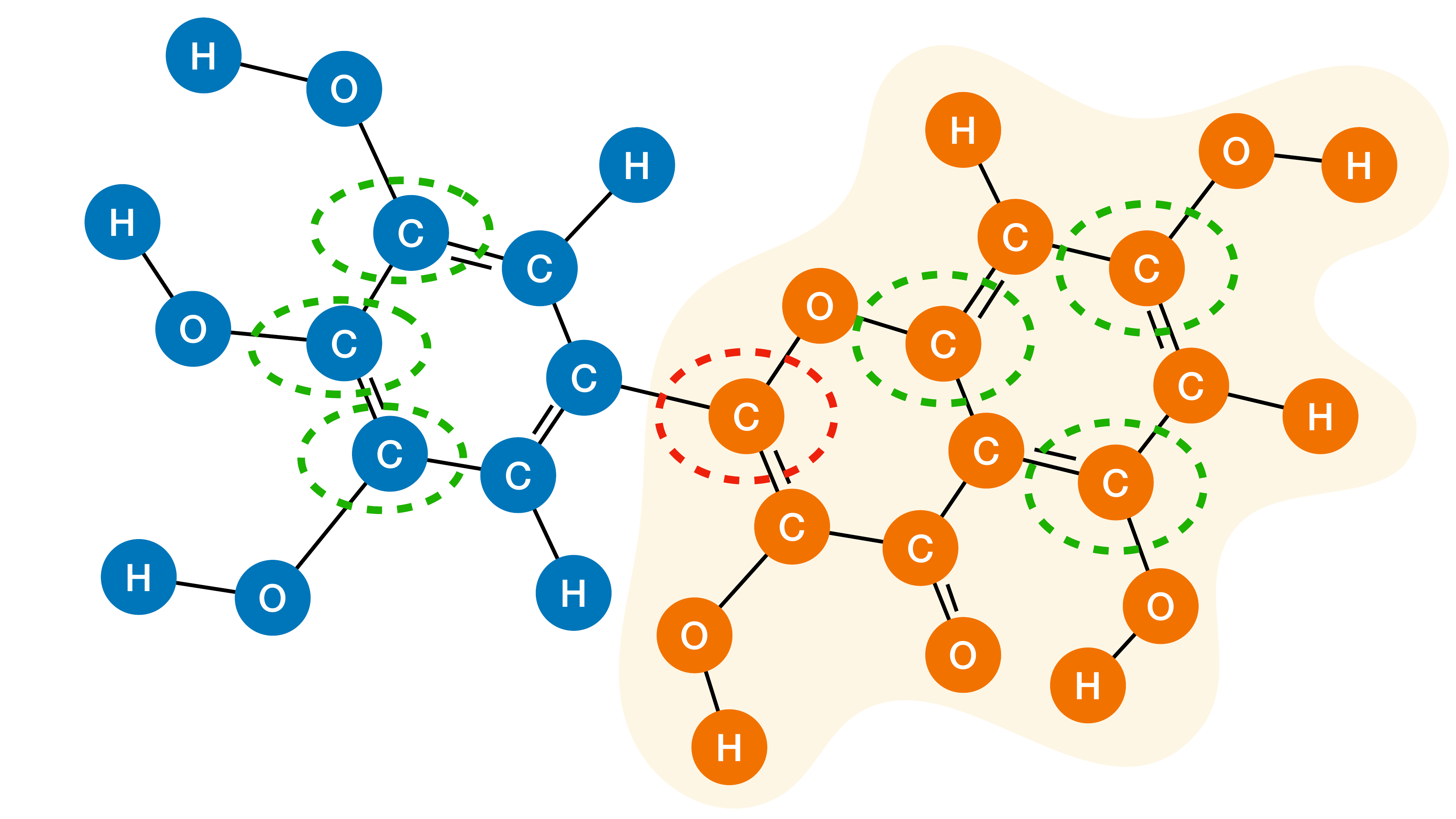}
    \caption{Interdependence in Graph.}
    \label{fig:graph_spatial_interdependency}
    \vspace{-10pt}
\end{wrapfigure}

In addition to images, language and time-series data, graphs are another representative example of data structures with extensive dependence relationships among nodes. Graphs can be represented as a set of nodes connected by links. In the right plot, we illustrate an example of the Myricetin molecule ({\ie} $C_{15}H_{10}O_8$), a member of the flavonoid class of polyphenolic compounds with antioxidant properties. By treating atoms as nodes and atomic bonds as links, the Myricetin molecule can be represented as a molecular graph structure. Single and double bonds can be represented as different types of links in the graph, rendering it heterogeneous. (Note: The distinction between homogeneous and heterogeneous graphs is slightly out of the scope of this paper and will not be discussed further.)

In the molecular graph, it is difficult to infer the roles or functions of individual nodes, such as the central carbon atom in the red dashed circle, based solely on the node itself. We must consider its surrounding nodes on which it depends. Unlike images, nodes' dependence relationships in graphs can span across the entire structure to distant nodes, and local neighbors may not provide sufficient information for inferring their functions in the molecule. For instance, in the plot, we highlight several other carbon nodes with identical surrounding neighbors connected by the same types of links. To infer the functions of the central carbon node, we may also need to consider the functional groups it is involved in, {\eg} the one highlighted in the orange background color in the plot.


\subsection{Data Interdependence Handling}\label{subsec:interdependence_matrix}

The diverse data interdependence relationships illustrated in the above examples play a critical role in the function learning tasks studied in this paper. In this section, we introduce two different approaches for handling such data interdependence relationships: \textit{interdependence padding} and \textit{interdependence aggregation}. Furthermore, we demonstrate that these two approaches can be unified under a shared representation, expressed as the multiplication of the data batch with an \textit{interdependence matrix} defined based on the input data.


\subsubsection{Interdependence Padding}

\begin{wrapfigure}{r}{0.32\textwidth}
    \vspace{-30pt}
    \centering
    \includegraphics[width=0.3\textwidth]{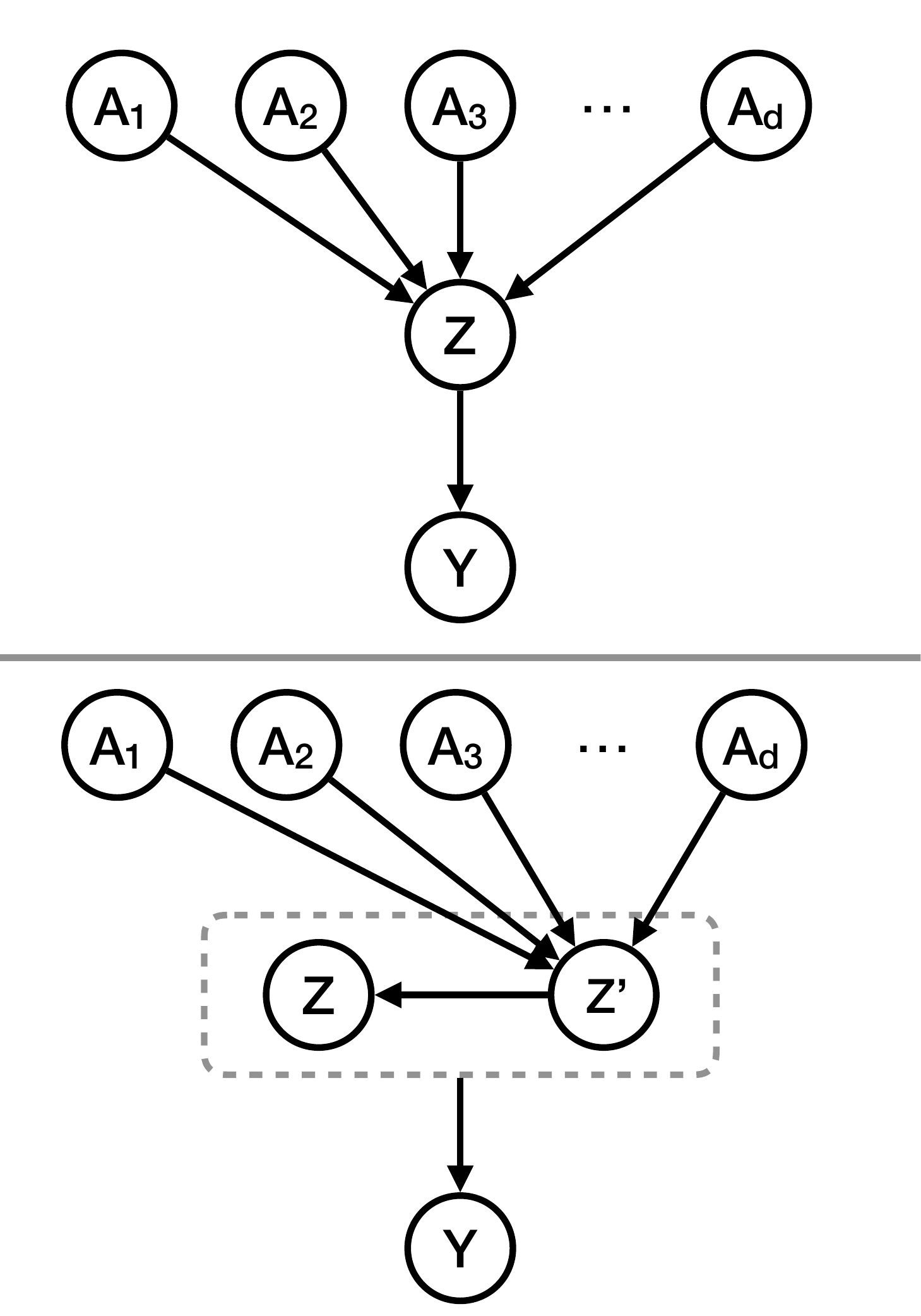}
    \caption{An Illustration of Variable Dependence Padding.}
    \label{fig:dependence_padding}
    \vspace{-10pt}
\end{wrapfigure}

Formally, given two variables $Z$ and $Y$ representing the input and output of a data instance in a function learning task, and a set of attribute variables $A_1, A_2, \ldots, A_d$ that $Z$ depends on, we can represent the dependence relationships among these variables with the top plot shown on the right, with notations borrowed from Bayesian networks. The notation $Z \to Y$ denotes the direct dependence relationship of variable $Y$ on variable $Z$; and $(A_1, A_2, \ldots, A_d) \to Z$ denotes the direct dependence of $Z$ on variables $A_1, A_2, \ldots, A_d$.

The \textit{interdependence padding} approach proposes to introduce one extra new variable $Z'$ to model the information from variables $A_1, A_2, \ldots, A_d$ that $Z$ depends on. The new variable $Z'$ acts as an intermediate bridge between $A_1, A_2, \ldots, A_d$ and $Z$. Also, according to the Bayesian network, the newly created variable $Z'$ renders $Z$ and $A_1, A_2, \ldots, A_d$ to be conditionally independent given $Z'$, {\ie} $Z {\ind} (A_1, A_2, \ldots, A_d) | Z'$.

There may exist different ways to define the new variable $Z'$. In this paper, we will define $Z'$ as a concatenation of the dependent variables $A_1, A_2, \ldots, A_d$, {\ie} $Z' = [A_1, A_2, \ldots, A_d]$. This new variable $Z'$ and the input variable $Z$ together will define the interdependence padding operator as

\begin{equation}
\text{padding}(Z | A_1, A_2, \cdots, A_d) = [Z, Z'].
\end{equation}

\begin{wrapfigure}{r}{0.35\textwidth}
    \vspace{-15pt}
    \centering
    \includegraphics[width=0.35\textwidth]{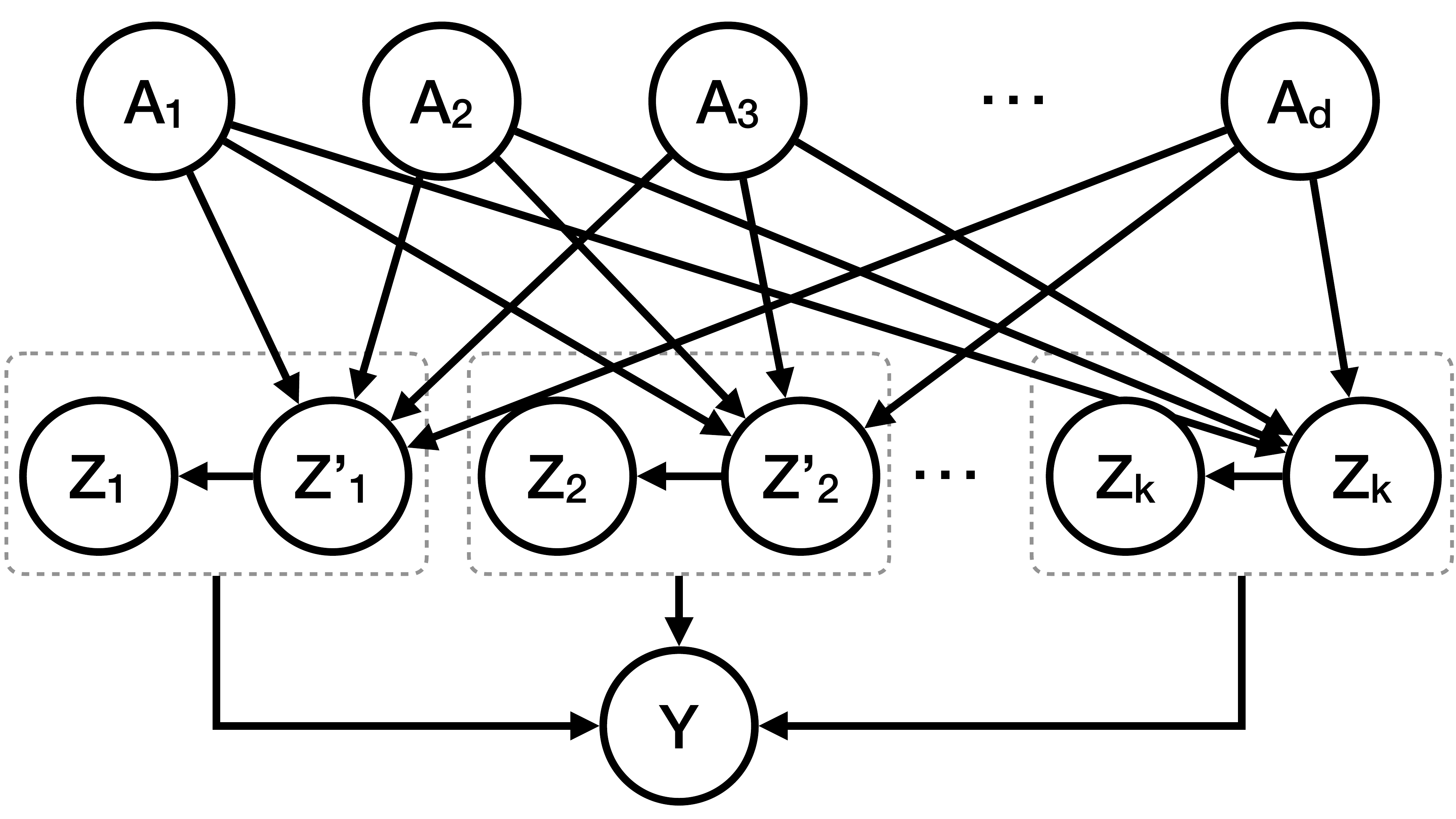}
    \caption{Redundancy in the Variable Interdependence Paddings.}
    \label{fig:redundancy_in_padding}
    \vspace{-5pt}
\end{wrapfigure}

For a single variable, the above interdependence padding-based approach may work well, as it includes all the dependent information into the padded new variable, which will be used for inferring the desired variable $Y$. However, in practice, there may exist multiple variables, such as $Z_1, Z_2, \ldots, Z_k$, which may all depend on $A_1, A_2, \ldots, A_d$, as illustrated in the right plot. To make the variables $Z_1, Z_2, \ldots, Z_k$ conditionally independent from those in $A_1, A_2, \ldots, A_d$, redundant interdependence padding can be applied to all the variables $Z_1, Z_2, \ldots, Z_k$ as follows:

\begin{equation}\label{equ:multi_variable_interdependence_padding}
\begin{cases}
&\text{padding}(Z_1 | A_1, A_2, \cdots, A_d) = [Z_1, Z_1'],\\
&\text{padding}(Z_2 | A_1, A_2, \cdots, A_d) = [Z_2, Z_2'],\\
& \cdots  \\
&\text{padding}(Z_k | A_1, A_2, \cdots, A_d) = [Z_k, Z_k'],
\end{cases}
\end{equation}

where the newly created padding variables $Z'_1, Z'_2, \cdots, Z'_k$ are all concatenations of the dependent variables $A_1, A_2, \cdots, A_d$. In other words, multiple duplicated copies of data vectors represented by variables $A_1, A_2, \cdots, A_d$ will be concatenated to those of $Z_1, Z_2, \cdots, Z_k$, which is actually how the existing models, like convolutional neural networks, handle the data interdependence.


\noindent \textbf{Discussions}: By default, we will only use the above interdependence padding for handling attribute interdependence relationships. Such duplicated and interdependence padding should not be a significant problem when the size or dimension of data that $A_1, A_2, \cdots, A_d$ represent is small. For example, if each variable represents a feature (such as a single pixel in images), duplicating these redundant features—similar to the approach in convolutional neural networks—does not pose significant learning challenges for small-sized input data. Additionally, the parameters used to handle these features can be shared across the newly created padding variables, thereby reducing the learning cost in terms of the number of parameters.

However, in real-world practice, such variables may also represent vectors of feature segments or data instances with high dimensions. Applying the above redundant interdependence padding-based approach will not only introduce much higher storage consumption but also require more learnable parameters to handle the longer variable list after padding. To address the problem, below, we will introduce an alternative data interdependence modeling method via the aggregation operator instead.


\subsubsection{Interdependence Aggregation}

Besides interdependence padding, another alternative approach to modeling such data interdependence relationships is via the interdependence aggregation operator, which incorporates information from all the variables via aggregation operators, such as weighted summation.

Formally, given the variable $Z$ and other variables $A_1, A_2, \ldots, A_d$ that it depends on, as illustrated by the upper plot of Figure~\ref{fig:dependence_padding}, the \textit{interdependence aggregation} approach proposes to integrate those variables as follows:

\begin{equation}
\text{aggregation}(Z | A_1, A_2, \cdots, A_d) = \alpha_0 \cdot Z + \alpha_1 \cdot A_1 + \alpha_2 \cdot A_2 + \cdots + \alpha_d \cdot A_d,
\end{equation}

where the scalar weights $\alpha_1, \ldots, \alpha_d \in \mathbbm{R}$ denote the interdependence strength of $Z$ on the corresponding variables in $A_1, A_2, \ldots, A_d$. Specifically, $\alpha_0$ denotes the interdependence of $Z$ on itself, which can also be referred to as self-dependence.

When it comes to the multi-variate cases as shown in Figure~\ref{fig:redundancy_in_padding}, similar operators can also be applied for the other variables $Z_1, Z_2, \ldots, Z_k$ which also depend on $A_1, A_2, \ldots, A_d$ as follows:

\begin{equation}\label{equ:multi_variable_interdependence_aggregation}
\begin{cases}
&\text{aggregation}(Z_1 | A_1, A_2, \cdots, A_d) = \alpha^{1}_0 \cdot Z_1 +  \alpha^{1}_1 \cdot A_1 + \alpha^{1}_2 \cdot A_2 + \cdots + \alpha^{1}_d \cdot A_d,\\
&\text{aggregation}(Z_2 | A_1, A_2, \cdots, A_d) = \alpha^{2}_0 \cdot Z_2 +  \alpha^{2}_1 \cdot A_1 + \alpha^{2}_2 \cdot A_2 + \cdots + \alpha^{2}_d \cdot A_d,\\
& \cdots  \\
&\text{aggregation}(Z_k | A_1, A_2, \cdots, A_d) = \alpha^{k}_0 \cdot Z_k +  \alpha^{k}_1 \cdot A_1 + \alpha^{k}_2 \cdot A_2 + \cdots + \alpha^{k}_d \cdot A_d.
\end{cases}
\end{equation}

There exist different ways to define the above scalars $\alpha_0, \alpha_1, \ldots, \alpha_d$ (with different superscripts). The simplest way is to assign them equal constants of $1$ (or $\frac{1}{d+1}$), which renders the above aggregation operator to a summation (or averaging) operator. To further distinguish and model the different roles of variables $A_1, A_2, \ldots, A_d$ on different variables $Z_1, Z_2, \ldots, Z_k$, we will introduce several different approaches to define the dependence strength weight parameters.

\noindent \textbf{Discussions}: Furthermore, besides weighted aggregation, some extra transformation operators (such as linear transformation, expansion, compression, and other more complex ones with learnable parameters) can also be applied to the variables prior to or after the aggregation. This provides the interdependence aggregation with greater modeling capacities for complex function learning problems. Some of these will be briefly discussed in the following part in this section, and more will be introduced in detail in Section~\ref{sec:function}.

Compared with the above interdependence padding method, the interdependence aggregation will consume less computational space and time, and provide greater learning capacities for modeling the interdependent data. However, there is no free lunch; interdependence aggregation also creates extra parameters (or hyper-parameters) which may need to be learned (or manually defined). Below, we will illustrate that these two different interdependence modeling approaches can be unified with the interdependence matrix.


\subsubsection{Data Interdependence Matrix}\label{subsubsec:interdependence_matrix}

Based on the above discussion, we introduce the concept of the \textit{interdependence matrix} in this paper, which can model both the attribute and instance interdependence relationships of the input data batch. Moreover, as mentioned above, the interdependence matrix can also unify the previously discussed interdependence padding and interdependence aggregation-based modeling approaches into one shared representation, which will be discussed as follows.

\begin{figure}[t]
    \centering
    \begin{subfigure}[b]{0.45\textwidth}
        \centering
        \includegraphics[width=\textwidth]{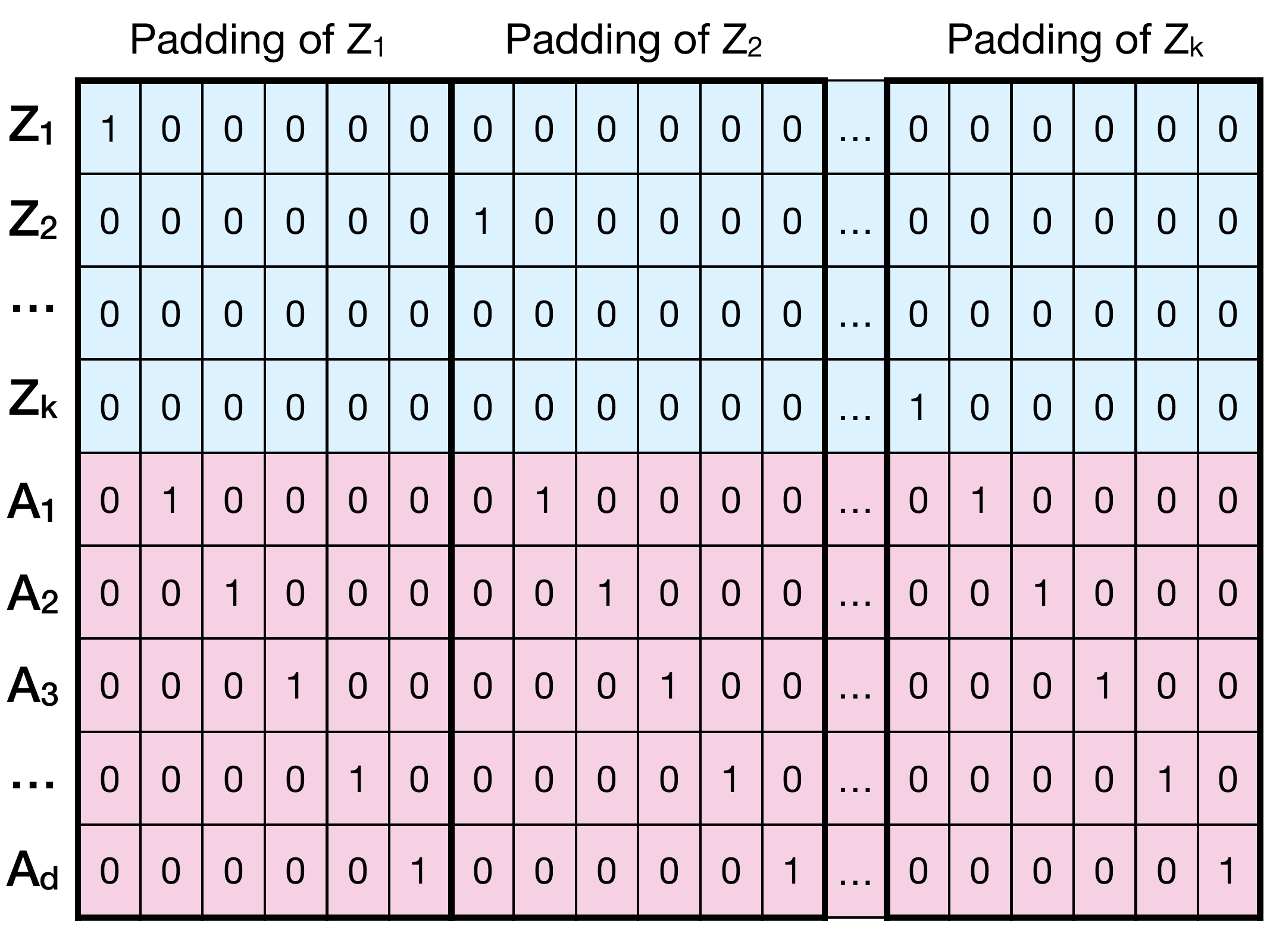}
    	\caption{Matrix $\mb{A}_{a}$ for Interdependence Padding.}
    	\label{fig:dependency_matrix_redundant_padding}
    \end{subfigure}
    \hfill
    \begin{subfigure}[b]{0.45\textwidth}
        \centering
        \includegraphics[width=\textwidth]{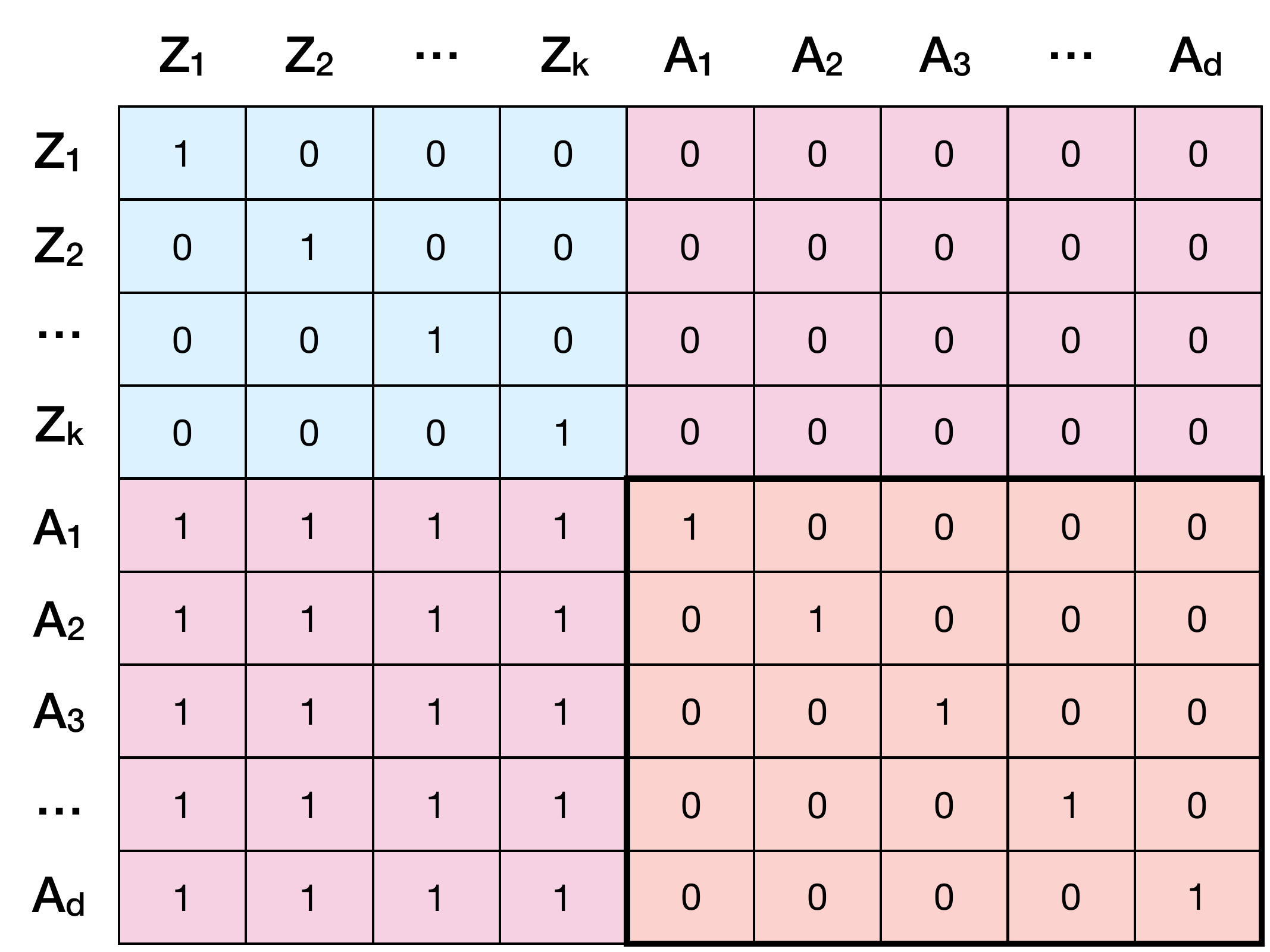}
    	\caption{Matrix $\mb{A}_{a}'$ for Interdependence Aggregation.}
    	\label{fig:dependency_matrix_interdependence_aggregation}
    \end{subfigure}
    \caption{An Illustration of Interdependence Matrix for Interdependence Padding and Interdependence Aggregation.}
    \label{fig:interdependence_matrix_example}
\end{figure}


\noindent \textbf{Attribute Interdependence Matrix}: Formally, given the variables $Z_1, Z_2, \ldots, Z_k$ shown in Figure~\ref{fig:redundancy_in_padding}, which depend on the attribute variables $A_1, A_2, \ldots, A_d$, we can group these variables and represent them as the data instance variable $X = [Z_1, Z_2, \ldots, Z_k, A_1, A_2, \ldots, A_d]$. To simplify the descriptions, we introduce a data instance vector $\mathbf{x} \in \mathbbm{R}^m$, where the dimension $m = k + d$ and the vector elements denote the values of variables $Z_1, Z_2, \ldots, Z_k$ and $A_1, A_2, \ldots, A_d$, respectively.

Based on the above notations, we can rewrite Equation~(\ref{equ:multi_variable_interdependence_padding}) for interdependence padding by multiplying the input data instance vector $\mathbf{x} \in \mathbbm{R}^m$ with the attribute interdependence matrix $\mathbf{A}_{a} \in \mathbbm{R}^{m \times (k \times (d+1))}$ shown in Figure~\ref{fig:dependency_matrix_redundant_padding} as follows:

\begin{equation}
\mb{x} = \mb{x} \mb{A}_{a} \in \mathbbm{R}^{k \times (d+1)}.
\end{equation}

The output vector is composed of the values corresponding to the variables $Z_1, Z_2, \ldots, Z_k$ with the paddings sequentially attached to them, whose length will be $k \times (d+1)$.

For the interdependence aggregation approach, with the interdependence matrix $\mathbf{A}'_{a} \in \mathbbm{R}^{m \times m}$ shown in Figure~\ref{fig:dependency_matrix_interdependence_aggregation}, we can also update the involved attributes data instance vector $\mathbf{x}$ by aggregating all the dependent conditions as follows:

\begin{equation}
\mb{x} = \mb{x} \mb{A}_{a}' \in \mathbbm{R}^{m}.
\end{equation}

The output vector has a length of $m$. For the entries corresponding to variables $Z_1, Z_2, \ldots, Z_k$ in the vector, they will be summed with their conditions; while for the entries corresponding to variables $A_1, A_2, \ldots, A_d$, they will remain unchanged in the output vector.

The above matrix $\mathbf{A}_a \in \mathbbm{R}^{m \times m'}$ (and $\mathbf{A}'_a$) describes the interdependence relationships among the attributes, which is named the attribute interdependence matrix. Depending on the choice of the interdependence matrix, the dimension term $m'$ denoting the output dimension of the data instance can take different values, which will be specified in the matrix definition.


\noindent \textbf{Instance Interdependence Matrix}: Similarly, for describing the data instance interdependence relationships within the data batch, we can also define the instance interdependence matrix $\mathbf{A}_{i} \in \mathbbm{R}^{b' \times b}$, where $b$ denotes the batch size and $b'$ is the output batch size after considering the instance interdependence relationships. For most cases, the term $b'$ will be equal to $b$, but we also provide flexibility and allow $b'$ to be a parameter determined by the matrix definition. Unlike $\mathbf{A}_{a}$ for attribute interdependence modeling ({\ie} the columns of data batch), the matrix $\mathbf{A}_{i}$ will operate on the data instances instead ({\ie} the rows).

Formally, given the input data batch $\mathbf{X} \in \mathbbm{R}^{b \times m}$ involving $b$ data instances, based on the data interdependence matrix $\mathbf{A}_{i}\in \mathbbm{R}^{b' \times b}$, we can represent the updated data batch that incorporates the data instance interdependence as follows:

\begin{equation}
\mb{X} = \mb{A}_{i} \mb{X} \in \mathbbm{R}^{b' \times m}.
\end{equation}

To consider both the attribute interdependence and instance interdependence relationships, we can transform the data batch $\mathbf{X}$ with both matrices $\mathbf{A}_{i}\in \mathbbm{R}^{b' \times b}$ and $\mathbf{A}_{a} \in \mathbbm{R}^{m \times m'}$ as follows:

\begin{equation}
\mb{X} = \mb{A}_{i} \mb{X} \mb{A}_{a} \in \mathbbm{R}^{b' \times m'}.
\end{equation}


\noindent \textbf{How to Define Data Interdependence Matrix}: In this section, we have demonstrated the use of data interdependence matrices for modeling attribute and instance dependence relationships within input data batches. The examples presented in Figure~\ref{fig:interdependence_matrix_example} and the quantitative metrics discussed in Section~\ref{subsec:quantitative_metrics} have illustrated various approaches to defining these interdependence matrices, which are applicable to a wide range of function learning tasks.

Meanwhile, it is important to note that these examples and metrics represent only specific instances of interdependence matrices. To enhance model flexibility and learning capacity in real-world applications, instead of manually pre-defining interdependence matrices, we introduce a new family of component functions called \textit{data interdependence functions} to define and learn such matrices instead. These functions can automatically compute and fine-tune data interdependence matrices based on the input data batch, additional contextual information, and optional learnable parameters.

Furthermore, the data interdependence functions in our newly developed {\toolkit} are designed to compute both attribute and instance interdependence matrices. These computations are guided by specified target dimension hyper-parameters ({\ie} ``attribute'' or ``instance''), allowing a single implementation to calculate interdependence matrices along any dimension of the data batch. The following sections will delve into these data interdependence functions and their integration into the redesigned {\our} model.

\section{{\our}: Enhanced {\old} with Data Interdependence Functions}\label{sec:method}

Building upon the previously discussed background knowledge of function learning tasks and the {\old} model, as well as the elucidated concepts of data interdependence, this section introduces the new {\our} (Reconciled Polynomial Network, version 2) model. 

This enhanced {\our} model incorporates data interdependence modeling capabilities through a suite of innovative component functions, collectively termed \textbf{data interdependence functions}. These incorporated functions enable {\our} to effectively address function learning tasks on diverse multi-modal data characterized by complex interdependence relationships.


\subsection{{\our}: Reconciled Polynomial Network 2}\label{subsec:rpn2_architecture}

Formally, given the underlying data distribution function $f: \mathbbm{R}^{m} \to \mathbbm{R}^{n}$ to be inferred, we can represent the new {\our} model with data interdependence modeling capabilities to approximate the underlying function $f$ as follows:

\begin{equation}
g(\mb{x} | \mb{w}) = \left \langle \kappa_{\xi} (\mb{x}), \psi(\mb{w}) \right \rangle + \pi(\mb{x}),
\end{equation}

where

\begin{itemize}

\item $\kappa_{\xi}: \mathbbm{R}^{m} \to \mathbbm{R}^{D}$ is named as the \textbf{data interdependent transformation function}. It is a composite function of the \textit{data transformation function} $\kappa$ and the \textit{data interdependence function} $\xi$. Notation $D$ denotes the target space dimension.


\item $\psi: \mathbbm{R}^l \to \mathbbm{R}^{n \times D}$ is named as the \textbf{parameter reconciliation function} defined on the parameters only. Notation $l$ denotes the learnable parameter vector space dimension.

\item $\pi: \mathbbm{R}^{m} \to \mathbbm{R}^{n}$ is named as the \textbf{remainder function}. 

\end{itemize}

Moreover, the data transformation function serves as a general term, encompassing both data expansion functions where $D \ge m$, and data compression functions where $D \le m$ (which will be introduced in the following Section~\ref{subsec:data_compression_function}). The data interdependence functions can model the relationships among both attributes and instances of the input data batch, whose detailed representation will be provided in the following subsection. The parameter reconciliation function fabricates a low-dimensional parameter vector of length $l$ to a high-dimensional parameter matrix of dimensions $n \times D$, where $l \ll n \times D$. It may also be referred to by its general name, the parameter fabrication function; these terms will be used interchangeably throughout this paper. The remainder function potentially provides complementary information and helps reduce the approximation errors. Figure \ref{fig:architecture} illustrates the {\our} model architecture, with its modules and components to be introduced in the following section.


\subsection{Data Interdependent Transformation Function}

The {data interdependence transformation function} $\kappa_{\xi}$, as introduced above, is a composition of the {data transformation function} $\kappa$ and the {data interdependence functions} $\xi$. As proposed in the previous {\old} paper \cite{zhang2024rpnreconciledpolynomialnetwork}, the data transformation function $\kappa$ efficiently projects input data into an intermediate vector space characterized by novel basis vectors, which is subsequently mapped to the output space through inner products with reconciled parameters. Meanwhile, the data interdependence functions $\xi$, newly introduced in this paper, capture the intricate interdependence relationships among data instances and attributes. These functions will extract nuanced information from the input data batch, operating both prior to and following the data projection facilitated by function $\kappa$.

\begin{definition}
(\textbf{Data Interdependence Function}): Formally, given an input data batch $\mb{X} \in \mathbbm{R}^{b \times m}$ (with $b$ instances and each instance with $m$ attributes), the attribute and instance data interdependence functions are defined as:

\begin{equation}
\xi_a: \mathbbm{R}^{b \times m} \to \mathbbm{R}^{m \times m'} \text{, and }
\xi_i: \mathbbm{R}^{b \times m} \to \mathbbm{R}^{b \times b'},
\end{equation}

where $m'$ and $b'$ denote the output dimensions of their respective interdependence functions, respectively.
\end{definition}

\begin{figure*}[t]
    \begin{minipage}{\textwidth}
    \centering
    	\includegraphics[width=1.0\linewidth]{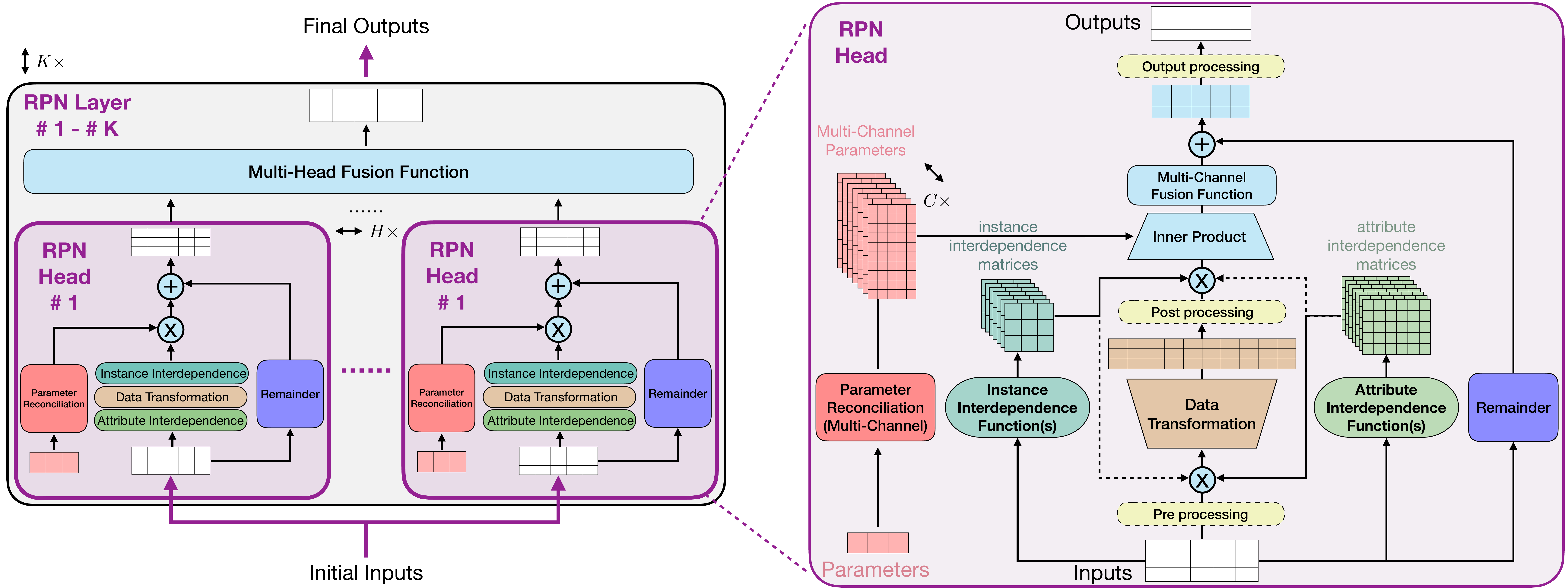}
    	\caption{An illustration of the {\our} framework. The left plot illustrates the multi-layer ($K$-layer) architecture of {\our}. Each layer involves multi-head for function learning, whose outputs will be fused together. The right plot illustrates the detailed architecture of the {\our} head, involving data transformation, multi-channel parameter reconciliation, remainder functions, and their internal operations. The attribute and instance interdependence functions calculate the interdependence matrices, which will be applied to the input data batch either prior or posterior to the data transformation function. The components in the rounded rectangle with yellow color in dashed lines denote the optional data processing functions ({\eg} activation functions and norm functions) for the inputs, expansions and outputs.}
    	\label{fig:architecture}
    \end{minipage}%
\end{figure*}

To elucidate the mechanisms of attribute and instance interdependence functions in defining the {data interdependence transformation function} $\kappa_{\xi}$, we shall consider a multi-instance input data batch $\mb{X} \in \mathbbm{R}^{b \times m}$ as an exemplar. Here, $b$ and $m$ denote the number of instances and attributes, respectively. Given this input data batch $\mb{X}$, as shown in the right plot of Figure~\ref{fig:architecture}, we can formulate the data interdependence transformation function $\kappa_{\xi}$ as follows:

\begin{equation}\label{equ:enhanced_data_transformation_function}
\kappa_{\xi}(\mb{X}) = \mb{A}^\top_{\xi_i} \kappa(\mb{X} \mb{A}_{\xi_a}) \in \mathbbm{R}^{b' \times D}.
\end{equation}

These attribute and instance interdependence matrices $\mb{A}_{\xi_a} \in \mathbbm{R}^{m \times m'}$ and $\mb{A}_{\xi_i} \in \mathbbm{R}^{b \times b'}$ are computed with the corresponding interdependence functions defined above, {\ie}

\begin{equation}\label{equ:enhanced_data_transformation_function_interdependence_matrix}
\mb{A}_{\xi_a} = \xi_a(\mb{X}) \in \mathbbm{R}^{m \times m'} \text{, and } \mb{A}_{\xi_i} = \xi_i(\mb{X}) \in \mathbbm{R}^{b \times b'}.
\end{equation}

The dimension of the target transformation space, denoted as $D$, is determined by the codomain dimension $m'$ of the attribute interdependence function. In most cases, the domain and codomain dimensions of the attribute and instance dependence functions analyzed in this paper are identical, {\ie} $m' = m$ and $b' = b$. However, for certain interdependence functions, adjustments to the codomain dimension are permitted, such as those incorporating padding modes discussed in Section~\ref{subsec:interdependence_matrix}. It is critical to note that the codomain dimensions $m'$ and $b'$ must are explicitly specified in the definitions of the functions $\xi_a$ and $\xi_i$, respectively.


\subsection{Versatile Data Interdependence Function}

The {\our} model features a versatile architecture. For the {\our} model implemented within the {\toolkit} toolkit, the default architecture adheres to the structure defined by Equation~\ref{equ:enhanced_data_transformation_function}. However, the {\toolkit} toolkit also provides users with the flexibility to modify the architecture to meet specific project requirements. 

In addition to the current interdependence matrices involved in Equation~\ref{equ:enhanced_data_transformation_function}, as depicted by the dashed lines connecting the attribute and instance interdependence matrices to the inputs and outputs of the data transformation function, the {\our} model enables the definition of attribute and instance interdependence matrices both prior and posterior to the data transformation operator. These configurations can be achieved with minor updates to the {\our} model architecture, as detailed below:

\begin{equation}
\kappa_{\xi}(\mb{X}) = 
\underbrace{\mb{A}^\top_{\xi_i^{\text{post}}}}_{\shortstack{\scriptsize \text{posterior instance} \\ \scriptsize \text{interdependence}}} 
\kappa\Big(
\underbrace{\mb{A}^\top_{\xi_i^{\text{prior}}}}_{\shortstack{\scriptsize \text{prior instance} \\ \scriptsize \text{interdependence}}} 
\mb{X} 
\underbrace{\mb{A}_{\xi_a^{\text{prior}}}}_{\shortstack{\scriptsize \text{prior attribute} \\ \scriptsize \text{interdependence}}} 
\Big) 
\underbrace{\mb{A}_{\xi_a^{\text{post}}}}_{\shortstack{\scriptsize \text{posterior attribute} \\ \scriptsize \text{interdependence}}} 
\in \mathbbm{R}^{b' \times D}.
\end{equation}

Furthermore, in practical implementations, all interdependence functions currently incorporated in the {\toolkit} toolkit establish interdependence relationships along the column dimension. With minor reshaping of inputs, these implementations can be applied across any dimension of the input data batch. In some implementations of interdependence functions, we allow for the inclusion of optional learnable parameters, denoted as $\mathbf{w}_{\xi_a} \in \mathbbm{R}^{l_{\xi_a}}$ and $\mathbf{w}_{\xi_i} \in \mathbbm{R}^{l_{\xi_i}}$, which slightly alter the function representations as follows:

\begin{equation}
\mb{A}_{\xi_a} = \xi_a(\mb{X} | \mb{w}_{\xi_a}) \in \mathbbm{R}^{m \times m'} \text{, and } \mb{A}_{\xi_i} = \xi_i(\mb{X}^\top | \mb{w}_{\xi_i}) \in \mathbbm{R}^{b \times b'}.
\end{equation}

By using $\mb{X}^\top$ as input to $\xi_i$, the column-based interdependence function implementation can also be leveraged to model interdependence relationships along the row dimension of the input data batch $\mb{X}$. By default, the interdependence matrix outputs incorporate optional pre- and post-processing operations, including but not limited to input data batch normalization, as well as the row and column normalizations of the output matrix.


\subsection{Wide and Deep Model Architectures}\label{subsec:wide_interdependence_functions}

Analogous to the previous {\old} model introduced in \cite{zhang2024rpnreconciledpolynomialnetwork}, the new {\our} model, enhanced with data interdependence functions, can also adopt a wide and deep architecture incorporating multi-head, multi-channel and multi-layer structures. These expansive architectures endow {\our} with greater model capacities for addressing function learning tasks on complex interdependent data.

\noindent \textbf{Wide Architecture}: As illustrated in the left plot of Figure~\ref{fig:architecture}, {\our} concurrently feeds the input batch to multiple heads, with each head possessing unique component functions. Simultaneously, as depicted in the right plot, each head within {\our} employs a multi-channel architecture. This architecture fabricates multiple copies of reconciled parameters and interdependence matrices to compute the desired output, specifically:

\begin{equation}\label{equ:wide_rpn2_model}
g(\mb{X} | \mb{w}, H, C) = \text{Fusion} \left( \left\{ \left\langle \kappa^{(h)}_{\xi^{(h),c}} \left( \mb{X} \right), \psi^{(h)} \left(\mb{w}^{(h), c}_{\psi} \right) \right\rangle + \pi^{(h)} \left(\mb{X} \right) \right\}_{h,c=1}^{H, C} \right),
\end{equation} 

where ``$\text{Fusion}(\cdot)$'' denotes the multi-head and multi-channel fusion functions (we will discuss it in the following subsection). Specifically, the data independent transformation function $\kappa^{(h)}_{\xi^{(h),c}} \left( \mb{X} \right)$ at the $h_{th}$-head and $c_{th}$-channel can be represented as

\begin{equation}
\kappa^{(h)}_{\xi^{(h),c}} \left( \mb{X} \right) = \left( \mb{A}_{\xi_i}^{(h),c} \right)^\top \kappa^{(h)} \left(\mb{X} \mb{A}_{\xi_a}^{(h),c} \right).
\end{equation}

The terms $\mb{A}_{\xi_a}^{(h), c}$ and $\mb{A}_{\xi_i}^{(h), c}$ represent the attribute and instance interdependence matrices for the $h_{th}$ head and $c_{th}$ channel of the model, respectively. Furthermore, in the above model notation $g(\mb{X} | \mb{w}, H, C)$, the learnable parameter vector $\mb{w}$ encompasses both the input parameters to the reconciliation functions, {\ie} $\left\{\mb{w}^{(h), c}_{\psi} \right\}_{h,c}$, and the (optional) parameters of the data interdependence functions, {\ie} $\left\{\mb{w}^{(h), c}_{\xi_a}, \mb{w}^{(h), c}_{\xi_i} \right\}_{h,c}$. For simplicity, we can represent the dimensions of these learnable parameter vectors $\mb{w}^{(h), c}_{\psi}$, $\mb{w}^{(h), c}_{\xi_a}$ and $\mb{w}^{(h), c}_{\xi_i}$ at different heads and channels to be $l_{\psi}$, $l_{\xi_a}$ and $l_{\xi_i}$, respectively, without indicating their head and channel indices.

The parameters of the data interdependence functions are, in fact, optional and not required for many of the interdependence functions to be introduced in the subsequent section. Throughout this paper, when discussing learnable parameters, we primarily refer to the parameters of the reconciliation functions, unless otherwise specified.

\noindent \textbf{Deep Architecture}: Similarly, by stacking multiple {\our} layers on top of each other, we may build a deep {\our} involving a deeper interdependence relationships among the features and data instances spanning across multiple layers:

\begin{equation}\label{equ:deep_rpn}
\begin{cases}
\text{ Input: } & \mb{h}_0  = \mb{X},\\
\text{ Layer 1: } & \mb{h}_1 = \left\langle \kappa_{\xi, 1} \left( \mb{X} \right), \psi_1(\mb{w}_{\psi, 1}) \right\rangle + \pi_1(\mb{X}),\\
\text{ Layer 2: } & \mb{h}_2 = \left\langle \kappa_{\xi, 2} \left( \mb{X} \right), \psi_2(\mb{w}_{\psi, 2}) \right\rangle + \pi_2(\mb{X}),\\
\cdots & \cdots \ \cdots\\
\text{ Layer K: } & \mb{h}_K = \left\langle \kappa_{\xi, K} \left( \mb{X} \right), \psi_K(\mb{w}_{\psi, K}) \right\rangle + \pi_K(\mb{X}),\\
\text{ Output: } & \hat{\mb{y}}  = \mb{h}_K.
\end{cases}
\end{equation}

In the aforementioned equations, we intentionally omitted the head and channel indices, as well as the fusion function, to streamline the notations. By default, each layer of {\our} may incorporate interdependence functions with multi-head and multi-channel architectures. Furthermore, each layer in the {\our} architecture can possess unique data interdependence functions, affording greater flexibility in model architecture design.


\subsection{Output Fusion Strategies}

To aggregate outputs from the wide architecture introduced above, {\our} introduces the ``Fusion($\cdot$)'' function as indicated in the above Equation~(\ref{equ:wide_rpn2_model}), which combines the learning results across multiple heads and channels. The {\toolkit} toolkit implements several different fusion strategies, which can be used within the {\our} model to consolidate outputs learned by this multi-head and multi-channel wide architecture. 

Some of the fusion strategies implemented in the {\toolkit} toolkit are briefly described as follows:

\begin{itemize}

\item \textbf{Summation}: This strategy directly aggregates the outputs learned by the multi-head and multi-channel model architecture via summation. It significantly enhances {\our}'s learning capacity without introducing substantial computational overhead.

\item \textbf{Average}: A variant of the summation fusion strategy, this approach calculates the average of outputs learned across different heads and channels of the {\our} model. Similar as the summation strategy, it treats those heads and channels with equal importance.

\item \textbf{Parameterized Fusion}: The parameterized fusion function strategy computes a weighted fusion ({\eg} summation, average) of the outputs from different heads and channels. The outputs from different heads and channels are assigned with different weights, which are defined as learnable parameters to be learned together with the model.

\item \textbf{Metric-based Fusion}: The metric-based fusion strategy aggregates outputs from different heads and channels using predefined numeric and statistical metrics, applied to elements at corresponding positions of the outputs across heads and channels. Examples of such metrics include, but are not limited to, maximum, minimum, product, median, and vector norms.

\item \textbf{Concatenation}: This method concatenates the learning results from the multi-head and multi-channel architecture for each instance along the column dimension, yielding an extended vector output for each data instance.

\item \textbf{Parameterized Concatenation}: To address potential space constraints arising from increased data batch sizes due to the concatenation fusion method, the parameterized concatenation fusion strategy incorporates a linear transformation function with learnable parameters. This function projects the concatenated outputs of the wide architecture into a dense vector for each data instance. To minimize the number of learnable parameters in the transformation, parameter reconciliation techniques from the previous study \cite{zhang2024rpnreconciledpolynomialnetwork}, such as low-rank approximations and dual low-rank hypercomplex multiplication, can be integrated into the fusion function.

\end{itemize}


\subsection{Computational Costs}

We previously have introduced the learning costs of the old {\old} without interdependence modeling capabilities in \cite{zhang2024rpnreconciledpolynomialnetwork} already. Here, we focus on analyzing the additional costs introduced by the data interdependence functions incorporated into the new {\our} model architecture. We assume the {\our} model has $K$ layers, $H$ heads, where each head involves $C$ channels. Each channel learns the attribute and instance interdependence matrices of dimensions $m \times m'$ and $b \times b'$ based on the input data batch of size $b \times m$ and (optional) learnable parameters of length $l_{\xi_a}$ and $l_{\xi_i}$. The learning cost of the attribute and instance interdependence functions are denoted as $t_a(m, m')$ and $t_i(b, b')$, respectively.

\begin{itemize}

\item \textbf{Space Cost}: The newly introduced storage requirements for the attribute and instance interdependence matrices, along with the (optional) extra learnable parameters, can be represented as $\mc{O}(K H (\underbrace{m m' + b b'}_{\text{space for matrices}} + \underbrace{C (l_{\xi_a} + l_{\xi_i})}_{\text{(optional) space for param.}}))$. In real practice, the interdependence matrices are typically sparse, resulting in substantially lower storage costs than the above notations $b'b$ and $m'm$ denoted above actually.

\item \textbf{Time Cost}: The additional computational cost for learning the attribute and instance interdependence matrices with the corresponding interdependence functions and multiplying them with the data batch can be represented as $\mc{O}(KH ( \underbrace{t_{a}(m, m')}_{\shortstack {\scriptsize \text{attribute matrix} \\ \scriptsize \text{computing}}} + \underbrace{t_i(b, b')}_{\shortstack {\scriptsize \text{instance matrix} \\ \scriptsize \text{computing}}} + \underbrace{b m m'}_{\shortstack {\scriptsize \text{attribute matrix} \\ \scriptsize \text{multiplication}}} + \underbrace{b' D b}_{\shortstack {\scriptsize \text{instance matrix} \\ \scriptsize \text{multiplication}}} ))$.

\item \textbf{Learnable Parameters}: Depending on the specific definitions of attribute and instance interdependence functions, some may involve additional learnable parameters. The number of involved learnable parameters can be represented as $\mc{O}(KHC (l_{\xi_a} + l_{\xi_i}))$, which is optional in real practice.

\end{itemize}


\section{Data Interdependence Functions}\label{sec:function}

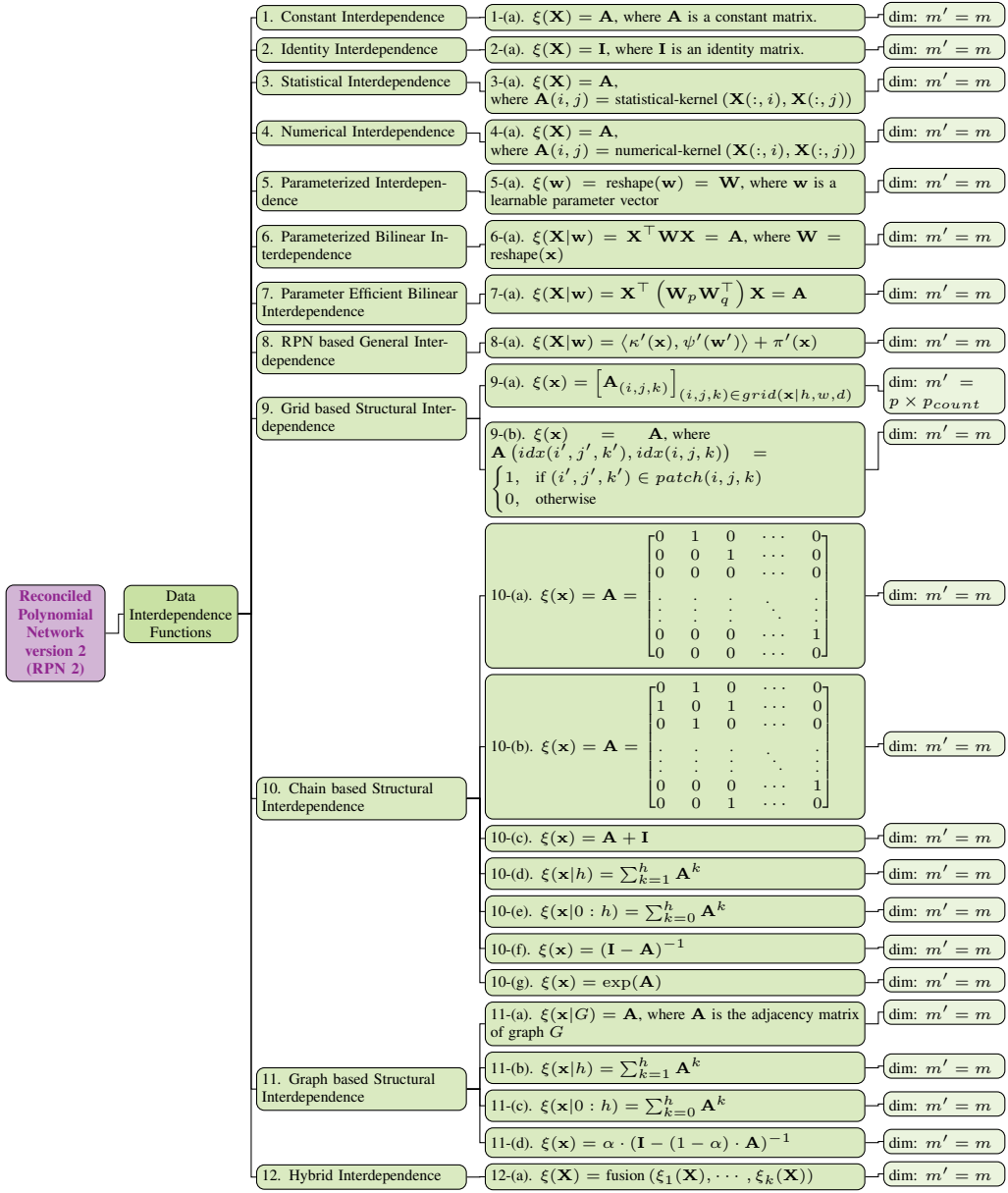
\begin{figure*}[t]
    \begin{center}
    \tiny
            \begin{forest}
                for tree={
                    forked edges,
                    grow'=0,
                    s sep=2.1pt,
                    draw,
                    rounded corners,
                    node options={align=center,},
                    text width=2.7cm,
                },
                [{\textcolor{Plum}{\textbf{Reconciled\\ Polynomial\\ Network\\ version 2\\ ({\our})}}}, fill=Plum!30, parent
                	    [Data\\ Interdependence\\ Functions, for tree={fill=LimeGreen!45, child}
                    	[1. Constant Interdependence, fill=LimeGreen!30, grandchild
                    		[{1-(a). $\xi(\mb{X}) = \mb{A}$, where $\mb{A}$ is a constant matrix.}, fill=LimeGreen!30, greatgrandchild
                                		[dim: {$m'=m$}, fill=LimeGreen!15, referenceblock]
                            	]
                        ]
                    	[2. Identity Interdependence, fill=LimeGreen!30, grandchild
                    		[{2-(a). $\xi(\mb{X}) = \mb{I}$, where $\mb{I}$ is an identity matrix.}, fill=LimeGreen!30, greatgrandchild
                                		[dim: {$m'=m$}, fill=LimeGreen!15, referenceblock]
                            	]
                        ]
                    	[3. Statistical Interdependence, fill=LimeGreen!30, grandchild
                    		[{3-(a). $\xi(\mb{X}) = \mb{A}$,\\ where $\mb{A}(i, j) = \text{statistical-kernel} \left(\mb{X}(:, i), \mb{X}(:, j)\right)$}, fill=LimeGreen!30, greatgrandchild
                                		[dim: {$m'=m$}, fill=LimeGreen!15, referenceblock]
                            	]
                        ]
                    	[4. Numerical Interdependence, fill=LimeGreen!30, grandchild
                    		[{4-(a). $\xi(\mb{X}) = \mb{A}$,\\ where $\mb{A}(i, j) = \text{numerical-kernel} \left(\mb{X}(:, i), \mb{X}(:, j)\right)$}, fill=LimeGreen!30, greatgrandchild
                                		[dim: {$m'=m$}, fill=LimeGreen!15, referenceblock]
                            	]
                        ]
                    	[5. Parameterized Interdependence, fill=LimeGreen!30, grandchild
                    		[{5-(a). $\xi(\mb{w}) = \text{reshape}(\mb{w}) = \mb{W}$, where $\mb{w}$ is a learnable parameter vector}, fill=LimeGreen!30, greatgrandchild
                                		[dim: {$m'=m$}, fill=LimeGreen!15, referenceblock]
                            	]
                        ]
                    	[6. Parameterized Bilinear Interdependence, fill=LimeGreen!30, grandchild
                    		[{6-(a). $\xi(\mb{X} | \mb{w}) = \mb{X}^\top \mb{W} \mb{X} = \mb{A}$, where $\mb{W} = \text{reshape}(\mb{x})$}, fill=LimeGreen!30, greatgrandchild
                                		[dim: {$m'=m$}, fill=LimeGreen!15, referenceblock]
                            	]
                        ]
                    	[7. Parameter Efficient Bilinear Interdependence, fill=LimeGreen!30, grandchild
                    		[{7-(a). $\xi(\mb{X} | \mb{w}) = \mb{X}^\top \left( \mb{W}_p \mb{W}_q^\top \right) \mb{X} = \mb{A}$}, fill=LimeGreen!30, greatgrandchild
                                		[dim: {$m'=m$}, fill=LimeGreen!15, referenceblock]
                            	]
                        ]
                    	[8. {\old} based General Interdependence, fill=LimeGreen!30, grandchild
                    		[{8-(a). $\xi(\mb{X} | \mb{w}) = \left\langle \kappa'(\mb{x}) , \psi'(\mb{w}') \right\rangle + \pi'(\mb{x})$}, fill=LimeGreen!30, greatgrandchild
                                		[dim: {$m'=m$}, fill=LimeGreen!15, referenceblock]
                            	]
                        ]
                    	[9. Grid based Structural Interdependence, fill=LimeGreen!30, grandchild
                    		[{9-(a). $\xi(\mb{x}) = \left[ \mb{A}_{(i, j, k)} \right]_{(i,j,k) \in grid(\mb{x} | h, w, d)}$}, fill=LimeGreen!30, greatgrandchild
                                		[dim: {$m'=p \times p_{count}$}, fill=LimeGreen!15, referenceblock]
                            	]
				[{9-(b). $\xi(\mb{x}) = \mb{A}$, where 
					$\mb{A}\left( {idx(i',j',k'), idx(i,j,k)} \right) =
						\begin{cases}
							1, & \text{if } (i', j', k') \in patch(i, j, k) \\
							0, & \text{otherwise}
						\end{cases}$
					}, fill=LimeGreen!30, greatgrandchild
                                		[dim: {$m'=m$}, fill=LimeGreen!15, referenceblock]
                            	]
                        ]
                    	[10. Chain based Structural Interdependence, fill=LimeGreen!30, grandchild
                    		[{10-(a). $\xi(\mb{x}) = \mb{A} =
							\begin{bmatrix}
							0 & 1 & 0 & \cdots & 0 \\
							0 & 0 & 1 & \cdots & 0 \\
							0 & 0 & 0 & \cdots & 0 \\
							\vdots & \vdots & \vdots & \ddots & \vdots \\
							0 & 0 & 0 & \cdots & 1\\
							0 & 0 & 0 & \cdots & 0
							\end{bmatrix}
						$}, fill=LimeGreen!30, greatgrandchild
                                		[dim: {$m'=m$}, fill=LimeGreen!15, referenceblock]
                            	]
				[{10-(b). $\xi(\mb{x}) = \mb{A} =
							\begin{bmatrix}
							0 & 1 & 0 & \cdots & 0 \\
							1 & 0 & 1 & \cdots & 0 \\
							0 & 1 & 0 & \cdots & 0 \\
							\vdots & \vdots & \vdots & \ddots & \vdots \\
							0 & 0 & 0 & \cdots & 1\\
							0 & 0 & 1 & \cdots & 0
							\end{bmatrix}
						$}, fill=LimeGreen!30, greatgrandchild
                                		[dim: {$m'=m$}, fill=LimeGreen!15, referenceblock]
                            	]
				[{10-(c). $\xi(\mb{x}) = \mb{A} + \mb{I}$}, fill=LimeGreen!30, greatgrandchild
                                		[dim: {$m'=m$}, fill=LimeGreen!15, referenceblock]
                            	]
				[{10-(d). $\xi(\mb{x}|h) = \sum_{k=1}^h \mb{A}^k$}, fill=LimeGreen!30, greatgrandchild
                                		[dim: {$m'=m$}, fill=LimeGreen!15, referenceblock]
                            	]
				[{10-(e). $\xi(\mb{x}| 0:h) = \sum_{k=0}^h \mb{A}^k$}, fill=LimeGreen!30, greatgrandchild
                                		[dim: {$m'=m$}, fill=LimeGreen!15, referenceblock]
                            	]
				[{10-(f). $\xi(\mb{x}) = (\mb{I} - \mb{A})^{-1}$}, fill=LimeGreen!30, greatgrandchild
                                		[dim: {$m'=m$}, fill=LimeGreen!15, referenceblock]
                            	]
				[{10-(g). $\xi(\mb{x}) = \exp(\mb{A})$}, fill=LimeGreen!30, greatgrandchild
                                		[dim: {$m'=m$}, fill=LimeGreen!15, referenceblock]
                            	]
                        ]
                        [11. Graph based Structural Interdependence, fill=LimeGreen!30, grandchild
                    		[{11-(a). $\xi(\mb{x} | G) = \mb{A}$, where $\mb{A}$ is the adjacency matrix of graph $G$}, fill=LimeGreen!30, greatgrandchild
                                		[dim: {$m'=m$}, fill=LimeGreen!15, referenceblock]
                            	]
				[{11-(b). $\xi(\mb{x}|h) = \sum_{k=1}^h \mb{A}^k$}, fill=LimeGreen!30, greatgrandchild
                                		[dim: {$m'=m$}, fill=LimeGreen!15, referenceblock]
                            	]
				[{11-(c). $\xi(\mb{x}| 0:h) = \sum_{k=0}^h \mb{A}^k$}, fill=LimeGreen!30, greatgrandchild
                                		[dim: {$m'=m$}, fill=LimeGreen!15, referenceblock]
                            	]
				[{11-(d). $\xi(\mb{x}) = \alpha \cdot \left( \mb{I} - (1- \alpha) \cdot {\mb{A}} \right)^{-1}$}, fill=LimeGreen!30, greatgrandchild
                                		[dim: {$m'=m$}, fill=LimeGreen!15, referenceblock]
                            	]
                        ]
                        [12. Hybrid Interdependence, fill=LimeGreen!30, grandchild
                    		[{12-(a). $\xi(\mb{X}) = \text{fusion} \left( \xi_1(\mb{X}), \cdots, \xi_k(\mb{X}) \right)$}, fill=LimeGreen!30, greatgrandchild
                                		[dim: {$m'=m$}, fill=LimeGreen!15, referenceblock]
                            	]
                        ]
                    ]
                ]
            \end{forest}
    \end{center}
    \caption{An overview of data interdependence, fusion, and data compression functions implemented in the {\toolkit} toolkit for constructing the {\our} model architecture. }\label{fig:interdependence_function_instances}
\end{figure*}

This section introduces a set of novel component functions for constructing the {\our} model, designed to address function learning tasks involving complex and interdependent data. We assume readers are familiar with the previous {\old} paper \cite{zhang2024rpnreconciledpolynomialnetwork} and the expansion, reconciliation, and remainder functions that have been already implemented in the {\oldtoolkit} toolkit, which will not be reiterated here. Readers seeking a concise overview of this section can also refer to Figure~\ref{fig:interdependence_function_instances}, which summarize the lists of interdependence functions to be introduced in this section.

Specifically, we introduce a new family of interdependence functions capable of modeling a wide range of interdependence relationships among both attributes and instances. These functions can be defined using input data batches, underlying geometric and topological structures, optional learnable parameters, or a hybrid combination of these elements. 

\subsection{Data Interdependence Functions}

The {data interdependence functions} are designed to model the complex relationships among attributes and data instances, corresponding to the columns and rows of the input data batch, respectively. This section presents several categories of interdependence functions, each firmly grounded in robust theoretical foundations, including linear algebra and statistical theory. The interdependence functions defined using computational geometry and topology approaches, which are based on input data batch side information such as shapes, structures, and spatial interconnections, will be introduced in the subsequent subsection instead.

Many of the interdependence functions presented herein are versatile, capable of computing interdependence matrices for both columns and rows of the input data batch, which correspond to the attribute and instances, respectively. However, to simplify the presentations below, we will focus on computing the interdependence matrices for the columns of the input data batch, {\ie} the attribute interdependence. It is worth noting that, as briefly discussed in the previous section, these function implementations can also be readily adapted to the alternative mode ({\eg} row-based analysis or instance interdependence) through the simple transposition of the input data batch.


\subsubsection{Constant Interdependence Function}

Among the array of interdependence functions to be introduced in this section, as one of the most basic function, the constant interdependence function generates an output interdependence matrix in the form of a constant matrix with a specified shape. 

Formally, based on the (optional) input data batch $\mb{X} \in \mathbbm{R}^{b \times m}$, we define the constant interdependence function as:

\begin{equation}
\xi(\mb{X}) = \mb{A} \in \mathbbm{R}^{m \times m'}.
\end{equation}

This function facilitates the definition of customized constant interdependence matrices, allowing for a manually defined matrix $\mb{A}$ to be provided as a hyper-parameter during function initialization. Two special cases warrant particular attention: when $\mb{A}_c$ consists entirely of zeros, it is designated as the \textit{zero interdependence matrix}, whereas a matrix of all ones is termed the \textit{one interdependence matrix}. Moreover, it is noteworthy that the constant interdependence function exclusively utilizes the shape information of the input data batch, specifically the dimension $m$. It does not incorporate any additional information from the data batch matrix $\mb{X}$ in constructing the output interdependence matrices. The second dimension, $m'$, can be manually specified when defining the interdependence matrix $\mb{A}$. In the absence of an explicit specification, by default, we will assign $m'=m$. More importantly, this function operates without learnable parameters, thereby incurring no additional learning costs during the model's training process.

As a standard feature shared with subsequent interdependence functions, this function accommodates the integration of pre- and post-processing operations, such as input batch normalization and activation, and the output matrix row and column normalizations. For brevity, this default capability will not be reiterated in the descriptions of subsequent functions.


\subsubsection{Identity Interdependence Function}

A notable special case of the aforementioned constant interdependence functions is the identity interdependence function. This function outputs the interdependence matrix as a diagonal constant identity (or eye) matrix, formally represented as:

\begin{equation}
\xi(\mb{X}) = \mb{I} \in \mathbbm{R}^{m \times m'},
\end{equation}

where the output interdependence matrix is, by default, a square matrix with $m'=m$.

The identity interdependence function proves particularly useful for modeling the \textit{independence} of attributes and data instances within the input data batch. Notably, all models developed based on the previous version of {\old}, as introduced in \cite{zhang2024rpnreconciledpolynomialnetwork}, can be precisely reduced to a special case of {\our} with identity interdependence functions for both attributes and instances, {\ie} both the instances and attributes are independent.

In practical applications, leveraging sparse matrix representation and multiplication techniques results in minimal additional storage and time costs introduced by the identity interdependence function, rendering these costs nearly negligible. For scenarios where even these minor costs are undesirable in modeling independence relationships, an alternative approach is to define the interdependence functions as ``None'' in the implementation, which will be properly handled by the {\toolkit} toolkit without incurring any extra space or time costs.


\subsubsection{Statistical Kernel Based Interdependence Function}

The previous Section~\ref{subsubsec:statistical_interdependence_matrix} has introduced several statistical kernels for quantifying interdependence relationships among vectors. Building upon this foundation, we now present an expansive family of interdependence functions derived from these kernels. Rooted in probability theory and statistical inference, these kernels offer the advantage of explicitly modeling and quantifying uncertainty in the interdependence calculations derived from the input data batch.

Formally, given a data batch $\mb{X} \in \mathbbm{R}^{b \times m}$, we can define the statistical kernel-based interdependence function as:

\begin{equation}
\xi(\mb{X}) = \mb{A} \in \mathbbm{R}^{m \times m'} \text{, where } \mb{A}(i, j) = \text{kernel} \left(\mb{X}(:, i), \mb{X}(:, j)\right),
\end{equation}

We now present a concise overview of several frequently employed statistical kernels that can be utilized to define the aforementioned interdependence function:

\noindent\begin{minipage}{.45\linewidth}
\begin{fleqn}
{
\begin{equation}
\begin{aligned}
&\underline{\textbf{(a) KL Divergence:}}\\[3pt]
&\text{kernel}(\mb{x}, \mb{y}) = \sum_i \mb{x}(i) \log \left(\frac{\mb{x}(i)}{\mb{y}(i)} \right),\\[3pt]
&\text{where $\mb{x}$ and $\mb{y}$ have been normalized}.
\end{aligned}
\end{equation}
}
\end{fleqn}
\end{minipage}
\hfill
\noindent\begin{minipage}{.52\linewidth}
\begin{fleqn}
{
\begin{equation}
\begin{aligned}
&\underline{\textbf{(b) Pearson Correlation:}}\\[3pt]
&\text{kernel}(\mb{x}, \mb{y}) = \frac{\sum_{i=1}^b (\frac{\mb{x}(i) - \mu_x}{\sigma_x} ) (\frac{\mb{y}(i) - \mu_y}{\sigma_y} )}{b} ,\\[3pt]
&\text{where $\mu_x, \mu_y, \sigma_x, \sigma_y$ are the mean and std.}
\end{aligned}
\end{equation}
}
\end{fleqn}
\end{minipage}

\noindent\begin{minipage}{.45\linewidth}
\begin{fleqn}
{
\begin{equation}
\begin{aligned}
&\underline{\textbf{(c) RV Coefficient:}}\\[3pt]
&\text{kernel}(\mb{x}, \mb{y}) = \frac{ tr(\bs{\Sigma}_{x,y} \bs{\Sigma}_{y,x} ) }{ \sqrt{ tr(\bs{\Sigma}_x^2) tr(\bs{\Sigma}_y^2) } } .\\[3pt]
\end{aligned}
\end{equation}
}
\end{fleqn}
\end{minipage}
\hfill
\noindent\begin{minipage}{.52\linewidth}
\begin{fleqn}
{
\begin{equation}
\begin{aligned}
&\underline{\textbf{(d) Mutual Information:}}\\[3pt]
&\text{kernel}(\mb{x}, \mb{y}) = \frac{1}{2} \log \left( \frac{ det(\bs{\Sigma}_x) det( \bs{\Sigma}_y) }{det \left( \bs{\Sigma} \right)} \right).\\[3pt]
\end{aligned}
\end{equation}
}
\end{fleqn}
\end{minipage}

We have previously provided detailed descriptions of the RV coefficient and mutual information kernels in Section~\ref{subsubsec:statistical_interdependence_matrix}. To further elucidate these concepts, let us consider an input data batch $\mb{X} \in \mathbbm{R}^{b \times m}$. We can compute the column-wise mean values as vector $\bs{\mu} \in \mathbbm{R}^m$, which allows us to define the centered data batch $\mb{X}'$ as follows:

\begin{equation}
\mb{X}' = \mb{X} - \begin{bmatrix} \bs{\mu} \\ \bs{\mu} \\ \cdots\\ \bs{\mu} \end{bmatrix} \in \mathbbm{R}^{b \times m}.
\end{equation}

Subsequently, we can define the column-wise covariance matrix $\bs{\Sigma}$ based on this centered data matrix:

\begin{equation}
\bs{\Sigma} = \frac{1}{b-1} \left( \mb{X}' \right)^\top \mb{X}' \in \mathbbm{R}^{m \times m}.
\end{equation}

This covariance matrix serves as a crucial component in calculating the aforementioned RV coefficient and mutual information kernels for computing the corresponding interdependence matrices.



\subsubsection{Numerical Kernel Based Interdependence Function}


In addition to the statistical metrics discussed above, we can also the define interdependence functions based on numerical metrics, some of which have been briefly introduced in Section~\ref{subsubsec:numerical_interdependence_matrix}. Similar as the above statistical kernel-based functions, these numerical kernel-based interdependence functions compute the pairwise numerical scores for column vectors within the input data batch, thereby constructing a comprehensive interdependence matrix.
 
Formally, given a data batch $\mb{X} \in \mathbbm{R}^{b \times m}$, we define the numerical metric-based interdependence function as:

\begin{equation}
\xi(\mb{X}) = \mb{A} \in \mathbbm{R}^{m \times m'} \text{, where } \mb{A}(i, j) = \text{kernel} \left(\mb{X}(:, i), \mb{X}(:, j)\right).
\end{equation}

By convention, the resulting matrix $\mb{A}$ is square, with dimensions $m' = m$. This section offers a diverse array of approaches for defining numerical metrics on vectors from the input data batch. We elucidate below a curated selection of these methods, encompassing those previously introduced in Section~\ref{subsubsec:numerical_interdependence_matrix}, as well as additional noteworthy new kernels:

\noindent\begin{minipage}{.48\linewidth}
\begin{fleqn}
{
\begin{equation}
\begin{aligned}
&\underline{\textbf{(a) Linear (Inner-Product) Kernel:}}\\[6pt]
&\text{kernel}(\mb{x}, \mb{y}) = \left\langle \mb{x}, \mb{y} \right \rangle,\\[6pt]
&\text{where $\mb{x}$ and $\mb{y}$ are the column vectors}\\
&\text{from the input data batch.}\\
\end{aligned}
\end{equation}
}
\end{fleqn}
\end{minipage}
\hfill
\noindent\begin{minipage}{.48\linewidth}
\begin{fleqn}
{
\begin{equation}
\begin{aligned}
&\underline{\textbf{(b) Polynomial Kernel:}}\\[6pt]
&\text{kernel}(\mb{x}, \mb{y} | c, d) = \left(\left\langle \mb{x}, \mb{y} \right \rangle +c \right)^d,\\[6pt]
&\text{where $c$ and $d$ are the hyper-parameters}\\
&\text{of the kernel function.}\\
\end{aligned}
\end{equation}
}
\end{fleqn}
\end{minipage}

\noindent\begin{minipage}{.48\linewidth}
\begin{fleqn}
{
\begin{equation}
\begin{aligned}
&\underline{\textbf{(c) Hyperbolic Tangent Kernel:}}\\[6pt]
&\text{kernel}(\mb{x}, \mb{y} | \alpha, c) = \text{tanh} \left( \alpha \left\langle \mb{x}, \mb{y} \right \rangle + c \right).\\[6pt]
\end{aligned}
\end{equation}
}
\end{fleqn}
\end{minipage}
\hfill
\noindent\begin{minipage}{.48\linewidth}
\begin{fleqn}
{
\begin{equation}
\begin{aligned}
&\underline{\textbf{(d) Exponential Kernel:}}\\[6pt]
&\text{kernel}(\mb{x}, \mb{y} | \gamma) = \exp \left(- \gamma \left\| \mb{x} - \mb{y} \right\|_1 \right).\\[6pt]
\end{aligned}
\end{equation}
}
\end{fleqn}
\end{minipage}

\noindent\begin{minipage}{.48\linewidth}
\begin{fleqn}
{
\begin{equation}
\begin{aligned}
&\underline{\textbf{(e) Cosine Similarity based Kernel:}}\\[6pt]
&\text{kernel}(\mb{x}, \mb{y}) = \frac{\left\langle \mb{x}, \mb{y} \right \rangle}{\left\|\mb{x}\right\| \cdot \left\| \mb{y} \right\|},\\[6pt]
\end{aligned}
\end{equation}
}
\end{fleqn}
\end{minipage}
\hfill
\noindent\begin{minipage}{.48\linewidth}
\begin{fleqn}
{
\begin{equation}
\begin{aligned}
&\underline{\textbf{(f) Minkowski Distance based Kernel:}}\\[6pt]
&\text{kernel}(\mb{x}, \mb{y}) = 1- \left\| \mb{x} - \mb{y} \right\|_p,\\[6pt]
&\text{where $p \in \{1, 2, \cdots, \infty\}$}.
\end{aligned}
\end{equation}
}
\end{fleqn}
\end{minipage}


\noindent\begin{minipage}{.48\linewidth}
\begin{fleqn}
{
\begin{equation}
\begin{aligned}
&\underline{\textbf{(g) Gaussian RBF Kernel:}}\\[6pt]
&\text{kernel}(\mb{x}, \mb{y} | \sigma) = \exp \left(- \frac{\left\| \mb{x} - \mb{y} \right\|^2_2}{2 \sigma^2} \right),\\[6pt]
\end{aligned}
\end{equation}
}
\end{fleqn}
\end{minipage}
\hfill
\noindent\begin{minipage}{.48\linewidth}
\begin{fleqn}
{
\begin{equation}
\begin{aligned}
&\underline{\textbf{(h) Laplacian Distance:}}\\[6pt]
&\text{kernel}(\mb{x}, \mb{y} | \sigma) = \exp \left(- \frac{\left\| \mb{x} - \mb{y} \right\|_1}{ \sigma} \right),\\[6pt]
\end{aligned}
\end{equation}
}
\end{fleqn}
\end{minipage}

\noindent\begin{minipage}{.48\linewidth}
\begin{fleqn}
{
\begin{equation}
\begin{aligned}
&\underline{\textbf{(i) Anisotropic RBF Kernel:}}\\[6pt]
&\text{kernel}(\mb{x}, \mb{y}) = \exp \left( - (\mb{x} - \mb{y}) \mb{A} (\mb{x} - \mb{y})^\top \right),\\[6pt]
&\text{where $\mb{A} = \text{diag}(\mb{a})$ is a diagonal matrix}\\
&\text{modeling vector element-specific scaling factors}.
\end{aligned}
\end{equation}
}
\end{fleqn}
\end{minipage}
\hfill
\noindent\begin{minipage}{.48\linewidth}
\begin{fleqn}
{
\begin{equation}
\begin{aligned}
&\underline{\textbf{(j) Custom Hybrid Kernels:}}\\[6pt]
&\text{kernel}(\mb{x}, \mb{y} | \alpha, \beta) = \alpha k_1(\mb{x}, \mb{y}) + \beta k_2(\mb{x}, \mb{y}),\\[6pt]
&\text{where $k_1$ and $k_2$ are custom designed kernel}\\
&\text{functions and $\alpha$, $\beta$ are their weights}.
\end{aligned}
\end{equation}
}
\end{fleqn}
\end{minipage}

The aforementioned Minkowski distance-based kernel metric serves as a generalized representation of several frequently employed distance metrics. Its versatility is evident in its ability to reduce to specific, well-known distances depending on the value of the parameter $p$ in the vector norm. Notable special cases include the Manhattan distance ($p=1$), Euclidean distance ($p=2$), and Chebyshev distance ($p=\infty$), which can be expressed as follows:

\begin{equation}
\begin{aligned}
& \text{kernel}(\mb{x}, \mb{y}) = 1- \left\| \mb{x} - \mb{y} \right\|_1 = 1- \sum_{i=1}^d\left| \mb{x}(i) - \mb{y}(i) \right|,\\
& \text{kernel}(\mb{x}, \mb{y}) = 1- \left\| \mb{x} - \mb{y} \right\|_2 = 1- \sqrt{ \sum_{i=1}^d \left(\mb{x}(i) - \mb{y}(i) \right)^2},\\
& \text{kernel}(\mb{x}, \mb{y}) = 1- \left\| \mb{x} - \mb{y} \right\|_{\infty} = 1- \max \left( \left\{ \left| \mb{x}(i) - \mb{y}(i) \right| \right\}_{i=1}^d \right).
\end{aligned}
\end{equation}


\subsubsection{Parameterized Interdependence Function}

In addition to the above interdependence function solely defined based on the input data batch, another category of fundamental interdependence functions is the parameterized interdependence function, which constructs the interdependence matrix exclusively from learnable parameters.

Formally, given a learnable parameter vector $\mb{w} \in \mathbbm{R}^{l_{\xi}}$, the parameterized interdependence function transforms it into a matrix of desired dimensions $m \times m'$ as follows:

\begin{equation}
\xi(\mb{w}) = \text{reshape}(\mb{w}) = \mb{W} \in \mathbbm{R}^{m \times m'}.
\end{equation}

This parameterized interdependence function operates independently of any data batch, deriving the output interdependence matrix solely from the learnable parameter vector $\mb{w}$, whose requisite length of vector $\mb{w}$ is $l_{\xi} = m \times m'$.

In addition to the above function, several parameter reconciliation techniques, as introduced in the preceding {\old} paper \cite{zhang2024rpnreconciledpolynomialnetwork}, can also be utilized to generate the interdependence matrix $\mb{W}$ of the desired shape from the input parameter vector $\mb{w}$. These methods, including {low-rank parameter reconciliation} (LoRR), {hypercomplex multiplication} (HM), {low-rank parameterized HM} (LPHM), and {dual LPHM}, also offer the advantages of reducing the number of required learnable parameters. For a comprehensive exposition of these techniques, readers are recommended to refer to the previous {\old} paper \cite{zhang2024rpnreconciledpolynomialnetwork}; we shall not revisit them in detail here.


\subsubsection{Parameterized Bilinear Interdependence Function}\label{subsubsec:parameterized_bilinear_interdependence_function}

In addition to the numerical metrics discussed above, we have previously introduced another quantitative measure, namely the \textit{bilinear form}, in Section~\ref{subsubsec:numerical_interdependence_matrix}. This measure enumerates all potential interactions between vector elements to compute interdependence scores. In this section, we propose defining the interdependence function based on the bilinear form with learnable parameters to model these element interactions.

Formally, given a data batch $\mb{X} \in \mathbbm{R}^{b \times m}$, we can represent the parameterized bilinear form-based interdependence function as follows:

\begin{equation}\label{equ:bilinear_interdependence_function}
\xi(\mb{X} | \mb{w}) = \mb{X}^\top \mb{W} \mb{X} = \mb{A} \in \mathbbm{R}^{m \times m},
\end{equation}

where $\mb{W} = \text{reshape}(\mb{w}) \in \mathbbm{R}^{b \times b}$ denotes the parameter matrix reshaped from the learnable parameter vector $\mb{w} \in \mathbbm{R}^{l_{\xi}}$ with length $l_{\xi} = b^2$.


\subsubsection{Parameter Efficient Bilinear Interdependence Function}

For input data batches with a large number of instances ({\ie} when $b$ is very large), the introduced parameter matrix $\mb{W}$ for the above parameterized bilinear interdependence function will also have a large dimension. To avoid introducing a large number of parameters for the interdependence function definition, we can employ parameter reconciliation techniques similar to those introduced in \cite{zhang2024rpnreconciledpolynomialnetwork}. These techniques, such as low-rank approximation, HM, LPHM, and Dual LPHM, can all be used to reduce the number of required learnable parameters. These methods have been implemented in the {\toolkit} toolkit and are all ready for use.

\noindent \textbf{Low-Rank Parameterized Bilinear Interdependence Function}: To illustrate, we present the bilinear interdependence function with low-rank parameter reconciliation, which will also be used to construct the Transformer model with the {\our} model architecture. More discussions on that will be provided in the following Section~\ref{sec:backbone_unification}.

Similar to the low-rank parameter reconciliation function, the parameter matrix $\mb{W} \in \mathbbm{R}^{b \times b}$ used in Equation~(\ref{equ:bilinear_interdependence_function}) can be represented as the product of two low-rank sub-matrices $\mb{W}_p \in \mathbbm{R}^{b \times r}$ and $\mb{W}_q \in \mathbbm{R}^{b \times r}$, both of rank $r$ (where $r \ll b$). This will transform the equation to:

\begin{equation}
\xi(\mb{X} | \mb{w}) = \mb{X}^\top \left( \mb{W}_p \mb{W}_q^\top \right) \mb{X} = (\mb{X}^\top \mb{W}_p)(\mb{X}^\top \mb{W}_q)^\top = \mb{A} \in \mathbbm{R}^{m \times m}.
\end{equation}

In the implementation, the parameter vector $\mb{w}$ is first partitioned into two sub-vectors and subsequently reshaped into two matrices $\mb{W}_p , \mb{W}_q \in \mathbbm{R}^{b \times r}$. This approach will reduce the number of the required learnable parameter vector $\mb{w}$ to $l_{\xi} = 2br$.


\subsubsection{{\old} based General Data Interdependence Function}

While it is beyond the scope of this paper to enumerate all the parameter-efficient reconciliation techniques introduced in previous work, it is worth noting that many of the data expansion techniques previously discussed in \cite{zhang2024rpnreconciledpolynomialnetwork} can also be potentially utilized in defining interdependence functions. For example, the inner-product and bilinear form based interdependence functions defined above can both be implemented based on Taylor's expansion of the input data batch.

To illustrate, consider the output interdependence matrix $\mb{A}$ generated by the bilinear interdependence function in Equation~(\ref{equ:bilinear_interdependence_function}), its $(i, i)_{th}$ element models the interactions of the $i_{th}$ column with itself:

\begin{equation}
\begin{aligned}
\mb{A}(i, i) &= \mb{X}(:, i)^\top \mb{W} \mb{X}(:, i) = \sum_{p=1}^b \sum_{q=1}^b \mb{X}(p, i) \mb{X}(q, i) \mb{W}(p, q),\\
&= \left \langle \mb{X}(:, i) \otimes \mb{X}(:, i), \text{flatten}(\mb{W}) \right \rangle,
\end{aligned}
\end{equation}

where $\otimes$ denotes the Kronecker product of the vectors and the $\text{flatten}(\cdot)$ operator will flatten the matrix to a vector for computing the inner product. In essence, the calculated $\mb{A}(i,i)$ contains the weighted summation of the second-order Taylor's expansion of the input data batch column vector $\mb{X}(:,i)$. Similar representations also hold for other entries of the interdependence matrix. 

To generalize this function definition, we propose to disentangle the interdependence function as a {\old}-head proposed in the previous paper \cite{zhang2024rpnreconciledpolynomialnetwork} composed of three components: the data expansion function, parameter reconciliation function, and remainder function (without further considering any interdependence relationships to avoid nested modeling). Formally, the interdependence function $\xi: \mathbbm{R}^{b \times m} \to \mathbbm{R}^{m \times m'}$ can be viewed as a mapping between the input vector space of dimension $(b \times m)$ and the output vector space of dimension $(m \times m')$. We propose to represent it as follows:

\begin{equation}
\xi(\mb{X} | \mb{w}) = \left\langle \kappa'(\mb{x}) , \psi'(\mb{w}') \right\rangle + \pi'(\mb{x}),
\end{equation}

where $\mb{x} = \text{flatten}(\mb{X})$ is the flattened vector of length $(b \times m)$ from the input data batch matrix $\mb{X} \in \mathbbm{R}^{b \times m}$. This is viewed as a single (independent) pseudo ``data instance'' for the above {\old}-layer. The notations are defined as:

 \begin{equation}
 \kappa': \mathbbm{R}^{(b \times m)} \to \mathbbm{R}^D \text{, } \psi': \mathbbm{R}^{l} \to \mathbbm{R}^{(m \times m') \times D} \text{, and }\pi_{\xi}: \mathbbm{R}^{(b \times m)} \to \mathbbm{R}^{(m \times m')},
 \end{equation}
 
which denote the data expansion function, parameter reconciliation function, and remainder function, respectively, as specifically defined for the interdependence function. The prime symbols attached to the function name notations are added to differentiate them from the component functions used for building the {\our} architecture.

By default, the component functions for data expansion, parameter reconciliation, and remainder introduced in previous work, as well as those to be introduced in this paper, can all be used to define the interdependence function illustrated above. To avoid over-complicating the interdependence function definition, we will only use the single-layer, single-head, and single-channel {\old}-layer here. Additionally, to prevent infinitely nested interdependence function definitions, we assume all attributes in this vector ``$\mb{x} = \text{flatten}(\mb{X})$'' to be independent.

The {\old}-layer based interdependence function described above provides readers with considerable flexibility in defining customized interdependence functions for diverse input data. These functions have been implemented in the {\toolkit} toolkit and are now ready for use.


\subsection{Structural Interdependence Functions}\label{subsec:structural_interdependence_functions}

This section delves into the definition of data interdependence functions based on the geometric and topological structural information inherent in the data batch. We examine diverse structural interdependence relationships among attributes and instances from computational geometry and topology perspectives, taking into account structural factors, such as spatial configurations, sequential orders, and interconnections, about the input data batch.

\noindent \textbf{Note}: This subsection contains numerous technical details accompanied by complex notations denoting grids, patches, and matrices. Despite its challenging nature, we strongly encourage readers to thoroughly engage with this material. The interdependence functions defined herein are crucial and will be instrumental in unifying Convolutional Neural Networks (CNNs), Recurrent Neural Networks (RNNs), and Graph Neural Networks (GNNs) into {\our}'s canonical representation. Omitting this subsection may impede readers' comprehension of the subsequent model unification and new technique designs.


\subsubsection{Modality Specific Structural Interdependence Relationships}

\begin{figure*}[t]
    \begin{minipage}{\textwidth}
    \centering
    	\includegraphics[width=1.0\linewidth]{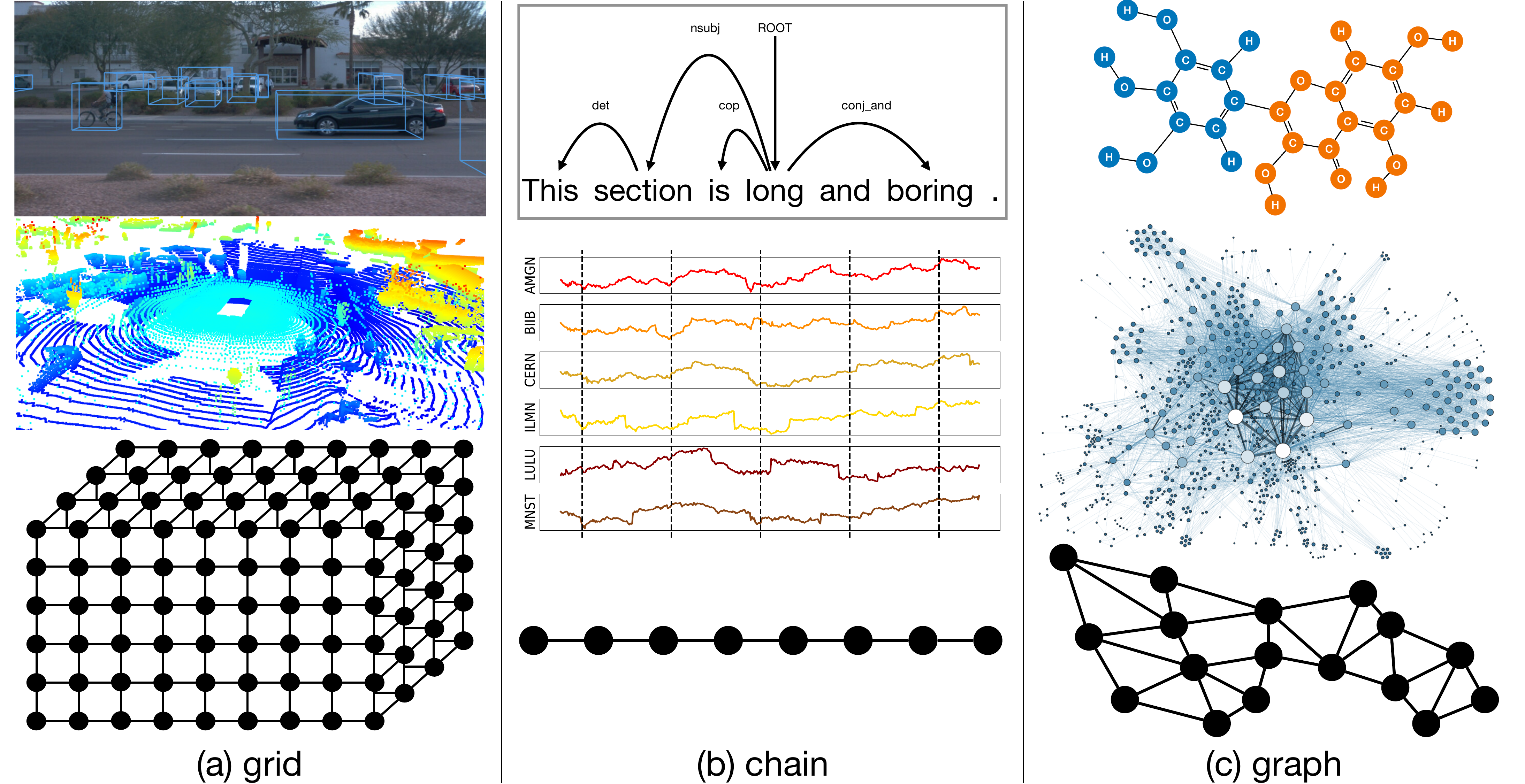}
    	\caption{An illustration of the underlying structures of multi-modal data. (a) Images and point clouds represented as grid-structured data, with nodes depicting pixels and voxels, and links representing spatial relationships; (b) Language and time series data represented as chain-structured data, with nodes depicting tokens and numerical values, and links representing sequential orders; (c) Molecule compounds and online social networks represented as graph-structured data, with nodes depicting atoms and users, and links representing atomic bonds and social connections.}
    	\label{fig:geometry_example}
    \end{minipage}%
\end{figure*}

\textit{Computational geometry and topology} are fields within computer science that focus on the study of algorithms expressible in terms of data geometry and topology structures. These disciplines involve the design, analysis, and implementation of algorithms to solve geometric and topological problems, often dealing with objects such as points, lines, and polygons, as well as concepts of connectedness, compactness, and continuity. Both computational geometry and topology play fundamental roles in numerous areas of computer science and machine learning, with their algorithms being crucial for processing, analyzing, and interpreting data in geometric and topological forms.

Figure~\ref{fig:geometry_example} illustrates examples of real-world data and their corresponding topological structures. Images and point clouds can be represented as grids in 2D and 3D space; language and time series data exhibit chain structures with sequential connections; and molecule compounds and social networks can be represented as graphs. These diverse data types demonstrate various structural interdependence relationships among instances and attributes in terms of spatial distributions, sequential orders, and extensive interconnections, respectively.

Such geometric and topological structure information of input data plays a crucial role in elucidating underlying data distribution patterns. However, when converting these data into mathematical representations ({\eg} vectors or matrices) and building models to fit them, much of this geometric and topological structure information is either lost or not fully utilized. In the following subsections, we will define diverse data interdependence functions to explicitly model these varied data interdependence relationships.

\noindent \textbf{Note}:  In addition to the grid, chain, and graph structures discussed here, there exist other data types (such as graphics mesh, hierarchical ontology, climate and space satellite sensor data) that can be represented with alternative or more complex structures. We plan to study these in future papers addressing concrete real-world application problems and will incorporate their corresponding data interdependence functions into the {\toolkit} toolkit as well.


\subsubsection{Topological Grid Structure and Geometric Patch Shapes}\label{subsubsec:data_grid_patch_structure}

This section introduces several structural interdependence functions based on geometric patches present in input data batches with grid structures. These geometric patch-based data structural interdependence functions are primarily applied to model the interdependence relationships among attributes, such as pixels and voxels in images and point clouds. Additionally, in practical applications, they can also be utilized to model relationships among data instances with fixed grid-structured interdependence relationships.

\begin{figure*}[t]
    \begin{minipage}{\textwidth}
    \centering
    	\includegraphics[width=0.9\linewidth]{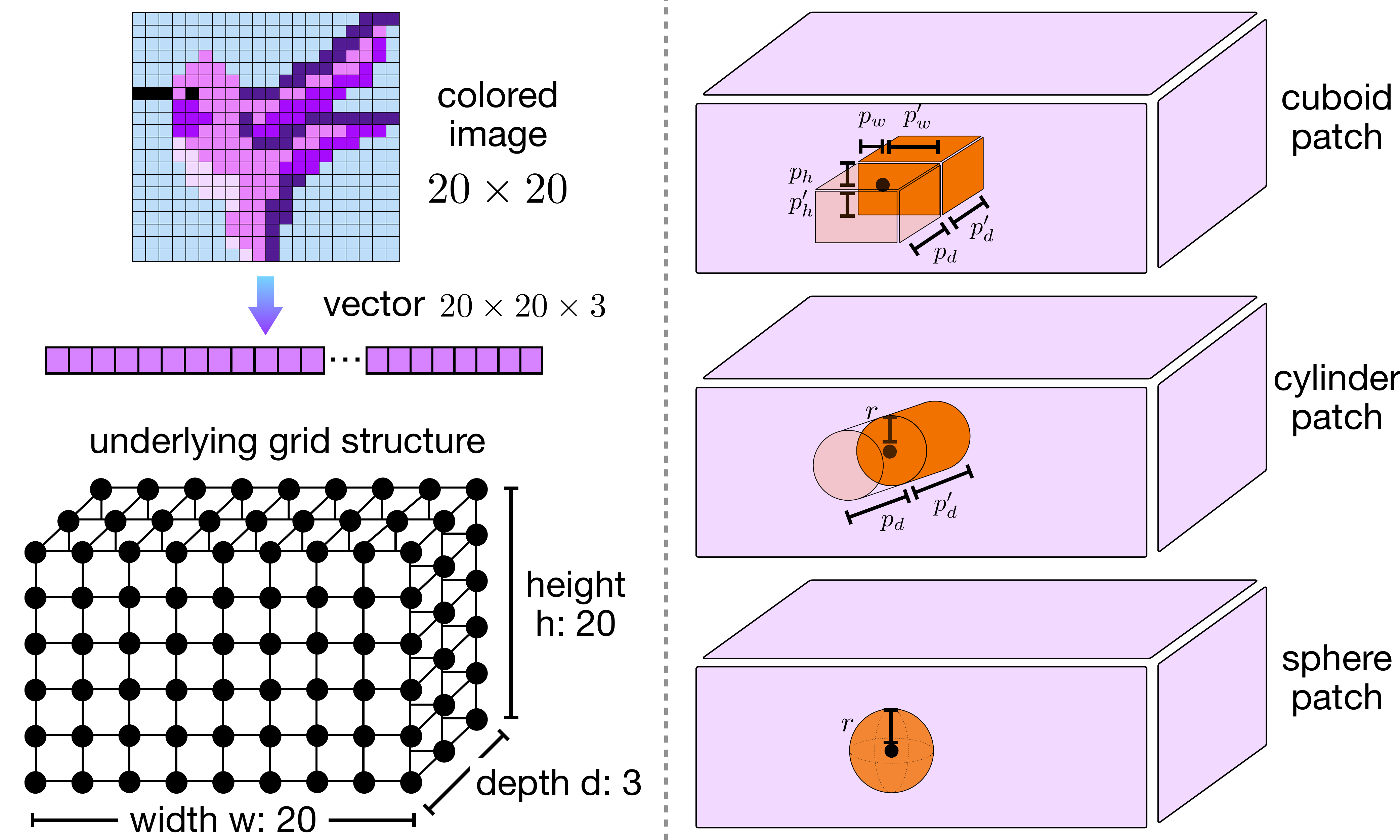}
    	\caption{An illustration of data grid structure and diverse patch shapes. (a) Left: Representation of an input data instance vector and its corresponding 3D grid structure. (b) Right: Illustration of three patch types defined on the grid: cuboid, cylinder, and sphere.}
    	\label{fig:patch_illustration}
    \end{minipage}%
\end{figure*}

\noindent \textbf{Geometric Grid}: Formally, given a data instance vector $\mb{x} \in \mathbbm{R}^{m}$, as illustrated in Figure~\ref{fig:patch_illustration}, we can represent its underlying 3D grid structure as an ordered list of coordinates denoting the attributes' locations in the grid:

\begin{equation}
grid(\mb{x} | h, w, d) = \left[ \left(i, j, k\right) \right]_{i \in \{0, 1, \cdots, h-1\}, j \in \{0, 1, \cdots, w-1\}, k \in \{0, 1, \cdots, d-1\}}.
\end{equation}

This paper focuses on cuboid-structured grids, which are prevalent as underlying structures of data instances in real-world applications, such as the aforementioned images and point clouds. The grid size is represented as $| grid(\mb{x} | h, w, d) | = h \times w \times d$, where $h, w, d$ denote the height, width, and depth dimensions of the grid, respectively. The grid size should equal the length of the data vector, {\ie} $m = h \times w \times d$, where $\mb{x} \in \mathbbm{R}^m$. By default, we can also use the input data instance vector dimension $m$ to represent the grid size.


\noindent \textbf{Attribute Index - Grid Coordinate Bijection}: A bijective mapping exists between the coordinates of the data instance vector $\mb{x}$ and its corresponding grid structure $ grid(\mb{x} | h, w, d) $. For each attribute in vector $\mb{x}$, we can identify its corresponding coordinates in its underlying grid; conversely, for each coordinate tuple from the grid, we can precisely locate the attribute from vector $\mb{x}$ by its index.

Formally, given the coordinate index tuple $(i, j, k)$ from $grid(\mb{x} | h, w, d)$, its corresponding attribute index in vector $\mb{x}$ can be represented as:

\begin{equation}\label{equ:idx_calculation}
idx(i, j, k) = i \cdot w \cdot d + j \cdot d + k, \forall i \in \{0, 1, \cdots, h-1\}, j \in \{0, 1, \cdots, w-1\}, k \in \{0, 1, \cdots, d-1\}.
\end{equation} 

Conversely, given an attribute with index $idx$ from vector $\mb{x}$, we can calculate its corresponding coordinate tuples $(i, j, k)$ in $grid(\mb{x} | h, w, d)$ as follows:

\begin{equation}
\begin{aligned}
i = \left \lfloor \frac{idx}{w \cdot d} \right \rfloor \text{ , } j = \left \lfloor \frac{idx \% (w \cdot d)}{d} \right \rfloor \text{ , } k = idx \% d,
\end{aligned}
\end{equation}

where notation $\%$ denotes the modulus operator and $\left \lfloor \cdot \right \rfloor$ is the floor operator.

These bijection mappings enable direct access to the attributes in the data instance $\mb{x}$ and the coordinate tuples in its underlying grid structure $grid(\mb{x} | h, w, d)$. For simplicity, in subsequent descriptions, we may treat these attributes and grid coordinates equivalently without distinguishing their differences.


\noindent \textbf{Geometric Cuboid Patch}: A geometric patch denotes a small-sized, localized region in the input data instance's underlying grid structure, facilitating the analysis of local data structures and details by breaking down complex data into manageable local parts. As illustrated in Figure~\ref{fig:patch_illustration}, the {\our} model allows patches of different shapes, such as \textit{cuboid}, \textit{cylinder}, and \textit{sphere}, each capturing distinct types of local structural interdependence relationships.

As shown in Figure~\ref{fig:patch_illustration}, these geometric patches of different shapes can be described with a set of hyper-parameters denoting the sizes, such as $(p_h, p_h'; p_w, p_w'; p_d, p_d')$ for cuboid patch, $(r; p_d, p_d')$ for cylinder patch, and $(r)$ for sphere patch. For cases where cuboid and cylinder patches have symmetric shapes along dimensions ({\ie} $p_h'=p_h$, $p_w'=p_w$, $p_d'=p_d$ for cuboid patch and $p_d' = p_d$ for cylinder patch), these shape hyper-parameters can be simplified to $(p_h; p_w; p_d)$ for cuboid patch and $(r; p_d)$ for cylinder patch.

Given a coordinate tuple $(i, j, k)$ in the grid, we can represent the patch, {\eg} a cuboid with shape $(p_h, p_h'; p_w, p_w'; p_d, p_d')$, centered at $(i, j, k)$ as an ordered list of coordinate tuples:

\begin{equation}\label{equ:cuboid_patch}
patch(i, j, k) = \left[ \left( i+ \nabla i, j + \nabla j, k + \nabla k \right) \right]_{\nabla i \in [-p_h, p_h'], \nabla j \in [- p_w, p_w'], \nabla k \in [- p_d, p_d']},
\end{equation}

Its size is represented as $\left| patch(i, j, k) \right| = (p_h + p_h' + 1) \times (p_w + p_w' + 1) \times (p_d + p_d' + 1)$, denoting the number of coordinate tuples covered by the patch. For simplicity, we introduce the term $p$ to represent the patch size, {\ie}

\begin{equation}
p = \left| patch(i, j, k) \right|,
\end{equation}

which is determined by the hyper-parameters and will be used frequently in the follow descriptions.


\noindent \textbf{Cylinder and Sphere Patches}: Similarly, we can represent cylinder patches of shape $(r; p_d, p_d')$ and sphere patches of shape $(r)$ centered at coordinates $(i, j, k)$ as follows:

\begin{equation}\label{equ:cylinder_sphere_patch}
\begin{aligned}
patch(i, j, k) &= \left[ \left( i+ \nabla i, j + \nabla j, k + \nabla k \right) \right]_{\nabla i, \nabla j \in [-r, r] \land \nabla i^2 + \nabla j^2 \le r^2, \nabla k \in [- p_d, p_d']},\\
patch(i, j, k) &= \left[ \left( i+ \nabla i, j + \nabla j, k + \nabla k \right) \right]_{\nabla i, \nabla j, \nabla k \in [-r, r] \land \nabla i^2 + \nabla j^2 + \nabla k^2 \le r^2},
\end{aligned}
\end{equation}

whose size is also represented by the term $p = \left| patch(i, j, k) \right|$ by default.

The main motivation for proposing the cylinder and sphere patch shapes in this paper is due to their advantages in maintaining their rotational invariances. Rotational transformations frequently occur in various types of input data during data sensing, collection, storage, clean and processing, particularly in images and point clouds. When considering cuboid patch shapes, although the data instance itself remains unaltered under rotation ({\eg} about an axis parallel to the depth), the grid nodes encompassed by these patches can undergo significant variations.

In contrast, cylinder and sphere patches exhibit superior rotational invariance characteristics:

\begin{itemize}

\item \textbf{Cylinder patches}: These maintain consistent attribute coverage when the data undergoes rotation around an axis parallel to the depth dimension.

\item \textbf{Sphere patches}: These demonstrate even greater invariance properties. The grid nodes encompassed by a sphere patch remain constant irrespective of the orientation of the rotation axis in three-dimensional space.

\end{itemize}

These inherent properties of cylinder and sphere patches provide robust alternatives to cuboid shapes, especially in scenarios where rotational invariance is crucial for data analysis and processing.


\subsubsection{Geometric Patch based Structural Interdependence Function}\label{subsubsec:patch_structural_interdependence_function}

\begin{figure*}[t]
    \begin{minipage}{\textwidth}
    \centering
    	\includegraphics[width=0.9\linewidth]{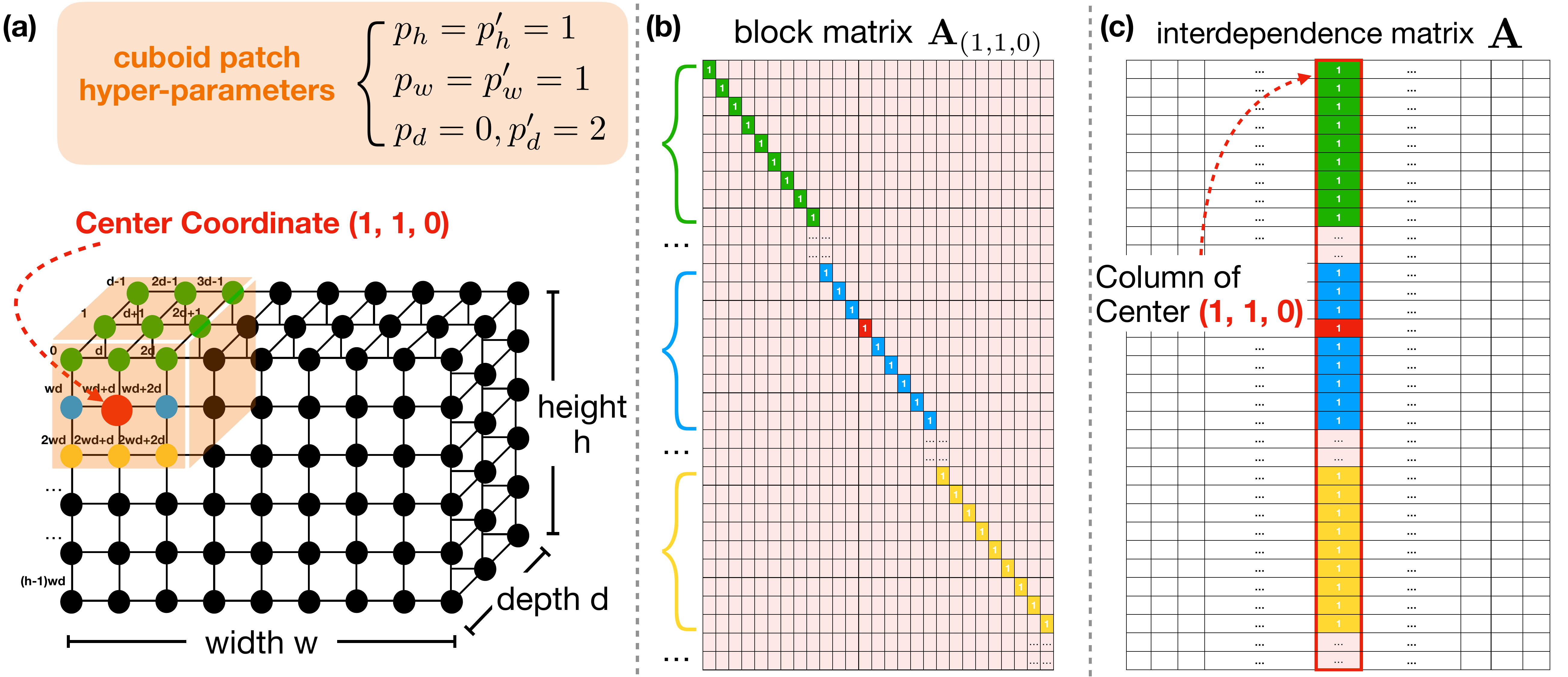}
    	\caption{An illustration of cuboid patch based structural interdependence matrices. (a) Underlying grid of input data instance and cuboid patch centered as $(1, 1, 0)$ with pre-defined sizes. (b) Block interdependence matrix $\mb{A}_{(1, 1, 0)}$ in the padding mode of center coordinate $(1, 1, 0)$. (c) Interdependence matrix $\mb{A}$ in the aggregation mode and the column corresponding to coordinate $(1, 1, 0)$.}
    	\label{fig:patch_interdependence_matrix_illustration}
    \end{minipage}%
\end{figure*}

Based on the preceding descriptions, we introduce the geometric patch-based structural interdependence function defined for input data instance $\mb{x} \in \mathbbm{R}^m$ as follows:

\begin{equation}\label{equ:patch_structural_interdependence_function}
\xi(\mb{x}) = \mb{A} \in \mathbbm{R}^{m \times m'}.
\end{equation}

As depicted in Plot (a) of Figure~\ref{fig:patch_interdependence_matrix_illustration}, the composition of the interdependence matrix $\mb{A}$ can be conceptualized as the systematic placement and translation of pre-defined patches along the dimensions of the data grid structure. In this representation, the surrounding nodes within each patch define the dependent conditions for the central node (highlighted in red) in the grid. This structural arrangement aligns with the two distinct operational modes of the interdependence function, namely the interdependence padding and interdependence aggregation modes, as elaborated in Section~\ref{subsec:interdependence_matrix}. These modes, while utilizing the same underlying patch structure, result in different compositions of the interdependence matrices $\mb{A}$, each capturing unique aspects of the data's structural interdependencies and leading to different learning performance.

 
\noindent \textbf{Padding Mode}: In the interdependence padding mode, the function composes matrix $\mb{A}$ as the concatenation of a sequence of block matrices:

\begin{equation}
\xi(\mb{x}) = \mb{A} = \left[ \mb{A}_{(i, j, k)} \right]_{(i,j,k) \in grid(\mb{x} | h, w, d)} \in \mathbbm{R}^{m \times m'}.
\end{equation}

For each coordinate tuple $(i, j, k) \in grid(\mb{x} | h, w, d)$ in the underlying grid structure of instance vector $\mb{x}$, a block sub-matrix $\mb{A}_{(i,j,k)} \in \mathbbm{R}^{m \times p}$ is defined. Specifically, the block sub-matrix $\mb{A}_{(i,j,k)}$ has $p$ columns, each corresponding to one of the coordinate tuples in the $patch(i, j, k)$ centered at coordinate $(i, j, k)$. Moreover, for the column of coordinate $(i', j', k') \in patch(i, j, k)$, all entries are filled with zeros except the entry with index $idx(i', j', k')$, which is filled with value 1.

Plot (b) in Figure~\ref{fig:patch_interdependence_matrix_illustration} provides a visual representation of this structure. It illustrates the sparse block sub-matrix $\mb{A}_{(1, 1, 0)}$, corresponding to the patch centered at coordinate tuple $(1, 1, 0)$. In this visualization, entries with value 1 are prominently highlighted, while the remaining entries, all zeros, form the background of the matrix.

The concatenation of these sub-matrices along the column dimension composes the sparse matrix $\mb{A} \in \mathbbm{R}^{m \times m'}$. The matrix dimension $m'$ can be represented as $m'= (h \times w \times d) \times p = m \times p$, which is proportional to the sizes of both the grid and patches.


\noindent \textbf{Aggregation Mode}: In contrast, the interdependence matrix defined in the aggregation mode is considerably denser:

\begin{equation}
\xi(\mb{x}) = \mb{A} \in \mathbbm{R}^{m \times m'}.
\end{equation}

In the underlying grid structure of instance vector $\mb{x}$, each coordinate tuple $(i, j, k) \in grid(\mb{x} | h, w, d)$ corresponds to a specific column in matrix $\mb{A}$. This column is uniquely identified by the index $idx(i,j,k)$. Plot (c) of Figure~\ref{fig:patch_interdependence_matrix_illustration} provides a visual representation of how the entries in this column are populated. The filling pattern is determined by the coordinates encompassed by the patch centered at $(i,j,k)$, as follows:

\begin{equation}
\mb{A}\left( {idx(i',j',k'), idx(i,j,k)} \right) =
\begin{cases}
1, & \text{if } (i', j', k') \in patch(i, j, k) \\
0, & \text{otherwise}
\end{cases},
\end{equation}

for all $(i, j, k) \in grid(\mb{x} | h, w, d)$. Furthermore, in this mode, the matrix dimension $m'$ equals the size of the grid, {\ie} $m' = m$, and is independent of the patch size.

The interdependence function exhibits remarkable versatility, accommodating patches of diverse geometric shapes through its two distinct compositional modes for interdependence matrices. When applied to a given input data instance, variations in patch shape yield different manifestations of the $patch(i, j, k)$ notation, as formally expressed in Equations~(\ref{equ:cuboid_patch}) and (\ref{equ:cylinder_sphere_patch}). These shape-induced variations subsequently engender distinct interdependence matrices, each capturing unique structural relationships within the data. The efficacy of different patch shapes, as well as the comparative performance of the function's two operational modes, will be rigorously assessed through empirical evaluation in the following Section~\ref{sec:experiments}.


\subsubsection{Cuboid Geometric Patch Packing based Structural Interdependence Function}\label{subsubsec:cuboid_patch}

\begin{figure*}[t]
    \begin{minipage}{\textwidth}
    \centering
    	\includegraphics[width=0.9\linewidth]{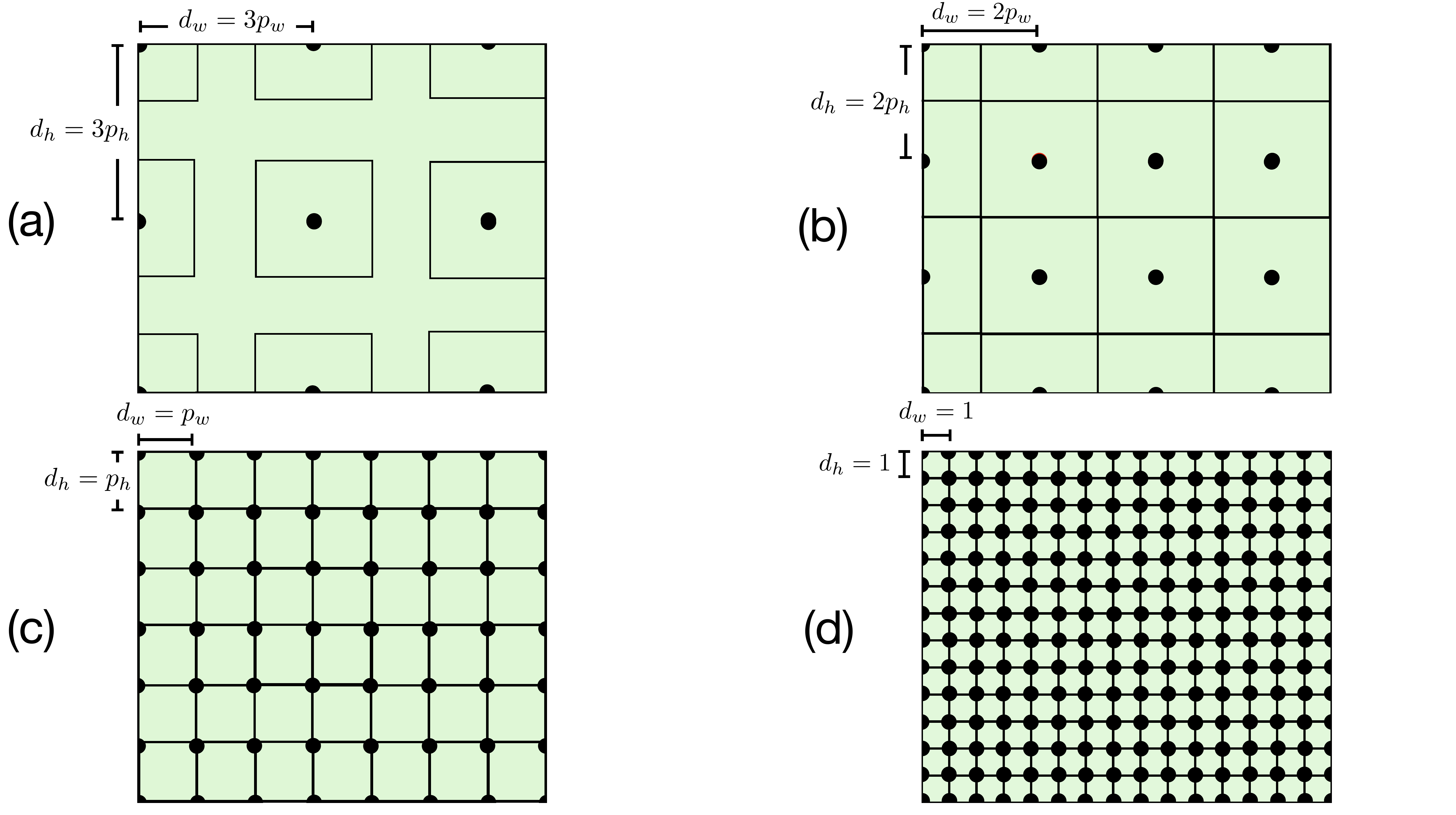}
    	\caption{An illustration of cuboid geometric patch with shape parameters $p_h'=p_h$, $p_w'=p_w$ (and $p_d'=p_d$) packing with varying patch center distance parameters $d_h$, $d_w$ (and $d_d$) in a 2D grid plane. (a) $d_h = 3p_h$, $d_w = 3p_w$: sparse packing with gaps between patches; (b) $d_h = 2p_h$, $d_w = 2p_w$: complete grid full coverage without patch overlaps; (c) $d_h = p_h$, $d_w = p_w$: complete grid coverage with full patch overlap between adjacent patches; and (d) $d_h = 1$, $d_w = 1$: densest packing configuration, where each grid node serves as the center of a patch. For the depth dimension that is not shown, the patch packing can be analyzed in a similar way.}
    	\label{fig:cuboid_packing}
    \end{minipage}%
\end{figure*}

In geometry, \textit{packing} and \textit{covering} problems constitute a specialized category of optimization challenges pertaining to geometric objects within a defined space. These problems typically involve the arrangement of identical geometric objects to maximize density without overlap, while ensuring the largest possible spatial coverage. The structural interdependence functions introduced in this study, however, deviate from conventional packing problems. Our approach permits the overlapping of patches when positioning them within the data's underlying grid structure, while simultaneously aiming to achieve the largest possible comprehensive grid coverage to minimize information loss.

The geometric patch-based structural interdependence functions defined earlier enumerate all nodes in the grid as patch centers for composing the interdependence matrices, resulting in the densest possible packing. This configuration is illustrated in Plot (d) of Figure~\ref{fig:cuboid_packing}, which demonstrates the concept in a 2D grid. From a computational cost perspective, such dense packing significantly impacts the output dimension $m'$ of these matrices, particularly in the padding mode where $m' = m \times p$, leading to considerably high dimensionality.

In this study, we introduce the concept of patch center distance hyper-parameters as a means to control the packing of patches that cover the grid structure. For cuboid patches, these parameters are denoted as $d_h$, $d_w$, and $d_d$, corresponding to the height, width, and depth dimensions, respectively. Figure~\ref{fig:cuboid_packing}, specifically Plots (a)-(c), provides a visual representation of cuboid patch packing in a two-dimensional example grid plane. These illustrations demonstrate the effect of varying center distance hyper-parameters on patches of identical shape, defined by patch sizes $p_h$ and $p_w$. By modulating these center distance parameters, we can precisely control two critical aspects of patch arrangement: the extent of grid coverage and the degree of overlap between adjacent patches, which also determine the information coverage and redundancy of the packing results.

Formally, given a grid of shape $(h, w, d)$ and patch center hyper-parameters $d_h$, $d_w$, and $d_d$, we can select the coordinates $(i, j, k)$ as patch centers from the following set, initializing from the coordinate $(0, 0, 0)$:

\begin{equation}\label{equ:patch_packing_coordinate}
\begin{aligned}
i & \in \left\{0, d_h, 2 \cdot d_h, \cdots, \left \lfloor \frac{h}{d_h} \right \rfloor \cdot d_h \right\}, \\
j & \in \left\{0, d_w, 2 \cdot d_w, \cdots, \left \lfloor \frac{w}{d_w} \right \rfloor \cdot d_w \right\},\\
k & \in \left\{0, d_d, 2 \cdot d_d, \cdots, \left \lfloor \frac{d}{d_d} \right \rfloor \cdot d_d \right\}.
\end{aligned}
\end{equation} 

This selection method facilitates the packing of a set of patches within the grid, whose size can be represented as

\begin{equation}\label{equ:patch_packing_count}
p_{count} = \left(1+ \left \lfloor \frac{h}{d_h} \right \rfloor \right)\left(1+ \left \lfloor \frac{w}{d_w} \right \rfloor \right)\left(1+ \left \lfloor \frac{d}{d_d} \right \rfloor \right)
\end{equation}

For patches that extend beyond the grid boundaries, a default value padding approach is employed. Specifically, dummy attribute values (such as $0$) are used to pad these boundary-exceeding patches, ensuring uniformity in patch sizes across the entire grid structure.

Using these selected patch centers, we can define the interdependence function as follows:

\begin{equation}\label{equ:patch_packing_interdependence_function}
\xi(\mb{x}) = \mb{A} \in \mathbbm{R}^{m \times m'}.
\end{equation}

Depending on the working modes, the output dimension $m'$ will be

\begin{equation}\label{equ:patch_packing_output_dimension}
m' = \begin{cases}
p \times p_{count}, & \text{ for the padding mode},\\
p_{count}, & \text{ for the aggregation mode}.
\end{cases}
\end{equation}

The structural interdependence function defined above is adaptable to patches of various shapes, including cylinder and sphere patches. In the following sections, we will examine the packing strategies for cylinder and sphere patches, with a focus on determining feasible patch center distance hyper-parameters $d_h$, $d_w$, and $d_d$. Both the packing strategy and the selected hyper-parameters play crucial roles in determining the patch center coordinates, the definition of the interdependence function, and the output dimension, as indicated by Equations (\ref{equ:patch_packing_coordinate})-(\ref{equ:patch_packing_output_dimension}).


\subsubsection{Cylinder Geometric Patch Packing based Structural Interdependence Function}\label{subsubsec:cylinder_patch}

\begin{figure*}[t]
    \begin{minipage}{\textwidth}
    \centering
    	\includegraphics[width=0.9\linewidth]{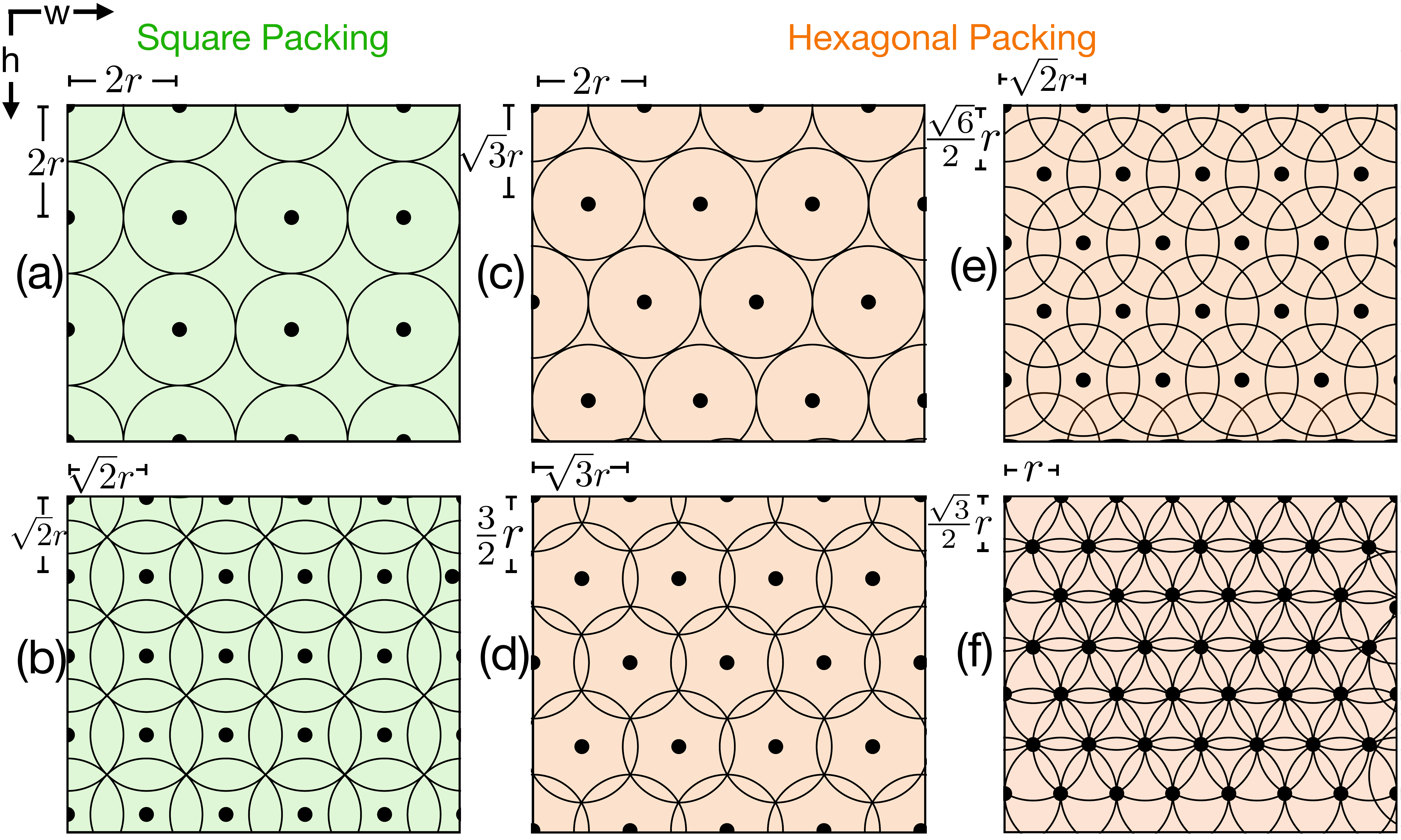}
    	\caption{An illustration of of cylinder geometric patch packing with patch shape parameters $r$ (and $p_d'=p_d$) and varying patch center distance parameters $d_h$, $d_w$ (and $d_d$) in a 2D grid plane. The depth dimension is omitted for clarity. 
(a) $d_h=2r$, $d_w=2r$: sparse square packing with inter-patch gaps;
(b) $d_h=\sqrt{2}r$, $d_w=\sqrt{2}r$: complete square packing;
(c) $d_h=\sqrt{3}r$, $d_w=2r$: sparse hexagonal packing with inter-patch gaps;
(d) $d_h=\frac{3}{2}r$, $d_w=\sqrt{3}r$: complete hexagonal packing with minimal overlap;
(e) $d_h = \frac{\sqrt{6}}{2}r$, $d_w = \sqrt{2}r$: complete hexagonal packing with increased overlap;
(f) $d_h = \frac{\sqrt{2}}{2}r$, $d_w = r$: complete hexagonal packing with full overlap.}
    	\label{fig:cylinder_packing}
    \end{minipage}%
\end{figure*}

The packing of cylinder-shaped geometric patches within a cuboid grid presents greater challenges compared to the cuboid patches discussed previously. While the packing along the depth dimension remains relatively straightforward, adhering to the principles established for cuboid patches, the arrangement of the circular surface on the plane composed of the height and width dimensions proves more complex.

Figure~\ref{fig:cylinder_packing} illustrates various packing strategies for these circular surfaces. These strategies can be categorized based on two primary criteria: the organization of patch centers and the coverage of the grid. Regarding the organization of patch centers, we observe square packings, depicted in Plots (a)-(b) in light green color, and hexagonal packings, shown in Plots (c)-(f) in light orange color. In terms of grid coverage, the strategies can be classified as sparse packings, exemplified in Plots (a) and (c), and complete packings, illustrated in Plots (b), (d)-(f). Each of these packing strategies offers unique characteristics and trade-offs between information coverage and redundancy.

\noindent \textbf{Sparse Packing Strategies}: The sparse packing strategies aim to maximize the coverage of the grid with packed patches, treating the patches as rigid bodies with no overlapping.

\begin{itemize}

\item \textbf{Square packing}: This strategy positions patch centers in rows and columns parallel to the height and width dimensions of the grid structure. As illustrated in Plot (a) of Figure~\ref{fig:cylinder_packing}, a square packing configuration with center distance hyper-parameters $d_h=2r$ and $d_w=2r$ (where $r$ is the circular surface radius of the cylinder patch) results in adjacent circular surfaces of the patches. However, this arrangement cannot achieve complete surface coverage, with a packing coverage rate of approximately $\frac{\pi}{4} = 0.785$. This indicates that a significant number of data elements are neither used as central nor dependent attributes.

\item \textbf{Hexagonal packing}: This approach employs a ``zig-zag'' arrangement of patch centers to minimize uncovered gaps and reduce information loss. Plot (c) of Figure~\ref{fig:cylinder_packing} demonstrates a hexagonal packing configuration with stride parameters $d_h=\sqrt{3}r$ and $d_w=2r$, which also results in adjacent patch placement. Notably, this method increases the packing coverage rate to $\frac{\pi}{2\sqrt{3}} = 0.907$, offering a substantial improvement over the square packing strategy and more effectively mitigating information loss during the packing process.

\end{itemize}

\noindent \textbf{Complete Packing Strategies}: The complete packing strategies aim to entirely cover the grid with packed patches, allowing for overlap among patches.

\begin{itemize}

\item \textbf{Square packing}: As illustrated in Plot (b) of Figure~\ref{fig:cylinder_packing}, reducing the patch center hyper-parameters to $d_h=\sqrt{2}r$ and $d_w=\sqrt{2}r$ results in overlapping circular surfaces that completely cover the grid surface. For each circle, the overlapping ratio with other circles is approximately $\frac{2\pi - 4}{\pi} = 0.726$, indicating that 72.6\% of the attributes covered by each cylinder patch are also involved in adjacent patches.

\item \textbf{Hexagonal packing}: Plot (d) of Figure~\ref{fig:cylinder_packing} demonstrates that reducing the center distance hyper-parameters to $d_h=\frac{3}{2}r$ and $d_w=\sqrt{3}r$ also achieves complete grid coverage. Compared to square packing, this configuration yields a lower overlapping ratio of approximately $\frac{2\pi - 3\sqrt{3}}{\pi} = 0.345$ per circle. Plots (e) and (f) illustrate further reductions in parameters to $d_h = \frac{\sqrt{6}}{2}r$, $d_w = \sqrt{2}r$ and $d_h = \frac{\sqrt{2}}{2}r$, $d_w = r$, respectively, both resulting in substantially higher overlapping ratios among patches.

\end{itemize}

All the cylinder patch packing strategies discussed above have been implemented in the {\toolkit} toolkit. Given the shape hyper-parameters $r$ and $p_d$ of the cylinder patch, the {\toolkit} toolkit can automatically compute the feasible patch center coordinates from the data grid structure for composing the interdependence matrices.


\subsubsection{Sphere Geometric Patch Packing based Structural Interdependence Function}\label{subsubsec:sphere_patch}

\begin{figure*}[t]
    \begin{minipage}{\textwidth}
    \centering
    	\includegraphics[width=0.9\linewidth]{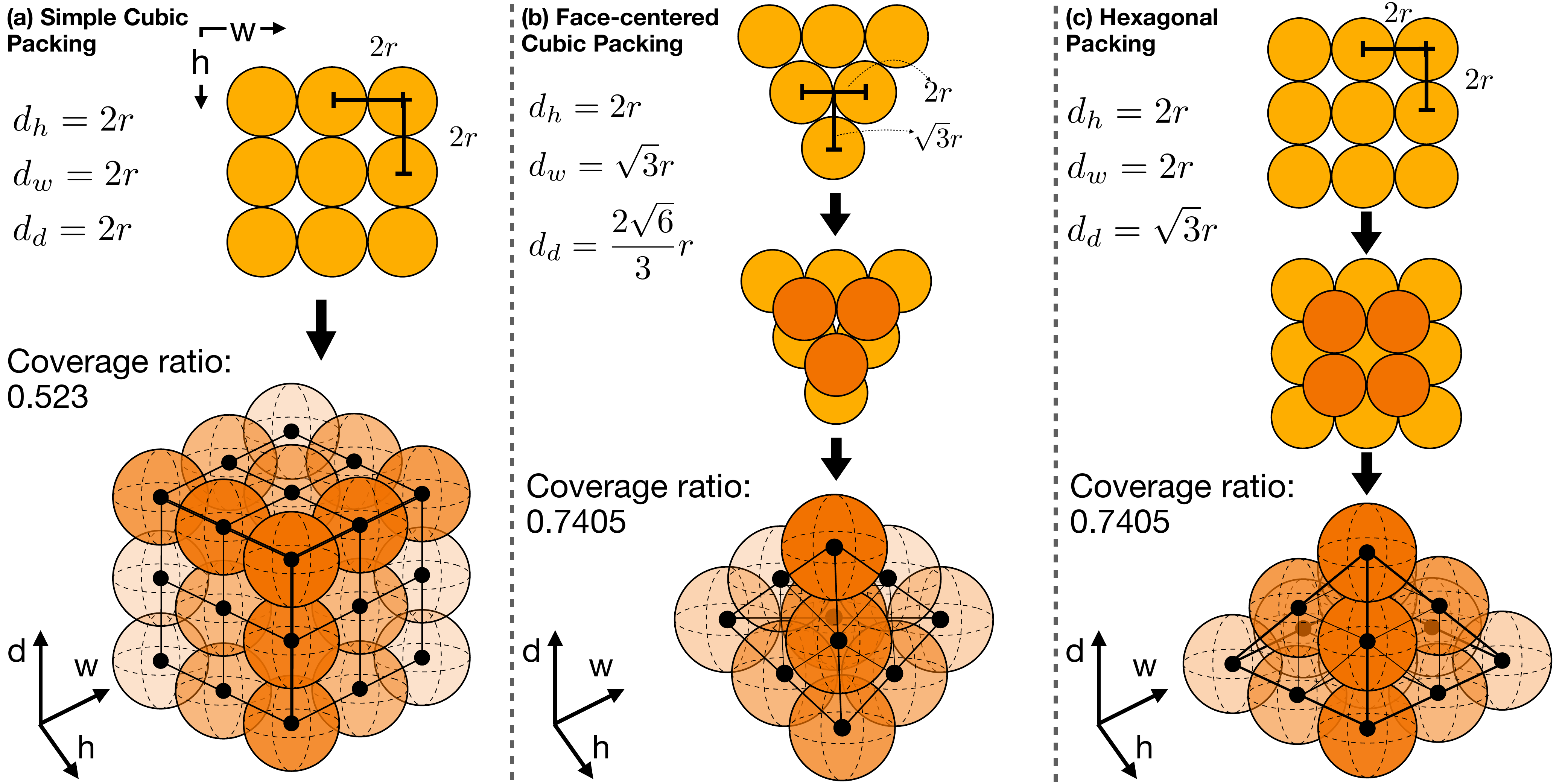}
    	\caption{An illustration of sphere geometric patch packing with radius $r$ in both 2D and 3D grid spaces, demonstrating varying patch center distance parameters $d_h$, $d_w$, and $d_d$. (a) $d_h = 2r, d_w=2r, d_d=2r$: Simple cubic sphere packing. (b) $d_h = \sqrt{3}r, d_w = 2r, d_d = \frac{2\sqrt{6}}{3}r$: Face-centered cubic sphere packing. (c) $d_h = 2r, d_w = 2r, d_d = \sqrt{3}r$: Hexagonal sphere packing. }
    	\label{fig:sphere_packing}
    \end{minipage}%
\end{figure*}

Sphere patches offer superior rotational invariance compared to cylinder patches, maintaining this property for rotation axes in three-dimensional space. This characteristic is crucial for defining interdependence functions for data that require preservation of rotational invariance properties. However, the packing of sphere-shaped patches within a cuboid grid presents significantly greater challenges than cylinder patch packing. As illustrated in Figure~\ref{fig:sphere_packing}, sphere patch packing strategies can be categorized into sparse and complete packing configurations, differentiated by their grid space coverage and patch overlapping ratios. These distinctions parallel those observed in cylinder patch packing, yet the three-dimensional nature of spheres introduces additional complexities in achieving optimal spatial arrangements.

\noindent \textbf{Sparse packing}: Formally, given a sphere patch with radius $r$, the sparse patch packing strategy positions sphere patches adjacent to each other without overlapping, inevitably resulting in gaps between patches. Figure~\ref{fig:sphere_packing} illustrates three distinct sphere patch packing strategies examined in this paper:

\begin{itemize}
\item \textbf{Simple cubic packing}: This straightforward approach to packing sphere patches within a cuboid grid is illustrated in Plot (a) of Figure~\ref{fig:sphere_packing}. With patch center distance hyper-parameters $d_h = 2r, d_w=2r, d_d=2r$, each sphere is stacked directly atop another, touching 6 other patches. The grid space coverage ratio for this simple cubic packing is approximately $\frac{\pi}{6} = 0.523$.

\item \textbf{Face-centered cubic packing}: As shown in Plot (b) of Figure~\ref{fig:sphere_packing}, this strategy places spheres diagonally adjacent to each other within each layer. Subsequent layers of spheres are positioned in the interstices between spheres of the underlying layer, with every third layer directly overlying its counterpart. This configuration, utilizing patch center distance hyper-parameters $d_h = \sqrt{3}r, d_w = 2r, d_d = \frac{2\sqrt{6}}{3}r$, enables each patch to contact 12 adjacent patches, yielding a coverage ratio of $\frac{\pi}{3\sqrt{2}} \approx 0.7405$.

\item \textbf{Hexagonal packing}: Plot (c) of Figure~\ref{fig:sphere_packing} depicts this packing strategy, where the bottom layer consists of spheres in direct contact. Spheres in subsequent layers are situated in the interstices of the underlying layer. With hyper-parameters $d_h = 2r, d_w = 2r, d_d = \sqrt{3}r$, alternate layers directly overlie each other. Each patch contacts 12 neighboring patches, achieving a coverage ratio of approximately $\frac{\pi}{3\sqrt{2}} \approx 0.7405$, identical to that of face-centered cubic packing.
\end{itemize}


\noindent \textbf{Complete packing}: The sphere patch packing strategies introduced earlier can be adapted to achieve complete grid space coverage by reducing the patch center distance hyper-parameters $d_h, d_w, d_d$ and allowing patch overlap. This approach gradually eliminates gaps between patches, ensuring comprehensive coverage of the grid space. Based on the sphere patch organizations illustrated in Figure~\ref{fig:sphere_packing}, we propose the following potential stride parameters $d_h, d_w, d_d$ to eliminate gaps and avoid information loss:

\begin{itemize}
\item \textbf{Simple cubic packing}: Complete coverage is achieved by setting the patch center distance hyper-parameters to $d_h = d_w = d_d = \frac{2\sqrt{3}}{3}r$. This configuration ensures that all attributes are encompassed within sphere patches.

\item \textbf{Face-centered cubic packing}: For this packing strategy, setting patch center distance parameters to $d_h = \sqrt{2} r$, $d_w = \frac{2\sqrt{6}}{3}r$, and $d_d = \frac{4}{3} r$ effectively eliminates space gaps between spheres.

\item \textbf{Hexagonal packing}: Complete coverage in hexagonal packing is attained with patch center distance parameters $d_h = d_w = \frac{2\sqrt{3}}{3}r$ and $d_d = r$. This configuration ensures that every attribute in the data instance is included within at least one sphere patch.
\end{itemize}

All these sphere packing strategies have been implemented within the new {\toolkit} toolkit, enabling direct application with the sphere patch packing-based structural interdependence function. This implementation facilitates the practical use of these advanced packing strategies in various data analysis tasks.


\subsubsection{Chain based Structural Interdependence Function}\label{subsubsec:chain_interdependence_function}

\begin{figure*}[t]
    \begin{minipage}{\textwidth}
    \centering
    	\includegraphics[width=0.9\linewidth]{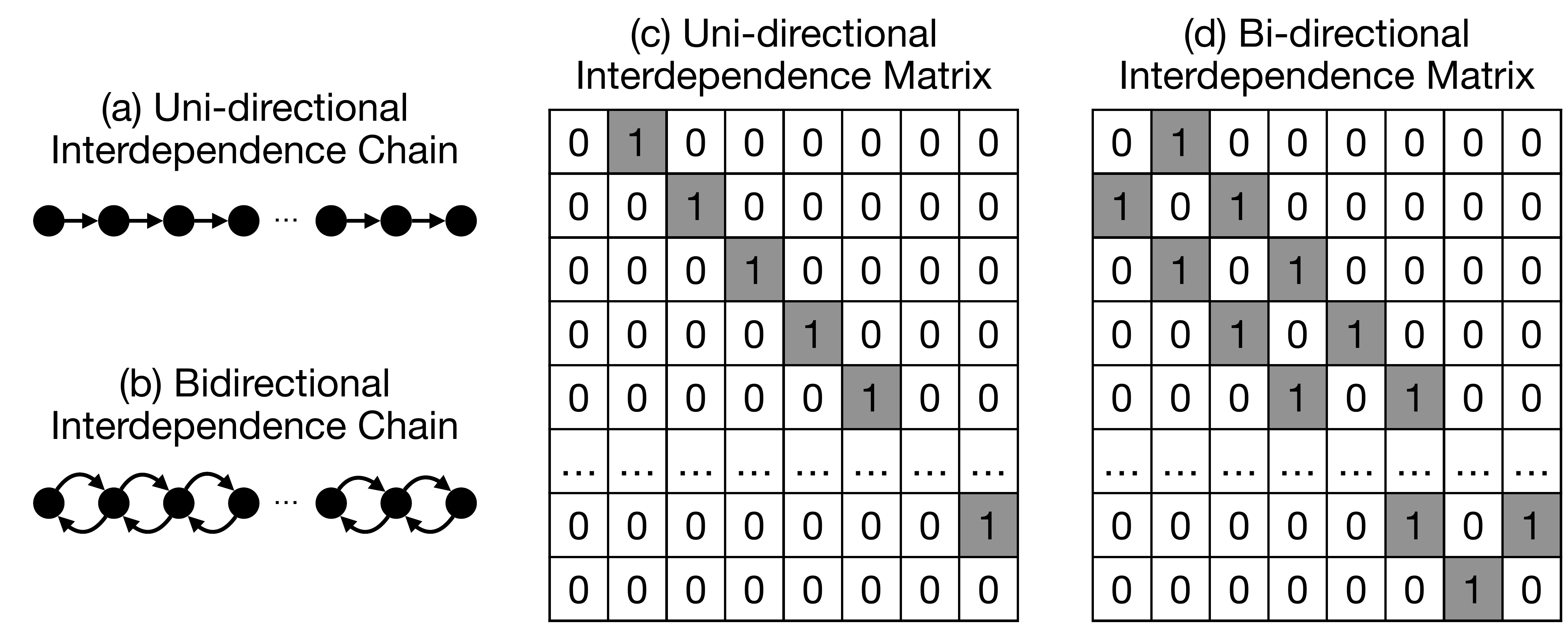}
    	\caption{An illustration of  uni- and bi-directional chain-structured topological interdependence relationships. (a) Uni-directional chain-structured interdependence relationship. (b) Bi-directional chain-structured interdependence relationship. (c) Interdependence matrix of the uni-directional chain. (d) Interdependence matrix of the bi-directional chain.}
    	\label{fig:chain_interdependence}
    \end{minipage}%
\end{figure*}

In addition to the grid structures discussed earlier, as shown in Figure~\ref{fig:geometry_example}, many real-world datasets can be represented as chains, reflecting their underlying topological structure and sequential interdependence relationships. Chain-structured interdependence refers to a series of interconnected dependencies where each element in the chain relies on the previous one (or the later one), creating a linear sequence of relationships. Examples of data with chain-structured interdependence relationships include, but are not limited to {natural languages}, {gene sequences}, {audio recordings} and {stock prices}. The chain interdependence function can define the interdependence matrix for both features and data instances. We will use data attribute interdependence relationships to illustrate this concept.

Formally, given a data batch $\mathbf{x} \in \mathbbm{R}^{m}$ with sequential chain-structured interdependence relationships among the attributes as illustrated in Plot (a) of Figure~\ref{fig:chain_interdependence}, we can define the corresponding unidirectional chain interdependence function as follows:

\begin{equation}
\xi(\mb{x}) = \mb{A}  \in \mathbbm{R}^{m \times m'},
\end{equation}

where $\mathbf{A}$ is the composed attribute interdependence matrix illustrated in Plot (c) of Figure~\ref{fig:chain_interdependence}. By default, the output dimension $m'$ equals the input instance dimension, {\ie} $m' = m$.

In many cases, we sum this interdependence matrix with an identity matrix to denote self-dependency:

\begin{equation}
\xi(\mb{x}) = \mb{A} + \mb{I}  \in \mathbbm{R}^{m \times m}.
\end{equation}

Here, $\mathbf{I} \in \mathbbm{R}^{m \times m}$ is a square diagonal identity matrix of size $m \times m$, allowing the function to model both interdependence and self-dependence with a single dependency function. This self-dependence can also be defined using the linear remainder term in {\our}, both of which contribute to defining sequential interdependence relationships.

For some input data batches, bidirectional chain-structured interdependence relationships may exist among the inputs. In this case, as illustrated in Plot (b) of Figure~\ref{fig:chain_interdependence}, each element in the chain relies on both the previous and subsequent elements. We can define a bidirectional chain interdependence function to compose the interdependence matrix $\mathbf{A} \in \mathbbm{R}^{m \times m}$.

As with the unidirectional case, this interdependence matrix can be summed with an identity matrix to represent self-dependency. The bidirectional interdependence matrix is particularly useful for modeling dependency relationships in sequence inputs, such as language data with complete context both before and after each token. This contextual information provides more comprehensive data for learning useful features of each token in the input language, potentially leading to more accurate learning results.


\subsubsection{Multi-Hop Chain based Structural Interdependence Function}\label{subsubsec:multihop_chain_interdependence_function}

According to the interdependence matrix (both unidirectional and bidirectional), for the last attribute in the input data instance $\mb{x}$ to access information from the beginning ones it depends on, it may require the multiplication with the data instance vector $\mathbf{x}$ with the interdependence matrix $\mathbf{A}$ at least $m-1$ times:

\begin{equation}
\underbrace{\mb{x} \mb{I}}_{\text{0-hop}}, \ \underbrace{\mb{x} \mb{A}}_{\text{1-hop}}, \ \underbrace{\mb{x} \mb{A} \mb{A}}_{\text{2-hop}}, \ \cdots , \underbrace{\mb{x} \mb{A}^{m-1}}_{\text{(m-1)-hop}},
\end{equation}

Such a step-wise information propagation along the chains denotes how current sequential models operate. However, this approach can be computationally expensive, especially for input data with extremely long chain-structured interdependence relationships. To reduce computational time and space costs, we introduce the multi-hop chain-based structural interdependence function:

\begin{equation}
\xi(\mb{x} | h) = \mb{A}^h \in \mathbbm{R}^{m \times m} \text{, for } h \in \{0, 1, 2, \cdots, m-1\},
\end{equation}

where $h$ is the hop parameter for modeling multi-hop interdependence relationships of the input data instance along the chain.

With this interdependence matrix, each attribute in the data instance can directly access information from those $h$-hops away along the chain structure. To accumulate all data instances within $h$-hops, we introduce the accumulative multi-hop chain-based structural interdependence function as follows:

\begin{equation}\label{equ:chain_accumulative}
\xi(\mb{x} | 0:h) = \mb{I} + \mb{A} + \mb{A}^2 + \mb{A}^3 + \cdots + \mb{A}^h = \sum_{i=0}^h \mb{A}^i \in \mathbbm{R}^{m \times m}.
\end{equation}

To further optimize the computations, we propose approximating the accumulative interdependence function using Taylor's polynomial expansion series. Considering the Taylor's expansions of the reciprocal function $\frac{1}{1-x}$ and the exponential function $\exp(x)$:

\begin{equation}
\begin{aligned}
\frac{1}{1-x} & = 1 + x + x^2 + x^3 + \cdots = \sum_{h=0}^{\infty} x^h.\\
\exp(x) &= 1 + x + \frac{x^2}{2!} + \frac{x^3}{3!} + \cdots = \sum_{h=0}^{\infty} \frac{x^h}{h!}.
\end{aligned}
\end{equation}

Based on them, we define the reciprocal structural interdependence function and exponential structural interdependence function for approximating the above multi-hop chain-structured topological interdependence relationships as:

\begin{equation}
\xi(\mb{x}) = (\mb{I} - \mb{A})^{-1} \in \mathbbm{R}^{m \times m} \text{, and } \xi(\mb{x}) = \exp(\mb{A}) \in \mathbbm{R}^{m \times m}.
\end{equation}

The matrix $\mb{A}$, as defined above, is a nilpotent matrix, with its power becoming zero for exponents greater than $m-1$ ({\ie} $\mb{A}^h = \mb{0}, \forall h > m-1$). Consequently, the functions described do not introduce interdependence relationships beyond $m$ hops. These calculations offer superior time and space efficiency compared to multi-hop based matrix power computations. The resulting interdependence matrix from these functions incorporates information from data instances within the input data batch. Notably, in the exponential function, data instances separated by $h$ hops are subject to a penalization factor of $\frac{1}{h!}$, effectively weighting the influence of more distant relationships.

Meanwhile, for the reciprocal interdependence function, the matrix ``$\mb{I} - \mb{A}$'' will be singular when the chain is bi-directional, which can be calculated with the above accumulative interdependence Equation~(\ref{equ:chain_accumulative}) with $h=m-1$ instead. In the implementation of these interdependence functions in the toolkit, the chain interdependence matrix can be optionally normalized in the columns (or row for instance interdependence) to avoid dramatically high values for matrix entries in computation.


\subsubsection{Graph based Structural Interdependence Function}\label{subsubsec:graph_interdependence_function}

\begin{figure*}[t]
    \begin{minipage}{\textwidth}
    \centering
    	\includegraphics[width=0.9\linewidth]{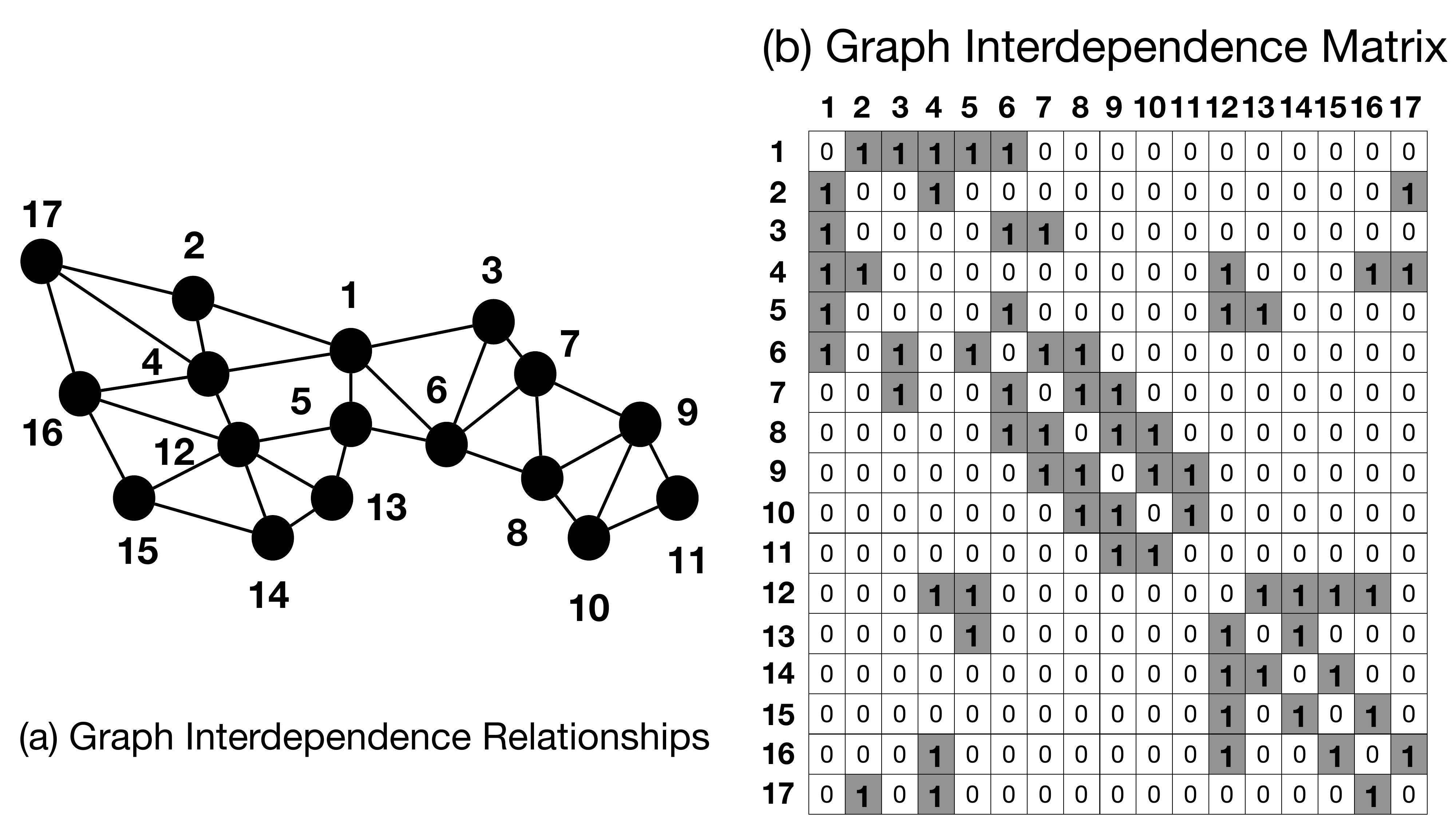}
    	\caption{An illustration of interdependence relationships in the graph topological structure and its corresponding structural interdependence matrix.}
    	\label{fig:graph_dependency}
    \end{minipage}%
\end{figure*}

In addition to chain based structural interdependence functions, some data structures exhibit more extensive interdependence relationships, such as graphs. Graph-structured interdependence relationships model complex dependencies between different features or data instances, where each element may depend on multiple other elements.

Given a data batch $\mathbf{x} \in \mathbbm{R}^{m}$ with extensively interdependent attributes, we can represent the interdependence relationships as a graph $G = (\mathcal{V}, \mathcal{E})$, where $\mathcal{V} = \{0, 1, \cdots, m-1\}$ is the node set and $\mathcal{E} = \{(i, j)\}_{i, j \in \{0, 1, \cdots, m-1\}}$ is the link set. For simplicity, we use the data instance's row index $i \in \{0, 1, \cdots, m-1\}$ to represent its corresponding node in the graph structure. The Plot (a) of Figure~\ref{fig:graph_dependency} illustrates an example of such graph-structured interdependence relationships among the attributes. In this example, these attributes are represented by the graph nodes, while their interdependence relationships are represented by the links connecting them.

Based on the graph structure, we define the graph interdependence function as:

\begin{equation}\label{equ:graph_interdependence_function}
\xi(\mb{x} | G) = \mb{A} \in \mathbbm{R}^{m \times m'},
\end{equation}

where the output dimension $m' = m$ by default.

In the output binary interdependence matrix $\mathbf{A}$, the entry $\mathbf{A}(i, j)$ is filled with value 1 if and only if the node pair $(i, j) \in \mathcal{E}$ is a link in the graph. In other words, the interdependence matrix $\mathbf{A}$ represents the adjacency matrix of the graph $G$. For the graph example discussed above, we illustrate its interdependence matrix at Plot (b) of Figure~\ref{fig:graph_dependency}.


\subsubsection{Graph PageRank based Structural Interdependence Function}

Similar to chain-structured interdependence relationships, multiplying the graph interdependence matrix with the data batch allows each instance to access information from its immediate neighbors (1-hop away). To retrieve information from neighbors that are $h$-hops away, we may need to multiply the interdependence matrix $h$ times with the data batch:

\begin{equation}
\underbrace{\mb{x} \mb{I}}_{\text{0-hop}}, \ \underbrace{\mb{x} \mb{A}}_{\text{1-hop}}, \ \underbrace{\mb{x} \mb{A} \mb{A}}_{\text{2-hop}}, \ \cdots , \underbrace{\mb{x} \mb{A}^h}_{\text{h-hop}}.
\end{equation}

To model multi-hop dependency relationships among data instances, we introduce the multi-hop graph interdependence function and the accumulative multi-hop graph interdependence function as follows:

\begin{equation}
\xi(\mb{x} | h) = \mb{A}^h \in \mathbbm{R}^{m \times m} \text{, and } \xi(\mb{x} | 0: h) = \sum_{i=0}^h \mb{A}^i \in \mathbbm{R}^{m \times m}.
\end{equation}

In addition to these formulas that calculate powers of matrix $\mathbf{A}$, existing graph studies also offer other approaches to calculate long-distance dependency relationships among data instances, such as the PageRank algorithm. Without delving into the step-wise derivation of PageRank updating equations, we define the PageRank multi-hop graph interdependence function using the convergence matrix representation from \cite{Zhang2020GraphBertOA}.

\begin{equation}
\xi(\mb{x}) = \alpha \cdot \left( \mb{I} - (1- \alpha) \cdot {\mb{A}} \right)^{-1} \in \mathbbm{R}^{m \times m}.
\end{equation}

Here, $\alpha \in [0, 1]$ is a hyper-parameter of the function, typically set to $0.15$ by default. Usually, matrix $\mathbf{A}$ is normalized before being used in this formula. Several frequently used matrix normalization approaches ({\eg} \textit{degree-based normalization}, \textit{mean-std-based normalization}, and \textit{softmax-based normalization}) have been implemented in the {\toolkit} toolkit, which can be directly applied in the above interdependence function definition for input graph structures.


\subsection{Hybrid Interdependence Functions}\label{subsec:hybrid_interdependence_function}

For the interdependence functions defined above, we also allow users to fuse them together to define more complex interdependence functions, which may be required for some special learning needs. For instance, for some input data, both the instance attributes and the underlying modality specific structures play a critical role in defining the interdependence relationships among the attributes (or instances), a fusion of the interdependence functions defined based on them will become necessary. 

In this part, we will discuss the hybrid interdependence functions that can integrate the above interdependence functions defined based on different information of the input data batch together for modeling the diverse interdependence relationships. Moreover, compared with the multi-channel and multi-head architecture discussed before in Section~\ref{subsec:wide_interdependence_functions}, the hybrid interdependence function fusion provides a lightweight and more flexible implementation to achieve similar purposes. Also, even within the {\our} model with one single head and channel, the hybrid interdependence function still allows the model to define and model complex interdependence relationships.


\subsubsection{Hybrid Interdependence Function Representation}

Formally, given the input data batch $\mb{X} \in \mathbbm{R}^{b \times m}$, we can define a set of data and structural interdependence functions $\xi_1, \xi_2, \cdots, \xi_k: \mathbbm{R}^{b \times m} \to \mathbbm{R}^{m \times m'}$ to measure the interdependence relationships among the attributes. These functions can be effectively fused together as follows:

\begin{equation}
\begin{aligned}
\xi(\mb{X}) &= \text{fusion} \left( \xi_1(\mb{X}), \xi_2(\mb{X}), \cdots, \xi_k(\mb{X}) \right)\\
&= \text{fusion} \left( \mb{A}_1, \mb{A}_2, \cdots, \mb{A}_k \right)\\
&= \mb{A} \in \mathbbm{R}^{m \times m'},
\end{aligned}
\end{equation}

where $\mb{A}_i = \xi_i(\mb{X})$ denotes the interdependence matrix obtained by function $\xi_i, \forall i \in \{1, 2, \cdots, k\}$. Different fusion strategies can be used to define the $\text{fusion}(\cdot)$ operator used above, which will be introduced in the following subsection specifically.

The hybrid interdependence function defined above is particularly valuable for modeling complex relationships that incorporate information from multiple sources, such as data batches and underlying topological structures. To demonstrate the application of hybrid interdependence functions, we will use graph data as an example below, explaining how to define an interdependence function for data batches with underlying graph structures. Similar hybrid functions can also be applied to data batches with underlying chain and grid structures.


\subsubsection{An Example: Hybrid Graph Interdependence Function}\label{subsubsec:hybrid_graph_interdependence}

Formally, given an input data batch $\mathbf{X} \in \mathbbm{R}^{b \times m}$ and its corresponding graph structure $G$, we define a hybrid interdependence function by combining the parameterized bilinear interdependence function with a one-hop graph interdependence function, as follows:

\begin{equation}
\xi(\mb{X}, G) = \text{fusion}(\mb{A}_{g} , \mb{A}_{b}) = \mb{A}_{g} \circ \mb{A}_{b} \in \mathbbm{R}^{m \times m},
\end{equation}

where the notations

\begin{equation}
\begin{aligned}
\text{Graph Interdependence: } \mb{A}_{g} &= \xi_{g}(\mb{X} | G) \in \mathbbm{R}^{m \times m},\\
\text{Bilinear Interdependence: }\mb{A}_{b} &= \xi_b(\mb{X} | \mb{w}) = \mb{X}^\top \mb{W} \mb{X} \in \mathbbm{R}^{m \times m},
\end{aligned}
\end{equation}

denote the attribute interdependence matrices derived from the graph structure and data batch, respectively. If we use the one-hop graph interdependence function described in Section~\ref{subsubsec:graph_interdependence_function}, the resulting graph interdependence matrix $\mathbf{A}_{g}$ serves as a binary mask, preserving the bilinear interdependence scores between nodes with direct connections. Assisted with the optional post-processing functions (such as softmax based interdependence matrix column normalizations), this hybrid interdependence function endows the {\our} model with capabilities akin to those of the Graph Attention Network (GAT) \cite{Velickovic2017GraphAN}, which uses a linear attention mechanism, and Graph-BERT \cite{Zhang2020GraphBertOA}, which utilizes a transformer-based attention mechanism.

The hybrid interdependence function offers significant flexibility in defining interdependence relationships between attributes and instances, using either identical or distinct input information. For complex datasets with diverse interdependence patterns, this approach enhances {\our}'s modeling capacity and performance robustness. The effectiveness of this hybrid interdependence will be further examined through experiments on real-world benchmark datasets in Section~\ref{sec:experiments}.

The fusion function, denoted as $\text{fusion}(\cdot)$ above, plays a critical role in defining hybrid interdependence by determining how different interdependence strategies are combined. This same function can also be used for effective integration of the outputs from multi-channel and multi-head components in our architecture. We will discuss these fusion functions in detail in the following section.

\begin{figure*}[t]
    \begin{center}
    \tiny
            \begin{forest}
                for tree={
                    forked edges,
                    grow'=0,
                    s sep=2.1pt,
                    draw,
                    rounded corners,
                    node options={align=center,},
                    text width=2.7cm,
                },
                [{\textcolor{Plum}{\textbf{Reconciled\\ Polynomial\\ Network\\ ({\our})}}}, fill=Plum!30, parent
                    [Data\\ Compression\\ Functions, for tree={fill=brown!45, child}
                    	[1. Identity Compression, fill=brown!30, grandchild
                    		[{1-(a). $\kappa(\mb{x}) = \mb{x}$}, fill=brown!30, greatgrandchild
                                		[dim: {$d=m$}, fill=brown!15, referenceblock]
                            	]
                            	[{1-(b). $\kappa(\mb{x}) = \sigma(\mb{x})$}, fill=brown!30, greatgrandchild
                                		[{dim: $d=m$}, fill=brown!15, referenceblock]
                            	]
                        ]
                    	[2. Linear Compression, fill=brown!30, grandchild
                    		[{2-(a). $\kappa(\mb{x}) = \mb{x} \mb{C}$}, fill=brown!30, greatgrandchild
                                		[dim: {$d=m$}, fill=brown!15, referenceblock]
                            	]
                        ]
                    	[3. Geometric Compression, fill=brown!30, grandchild
                    		[{3-(a). $\kappa(\mb{x}) = \left[ \phi(\mb{p}_1), \phi(\mb{p}_2), \cdots, \phi(\mb{p}_{|\mc{P}|}) \right]$}, fill=brown!30, greatgrandchild
                                		[dim: {$d = |\mathcal{P}| \times d_{\phi}$}, fill=brown!15, referenceblock]
                            	]
                        ]
                    	[4. Feature Selection Compression, fill=brown!30, grandchild
                    		[{4-(a). $\kappa(\mb{x}) = \text{feature-selection}(\mb{x})$}, fill=brown!30, greatgrandchild
                                		[dim: {$d=k$}, fill=brown!15, referenceblock]
                            	]
                        ]
                    	[5. Dimension Reduction Compression, fill=brown!30, grandchild
                    		[{5-(a). $\kappa(\mb{x}) = \mb{x} \mb{V}_k$}, fill=brown!30, greatgrandchild
                                		[dim: {$d=k$}, fill=brown!15, referenceblock]
                            	]
				[{5-(b). $\kappa(\mb{x}) = \mb{x} \mb{R}$}, fill=brown!30, greatgrandchild
                                		[dim: {$d=k$}, fill=brown!15, referenceblock]
                            	]
                        ]
                    	[6. Probabilistic Compression, fill=brown!30, grandchild
                    		[{6-(a). $\kappa(\mb{x}) = \mb{t}$, where $\mb{t} | \mb{x} \sim \mc{N}(\bs{\mu}, \bs{\Sigma})$}, fill=brown!30, greatgrandchild
                                		[dim: {$d=k$}, fill=brown!15, referenceblock]
                            	]
				[{6-(b). $\kappa(\mb{x}) = \log P(\mb{t} | \bs{\theta})$}, fill=brown!30, greatgrandchild
                                		[dim: {$d=k$}, fill=brown!15, referenceblock]
                            	]
				[{6-(c). $\kappa(\mb{x} | k) = \left[ {\mb{x} \choose 1}, {\mb{x} \choose 2}, \cdots, {\mb{x} \choose k} \right]$}, fill=brown!30, greatgrandchild
                                		[dim: {$d=\sum_{i=1}^k {m \choose i}$}, fill=brown!15, referenceblock]
                            	]
				[{6-(d). $\kappa(\mb{x}) = \log P\left( {\kappa(\mb{x} | k) \choose d} | \bs{\theta} \right)$}, fill=brown!30, greatgrandchild
                                		[dim: {$d=k$}, fill=brown!15, referenceblock]
                            	]
                        ]
                    ]
                    [Data\\ Fusion\\ Functions, for tree={fill=Cerulean!45, child}
                    	[1. Weighted Summation based Fusion, fill=Cerulean!30, grandchild
                    		[{1-(a). $\text{fusion}(\mb{A}_1, \cdots, \mb{A}_k) = \sum_{i=1}^k \alpha_i \mb{A}_i$}, fill=Cerulean!30, greatgrandchild
                                		[dim: {$m \times n$}, fill=Cerulean!15, referenceblock]
                            	]
                        ]
                    	[2. Numerical Operators based Fusion, fill=Cerulean!30, grandchild
                    		[{2-(a). $\text{fusion}(\mb{A}_1, \cdots, \mb{A}_k) = \mb{A}$, where\\ $\mb{A}(i, j) = \max \left( \mb{A}_1(i,j), \cdots, \mb{A}_k(i,j)  \right)$}, fill=Cerulean!30, greatgrandchild
                                		[dim: {$m \times n$}, fill=Cerulean!15, referenceblock]
                            	]
                        ]
                    	[3. Hadamard Product based Fusion, fill=Cerulean!30, grandchild
                    		[{3-(a). $\text{fusion}(\mb{A}_1, \cdots, \mb{A}_k) = \mb{A}_1 \circ \cdots \circ \mb{A}_k$}, fill=Cerulean!30, greatgrandchild
                                		[dim: {$m \times n$}, fill=Cerulean!15, referenceblock]
                            	]
                        ]
                    	[4. Concatenation based Fusion, fill=Cerulean!30, grandchild
                    		[{4-(a). $\text{fusion}(\mb{A}_1, \cdots, \mb{A}_k) = \mb{A}_1 \sqcup \cdots \sqcup \mb{A}_k$}, fill=Cerulean!30, greatgrandchild
                                		[dim: {$m \times (\sum_{i=1}^k n_i)$}, fill=Cerulean!15, referenceblock]
                            	]
				[{4-(b). $\text{fusion}(\mb{A}_1, \cdots, \mb{A}_k) = ( \sqcup_{i=1}^k \mb{A}_i ) \mb{W}$}, fill=Cerulean!30, greatgrandchild
                                		[dim: {$m \times n$}, fill=Cerulean!15, referenceblock]
                            	]
				[{4-(c). $\text{fusion}(\mb{A}_1, \cdots, \mb{A}_k) = ( \sqcup_{i=1}^k \mb{A}_i ) ( \mb{P} \mb{Q}^\top )$}, fill=Cerulean!30, greatgrandchild
                                		[dim: {$m \times n$}, fill=Cerulean!15, referenceblock]
                            	]
                        ]
                    ]
                [Data\\ Compression\\ Functions, for tree={fill=brown!45, child}
                    	[1. Hermite Expansion, fill=brown!30, grandchild
                    		[{1-(a). $\kappa(\mb{x} | d)  = \left[ He_1(\mb{x}), He_2(\mb{x}), \cdots, He_d(\mb{x}) \right]$}, fill=brown!30, greatgrandchild
                                		[dim: {$D=md$}, fill=brown!15, referenceblock]
                            	]
                        ]
                    	[2. Laguerre Expansion, fill=brown!30, grandchild
                    		[{2-(a). $\kappa(\mb{x} | d, \alpha) = \left[ P^{(\alpha)}_1(\mb{x}), P^{(\alpha)}_2(\mb{x}), \cdots, P^{(\alpha)}_d(\mb{x}) \right]$}, fill=brown!30, greatgrandchild
                                		[dim: {$D=md$}, fill=brown!15, referenceblock]
                            	]
                        ]
                    	[3. Legendre Expansion, fill=brown!30, grandchild
                    		[{3-(a). $\kappa(\mb{x} | d) = \left[ P_1(\mb{x}), P_2(\mb{x}), \cdots, P_d(\mb{x}) \right]$}, fill=brown!30, greatgrandchild
                                		[dim: {$D=md$}, fill=brown!15, referenceblock]
                            	]
                        ]
                    	[4. Gegenbauer Expansion, fill=brown!30, grandchild
                    		[{4-(a). $\kappa(\mb{x} | d, \alpha) = \left[ P^{(\alpha)}_1(\mb{x}), P^{(\alpha)}_2(\mb{x}), \cdots, P^{(\alpha)}_d(\mb{x}) \right]$}, fill=brown!30, greatgrandchild
                                		[dim: {$D=md$}, fill=brown!15, referenceblock]
                            	]
                        ]
                    	[5. Bessel Expansion, fill=brown!30, grandchild
                    		[{5-(a). $\kappa(\mb{x} | d) = \left[ B_1(\mb{x}), B_2(\mb{x}), \cdots, B_d(\mb{x}) \right]$}, fill=brown!30, greatgrandchild
                                		[dim: {$D=md$}, fill=brown!15, referenceblock]
                            	]
				[{5-(b). $\kappa(\mb{x} | d) = \left[ R_1(\mb{x}), R_2(\mb{x}), \cdots, R_d(\mb{x}) \right]$}, fill=brown!30, greatgrandchild
                                		[dim: {$D=md$}, fill=brown!15, referenceblock]
                            	]
                        ]
                    	[6. Fibonacci Expansion, fill=brown!30, grandchild
                    		[{6-(a). $\kappa(\mb{x} | d) = \left[ F_1(\mb{x}), F_2(\mb{x}), \cdots, F_d(\mb{x}) \right]$}, fill=brown!30, greatgrandchild
                                		[dim: {$D=md$}, fill=brown!15, referenceblock]
                            	]
                        ]
                    	[7. Lucas Expansion, fill=brown!30, grandchild
                    		[{7-(a). $\kappa(\mb{x} | d) = \left[ L_1(\mb{x}), L_2(\mb{x}), \cdots, L_d(\mb{x}) \right]$}, fill=brown!30, greatgrandchild
                                		[dim: {$D=md$}, fill=brown!15, referenceblock]
                            	]
                        ]
                    	[8. Lucas Expansion, fill=brown!30, grandchild
                    		[{8-(a). $\kappa(\mb{x} | d=1) = \left[ \phi_{0, 0}(\mb{x}), \phi_{0, 1}(\mb{x}), \cdots, \phi_{s, t}(\mb{x}) \right]$}, fill=brown!30, greatgrandchild
                                		[dim: {$D=s \cdot t \cdot m$}, fill=brown!15, referenceblock]
                            	]
				[{8-(b). $\kappa(\mb{x} | d=2) = \kappa(\mb{x} | d=1) \otimes \kappa(\mb{x} | d=1)$}, fill=brown!30, greatgrandchild
                                		[dim: {$D=(s \cdot t \cdot m)^2$}, fill=brown!15, referenceblock]
                            	]
                        ]
                    ]                    
                    [Parameter Reconciliation\\ Functions, for tree={fill=red!45,child}
                    	[1. Random Matrix Adaption based Reconciliation, fill=red!30, grandchild
				[{1-(a). $\psi(\mb{w}) = \bs{\Lambda}_1 \mb{A} \bs{\Lambda}_1 \mb{B}^\top$}, fill=red!20, greatgrandchild
                                		[{$l = n + r$}, fill=red!10, referenceblock]
                            	]
			]
                    	[2. Random Matrix based Hypernet Reconciliation, fill=red!30, grandchild
				[{2-(a). $\psi(\mb{w}) = \left( \sigma \left( (\mb{w} \mb{P}) \mb{Q}^\top \right) \mb{S} \right) \mb{T}^\top$}, fill=red!20, greatgrandchild
                                		[{$l$ manual setup}, fill=red!10, referenceblock]
                            	]
			]
		]                    
		[Remainder Functions, for tree={fill=blue!45, child}
                        [1. \sout{Expansion Remainder}, fill=blue!30, grandchild
				[{1-(a). $\msout { \pi(\mb{x} | \mb{w}') = \left\langle \kappa'(\mb{x}), \psi'(\mb{w}') \right\rangle + \pi'(\mb{x})}$}, fill=blue!20, greatgrandchild
                                		[\sout{requires parameter}, fill=blue!10, referenceblock]
                            	]
			]
                ]
            ]
            \end{forest}
    \end{center}
    \caption{An overview of the newly added data compression and fusion functions, and the updated data expansion, parameter reconciliation, and remainder functions implemented in the {\toolkit} toolkit for constructing the {\our} model architecture. The cross-line added to the ``Expansion Remainder'' denotes the function has been deleted from the toolkit.}\label{fig:compression_fusion_function}
\end{figure*}
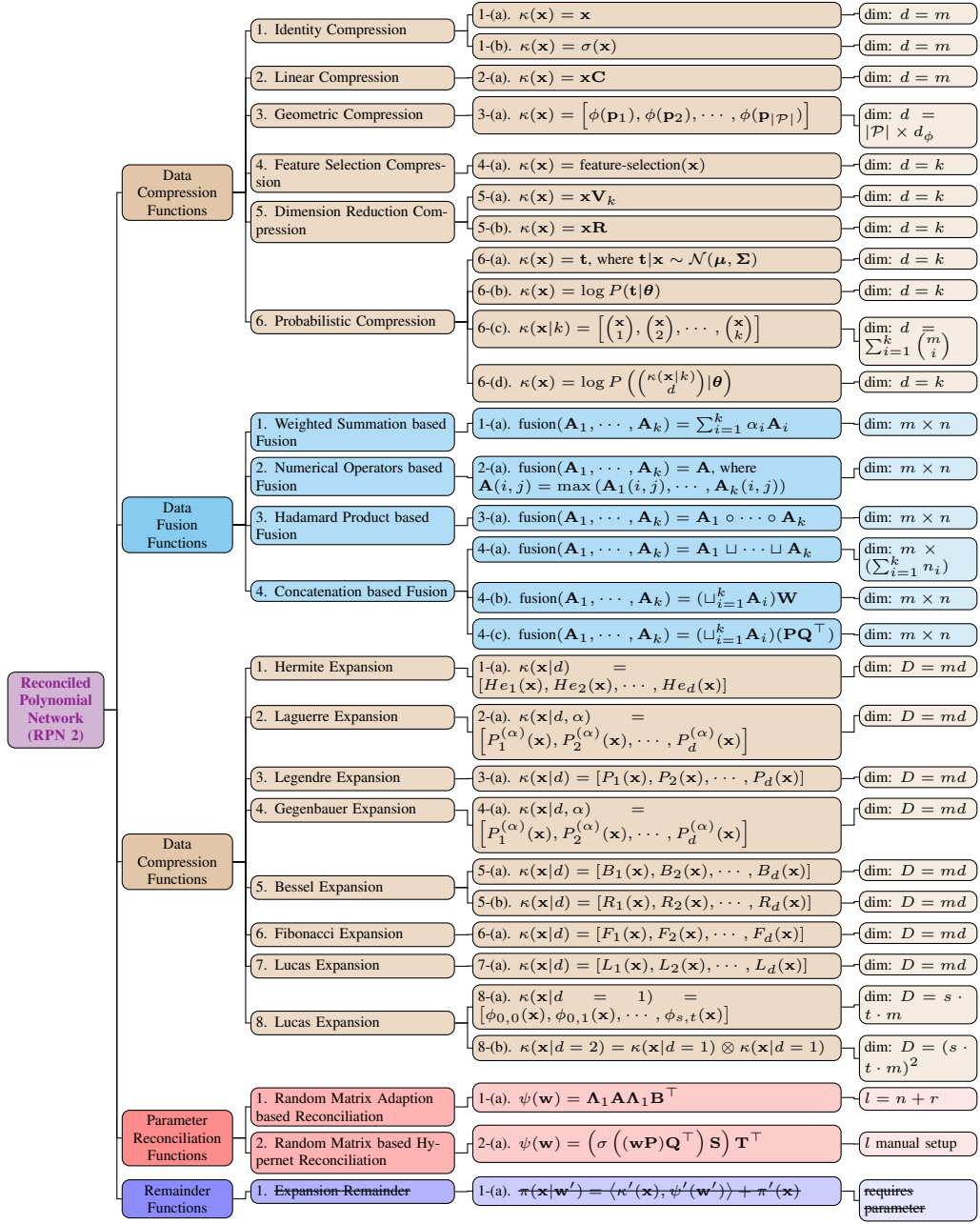


\section{Data Compression Function and Fusion Function}\label{sec:function_2}

Beyond interdependence functions, we also expand the scope of data transformation capabilities by presenting a new family of data compression functions. Together with previously introduced data expansion functions \cite{zhang2024rpnreconciledpolynomialnetwork}, they will define a comprehensive set of methods for transforming data across vector spaces. Additionally, to integrate outputs learned from different heads and channels, we propose a family of fusion functions based on various numerical and statistical metrics and aggregation techniques, which will be utilized in the wide architecture of the {\our} model.

Readers seeking for a concise review of these component functions may also refer to Figure~\ref{fig:compression_fusion_function}, which provides a summary of the functions to be introduced in this and the following sections. 


\subsection{Data Compression Functions}\label{subsec:data_compression_function}

In addition to the aforementioned data interdependence functions, which are defined based on either the input data batch or the underlying structural information, we introduce a new family of component functions in this paper: the \textit{data compression function}. This new data compression function, together with the data expansion functions introduced in our previous work \cite{zhang2024rpnreconciledpolynomialnetwork}, will comprise the family of data transformation functions.

As briefly mentioned in Section~\ref{subsec:rpn2_architecture}, the data transformation function $\kappa: \mathbbm{R}^m \to \mathbbm{R}^D$ projects data instances from a space of dimension $m$ to a target space of dimension $D$. Depending on the relative sizes of $m$ and $D$, this function can be further categorized into two types: \textit{data expansion functions} ($D \ge m$) and \textit{data compression functions} ($D \le m$). We have introduced various data expansion functions in the previous article \cite{zhang2024rpnreconciledpolynomialnetwork}, and readers are encouraged to refer to that paper for more information before proceeding with this subsection. Several new data expansion functions will also be introduced in the following subsection, which expands data instances with orthogonal polynomials and wavelets.

In this subsection, we focus on introducing the data compression functions that can be incorporated into the new {\our} model. Since the data interdependence relationships have been adequately addressed by the previously introduced data interdependence functions, we will treat all data instances and attributes as independent by default for the compression functions discussed below.


\subsubsection{Identity and Reciprocal Compression Function}

The identity and reciprocal functions, previously introduced in \cite{zhang2024rpnreconciledpolynomialnetwork}, can serve as both expansion and compression functions. For a given data instance $\mb{x} \in \mathbbm{R}^m$, we can represent the identity and reciprocal data compression functions as follows:

\begin{equation}
\kappa(\mb{x}) = \mb{x} \in \mathbbm{R}^d \text{, and } \kappa(\mb{x}) = \frac{1}{\mb{x}} \in \mathbbm{R}^d.
\end{equation}

To distinguish these from data expansion functions, we use the lower-case notation $d$ to represent the compression space dimension. As with expansion functions, optional pre- and post-processing functions ({\eg} activation or normalization functions) can be applied to the data instance before and after the compression function. It's worth noting that in these cases, the output dimension equals the input dimension, {\ie} $d = m$.


\subsubsection{Linear Compression Function}

We can also define a linear data compression function based on a pre-defined constant matrix $\mb{C} \in \mathbbm{R}^{m \times d}$. This function compresses a data instance from the input space of dimension $m$ to the compression space of dimension $d$ as follows:

\begin{equation}
\kappa(\mb{x}) = \mb{x} \mb{C} \in \mathbbm{R}^d.
\end{equation}

It's worth noting that this function was also defined as a data expansion function in our previous work \cite{zhang2024rpnreconciledpolynomialnetwork}. The linear transformation matrix $\mb{C}$ is pre-computed and provided as an input at the function definition. Additionally, the compression space dimension $d$ is a manually set hyper-parameter.

This linear compression function can be particularly effective when dealing with sparse input data. By reducing the vector dimensions, it helps decrease both the model space, time and learning costs, often with minimal information loss.


\subsubsection{Geometric Patch based Compression Function}\label{subsec:patch_compression_function}

In Section~\ref{subsec:structural_interdependence_functions}, we introduced several approaches to obtain geometric patches from the underlying grid structures of the input data batch. These methods can also be utilized to define data compression functions.

Formally, given a data instance $\mb{x}$ and its underlying grid structure, we can extract a set of patches denoted as $\mc{P} = \{{p}_1, {p}_2, \cdots {p}_{|\mc{P}|}\}$. The size of this patch set ({\ie} $|\mc{P}|$) is determined by three factors: the original size of $\mb{x}$, the patch shapes, and the patch center distance hyper-parameters used in the selected packing strategies. Each patch ${p}_i \in \mc{P}$ is represented as an ordered list of coordinates covered by the patch in the data grid structure. These coordinates can be used to retrieve the corresponding attribute values from the data instance vector $\mb{x}$ via the index-coordinate bijection introduced in the previous Section~\ref{subsubsec:data_grid_patch_structure}.

For simplicity, we use the notation $\mb{p}_i = \mb{x}(p_i) \in \mathbbm{R}^p$ to represent the attribute elements covered by patch ${p}_i \in \mc{P}$ from the input data instance vector $\mb{x}$, where $p$ denotes the patch size as introduced in Section~\ref{subsubsec:data_grid_patch_structure}. In this paper, we propose to compress the patch vector $\mb{p}_i$ using a mapping $\phi: \mathbbm{R}^p \to \mathbbm{R}^{d_{\phi}}$, which transforms it into a dense representation of length $d_{\phi}$. This mapping defines our patch-based data compression function as follows:

\begin{equation}
\kappa(\mb{x}) = \left[ \phi(\mb{p}_1), \phi(\mb{p}_2), \cdots, \phi(\mb{p}_{|\mc{P}|}) \right] \in \mathbbm{R}^{d}.
\end{equation}

In the above compression function, its compression output vector dimension is $d = |\mathcal{P}| \times d_{\phi}$. The dimension parameter $d_{\phi}$ must be manually specified when defining the patch vector compression mapping $\phi$. For the majority of mappings $\phi$ studied in this paper, the output is typically a scalar, {\ie} the dimension $d_{\phi} = 1$.

In practice, there are various ways to define the patch vector compression mapping $\phi$. We illustrate several of these approaches as follows:

\noindent\begin{minipage}{.45\linewidth}
\begin{fleqn}
{
\begin{equation}\label{equ:vector_norm_compression_mapping}
\begin{aligned}
&\underline{\textbf{(a) Vector Norms:}}\\[6pt]
&\phi(\mb{p}) = \left\| \mb{p} \right\|_p \in \mathbbm{R},\\[6pt]
&\text{where } p \in \{1, 2, \cdots, \infty\}.\\
\end{aligned}
\end{equation}
}
\end{fleqn}
\end{minipage}
\hfill
\noindent\begin{minipage}{.45\linewidth}
\begin{fleqn}
{
\begin{equation}\label{equ:entropy_compression_mapping}
\begin{aligned}
&\underline{\textbf{(b) Entropy:}}\\[2pt]
&\phi(\mb{p}) = - \sum_{i} (\mb{p}(i) \log \mb{p}(i))  \in \mathbbm{R},\\[0pt]
&\text{where vector $\mb{p}$ needs to be positive}.\\
\end{aligned}
\end{equation}
}
\end{fleqn}
\end{minipage}

\noindent\begin{minipage}{.45\linewidth}
\begin{fleqn}
{
\begin{equation}\label{equ:statistical_metric_compression_mapping}
\begin{aligned}
&\underline{\textbf{(c) Statistical Metrics:}}\\[6pt]
&\phi(\mb{p}) = \text{metric} (\mb{p}) \in \mathbbm{R}.\\[6pt]
\end{aligned}
\end{equation}
}
\end{fleqn}
\end{minipage}
\hfill
\noindent\begin{minipage}{.45\linewidth}
\begin{fleqn}
{
\begin{equation}\label{equ:numerical_compression_mapping}
\begin{aligned}
&\underline{\textbf{(d) Numerical Operator:}}\\[6pt]
&\phi(\mb{p}) = \text{operator} (\mb{p}) \in \mathbbm{R}.\\[6pt]
\end{aligned}
\end{equation}
}
\end{fleqn}
\end{minipage}

For the vector norm-based patch compression mapping, the $L_p$ norm can be selected with $p$ values from the set $\{1, 2, \cdots, \infty\}$. In the case of statistical metric-based patch compression functions, we employ the general function notation ``metric($\cdot$)'' to represent various statistical measures, including but not limited to \textit{variance}, \textit{standard deviation}, and \textit{skewness}. For numerical operator-based patch compression functions, the ``operator($\cdot$)'' notation encompasses different numerical operators, such as \textit{maximum}, \textit{minimum}, \textit{sum}, and \textit{product}, as well as various averaging functions like \textit{arithmetic mean}, \textit{geometric mean}, \textit{harmonic mean}, \textit{median}, and \textit{mode}. All of these diverse patch compression mappings have been implemented in the {\toolkit} toolkit.


\subsubsection{Feature Selection based Compression Function}

Several conventional machine learning techniques can also be employed to define data compression functions, notably \textit{feature selection} and \textit{dimension reduction}. \textit{Feature selection} is a process in machine learning and data analysis that aims to choose a subset of relevant features from a larger set of available features for model construction. \textit{Dimensionality reduction} techniques, on the other hand, seek to reduce the number of features or variables in a dataset while preserving as much important information as possible. In classic machine learning models, both feature selection and dimension reduction offer numerous advantages, such as improving model performance, reducing overfitting, simplifying models for easier interpretation, and decreasing training time.

Unlike the aforementioned compression functions, which can be applied directly to any input data instances or mini-batches, most existing feature selection and dimensionality reduction methods are primarily designed as pre-processing steps for static inputs on complete datasets. However, the multi-layered {\our} model involves dynamic inputs at each layer, where input and output values continuously change as new model parameters are learned. Consequently, most of these conventional static feature selection and dimensionality reduction methods can hardly be incorporated into the layers or adapted to such dynamic learning settings.

In this paper, we propose utilizing several incremental feature selection and dimension reduction approaches for defining the data compression functions at each layer. Simultaneously, we call for the development of novel online (or incremental) feature selection and dimension reduction methods from the research community, specifically those that can be adapted into the layers for deep function learning tasks.

Formally, given an input data instance $\mb{x} \in \mathbbm{R}^m$, we can represent the feature selection-based data compression function as follows:

\begin{equation}
\kappa(\mb{x}) = \text{feature-selection}(\mb{x}) \in \mathbbm{R}^d.
\end{equation}

Here, the function ``$\text{feature-selection}(\cdot)$'' represents an incremental feature selection approach capable of processing any input data instance $\mb{x}$ during the learning process. The output dimension $d$ must be manually specified when defining this function. Several examples of such approaches, including the \textit{incremental variance threshold} method and \textit{incremental feature clustering} method, will be introduced in detail in the following sections.


\noindent \textbf{Incremental Variance Threshold Method}: Slightly different from the notations used in the above compression functions, the incremental variance threshold method needs to calculate the variance for each attribute based on the inputs, which can only operate on the data batches instead of the individual instances discussed above. 

Formally, given an input data batch $\mb{X} \in \mathbbm{R}^{b \times m}$, we calculate its attribute variance vector as follows:

\begin{equation}
var(\mb{X}) = \mb{v} \in \mathbbm{R}^m.
\end{equation}

The $i_{th}$ entry of vector $\mb{v}$ denotes the variance calculated for the $i_{th}$ attribute of data batch $\mb{X}$, {\ie} the column vector $\mb{X}(:, i)$.

For small-sized input data batches, the calculated variance vector may change dramatically between batches, potentially leading to unstable performance. To address this issue, the incremental variance threshold method maintains a record of all historical variance vectors calculated since the first batch. This record is incrementally updated with new variance vectors as follows:

\begin{equation}\label{equ:incremental_updating}
\bar{\mb{v}} = \frac{(t-1) \cdot \bar{\mb{v}} + \mb{v}}{t} \in \mathbbm{R}^m,
\end{equation}

where $t$ denotes the counter index of the current batch and $\bar{\mb{v}}$ represents the historical variance record vector for all attributes. Furthermore, during both training and inference, attributes with variance values greater than a specified threshold are selected for the output:

\begin{equation}
\text{feature-selection}(\mb{X}) = \left[ \mb{X}(:, i) \right]_{i \in \{1, 2, \cdots, m\} \land {\bar{\mb{v}}(i) \ge p}},
\end{equation}

where $p$ denotes the provided variance threshold hyper-parameter.

In practice, rather than setting the variance threshold hyper-parameter $p$ (which may lead to varying numbers of selected attributes and cause dimension inconsistency between different data batches and training epochs), it is more common to select the top-$k$ attributes with the largest variances. This approach precisely determines the output dimension as $d = k$.


\noindent \textbf{Incremental Feature Clustering Method}: For the incremental feature clustering-based method, given an input data batch $\mb{X} \in \mathbbm{R}^{b \times m}$, the algorithm partitions the $m$ attributes into $k$ non-overlapping clusters based on pairwise similarities ({\eg} cosine similarity) calculated from the input data batches. Formally, we can represent these attribute clusters as:

\begin{equation}
\mc{C} = \{\mc{C}_1, \mc{C}_2, \cdots, \mc{C}_k\} \text{, where } \bigcup_{i} \mc{C}_i = \{1, 2, \cdots, m\} \land \mc{C}_i \cap \mc{C}_j = \emptyset, \forall i, j \in \{1, 2, \cdots, k\}.
\end{equation}

As new data batches arrive, this method incrementally updates the similarity scores among the attributes (similar to Equation~(\ref{equ:incremental_updating})), creating new partitions of the attributes. Within each identified attribute cluster, the most representative features are selected, typically those with the highest variance scores (or based on other selection criteria). This process can be represented as:

\begin{equation}
\text{feature-selection}(\mb{X}) = \left[ \mb{X}(:, i) \right]_{i \in \mc{C}_j \land \mc{C}_j \in \mc{C} \land \bar{\mb{v}}(i) \ge p},
\end{equation}

where $p$ is a threshold parameter and it can be set as the highest variance scores for each cluster ({\eg} the top-$1$ attribute with the largest variance in each cluster). Consequently, for the incremental feature clustering-based compression function, the output compression space dimension equals the provided cluster number hyper-parameter, {\ie} $d=k$.

For most input data batches, the number of attributes is typically not very large, especially compared to the number of instances. Therefore, recording attribute variances and learning attribute clusters does not incur significant space or time costs. Moreover, as the {\our} model's performance converges during the learning process, changes in the input data batches for each layer between sequential epochs become minor, resulting in minimal changes to attribute variances or clusters.

To further optimize the feature selection-based compression function, we can implement an early-stopping parameter. This parameter halts the updating of variance metrics and clustering results after a specified number of training epochs, helping to reduce the computational costs associated with tuning these incremental feature selection methods.


\subsubsection{Dimension Reduction based Compression Function}

Similar to feature selection, dimension reduction is another frequently used technique in classic machine learning for transforming high-dimensional data into a lower-dimensional space. In this paper, we propose the use of two dimensional reduction methods to define the data compression function, including the \textit{incremental principal component analysis} and \textit{random projection}.


\noindent \textbf{Incremental PCA}: Incremental Principal Component Analysis (PCA) \cite{Ross2008IncrementalLF} is a technique used for dimensionality reduction when dealing with large or continuously growing datasets. It extends traditional PCA to handle data incrementally by updating the principal components as new data arrives, rather than recomputing them from scratch.

Formally, given the input data batch $\mb{X} \in \mathbbm{R}^{b \times m}$, its singular value decomposition (SVD) can be represented as:

\begin{equation}
\mb{X} = \mb{U} \bs{\Sigma} \mb{V}^\top,
\end{equation}

where $\mb{U} \in \mathbbm{R}^{b \times b}$ and $\mb{V} \in \mathbbm{R}^{m \times m}$ are both orthogonal matrices. The matrix $\bs{\Sigma} = diag(\sigma_1, \sigma_2, \cdots, \sigma_r, 0, \cdots, 0) \in \mathbbm{R}^{b \times m}$ is a rectangular diagonal matrix with $r$ singular values $\sigma_1 \ge \sigma_2 \ge \cdots \ge \sigma_r > 0$ and zeros on the diagonal. The number of non-zero singular values on the diagonal of matrix $\bs{\Sigma}$ also defines the rank of matrix $\mb{X}$.

The column vectors of matrix $\mb{U}$ are orthogonal and called the \textit{left singular vectors}, while the orthogonal column vectors of matrix $\mb{V}$ are called the \textit{right singular vectors}. In the right singular matrix $\mb{V}$, we denote the columns corresponding to the $k$ largest singular values as $\mb{V}_{k} \in \mathbbm{R}^{m \times k}$. This helps calculate the principal components for each data instance $\mb{x} \in \mathbbm{R}^m$ in the data batch as:

\begin{equation}
\kappa(\mb{x}) = \mb{x} \mb{V}_k \in \mathbbm{R}^{d},
\end{equation}

where the instance output dimension $d = k$.

As a new data batch $\mb{X}' \in \mathbbm{R}^{b \times m}$ arrives, we need to efficiently calculate the SVD of the concatenation of $\mb{X}$ and $\mb{X}'$. The resulting $\mb{V}'$ matrix will help update the principal components from the new data batch $\mb{X}'$. In \cite{Ross2008IncrementalLF}, the authors introduce an efficient approach to update the previous SVD decomposition results ({\ie} $\mb{U}$, $\bs{\Sigma}$, $\mb{V}$ of $\mb{X}$) to calculate the new SVD decomposition matrices $\mb{U}'$, $\bs{\Sigma}'$, $\mb{V}'$ by incorporating the new data batch $\mb{X}'$. In this paper, we will use this method of incremental data batch dimension reduction for defining the data compression function.


\noindent \textbf{Random Projection}: Besides incremental PCA, random projection is another computationally efficient method for dimensionality reduction that can be used incrementally. While not as commonly used as PCA for incremental learning, it has unique advantages, especially for very high-dimensional data. The random projection method was proposed based on the \textit{Johnson-Lindenstrauss lemma} \cite{Johnson1984ExtensionsOL}, which states that ``\textit{a set of points in a high-dimensional space can be projected onto a lower-dimensional space while preserving pairwise distances}''.

Formally, given the input data instance $\mb{x} \in \mathbbm{R}^{m}$, random projection proposes to generate a random matrix $\mb{R} \in \mathbbm{R}^{m \times k}$ of size $m \times k$ to project the input data instance into a lower-dimensional space as follows:

\begin{equation}
\kappa(\mb{x}) = \mb{x} \mb{R} \in \mathbbm{R}^k.
\end{equation}

There exist different approaches to generate the random matrix $\mb{R}$, such as sparse random projection and Gaussian random projection.

For the \textit{sparse random projection} method, the random matrix $\mb{R}$ elements are generated subject to the density parameter $s \in [0, 1]$. For instance, each matrix element $\mb{R}(i, j)$ may take values:

\begin{equation}
\mb{R}(i, j) = 
\begin{cases}
-\sqrt{\frac{1}{s \cdot k}} & \text{ with probability } \frac{s}{2},\\
0 & \text{ with probability } 1-s,\\
+\sqrt{\frac{1}{s \cdot k}} & \text{ with probability } \frac{s}{2}.
\end{cases}
\end{equation}

Meanwhile, for the \textit{Gaussian random projection}, the random matrix elements, {\eg} $\mb{R}(i, j)$, are randomly drawn from the Gaussian distribution as follows:

\begin{equation}
\mb{R}(i, j) \sim \mc{N} \left(0, \frac{1}{k} \right).
\end{equation}

Similar to the feature selection-based compression functions, an early-stop parameter can also be applied to these dimension reduction-based compression functions. This parameter will halt the updating and tuning of the function's internal components as the {\our} model performance stabilizes with training epochs.

In addition to the dimension reduction methods based on \textit{incremental PCA} and \textit{random projection}, we have implemented several manifold-based techniques, including Isomap, t-SNE, Locally Linear Embedding, MDS, Spectral Embedding, and SMACOF. While we will not delve into the specifics of these manifold-based compression functions here, readers are encouraged to select the most suitable methods for their particular project and function learning tasks.


\subsubsection{Probabilistic Compression Function}

Building upon our previous work \cite{zhang2024rpnreconciledpolynomialnetwork}, which introduced probabilistic expansion functions for expanding input data instance vectors into their log-likelihood in naive or combinatorial modes, we now introduce a novel category of data compression functions based on probability distributions, termed probabilistic compression functions. These functions aim to compress data instances into vectors using probabilistic sampling methods.

Formally, given a data instance $\mb{x} \in \mathbbm{R}^m$, we define the probabilistic compression function based on probabilistic sampling as:

\begin{equation}\label{equ:probabilistic_compression_function}
\kappa(\mb{x}) = \mb{t} \in \mathbbm{R}^d,
\end{equation}

where the output vector $\mb{t}$ is conditionally dependent on $\mb{x}$ following certain distributions. For example, using a Gaussian distribution:

\begin{equation}\label{equ:compression_gaussian_distribution}
\mb{t} | \mb{x} \sim \mc{N}(\bs{\mu}, \bs{\Sigma}).
\end{equation}

The dimension $d$ of the output vector $\mb{t}$ is a hyper-parameter requiring manual setup. While we use the Gaussian distribution $\mc{N}(\bs{\mu}, \bs{\Sigma})$ as an example, $\mb{t} | \mb{x}$ can follow other distributions, such as Cauchy or Laplace distributions, which have been employed in defining probabilistic expansion functions in our previous work \cite{zhang2024rpnreconciledpolynomialnetwork}.

Similar to the previously introduced expansion functions, probabilistic compression functions can operate in both the naive and combinatorial modes.


\noindent \textbf{Naive Probabilistic Compression Function}: This naive probabilistic compression function assumes independence among attributes in the input data instance vector. It performs sequential random sampling of attributes without replacement from the data instance to compose the output vector of the desired length. Beyond simple random sampling with uniform distributions (implemented in the {\toolkit} toolkit as well), we can also sample attributes based on instance attribute values as follows.

Formally, given a data instance vector $\mb{x} \in \mathbbm{R}^m$, the sampling probabilities of all available attributes in the first sampling step can be represented as:

\begin{equation}\label{equ:naive_instance_probability_compression_function}
P( f(\mb{x}_i) | {\theta}_i), \forall i \in \{1, 2, \cdots, m\},
\end{equation}

where ${\theta}_i$ denotes the hyper-parameters of the distribution corresponding to the $i_{th}$ attribute. For Gaussian distributions, it can represent the mean $\mu_i$ and standard deviation $\sigma_i$ of the $i_{th}$ attribute, as indicated in Equation~(\ref{equ:compression_gaussian_distribution}). In practice, we may set equal-valued hyper-parameters ${\theta}$ for all attributes, {\ie} ${\theta}_i = {\theta}$, $\forall i \in {1, 2, \cdots, m}$.

The mapping $f(\cdot)$ used above projects the attribute value $\mb{x}_i$ to a scalar compatible with the distribution. For normalized attribute values, we can simply define $f(\mb{x}_i) = \mb{x}_i$; otherwise, we can define it as a normalization mapping, {\ie} $f(\mb{x}_i) = \text{normalize}(\mb{x}_i | {\theta}_i)$. Additionally, we can define the mapping to extract insightful numerical or statistical metrics about the attributes, similar to those introduced in Section~\ref{subsec:patch_compression_function}.

Furthermore, to maintain consistency with probabilistic expansion function outputs in certain learning scenarios, we can project the sampled attributes to their log-likelihoods, redefining the naive probabilistic compression function as:

\begin{equation}
\kappa(\mb{x}) = \log P(\mb{t} | \bs{\theta}) \in \mathbbm{R}^d.
\end{equation}


\noindent \textbf{Combinatorial Probabilistic Compression Function}: Building on the combinatorial expansions introduced in our previous work \cite{zhang2024rpnreconciledpolynomialnetwork}, we now present the combinatorial probabilistic compression function. Unlike naive probabilistic compression functions, this approach considers relationships among variables in multivariate distributions, enabling better modeling of complex data distributions.

Formally, given a data instance vector $\mb{x} \in \mathbbm{R}^m$, we can represent the combination of $k$ selected attributes from $\mb{x}$ as $\mb{x} \choose k$, for $k \in \{1, 2, \cdots, m\}$. For an input vector $\mb{x}$ of length $m$, the notation $\mb{x} \choose k$ represents a set containing $m \choose k$ attribute combination tuples, {\eg}

\begin{equation}
\begin{aligned}
k=1: {\mb{x} \choose 1} &= \left\{(\mb{x}_1), (\mb{x}_2), \cdots, (\mb{x}_m) \right\},\\
k=2: {\mb{x} \choose 2} &= \left\{(\mb{x}_1, \mb{x}_2), (\mb{x}_1, \mb{x}_3), \cdots, (\mb{x}_{m-1}, \mb{x}_m) \right\},\\
k=3: {\mb{x} \choose 2} &= \left\{(\mb{x}_1, \mb{x}_2, \mb{x}_3), (\mb{x}_1, \mb{x}_2, \mb{x}_4), \cdots, (\mb{x}_{m-2}, \mb{x}_{m-1} \mb{x}_m) \right\}.
\end{aligned}
\end{equation}

For all attribute combination tuples in $\mb{x} \choose d$, using multivariate distributions ({\eg} $\mc{N}(\bs{\mu}, \bs{\Sigma})$ defined on $d$ variables), one approach to compose the output attribute $\mb{t} \in \mathbbm{R}^d$ is to randomly select one combination tuple from $\mb{x} \choose d$ according to tuple-wise probability scores, similar to Equation~(\ref{equ:naive_instance_probability_compression_function}).

However, this sampling-based method can be challenging to implement due to the exponential size growth of $\mb{x} \choose d$ as $d$ increases. For large $d$ values, both enumerating $d$-sized attribute combinations in $\mb{x} \choose d$ and computing multivariate distributions with $d$ variables become computationally expensive.

To address this, we can utilize combinatorial expansion functions, consistent with the probabilistic expansion functions introduced in \cite{zhang2024rpnreconciledpolynomialnetwork}:

\begin{equation}\label{equ:combinatorial_expansion}
\kappa(\mb{x} | k) = \left[ {\mb{x} \choose 1}, {\mb{x} \choose 2}, \cdots, {\mb{x} \choose k} \right].
\end{equation}

This expansion output contains $\sum_{i=1}^k {m \choose i}$ tuples of varying lengths. Based on this expansion, we can define the combinatorial probabilistic compression function by sampling $d$ tuples from $\kappa(\mb{x} | 1:k)$, treating the tuples as independent ``items''. Depending on tuple length, a corresponding multivariate distribution can be applied to compute the log-likelihood of the tuples:

\begin{equation}
\kappa(\mb{x}) = \log P\left( {\kappa(\mb{x} | k) \choose d} | \bs{\theta} \right) \in \mathbbm{R}^d.
\end{equation}

In practice, due to the exponential growth of output dimensions in the combinatorial expansion function denoted by Equation~(\ref{equ:combinatorial_expansion}), the hyper-parameter $k$ is typically set to a small value, {\eg} $k=2$ or $k=3$. This ensures both the efficiency of this probabilistic compression function and also the output value consistency with the previous probabilistic expansion function outputs.


\subsection{Fusion Functions}\label{subsec:data_fusion_function}

The wide architecture with multi-head and multi-channel design endows {\our} with enhanced learning capacities. In the previous {\old} model, we used a summation-based fusion strategy that assigned equal importance to each head and channel. However, for more complex function learning tasks, this simple summation approach may no longer be adequate. In this section, we introduce several advanced fusion strategies that can more effectively aggregate the outputs from the wide architectures. Moreover, these fusion functions are versatile, supporting not only the construction of hybrid interdependence functions but also other essential functions, thus offering {\our} significant flexibility in architectural design.

To simplify the notations, we represent the inputs to the fusion function as matrices $\mb{A}_1, \mb{A}_2, \cdots, \mb{A}_k$, with the fusion output denoted by

\begin{equation}
\mb{A} = \text{fusion}(\mb{A}_1, \mb{A}_2, \cdots, \mb{A}_k).
\end{equation}

The dimensions of the input matrices $\mb{A}_1, \mb{A}_2, \cdots, \mb{A}_k$ may be identical or vary, depending on the specific definition of the fusion function. We will specify their dimensions in detail when introducing the concrete functions below.


\subsubsection{Weighted Summation based Fusion Function}

The weighted summation-based fusion function requires that all input matrices have identical dimensions. Formally, given interdependence matrices $\mb{A}_1, \mb{A}_2, \ldots, \mb{A}_k \in \mathbbm{R}^{m \times n}$ of dimension $m \times n$, we can combine them through a weighted summation as follows:

\begin{equation}
\text{fusion}(\mb{A}_1, \mb{A}_2, \cdots, \mb{A}_k) = \sum_{i=1}^k \alpha_i \mb{A}_i \in \mathbbm{R}^{m \times n},
\end{equation}

where $\alpha_i$ represents the weight assigned to matrix $\mb{A}_i$ for each $i \in \{1, 2, \cdots, k\}$. 

These weights can either be provided manually, leveraging domain expertise, or initialized as learnable parameters. Alternatively, we may set them to fixed values, such as 1 or $\frac{1}{k}$, simplifying the fusion function to a straightforward summation or averaging of the input matrices, respectively.


\subsubsection{Numerical Operators based Fusion Function}

In addition to weighted summation, the fusion operator can be defined using various numerical operators introduced in Section~\ref{subsec:patch_compression_function}, such as maximum, minimum, and different averaging functions. Like the summation-based fusion function, these numerical operator-based fusion functions also require that the input matrices have identical dimensions.

For example, we can define the fusion operator using the maximum operator, combining the input interdependence matrices $\mb{A}_1, \mb{A}_2, \ldots, \mb{A}_k \in \mathbbm{R}^{m \times n}$ as follows:

\begin{equation}
\text{fusion}(\mb{A}_1, \mb{A}_2, \cdots, \mb{A}_k) = \mb{A}  \in \mathbbm{R}^{m \times n},
\end{equation}

and the entry $\mb{A}(i, j)$ (for $i \in \{1, 2, \cdots, m\}$ and $j \in \{1, 2, \cdots, n\}$) can be represented as

\begin{equation}
\mb{A}(i, j) = \max \left( \mb{A}_1(i,j), \mb{A}_2(i,j), \cdots, \mb{A}_k(i,j)  \right).
\end{equation}


\subsubsection{Hadamard Product based Fusion Function}

In this paper, we extend the Hadamard product, originally defined for two matrices, to handle $k$ matrices, $\mb{A}_1, \mb{A}_2, \cdots, \mb{A}_k \in \mathbbm{R}^{m \times n}$, enabling element-wise fusion across multiple inputs:

\begin{equation}
\mb{A} = \text{fusion}(\mb{A}_1, \mb{A}_2, \cdots, \mb{A}_k) = \mb{A}_1 \circ \mb{A}_2 \circ \cdots \circ \mb{A}_k  \in \mathbbm{R}^{m \times n},
\end{equation}

where for the $(i, j)_{th}$ element is calculated as

\begin{equation}
\mb{A}(i, j) = \mb{A}_1(i,j) \times \mb{A}_2(i,j) \times \cdots \times \mb{A}_k(i,j).
\end{equation}

The Hadamard product-based fusion function is particularly beneficial for hybrid interdependence functions that incorporate both structural and data batch information. Here, structural information can serve as a filtering mask, selectively extracting relevant information from data batches. We provide an example of defining the hybrid interdependence function for graph-structured data in Section~\ref{subsubsec:hybrid_graph_interdependence}.


\subsubsection{Concatenation and Linear Transformation based Fusion Function}

Similar to the wide architectures discussed earlier, we can concatenate the input interdependence matrices and then reduce them to the desired dimensions using a linear transformation. This fusion function requires only that the input matrices have the same number of rows, allowing for differing numbers of columns.

Formally, given input interdependence matrices $\mb{A}_1, \mb{A}_2, \ldots, \mb{A}_k$, where each matrix $\mb{A}_i \in \mathbbm{R}^{m \times n_i}$ has $m$ rows and $n_i$ columns, we define the fusion operator as follows:

\begin{equation}
\begin{aligned}
\mb{A} &= \text{fusion}(\mb{A}_1, \mb{A}_2, \cdots, \mb{A}_k) \\
&= \left( \mb{A}_1 \sqcup \mb{A}_2 \sqcup \cdots \sqcup \mb{A}_k \right) \mb{W} \in \mathbbm{R}^{m \times n}
\end{aligned}
\end{equation} 

where $\sqcup$ denotes the row-wise concatenation of the matrices. The term $\mb{W} \in \mathbbm{R}^{(\sum_{i=1}^k n_i) \times n}$ represents a learnable parameter matrix that projects the concatenated matrix to a dimension of $n$.

The concatenation of these interdependence matrices results in a relatively large dimension, specifically $\sqcup_{i=1}^k \mb{A}i \in \mathbbm{R}^{m \times (\sum{i=1}^k n_i)}$. To reduce the number of learnable parameters, we can also apply low-rank parameter reconciliation, allowing us to rewrite the fusion function as follows:

\begin{equation}
\begin{aligned}
\mb{A} &= \text{fusion}(\mb{A}_1, \mb{A}_2, \cdots, \mb{A}_k) \\
&= \left( \sqcup_{i=1}^k \mb{A}_i \right) \left( \mb{P} \mb{Q}^\top \right) \in \mathbbm{R}^{m \times n},
\end{aligned}
\end{equation} 

where the parameter matrix $\mb{W}$ is factorized into the inner product of two sub-matrices, specifically $\mb{W} = \mb{P} \mb{Q}^\top \in \mathbbm{R}^{(\sum_{i=1}^k n_i) \times n}$, where $\mb{P} \in \mathbbm{R}^{(\sum_{i=1}^k n_i) \times r}$ and $\mb{Q} \in \mathbbm{R}^{n \times r}$ represent the low-rank factorization sub-matrices of $\mb{W}$.

Beyond the operators discussed above, several other fusion operators can also be employed to define fusion strategies for the hybrid interdependence function. While we illustrated this with attribute interdependence functions as an example of hybrid interdependence, similar functions can also be defined and fused for instance interdependence relationships, though these will not be covered in this section.

\section{Other Component Function Updates in the Enhanced RPN 2 Model}\label{sec:function_3}

Moreover, to enhance the existing suite of data expansion and parameter reconciliation functions, we introduce several new implementations designed to augment these components. These functions, in combination with others mentioned earlier, facilitate the creation of more versatile, efficient, and effective architectures within {\our}. In refining the {\toolkit} toolkit, we deliberately deleted the complementary expansion-based remainder function proposed in our previous work \cite{zhang2024rpnreconciledpolynomialnetwork}. This decision was made to eliminate redundancy in {\our}, as such complementary expansion-based remainders can be equivalently implemented using the multi-head and multi-channel mechanisms, which are now the default strategy in building {\our}. These updated component functions introduced in this section have also been summarized in the previous Figure~\ref{fig:compression_fusion_function}.



\subsection{Data Expansion Functions}

In addition to the aforementioned interdependence and compression functions, this paper also expands the existing list of data expansion functions proposed in our previous work \cite{zhang2024rpnreconciledpolynomialnetwork}. In our previous {\old} paper \cite{zhang2024rpnreconciledpolynomialnetwork}, we introduced several orthogonal polynomials, such as the Chebyshev and Jacobi polynomials. In mathematics, an orthogonal polynomial sequence is a family of polynomials where any two distinct polynomials in the sequence are orthogonal to each other under some inner product. Building on our previous work, we will define several new orthogonal polynomials that can perform different data expansions for extracting useful information from the input data batch.

In addition to these new orthogonal polynomials, we also introduce a novel data expansion function based on wavelets, designed to expand multi-modal data such as audio signals and images into new spaces. In signal processing, a wavelet is formally defined as a wave-like oscillation with an amplitude that begins at zero, increases or decreases, and then returns to zero one or more times. Compared to the transformations introduced in \cite{zhang2024rpnreconciledpolynomialnetwork}, such as Fourier transformation, wavelet transformation replaces trigonometric basis functions with wavelet-based basis functions, which offers greater advantages in terms of {multi-resolution analysis}, {adaptability}, and {computational efficiency}.

In this section, we will introduce additional orthogonal polynomials and new wavelet transformation techniques that can be utilized to define data expansion functions. These newly added component functions significantly enrich the foundational building blocks available for {\our} model design, potentially enhancing its learning performance across diverse function learning tasks.



\subsubsection{{Hermite Polynomials based Expansion Function}}


Hermite polynomials, first defined by Pierre-Simon Laplace in 1810 and studied in detail by Pafnuty Chebyshev in 1859, were later named after Charles Hermite, who published work on these polynomials in 1864. The Hermite polynomials can be defined in various forms:

\noindent \underline{\textit{Probabilist's Hermite polynomials}}: 

\begin{equation}
He_n(x) = (-1)^n \exp \left(\frac{x^2}{2} \right) \frac{\mathrm{d}^n}{\mathrm{d}x^n} \exp \left(- \frac{x^2}{2} \right).
\end{equation}

\noindent \underline{\textit{Physicist's Hermite polynomials}}: 

\begin{equation}
H_n(x) = (-1)^n \exp \left(x^2 \right) \frac{\mathrm{d}^n}{\mathrm{d}x^n} \exp \left(- x^2 \right).
\end{equation}

These two forms are not identical but can be reduced to each via rescaling:
\begin{equation}
H_n(x) = 2^{\frac{n}{2}} He_n (\sqrt{2}x) \text{, and } He_n(x) = 2^{-\frac{n}{2}} H_n \left(\frac{x}{\sqrt{2}} \right).
\end{equation}

In this paper, we will use the Probabilist's Hermite polynomials for to define the data expansion function by default, which can be formally defined as the following recursive representations:

\begin{equation}
He_{n+1}(x)  = x He_n(x) - n He_{n-1}(x), \forall n \ge 1.
\end{equation}

Some examples of the Probabilist's Hermite polynomials are also illustrated as follows:

\begin{equation}
\begin{aligned}
He_{0}(x) &= 1;\\
He_{1}(x) &= x;\\
He_{2}(x) &= x^2 - 1;\\
He_{3}(x) &= x^3 - 3x;\\
He_{4}(x) &= x^4 - 6x^2 + 3.\\
\end{aligned}
\end{equation}

Based on the Probabilist's Hermite polynomials, we can define the data expansion function with order $d$ as follows:

\begin{equation}
\kappa(\mb{x} | d)  = \left[ He_1(\mb{x}), He_2(\mb{x}), \cdots, He_d(\mb{x}) \right] \in \mathbbm{R}^D,
\end{equation}

where $d$ is the order hyper-parameter and the output dimension $D = md$. Similar as the data expansion functions introduced in the previous paper \cite{zhang2024rpnreconciledpolynomialnetwork}, the constant term $He_{0}(x)$ is not included in the expansion outputs.


\subsubsection{{Laguerre Polynomials based Expansion Function}}


In mathematics, the Laguerre polynomials, named after Edmond Laguerre, are the nontrivial solutions of Laguerre's differential equation:

\begin{equation}
x y'' + (\alpha + 1 -x) y' + d y = 0,
\end{equation}

where $y = y(x)$ is a function of variable $x$. Notations $y'$ and $y''$ denote first- and second-order derivatives of function $y$ with respect to variable $x$. Term $d \in \mathbbm{N}$ is a non-negative integer and $\alpha \in \mathbbm{R}$ is a hyper-parameter.

The closed-form of the Laguerre polynomials can be represented as follows:

\begin{equation}
P^{(\alpha)}_n(x) = \frac{e^x}{n!} \frac{\mathrm{d}^n}{\mathrm{d} x^n} (e^{-x} x^n) = \frac{x^{-\alpha}}{n!} \left( \frac{\mathrm{d}}{\mathrm{d} x} - 1 \right)^n x^{n + \alpha},
\end{equation}

where $\frac{\mathrm{d}}{\mathrm{d} x}$ denotes the derivative operator.

In practice, the Laguerre polynomials can be recursively defined as follows, which will be used for defining the data expansion function below. Specifically, when $\alpha = 0$, the above Laguerre polynomials are also known as simple Laguerre polynomials. 

\noindent \underline{\textit{Base cases $n=0$ and $n=1$}}:

\begin{equation}
P^{(\alpha)}_0(x) = 1 \text{, and } P^{(\alpha)}_1(x) = 1 + \alpha -  x.
\end{equation}

\noindent \underline{\textit{High-order cases with degree $n \ge 2$}}:

\begin{equation}
P^{(\alpha)}_n(x) = \frac{(2n-1+\alpha-x) P^{(\alpha)}_{n-1}(x) - (n-1+\alpha) P^{(\alpha)}_{n-2}(x) }{n}
\end{equation}

The recursive-form representations of the Laguerre polynomials can be used to define the data expansion function as follows:

\begin{equation}
\kappa(\mb{x} | d, \alpha) = \left[ P^{(\alpha)}_1(\mb{x}), P^{(\alpha)}_2(\mb{x}), \cdots, P^{(\alpha)}_d(\mb{x}) \right] \in \mathbbm{R}^D,
\end{equation}

where $d$ and $\alpha$ are the function hyper-parameters and the output dimension $D = md$.


\subsubsection{{Legendre Polynomials based Expansion Function}}


The Legendre polynomials, named after mathematician Adrien-Marie Legendre, are defined as an orthogonal system over the interval $[-1, 1]$, where the polynomial term $P_n(x)$ of degree $n$ satisfies the following equation:

\begin{equation}
\int_{-1}^{+1} P_m(x) P_n(x) dx = 0, \text{ if } m \neq n.
\end{equation}

Specifically, according to Bonnet's formula, the Legendre polynomials can be recursively represented as follows:

\noindent \underline{\textit{Base cases $n=0$ and $n=1$}}:

\begin{equation}
P_0(x) = 1 \text{, and } P_1(x) = x.
\end{equation}

\noindent \underline{\textit{High-order cases with degree $n \ge 2$}}:

\begin{equation}
P_n(x) = \frac{x(2n-1) P_{n-1}(x) - (n-1) P_{n-2}(x) }{n}
\end{equation}

The Legendre polynomials help define the data expansion function as follows:

\begin{equation}
\kappa(\mb{x} | d) = \left[ P_1(\mb{x}), P_2(\mb{x}), \cdots, P_d(\mb{x}) \right] \in \mathbbm{R}^D,
\end{equation}

where the output dimension $D = md$.


\subsubsection{{Gegenbauer Polynomials based Expansion Function}}


The Gegenbauer polynomials, named after mathematician Leopold Gegenbauer, are orthogonal polynomials that generalize both the Legendre and Chebyshev polynomials, and are special cases of Jacobi polynomials.

Formally, the Gegenbauer polynomials are particular solutions of the Gegenbauer differential equation:

\begin{equation}
(1 - x^2) y'' - (2 \alpha + 1) x y' + d(d+2 \alpha) y = 0,
\end{equation}

where $y = y(x)$ is a function of variable $x$ and $d \in \mathbbm{N}$ is a non-negative integer.

When $\alpha = \frac{1}{2}$, the Gegenbauer polynomials reduce to the Legendre polynomials introduced earlier; when $\alpha = 1$, they reduce to the Chebyshev polynomials of the second kind.

The Gegenbauer polynomials can be recursively defined as follows:

\noindent \underline{\textit{Base cases $n=0$ and $n=1$}}:

\begin{equation}
P^{(\alpha)}_0(x) = 1 \text{, and } P^{(\alpha)}_1(x) = 2 \alpha x.
\end{equation}

\noindent \underline{\textit{High-order cases with degree $n \ge 2$}}:

\begin{equation}
P^{(\alpha)}_n(x) = \frac{2x(n-1+\alpha) P^{(\alpha)}_{n-1}(x) - (n+2\alpha -2) P^{(\alpha)}_{n-2}(x) }{n}
\end{equation}

Based on the Gegenbauer polynomials, we can define the expansion function as follows:

\begin{equation}
\kappa(\mb{x} | d, \alpha) = \left[ P^{(\alpha)}_1(\mb{x}), P^{(\alpha)}_2(\mb{x}), \cdots, P^{(\alpha)}_d(\mb{x}) \right] \in \mathbbm{R}^D,
\end{equation}

where the output dimension $D = md$.


\subsubsection{{Bessel and Reverse Bessel Polynomials based Expansion Functions}}


Formally, the Bessel polynomials are an orthogonal sequence of polynomials with the following closed-form representation:

\begin{equation}
B_n(x) = \sum_{k=0}^n \frac{(n+k)!}{(n-k)! k!} \left( \frac{x}{2}\right)^k.
\end{equation}

Another definition, favored by electrical engineers, is sometimes known as the reverse Bessel polynomials, with the following closed-form representation:

\begin{equation}
R_n(x) = x^n B_n \left( \frac{1}{x} \right) = \sum_{k=0}^n \frac{(n+k)!}{(n-k)! k!} \frac{x^{n-k}}{2^k} .
\end{equation}

Both the Bessel and reverse Bessel polynomials can be recursively defined as follows:

\noindent \underline{\textit{Base cases $n=0$ and $n=1$}}:

\begin{equation}
\begin{aligned}
\textit{Bessel: } &B_0(x) = 1 \text{, and } B_1(x) = x + 1;\\
\textit{Reverse Bessel: } &R_0(x) = 1 \text{, and } R_1(x) = x + 1.
\end{aligned}
\end{equation}

\noindent \underline{\textit{High-order cases with degree $n \ge 2$}}:

\begin{equation}
\begin{aligned}
\textit{Bessel: } &B_n(x) = (2n - 1) x B_{n-1}(x) + B_{n-2}(x);\\
\textit{Reverse Bessel: } &R_n(x) = (2n - 1) B_{n-1}(x) + x^2 B_{n-2}(x).
\end{aligned}
\end{equation}

Both the Bessel and reverse Bessel polynomials can be used to define the data expansion functions as follows:

\begin{equation}
\begin{aligned}
\textit{Bessel: }  \kappa(\mb{x} | d) &= \left[ B_1(\mb{x}), B_2(\mb{x}), \cdots, B_d(\mb{x}) \right] \in \mathbbm{R}^D,\\
\textit{Reverse Bessel: }  \kappa(\mb{x} | d) &= \left[ R_1(\mb{x}), R_2(\mb{x}), \cdots, R_d(\mb{x}) \right] \in \mathbbm{R}^D,
\end{aligned}
\end{equation}

where the output dimension $D = md$.


\subsubsection{{Fibonacci and Lucas Polynomials based Expansion Functions}}

Formally, the Fibonacci polynomials are a polynomial sequence that can be considered a generalization of the Fibonacci numbers. Similarly, Lucas polynomials are generated from the Lucas numbers in an analogous manner.

Both Fibonacci and Lucas polynomials can be defined recursively. The Lucas polynomials can be viewed as identical to the Fibonacci polynomials but with different base case representations:

\noindent \underline{\textit{Base cases $n=0$ and $n=1$}}:

\begin{equation}
\begin{aligned}
\textit{Fibonacci: } &F_0(x) = 0 \text{, and } F_1(x) = 1;\\
\textit{Lucas: } &L_0(x) = 2 \text{, and } L_1(x) = x.
\end{aligned}
\end{equation}

\noindent \underline{\textit{High-order cases with degree $n \ge 2$}}:

\begin{equation}
\begin{aligned}
\textit{Fibonacci: } &F_n(x) = x F_{n-1}(x) + F_{n-2}(x);\\
\textit{Lucas: } &L_n(x) = x L_{n-1}(x) + L_{n-2}(x).
\end{aligned}
\end{equation}

\begin{example}

Based on these recursive representations, we can illustrate some examples of the Fibonacci and Lucas polynomials as follows:

\begin{multicols}{2}
\noindent \textit{Fibonacci Polynomials}: 
\begin{equation}
\begin{aligned}
F_0(x) &= 0 \\
F_1(x) &= 1 \\
F_2(x) &= x \\
F_3(x) &= x^2 + 1 \\
F_4(x) &= x^3 + 2x \\
F_5(x) &= x^4 + 3x^2 + 1 \\
\end{aligned}
\end{equation}
\vfill
\noindent \textit{Lucas Polynomials}: 
\begin{equation}
\begin{aligned}
L_0(x) &= 2 \\
L_1(x) &= x \\
L_2(x) &= x^2 + 2 \\
L_3(x) &= x^3 + 3x \\
L_4(x) &= x^4 + 4x^2 + 2 \\
L_5(x) &= x^5 + 5x^3 + 5x \\
\end{aligned}
\end{equation}
\end{multicols}
\end{example}

Both the Fibonacci and Lucas polynomials can be used to define the data expansion functions as follows:

\begin{equation}
\begin{aligned}
\textit{Fibonacci: }  \kappa(\mb{x} | d) &= \left[ F_1(\mb{x}), F_2(\mb{x}), \cdots, F_d(\mb{x}) \right] \in \mathbbm{R}^D,\\
\textit{Lucas: }  \kappa(\mb{x} | d) &= \left[ L_1(\mb{x}), L_2(\mb{x}), \cdots, L_d(\mb{x}) \right] \in \mathbbm{R}^D,
\end{aligned}
\end{equation}

where the output dimension $D = md$.


\subsubsection{Wavelet based Expansion Functions}

In our previous paper \cite{zhang2024rpnreconciledpolynomialnetwork}, we introduced the Fourier series for expanding data instances into a sequence of trigonometric functions. Fourier series are closely related to the Fourier transform, which can be used to find frequency information for non-periodic functions. In fact, the Fourier transform can be viewed as a special case of the continuous wavelet transform. While the standard Fourier transform is only localized in frequency, wavelets are localized in both time and frequency, making them particularly useful for non-stationary signals where frequency components change over time.

Wavelet transform represents the input data as a summation of basis functions, known as wavelets. Specifically, the basis functions in wavelet transformation can be categorized into the \textit{mother wavelet} and \textit{father wavelet}, which are orthogonal and can both be derived from the \textit{child wavelet} via rescaling and translation operators.

Formally, given the input variable $\mb{x} \in \mathbbm{R}^{m}$, to approximate the underlying mapping $f: \mathbbm{R}^m \to \mathbbm{R}^n$ with wavelet analysis, we can define the approximated output as

\begin{equation}\label{equ:wavelet_approximation}
f(\mb{x}) \approx \sum_{s, t} \left \langle f(\mb{x}), \phi_{s, t} (\mb{x} | a, b) \right \rangle \cdot \phi_{s, t} (\mb{x} | a, b),
\end{equation}

where $\phi_{s, t} (\cdot | a, b)$ denotes the child wavelet defined by hyper-parameters $a > 1$ and $b > 0$:

\begin{equation}
\phi_{s, t}(x | a, b) = \frac{1}{\sqrt{a^s}} \phi \left( \frac{x - t \cdot b \cdot a^s}{a^s} \right).
\end{equation}

Specifically, the functions $\left\{ \phi_{s, t}\right\}_{ s, t \in \mathbbm{Z}}$ defines the orthonormal basis of the space and the mapping $\phi(\cdot)$ used in the child wavelet may have different representations:

\noindent\begin{minipage}{.45\linewidth}
\begin{fleqn}
{
\begin{equation}
\begin{aligned}
&\underline{\textbf{(a) Haar wavelet:}}\\[6pt]
&\phi(\tau) = \begin{cases}
1, & 0 \le \tau < \frac{1}{2},\\
-1, & \frac{1}{2} \le \tau < 1,\\
0, & \text{ otherwise}.
\end{cases},\\[6pt]
\end{aligned}
\end{equation}
}
\end{fleqn}
\end{minipage}
\hfill
\noindent\begin{minipage}{.45\linewidth}
\begin{fleqn}
{
\begin{equation}
\begin{aligned}
&\underline{\textbf{(b) Beta wavelet:}}\\[6pt]
&\phi(\tau | \alpha, \beta) = \frac{1}{B(\alpha, \beta)} \tau^{\alpha - 1} (1-\tau)^{\beta -1},\\[6pt]
&\text{where } \alpha, \beta \in [1, \infty].
\end{aligned}
\end{equation}
}
\end{fleqn}
\end{minipage}

\noindent\begin{minipage}{.5\linewidth}
\begin{fleqn}
{
\begin{equation}
\begin{aligned}
&\underline{\textbf{(c) Ricker wavelet:}}\\[6pt]
&\phi(\tau) = \frac{2 \left( 1 - \left( \frac{\tau}{\sigma} \right)^2 \right)}{\sqrt{3 \sigma} \pi^{\frac{1}{4}}} \exp\left(- \frac{\tau^2}{2 \sigma^2} \right). \\[6pt]
\end{aligned}
\end{equation}
}
\end{fleqn}
\end{minipage}
\hfill
\noindent\begin{minipage}{.45\linewidth}
\begin{fleqn}
{
\begin{equation}
\begin{aligned}
&\underline{\textbf{(d) Shannon wavelet:}}\\[6pt]
&\phi(\tau) = \frac{\sin(2\pi \tau) - \sin(\pi \tau)}{\pi \tau}.\\[6pt]
\end{aligned}
\end{equation}
}
\end{fleqn}
\end{minipage}

\noindent\begin{minipage}{.5\linewidth}
\begin{fleqn}
{
\begin{equation}
\begin{aligned}
&\underline{\textbf{(e) Difference of Gaussians:}}\\[6pt]
&\phi(\tau | \sigma_1, \sigma_2) = {P}(\tau | 0, \sigma_1) - {P}(\tau | 0, \sigma_2), \\[6pt]
&\text{where } {P}(\cdot | 0, \sigma_1) \text{ denotes the PDF of the}\\
&\text{Gaussian distribution}.
\end{aligned}
\end{equation}
}
\end{fleqn}
\end{minipage}
\hfill
\noindent\begin{minipage}{.45\linewidth}
\begin{fleqn}
{
\begin{equation}
\begin{aligned}
&\underline{\textbf{(f) Meyer wavelet:}}\\[6pt]
&\phi(\tau) = 
\begin{cases}
\frac{2}{3} + \frac{4}{3\pi} & \tau = 0,\\
\frac{ \sin(\frac{2 \pi}{3} \tau) + \frac{4}{3} \tau \cos( \frac{4 \pi}{3} \tau) }{ \pi \tau - \frac{16 \pi}{9} \tau^3 } & \text{otherwise}.
\end{cases}
\end{aligned}
\end{equation}
}
\end{fleqn}
\end{minipage}

To apply the aforementioned wavelets for data expansion, we need to re-examine Equation~(\ref{equ:wavelet_approximation}) introduced earlier. This equation can be interpreted in various ways within the context of {\our}:

\begin{equation}
\sum_{s, t} \underbrace{\left \langle f(\mb{x}), \phi_{s, t} (\mb{x} | a, b) \right \rangle}_{\text{coefficients}} \cdot \underbrace{\phi_{s, t} (\mb{x} | a, b)}_{\text{the expansion}},
\end{equation}
and
\begin{equation}
\sum_{s, t} \langle \underbrace{f(\mb{x})}_{\text{coefficients}}, \underbrace{\phi_{s,t} (\mb{x} | a, b) \rangle \cdot \phi_{s,t} (\mb{x} | a, b)}_{\text{the expansion}}.
\end{equation}

Based on these above two representations, we can introduce the $1_{st}$-order and $2_{nd}$-order wavelet data expansion functions as follows:
\begin{equation}
\kappa(\mb{x} | d=1) = \left[ \phi_{0, 0}(\mb{x}), \phi_{0, 1}(\mb{x}), \cdots, \phi_{s, t}(\mb{x}) \right] \in \mathbbm{R}^{D_1}.
\end{equation}
and
\begin{equation}
\kappa(\mb{x} | d=2) = \kappa(\mb{x} | d=1) \otimes \kappa(\mb{x} | d=1) \in \mathbbm{R}^{D_2}.
\end{equation}

The output dimensions of the order-1 and order-2 wavelet expansions are $D_1 = s \cdot t \cdot m$ and $D_2 = (s \cdot t \cdot m)^2$, respectively.


\subsection{Parameter Reconciliation Function}

We have introduced several different categories of parameter reconciliation functions in the previous paper \cite{zhang2024rpnreconciledpolynomialnetwork} already. In this part, we will introduce a new category of parameter reconciliation functions defined based on the random matrices.


\subsubsection{{Random Matrix Adaption based Parameter Reconciliation Function}}

According to the low-rank reconciliation (LoRR) function introduced in our previous paper \cite{zhang2024rpnreconciledpolynomialnetwork}, given a parameter vector $\mb{w} \in \mathbbm{R}^l$ of length $l$, we can fabricate it into a parameter matrix of shape $n \times D$ as follows:

\begin{equation}
\psi(\mb{w}) = \mb{A} \mb{B}^\top \in \mathbbm{R}^{n \times D},
\end{equation}

where $\mb{A} \in \mathbbm{R}^{n \times r}$ and $\mb{B} \in \mathbbm{R}^{D \times r}$ are the low-rank parameter sub-matrices of rank $r$ reshaped from the input parameter vector $\mb{w}$. The length of the input parameter is determined by the rank hyper-parameter $r$, {\ie} $l = (n + D) \cdot r$.

The recent VeRA paper \cite{Kopiczko2023VeRAVR} proposes to freeze the sub-matrices $\mb{A}$ and $\mb{B}$ as random constants, {\eg}

\begin{equation}
\mb{A}, \mb{B} \sim \mc{N}(\mb{0}, \mb{I}),
\end{equation}

where the sub-matrix elements are randomly sampled from the Gaussian distribution $\mc{N}(\mb{0}, \mb{I})$.

The learnable parameters can be added as the diagonal matrices $\bs{\Lambda}_1 = diag(\bs{\lambda}_1) \in \mathbbm{R}^{n \times n}$ and $\bs{\Lambda}_2 = diag(\bs{\lambda}_2) \in \mathbbm{R}^{r \times r}$, where vectors $\bs{\lambda}_1 \in \mathbbm{R}^n$ and $\bs{\lambda}_2 \in \mathbbm{R}^r$ are split from the input parameter $\mb{w}$. This defines the random matrix adaptation-based parameter reconciliation function as follows:

\begin{equation}
\psi(\mb{w}) = \bs{\Lambda}_1 \mb{A} \bs{\Lambda}_1 \mb{B}^\top \in \mathbbm{R}^{n \times D},
\end{equation}

where the required number of learnable parameters is $l = n + r$.


\subsubsection{{Random Matrix based Hypernet Parameter Reconciliation Function}}

We introduced the Hypernet parameter reconciliation function in our previous paper \cite{zhang2024rpnreconciledpolynomialnetwork}. Given the input parameter vector $\mb{w} \in \mathbbm{R}^l$, the function can be represented as:

\begin{equation}
\psi(\mb{w}) = \text{Hypernet}(\mb{w}) \in \mathbbm{R}^{n \times D},
\end{equation}

where ``Hypernet($\cdot$)'' can be defined with different models, such as MLP, with randomized and frozen parameters. In the previous paper \cite{zhang2024rpnreconciledpolynomialnetwork}, we implemented this function with a 2-layered MLP model of dimensions $(l, d, n \cdot D)$, {\ie}

\begin{equation}
\text{Hypernet}(\mb{w}) = \sigma(\mb{w} \mb{H}_1) \mb{H}_2 \in \mathbbm{R}^{n \times D},
\end{equation}

where $\mb{H}_1 \in \mathbbm{R}^{l \times d}$ and $\mb{H}_2 \in \mathbbm{R}^{d \times (n \cdot D)}$ are the randomly initialized frozen parameters of the MLP. Notation $d$ denotes the middle hidden layer dimension and $\sigma(\cdot)$ denotes the sigmoid function. By default, we have the middle dimension $l < d < n \cdot D$.

In experimental testing, we encountered implementation challenges with this reconciliation function. For some data expansion functions, the expansion space $D$ can be very large, and the hypernet initialization may consume space of $\mc{O}\left(d \cdot (l + n \cdot D) \right)$.

In this paper, based on the aforementioned random matrix adaptation techniques, we propose replacing the two large-sized frozen parameter matrices $\mb{H}_1$ and $\mb{H}_2$ with their low-rank representations:

\begin{equation}
\begin{aligned}
\text{Hypernet}(\mb{w}) &= \sigma \left( \mb{w} (\mb{P} \mb{Q}^\top) \right) \left( \mb{S} \mb{T}^\top \right)\\
&= \left( \sigma \left( (\mb{w} \mb{P}) \mb{Q}^\top \right) \mb{S} \right) \mb{T}^\top \in \mathbbm{R}^{n \times D},
\end{aligned}
\end{equation}

where $\mb{P} \in \mathbbm{R}^{l \times r}$, $\mb{Q} \in \mathbbm{R}^{d \times r}$, $\mb{S} \in \mathbbm{R}^{d \times r}$ and $\mb{T} \in \mathbbm{R}^{(n \times D) \times r}$ are the low-rank random and frozen sub-matrices that can compose the matrices $\mb{H}_1$ and $\mb{H}_2$ of the hypernet. Moreover, by leveraging the associative law of matrix multiplication, we can avoid explicitly calculating and storing $\mb{H}_1$ and $\mb{H}_2$ as indicated by the above equation. These low-rank random matrix representations reduce the space consumption of this function to $\mc{O}\left(r \cdot (l + 2d + n \cdot D)\right)$.

\section{Unifying Existing Backbones with {\our}}\label{sec:backbone_unification}

The incorporation of new interdependence functions and compression functions significantly enhances {\our}'s capabilities, offering both improved modeling power and increased learning efficiency when dealing with complex data characterized by diverse underlying interdependence relationships. By strategically selecting these component functions based on the specific input data, we can construct highly versatile model architectures using {\our}. Furthermore, {\our} provides a unified framework capable of representing the many influential contemporary backbone architectures, including but not limited to Convolutional Neural Networks (CNNs), Recurrent Neural Networks (RNNs), Graph Neural Networks (GNNs), and Transformers.

\noindent \textbf{Motivations of Backbone Unification}: While unifying existing backbone models is not the primary focus of this paper, we propose this unification with several key goals. First, we aim to uncover the foundational architectural similarities and highlight the critical differences among current backbone models (to be discussed in this section). Second, a unified architectural representation enables the theoretical analyses and comparisons of learning performance for these existing structures (to be discussed in the following Section~\ref{sec:interpretation}). Finally, and most importantly, this unification allows us to pinpoint the core weaknesses within these models, opening up pathways for potential enhancements or the creation of novel ``Transformer-Next'' architectures—an exploration we will pursue in our future papers.

The illustrations of the unified representation of CNN, RNN, GNN and Transformer with {\our} are also provided in the following Figure~\ref{fig:backbone_representation_1} and Figure \ref{fig:backbone_representation_2}, respectively.


\subsection{Unifying CNNs with {\our}}

Convolutional Neural Networks (CNNs) \cite{726791} have long served as the backbone model for image processing tasks. Over time, their application has expanded beyond image data, encompassing a diverse range of modalities. With appropriate model extensions, CNNs have been successfully adapted to process data in other modalities, such as point clouds \cite{Qi2016PointNetDL}, textual data \cite{Kim2014ConvolutionalNN}, time series \cite{10.5555/303568.303704}, and graph structures \cite{Defferrard2016ConvolutionalNN}. In this section, we will introduce the Convolutional Neural Networks (CNNs) initially designed for images with the convolutional and pooling operators, and discuss how to represent CNNs into the unified representation of the {\our} model based on the component functions introduced above.


\subsubsection{Convolutional Neural Network (CNN)}

In this part, we will examine the architecture of CNN model, with particular emphasis on two crucial components in the model, {\ie} the convolutional operator and the pooling operator. These operators play pivotal roles in the CNN's ability to effectively extract features from input data, forming the foundation of the model's success in various tasks, particularly in image processing.

\begin{wrapfigure}{r}{0.55\textwidth}
    \vspace{-20pt}
    \centering
    \includegraphics[width=0.55\textwidth]{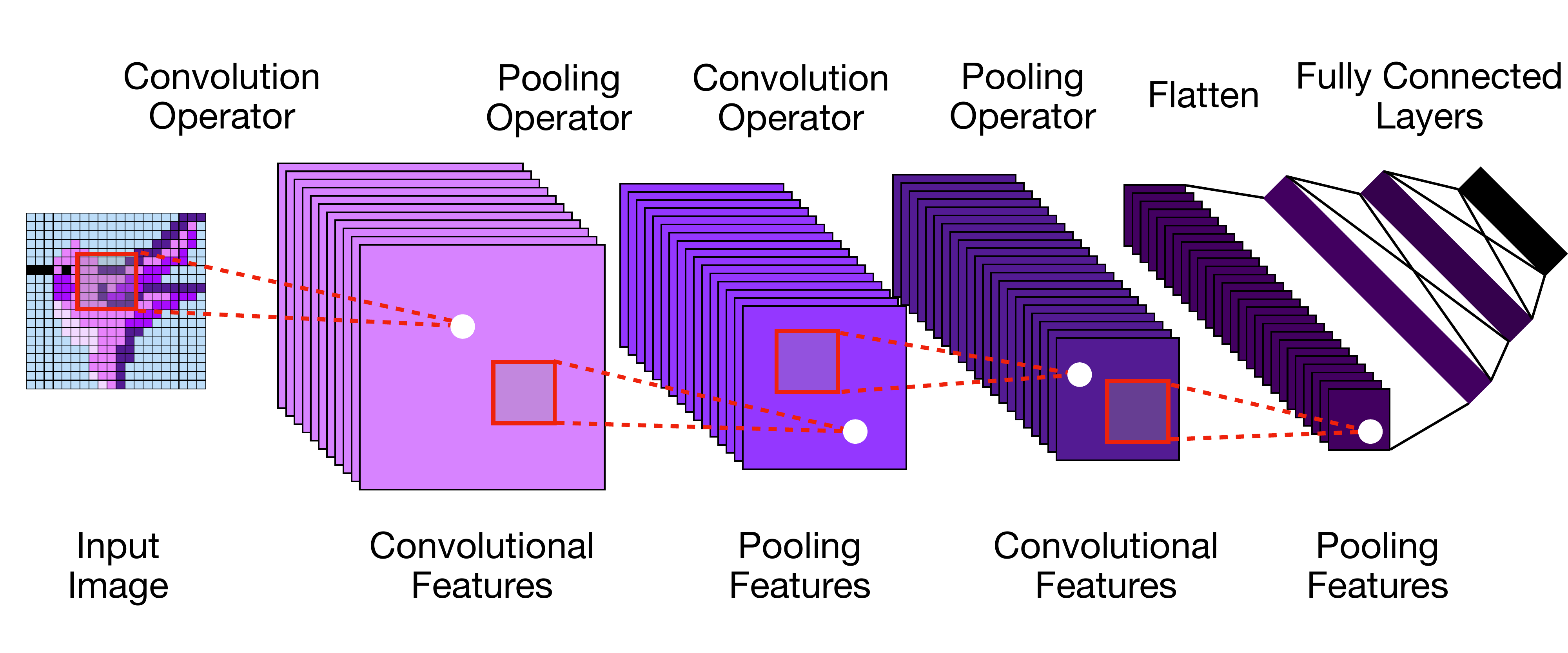}
    \caption{An illustration of CNN model.}
    \label{fig:cnn_architecture}
    \vspace{-10pt}
\end{wrapfigure}


\noindent \textbf{CNN Model Architecture}: The right plot illustrates the architecture of a classic Convolutional Neural Network (CNN) model, comprising convolutional operators, pooling operators, and fully connected layers. Given an input image of a ``hummingbird'', the CNN processes it through a series of layers of different operators. Initially, the convolutional layers shift convolutional kernels ({\ie} parameter matrices or tensors) across the image to compute convolutional feature maps. These features then pass through pooling layers, where pooling kernels extract salient features and compress the maps through operations such as max-pooling, resulting in compressed feature maps. CNNs often employ a deep architecture by stacking multiple convolutional and pooling layers sequentially, allowing for the extraction of increasingly abstract features. After the final convolutional or pooling layer, the learned feature maps are flattened into a dense feature vector. This flattened vector is then processed by fully connected layers, which perform the final classification of the input image into the appropriate label category, in this case, ``Hummingbird''.


\noindent \textbf{Convolutional Operator}: Convolution has been a cornerstone in conventional image processing tasks, employed for tasks such as blurring, sharpening, embossing, and edge detection. While conventional convolution operators often rely on manually defined kernel matrices, Convolutional Neural Networks (CNNs) innovate by learning these kernels as parameters. This approach endows CNNs with superior modeling capacity and flexibility compared to traditional image processing techniques, as the learned kernels can capture diverse useful image feature patterns.

Formally, given an input image $\mb{X} \in \mathbbm{R}^{h \times w \times d}$ with height $h$, width $w$, and depth $d$, the convolutional operator shifts a kernel parameter matrix $\mb{W} \in \mathbbm{R}^{(p_h + p_h'+1) \times (p_w+ p_w' +1) \times (p_d + p_d' +1)}$ of sizes $p_h + p_h'+1$, $p_w+ p_w' +1$, and $p_d + p_d' +1$ along the rows, columns, and depth dimensions. For a sub-image centered at pixel coordinates $(i, j, k)$, we can represent it as $\bar{\mb{X}} = \mb{X}(i-p_h:i+p_h', j-p_w: j + p_w', k-p_d:k+p_d')$, whose convolution with the kernel is calculated as:

\begin{equation}\label{equ:cnn_convolution_formula}
\bar{\mb{X}} \ast \mb{W} = \sum_{r=0}^{p_h+p_h'} \sum_{s=0}^{p_w+p_w'} \sum_{t=0}^{p_d+p_d'} \bar{\mb{X}}(r, s, t) \cdot \mb{W}(p_h+p_h'-r, p_w+p_w'-s, p_d+p_d'-t),
\end{equation}

where $\ast$ denotes the convolutional operator. In practice, many algorithms and toolkits substitute this convolutional operator with the cross-correlation operator instead:

\begin{equation}\label{equ:cnn_cross_correlation_formula}
\bar{\mb{X}} \circ \mb{W} = \sum_{r=0}^{p_h+p_h'} \sum_{s=0}^{p_w+p_w'} \sum_{t=0}^{p_d+p_d'}  \bar{\mb{X}}(r, s, t) \cdot \mb{W}(r, s, t).
\end{equation}

The convolutional operator $\bar{\mb{X}} \ast \mb{W}$ and the cross-correlation operator $\bar{\mb{X}} \circ \mb{W}$ are mathematically different but equivalent in practical model learning, with the distinction that the kernel matrix $\mb{W}$ is flipped in all dimensions for convolution. In practice, this substitution of convolution with cross-correlation does not impact the model's learning performance, as $\mb{W}$ is a learned parameter. Moreover, using cross-correlation can enhance computational efficiency on backend hardware without the redundant costs for the tensor reshaping and flipping at the memory.

Beyond the kernel sizes (or the patch shape and size in {\our}), CNNs incorporate additional hyper-parameters that significantly influence their behavior and learning performance. A key example is the \textit{stride}, which defines the steps for shifting the kernel across the input. The choice of stride directly affects the spatial dimensions of the resulting feature maps, thereby impacting the model's receptive field and the level of detail preserved in the output. The stride parameters used in CNNs can be precisely determined by the patch packing strategies in {\our} as discussed in the previous Sections~\ref{subsubsec:cuboid_patch}-\ref{subsubsec:sphere_patch}.


\noindent \textbf{Pooling Operator}: The pooling operator, devoid of learnable parameters, serves to compress the convolution feature map by extracting salient features. Various pooling methods can be employed based on the learning context, including max-pooling, mean-pooling, and min-pooling, which extract the maximum, average, and minimum values from feature map regions, respectively.

\begin{wrapfigure}{r}{0.47\textwidth}
    \vspace{-15pt}
    \centering
    \includegraphics[width=0.42\textwidth]{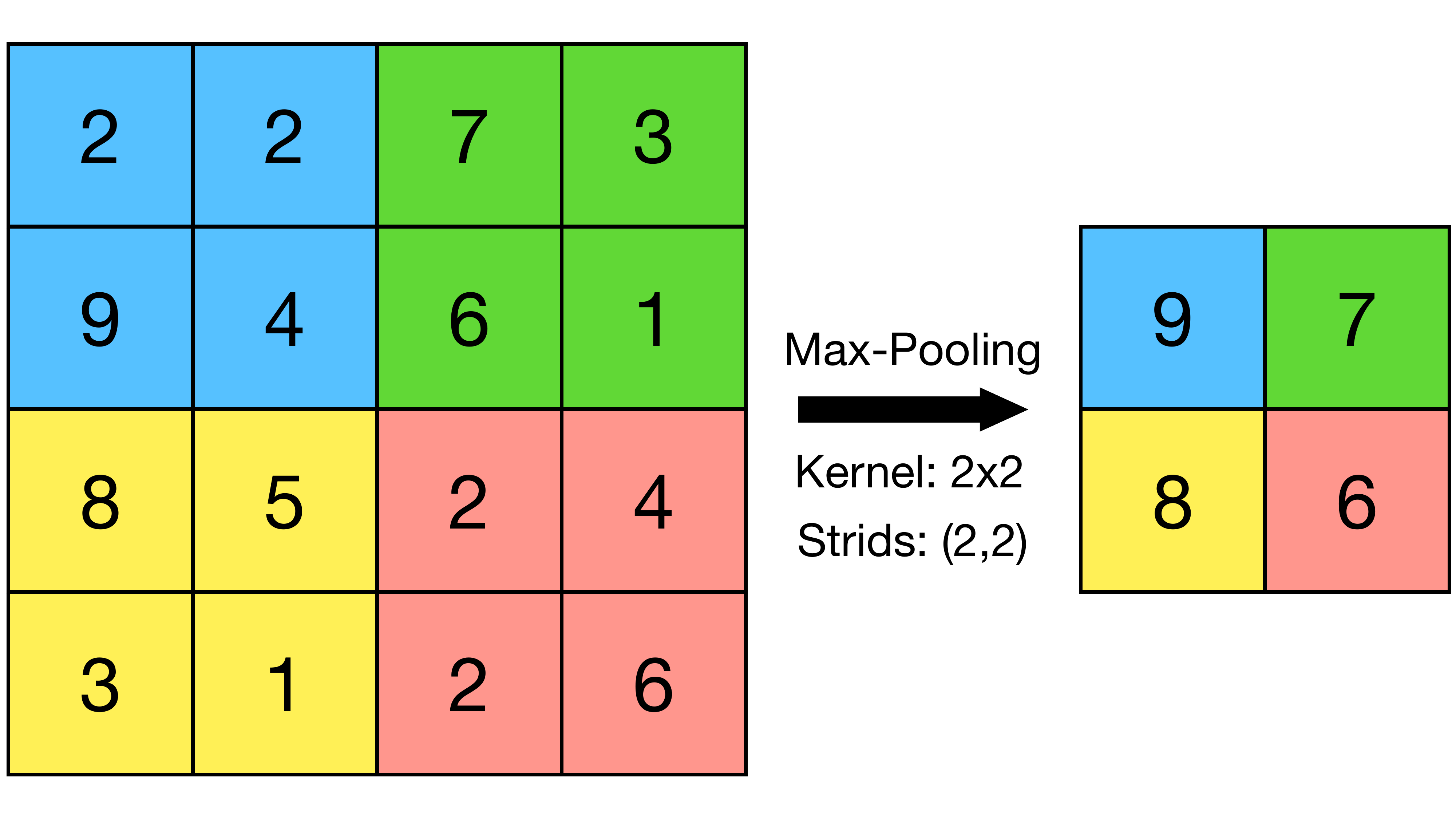}
    \vspace{-5pt}
    \caption{An illustration of pooling operator.}
    \label{fig:cnn_pooling}
    \vspace{-10pt}
\end{wrapfigure}

Similar to the convolutional operator, the pooling operator utilizes kernels of dimensions $p_h$, $p_w$, and $p_d$ (and also $p_h'$, $p_w'$, $p_d'$), which traverse the convolutional feature along its row, column, and depth dimensions with specified stride lengths. However, unlike convolution kernels, pooling kernels contain no learnable parameters and merely delineate the input regions for pooling operations. As depicted in Figure~\ref{fig:cnn_pooling}, a $4 \times 4$ feature map subjected to a $2 \times 2$ max-pooling kernel with stride $(2, 2)$ yields a condensed $2 \times 2$ feature map, where each value represents the maximum from the corresponding input region.


\begin{figure*}[t]
    \begin{minipage}{\textwidth}
    \centering
    	\includegraphics[width=\linewidth]{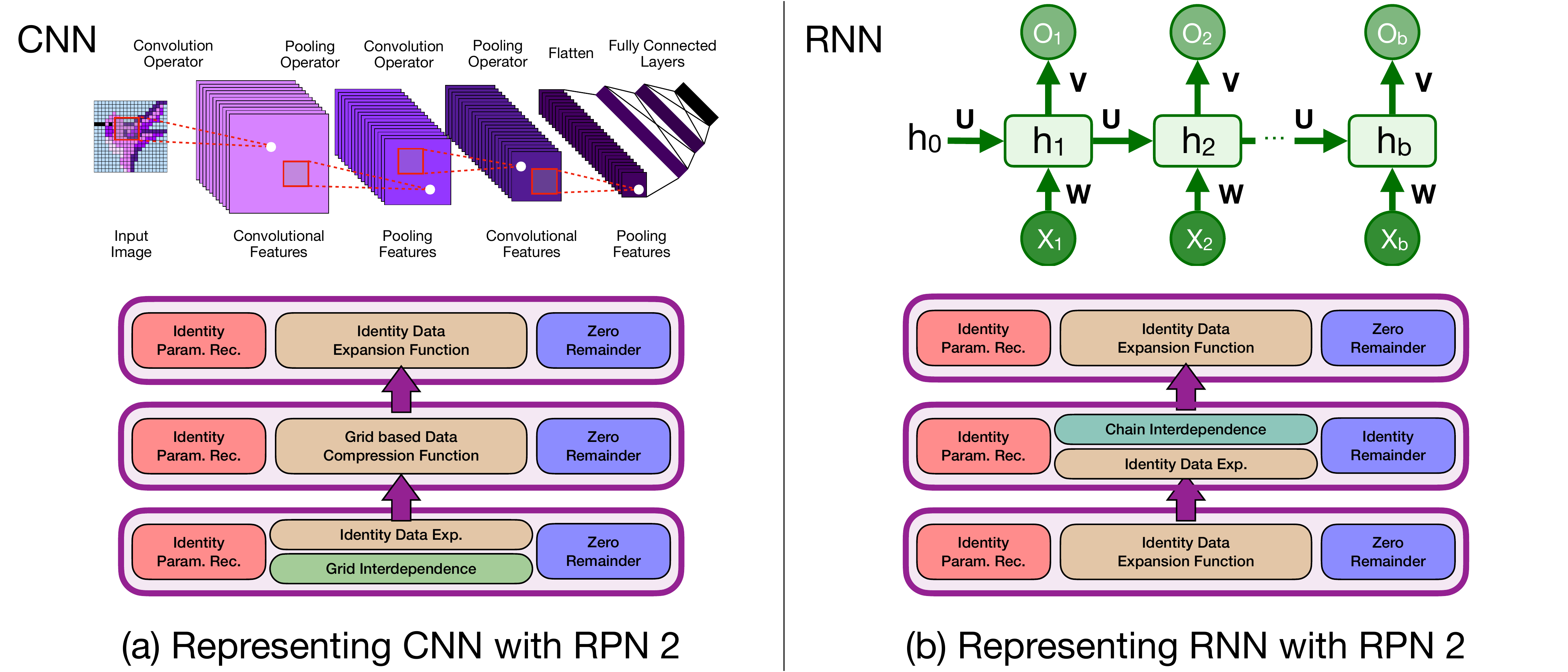}
    	\caption{An illustration of representing CNN and RNN with {\our}.}
    	\label{fig:backbone_representation_1}
    \end{minipage}%
\end{figure*}

\subsubsection{Representing CNN with {\our}}

Before formalizing the representation of CNN architecture within {\our}, we will first discuss how to represent the convolutional and pooling operators using the component functions introduced in previous sections.


\noindent \textbf{Representing Convolution with Grid Structural Interdependence Function}: The convolutional operator in CNNs utilizes learnable kernel parameters to extract information from input images. These kernels shift along the dimensions of the input image tensor, operating on sub-image patches of equal size to the kernels. The feature extraction process can be represented as the inner product of flattened sub-image patches with the kernel parameter vector, where identical kernel parameters are shared across all sub-image patches.

Formally, for an input image $\mb{X} \in \mathbbm{R}^{h \times w \times d}$ with height $h$, width $w$, and depth $d$, we can represent its flattened form as vector $\mb{x} = \text{reshape}(\mb{X}) \in \mathbbm{R}^{(h \times w \times d)}$. The original image modality specific topological structure can be represented as $grid(\mb{x} | h, w, d)$, as described in Section~\ref{subsubsec:data_grid_patch_structure}. The sub-image matrix $\bar{\mb{X}}$ from Equation~(\ref{equ:cnn_convolution_formula}) actually corresponds to a cuboid patch $patch(i, j, k)$ centered at coordinate $(i, j, k)$ in the grid, with patch shape hyper-parameters $p_h, p_h', p_w, p_w', p_d,$ and $p_d'$ defined in Section~\ref{subsubsec:data_grid_patch_structure}.

The cross-correlation (or convolution) based feature extraction defined in above Equation~(\ref{equ:cnn_cross_correlation_formula}) can be equivalently represented as:

\begin{equation}
\bar{\mb{X}} \circ \mb{W} = \left \langle \mb{p}, \mb{w} \right \rangle = \sum_{i=0}^{p-1} \mb{p}(i) \cdot \mb{w}(i),
\end{equation}

where $\mb{p} = \mb{x}(patch(i, j, k)) = \text{reshape}(\bar{\mb{X}}) \in \mathbbm{R}^p$ is the flattened patch vector representation of the sub-image, with $p = (p_h + p_h'+1) \times (p_w+ p_w' +1) \times (p_d + p_d' +1)$ denoting the patch size. Notation $\mb{w} = \text{reshape}(\mb{X})$ denotes the flattened kernel parameter vector of equal length $p$.

The feature extraction for the entire input image can be calculated concurrently using patch packing strategies discussed in the previous Sections~\ref{subsubsec:cuboid_patch}-\ref{subsubsec:sphere_patch}. Using the structural interdependence function from Section~\ref{subsubsec:patch_structural_interdependence_function} with patch center distance hyper-parameters $d_h, d_w, d_d$, we can represent the extracted feature map as:

\begin{equation}
\left \langle \kappa_{\xi}(\mb{x}), \psi(\mb{w}) \right \rangle = \left \langle \kappa \left(\mb{x} \xi(\mb{x}) \right), \psi(\mb{w}) \right \rangle = \left \langle \mb{x} \mb{A}, \mb{c} \otimes \mb{w} \right \rangle,
\end{equation}

where 

\begin{itemize}
\item $\xi(\mb{x}) = \mb{A} \in \mathbbm{R}^{m \times m'}$: This denotes the structural interdependence function with the padding mode.
\item $\kappa(\mb{x}) = \mb{x} \in \mathbbm{R}^m$: This denotes the identity data expansion function.
\item $\psi(\mb{w}) = \mb{I} \otimes \mb{w} = \mb{W} \in \mathbbm{R}^{(p \times p_{count}) \times p_{count}}$: This represents the duplicated padding-based parameter reconciliation function with diagonal block matrix outputs.
\end{itemize}

The structural interdependence function has been introduced in the previous Equation~(\ref{equ:patch_structural_interdependence_function}), while the identity data expansion and duplicated padding-based parameter reconciliation function were introduced in the previous paper \cite{zhang2024rpnreconciledpolynomialnetwork}. Specificall, the matrix dimension term $m'$ is defined as $m' = p \times p_{count}$, as indicated in Equation~(\ref{equ:patch_packing_output_dimension}), where $p_{count}$ represents the number of patches in the grid, as defined in the previous Equation~(\ref{equ:patch_packing_count}). The term $\mb{I}$ is a constant identity matrix defined as $\mb{I} = diag(\left[1, 1, \cdots, 1 \right]) \in \mathbbm{R}^{p_{count} \times p_{count}}$, and $\mb{I} \otimes \mb{w}$ creates $p_{count}$ duplicated paddings of parameter vector $\mb{w}$ along the diagonal, which can be extremely sparse.

In practice, an alternative, more flexible, and efficient representation of the convolution operator with {\our} involves applying a ``\textit{reshape}($\cdot$)'' post-processing operation to the data expansion function. This operation reshapes each expanded instance vector $\kappa \left(\mb{x} \xi(\mb{x}) \right) = \mb{x} \mb{A}$, of length $m'$, into a two-dimensional matrix of shape $p_{count} \times p$. Under this representation, the parameter reconciliation function can be simplified to an identity function, {\ie} $\psi(\mb{w}) = \mb{w} \in \mathbbm{R}^{p}$, where the inner product with the expanded data matrix is subsequently processed by another ``\textit{reshape}($\cdot$)'' output-processing operator, converting the result back into vectors. This method significantly reduces storage and computational costs compared to the previously discussed representation.


\noindent \textbf{Representing Pooling with Compression Function}: The pooling operator in CNNs can be precisely represented using the geometric patch-based compression functions introduced in Section~\ref{subsec:patch_compression_function}. In CNNs, these geometric patches typically have a cuboid shape, with pooling strides corresponding to the patch center distance hyper-parameters. Different pooling approaches can be implemented using various patch compression mappings defined in the previous Section~\ref{subsec:patch_compression_function}.


\noindent \textbf{Representing CNN with {\our}}: Based on the above analyses, as illustrated in the Plot (a) of Figure~\ref{fig:backbone_representation_1}, we can represent the CNN model within {\our} using the following layers. The representation of the fully connected layers ({\ie} MLP) has been introduced in the previous paper \cite{zhang2024rpnreconciledpolynomialnetwork} already, which will not repeated here.

\begin{itemize}

\item \textbf{Convolutional Layer}: Represented as a single-head layer in {\our} with:
(1) identity data expansion function;
(2) cuboid patch-based structural interdependence function (in padding mode); 
(3) identity parameter reconciliation function; 
(4) zero remainder function;
and (5) reshape functions for both expansion post-processing and output-processing.
For multi-channel CNNs, corresponding channel numbers can be used to define parameters in {\our}. For models with skip-layer residual connections ({\eg} ResNet \cite{He2015DeepRL}), a linear remainder function can be used instead of the zero remainder function.

\item \textbf{Pooling Layer}: Represented as an {\our} layer with:
(1) cuboid patch-based data compression function (using numerical operator-based patch compression mappings);
(2) identity interdependence function;
(3) constant parameter reconciliation function;
and (4) zero remainder function.

\end{itemize}

An example of the CNN's representation using {\our} is shown in Plot (a) of Figure~\ref{fig:backbone_representation_1}. This configuration includes one convolutional layer, one pooling layer, and one feed-forward layer. A deeper CNN architecture can be similarly represented by adding multiple layers to this structure.


\subsection{Unifying RNN with {\our}}

Recurrent neural networks (RNNs) denote a family of deep models that capture the internal transitional states of data sequences, which have been extensively used for the modeling of language \cite{Zaremba2014RecurrentNN}, time series \cite{Che2016RecurrentNN}, and video sequences \cite{Venugopalan2015SequenceTS}. In this section, we will investigate to unify RNN with {\our}'s representation based on the component functions introduced in the previous sections.


\subsubsection{Recurrent Neural Network (RNN)}

In this part, we will examine the architecture of the RNN model, with a particular focus on the crucial recurrent state updating operator. This operator forms the cornerstone of RNNs, enabling them to process sequential data by maintaining and updating internal states across time steps.

\begin{minipage}{\linewidth}
\begin{wrapfigure}{r}{0.42\textwidth}
    \vspace{-15pt}
    \centering
    \includegraphics[width=0.42\textwidth]{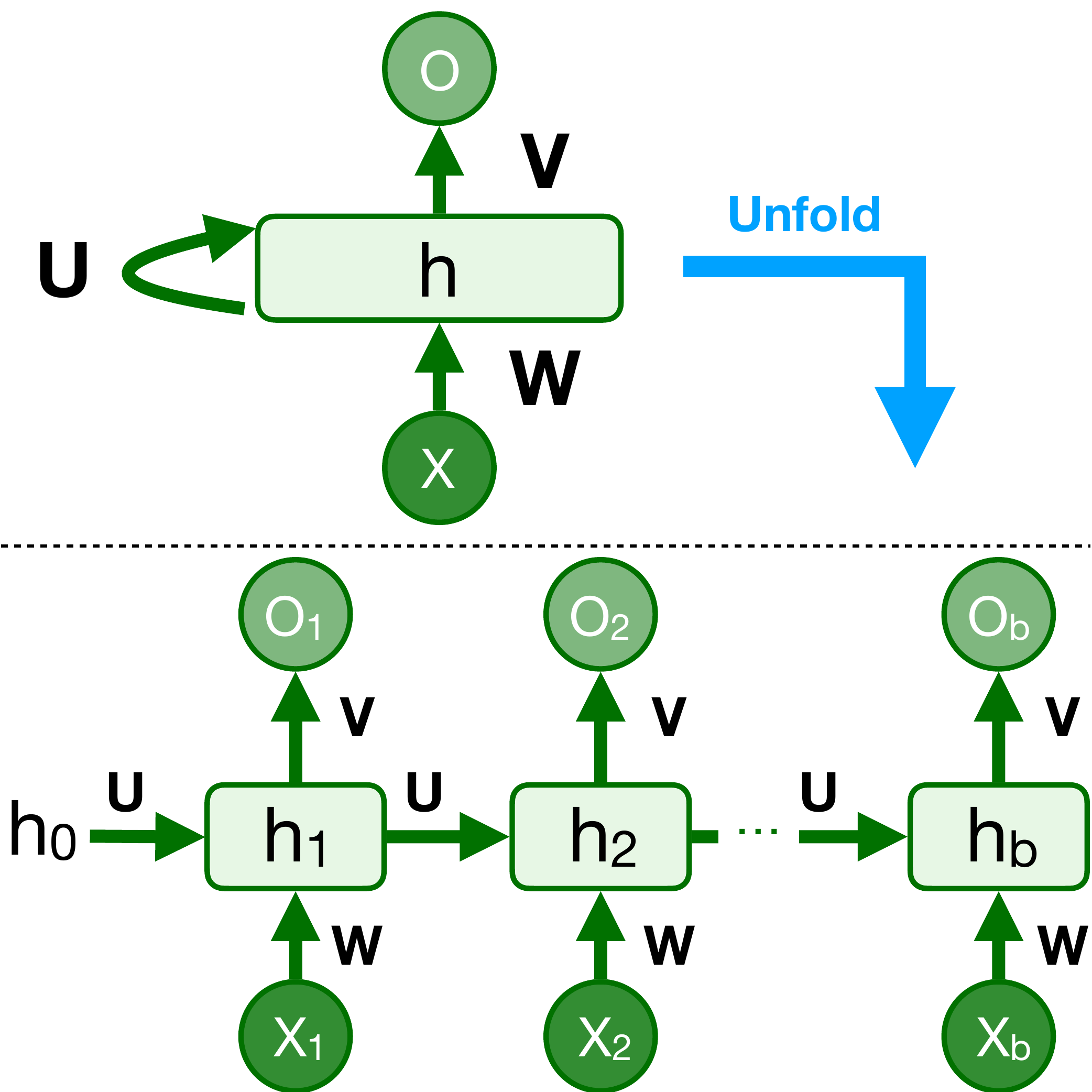}
    \caption{An illustration of RNN model.}
    \label{fig:rnn_architecture}
    \vspace{-10pt}
\end{wrapfigure}


\noindent \textbf{RNN Model Architecture}: As depicted in the right plot, the Recurrent Neural Network (RNN) model can be represented as a multi-layer architecture with full connections. In this structure, $\mb{x}_i \in \mathbbm{R}^{m}$ with different subscripts represent the sequence inputs, $\mb{h}_i \in \mathbbm{R}^{d_h}$ denotes the learned embeddings of the inputs, and $\textbf{o}_i \in \mathbbm{R}^{n}$ represents the corresponding outputs. The model's parameters are defined by three key matrices: $\mb{W} \in \mathbbm{R}^{m \times d_h}$, which represents the parameters of the fully connected layer between input and hidden layers; $\textbf{V} \in \mathbbm{R}^{d_h \times n}$, which denotes the parameters of the fully connected layer between hidden and output layers; and $\mb{U} \in \mathbbm{R}^{d_h \times d_h}$, which is the state transitional parameter unique to RNNs, allowing embedding state vectors to transition along the sequence. This recurrent structure, particularly the state transition facilitated by parameter $\mb{U}$, distinguishes RNNs from traditional feedforward models, enabling them to capture sequential dependencies in the data. By incorporating this feedback loop, RNNs can maintain and utilize information from previous time steps, making them particularly well-suited for processing sequential data such as time series or natural language.

\end{minipage}


\noindent \textbf{Recurrent State Updating Operator}: The RNN model architecture depicted in Figure~\ref{fig:rnn_architecture} can be unfolded into a sequential structure. In this unfolded representation, $\mb{x}_1, \mb{x}_2, \cdots, \mb{x}_b \in \mathbbm{R}^m$ denote the sequence inputs, while $\mb{h}_1, \mb{h}_2, \cdots, \mb{h}_b \in \mathbbm{R}^{d_h}$ represent the learned hidden state vectors connected via fully connected layers. These hidden state vectors are then projected to the corresponding outputs $\mb{o}_1, \mb{o}_2, \cdots, \mb{o}_b \in \mathbbm{R}^n$. Notably, the parameters $\mb{W} \in \mathbbm{R}^{m \times d_h}$, $\mb{U} \in \mathbbm{R}^{d_h \times d_h}$, and $\mb{V} \in \mathbbm{R}^{d_h \times n}$ are shared across all time steps, maintaining consistency in the network's behavior throughout the sequence.

Formally, for the $i_{th}$ input vector $\mb{x}_i$, we can express its learned embedding vector $\mb{h}_i$ based on input $\mb{x}_i$ as follows:

\begin{equation}\label{equ:rnn_input_state}
\mb{h}_i = \mb{x}_i \mb{W} \in \mathbbm{R}^{d_h}, \forall i \in \{1, 2, \cdots, b\}.
\end{equation}

The primary distinction between Recurrent Neural Networks (RNNs) and other models discussed in both our previous work \cite{zhang2024rpnreconciledpolynomialnetwork} and the current paper lies in the sequential dependence among inputs. In RNNs, the state of a later input instance is not solely dependent on itself, but also on the states of preceding inputs. This characteristic can be formally represented as follows:

\begin{equation}\label{equ:rnn_hidden_state}
\mb{h}_i' = \sigma \left( \mb{h}_{i-1} \mb{U} + \mb{h}_i \right), \forall i \in \{1, 2, \cdots, b\}.
\end{equation}

We use the notations with the prime symbols, {\ie} $\mb{h}_i'$, to represent the input state vectors updated with their dependent conditions. For the initial input of the sequence, the embedding input vector $\mb{h}_0$ is typically assigned a dummy vector, such as a zero vector or a vector with random values. Subsequently, based on the learned embedding vector $\mb{h}_i$, the corresponding output vector $\mb{o}_i$ (for $\forall i \in {1, 2, \cdots, b}$) is computed as:

\begin{equation}\label{equ:rnn_output_state}
\mb{o}_i = \text{softmax} \left( \mb{h}_i' \mb{V} \right) \in \mathbbm{R}^{n}.
\end{equation}

This formulation elegantly captures the essence of RNNs: their ability to process sequential data by maintaining a state that gets updated at each time step, allowing the network to retain information from previous inputs.


\subsubsection{Representing RNN with {\our}}

In this part, we will explore how to represent the RNN model within the {\our} framework. Our primary focus will be on the hidden state recurrent updating process, as described in Equation~(\ref{equ:rnn_hidden_state}). We will investigate how to represent this crucial step using the component functions introduced in previous sections, thereby demonstrating how the unique sequential nature of RNNs can be captured within our unified framework.

\noindent \textbf{Representing Recurrent Operator with Chain Structural Interdependence Function}:  
In the recurrent state update equation ({\ie} Equation~(\ref{equ:rnn_hidden_state})), updating each state vector $\mb{h}_i'$ (for $\forall i \in {1, 2, \cdots, b}$) requires both $\mb{h}_{i-1}$ and $\mb{h}_i$ as inputs. The state vectors across the data batch can be arranged into a matrix $\mb{H} = [\mb{h}_1, \mb{h}_2, \cdots, \mb{h}_b] \in \mathbbm{R}^{b \times m}$, enabling concurrent updates as follows:

\begin{equation}
\mb{H}' = \left\langle \xi_i(\mb{H})^\top \kappa(\mb{H}), \psi(\mb{w}) \right \rangle + \pi(\mb{h}_i) = \sigma(\mb{A}_{\xi_i}^\top \mb{H} \mb{U} + \mb{H}),
\end{equation}

where

\begin{itemize}
\item $\kappa(\mb{H}) = \mb{H}$: This denotes the identity data transformation function.
\item $\xi_i(\mb{H}) = \mb{A}_{\xi_i} = \begin{bmatrix}
0 & 1 & 0 & \cdots & 0 \\
0 & 0 & 1 & \cdots & 0 \\
0 & 0 & 0 & \cdots & 0 \\
\vdots & \vdots & \vdots & \ddots & \vdots \\
0 & 0 & 0 & \cdots & 1 \\
0 & 0 & 0 & \cdots & 0 \\
\end{bmatrix}$: This denotes the uni-directional chain structural interdependence function.
\item $\psi(\mb{w}) = \mb{U} \in \mathbbm{R}^{d_h \times d_h}$: This denotes the identity parameter reconciliation function that reshapes the input parameter vector into a matrix.
\item $\pi(\mb{H})=\mb{H}$: This denotes the linear remainder function (without dimension adjustment), and can be viewed as an identity function.
\end{itemize}

The output is typically processed through an activation function, such as $\sigma(\cdot)$ shown above. In this representation, the interdependence matrix models the uni-directional dependencies. For a bi-directional RNN, the interdependence matrix will also include ones in the lower off-diagonal entries, capturing the reverse directional relationships.

\noindent \textbf{Representing RNN with {\our}}: Based on this analysis, as illustrated in Plot (b) of Figure~\ref{fig:backbone_representation_1}, we propose the following representation of the RNN model using {\our}. For an input batch with $b$ sequential instances, {\our} defines the chain structural interdependence matrix of dimensions $d \times d$ to model their interdependence relationships:

\begin{itemize}
\item \textbf{Recurrent Layer}: Represented as a single-head and single-channel layer in {\our} with: (1) identity data expansion function; (2) chain structural interdependence function; (3) identity parameter reconciliation function; and (4) identity remainder function. The output of each layer is processed with an activation function.
\end{itemize} 

The unified representation of RNN with {\our} is also illustrated in the Plot (b) of Figure~\ref{fig:backbone_representation_1}, which involves one recurrent state updating layer. For the RNN with multi-layers, we can stack the above recurrent layers on top of each with perceptron layers inserted between them. Also both the input and output processing layers can be represented by the perceptron layer, which has been introduced in the previous paper \cite{zhang2024rpnreconciledpolynomialnetwork} and will not be detailed again here.

\subsection{Unifying GNN with {\our}}

\begin{figure*}[t]
    \begin{minipage}{\textwidth}
    \centering
    	\includegraphics[width=\linewidth]{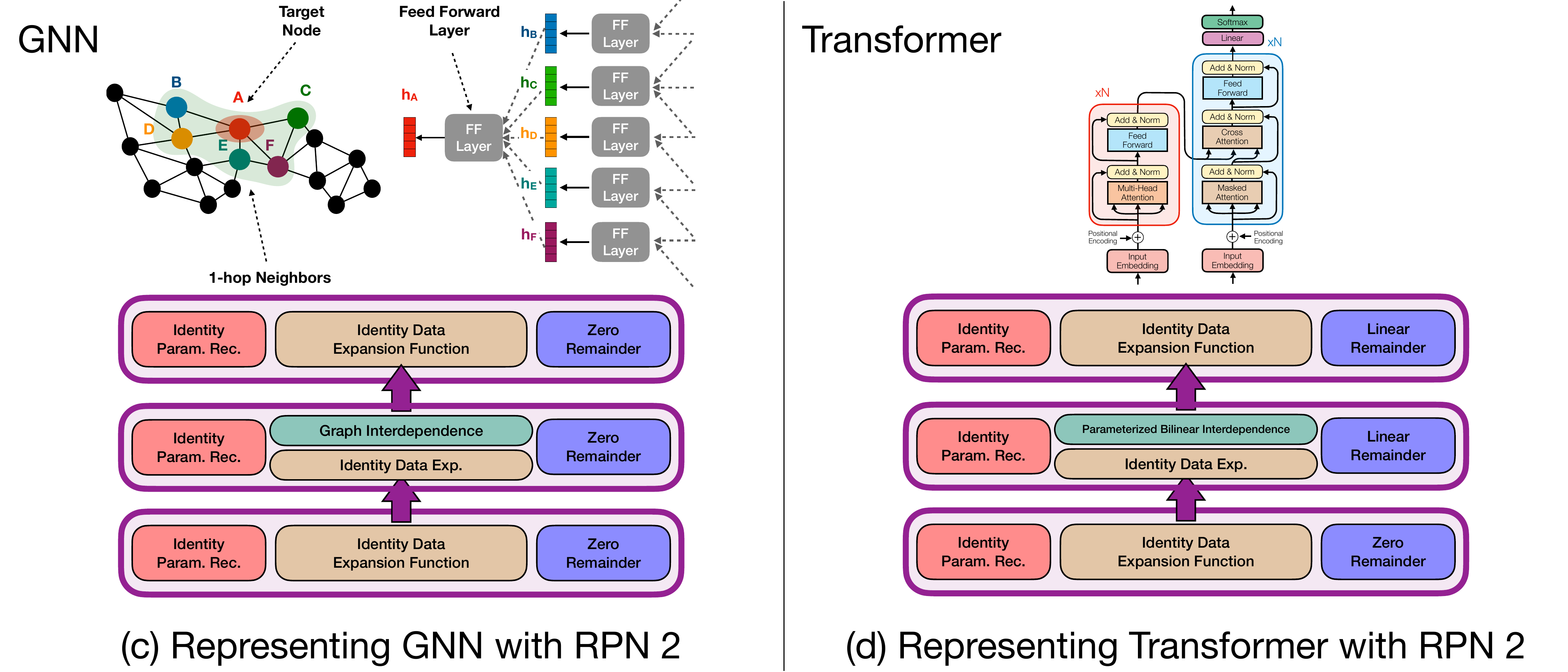}
    	\caption{An illustration of representing GNN and Transformer with {\our}.}
    	\label{fig:backbone_representation_2}
    \end{minipage}%
\end{figure*}

In addition to the image and sequence data discussed previously, graph-structured data are prevalent in the real world, with notable examples including molecular graphs, online social networks, and interconnected websites. To address the unique challenges posed by such data, Graph Neural Networks (GNNs) have emerged as a specialized family of deep learning models designed to handle graph-structured data with extensive connections. In this section, we will introduce GNN models and investigate how to unify them within {\our} canonical representation framework.


\subsubsection{Graph Neural Network (GNN)}

In this part, we will explore the architecture of the GNN models and delve into the spectral graph convolutional operator. This operator plays a crucial role in updating node representations by aggregating information from neighboring nodes within graph-structured data.

\begin{minipage}{\linewidth}
\begin{wrapfigure}{r}{0.54\textwidth}
    \vspace{-15pt}
    \centering
    \includegraphics[width=0.54\textwidth]{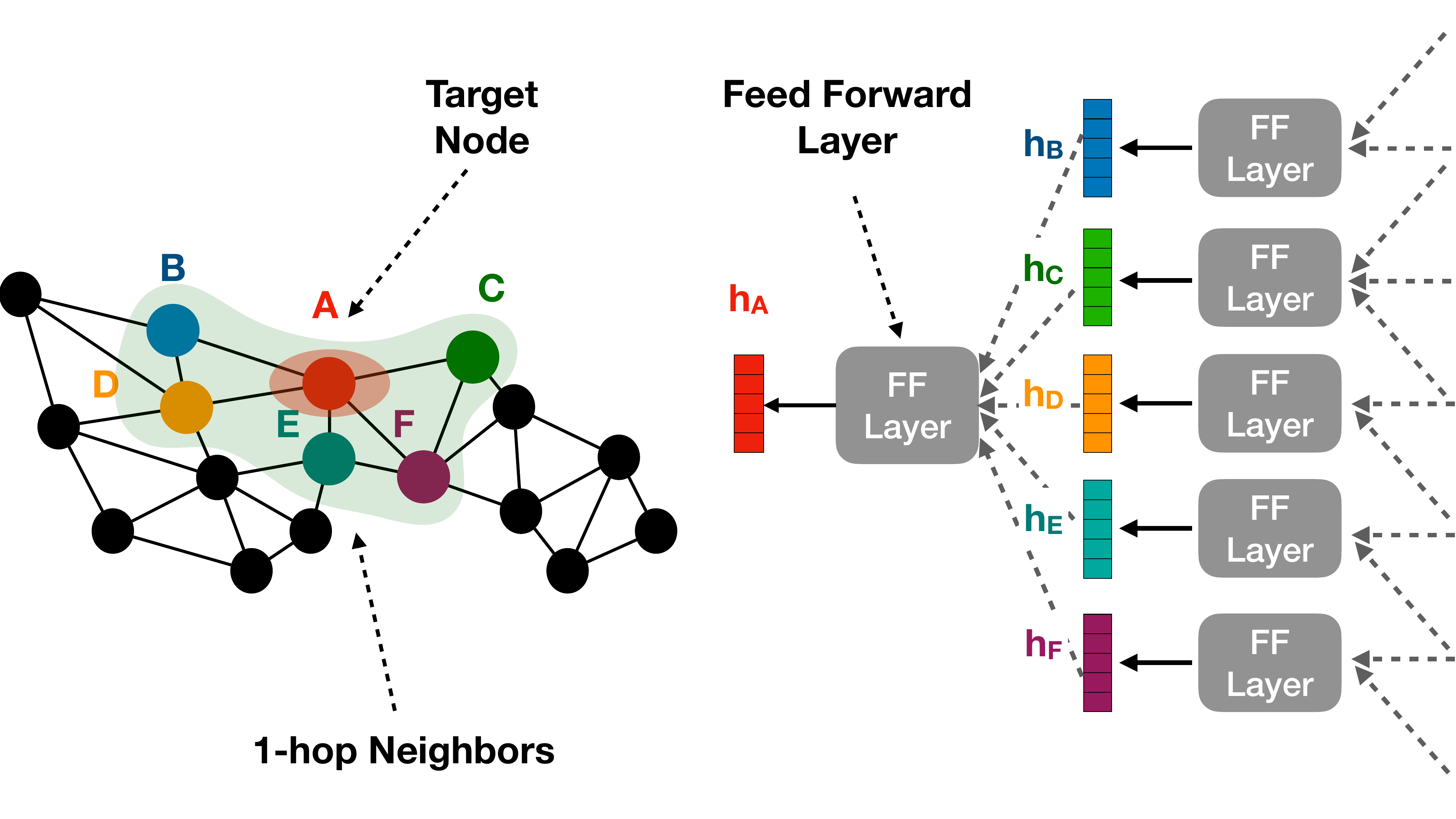}
    \caption{An illustration of GNN model.}
    \label{fig:gnn_architecture}
    \vspace{-10pt}
\end{wrapfigure}


\noindent \textbf{GNN Model Architecture}: Graph Neural Networks (GNNs) can be conceptualized as a generalization of Recurrent Neural Networks (RNNs), where the input structure evolves from a linear chain to an extensively interconnected graph. The right plot illustrates this concept with an example input graph, showcasing multiple nodes and their complex interconnections. Additionally, it depicts a potential deep GNN architecture designed for learning the embeddings of a target node within the graph. The process of learning the embedding vector for a target node in a GNN involves aggregating information from its surrounding neighbors. These neighboring nodes serve as inputs to a fully connected feed-forward layer in the network. Crucially, the embeddings of these neighbor nodes are themselves learned through a similar process, recursively incorporating information from their own neighbors. This recursive nature of information aggregation allows GNNs to capture complex, multi-hop relationships within the graph data.
\end{minipage}


\noindent \textbf{Spectral Graph Convolution (SGC) Operator}: In the aforementioned graph neural network architecture, the neighborhood aggregation operator (illustrated as the fully connected feedforward module in Figure~\ref{fig:gnn_architecture}) is formally known as the spectral graph convolutional operator.

Formally, given a graph $G = (\mc{V}, \mc{E})$ with node set $\mc{V}$ and link set $\mc{E}$, we define $\Gamma(v) = \{u | u \in \mc{V} \land (u, v) \in \mc{E} \}$ as the set of neighbors for a target node $v \in \mc{V}$. Each node $v \in \mc{V}$ is initially represented by an input vector $\mb{x}_v \in \mathbbm{R}^{m}$. The Graph Convolutional Network (GCN) model learns node embeddings by aggregating neighbor information through multiple layers of the spectral graph convolutional (SGC) operator as follows:

\begin{equation}\label{equ:sgc_operator}
\mb{h}_v = \text{SGC}(\mb{x}_v | \Gamma(u)) = \sigma \left( \sum_{u \in \Gamma(v)} \frac{1}{|\Gamma(u)|} \mb{x}_u \mb{W} + \mb{x}_v \mb{W} \right) \in \mathbbm{R}^n,
\end{equation}

where $\mb{W} \in \mathbbm{R}^{m \times n}$ represents the learnable parameters, and $\sigma(\cdot)$ denotes an activation function, such as sigmoid or ReLU.

Notably, the contribution of each neighbor node $u \in \Gamma(v)$ is weighted by the inverse of its degree, $\frac{1}{|\Gamma(u)|}$. This normalization scheme ensures that nodes with numerous connections ({\ie} the neighboring node $u$ with a large node degree $|\Gamma(u)|$) contribute less to the aggregation, preventing highly connected nodes from dominating the information flow. This degree-based weighting mechanism helps balance the influence of nodes with varying connectivity levels in the graph structure.


\subsubsection{Representing GNN with {\our}}

In this part, we will demonstrate how to represent the GNN model architecture using {\our}. Specifically, we will interpret the spectral graph convolutional (SGC) operator in GNNs as a structural interdependence function defined for the graph data modality.

\noindent \textbf{Representing SGC Operator with Graph Structural Interdependence Function}: In GNN models, node representations can be learned concurrently in a data batch, which can significantly reduce computational time compared to individual node representation learning.

Formally, given an input graph $G = (\mc{V}, \mc{E})$ with node set $\mc{V}$, we can organize all node raw features as a data batch $\mb{X} \in \mathbbm{R}^{b \times m}$, where $b = |\mc{V}|$ if the batch contains all nodes in the graph. The spectral graph convolutional (SGC) operator for all node representations can be represented as:

 \begin{equation}
 \mb{H} = \text{SGC}(\mb{X} | G) = \sigma \left( {\mb{A}} \mb{X} \mb{W} \right) \in \mathbbm{R}^{b \times n},
 \end{equation}
 
where ${\mb{A}} \in \mathbbm{R}^{b \times b}$ is the row-normalized adjacency matrix describing connections among nodes within the graph. It is calculated as ${\mb{A}} = \mb{D}^{-1} \hat{\mb{A}} + \mb{I}$, where $\hat{\mb{A}}$ is the graph's adjacency matrix and $\mb{D}$ is the diagonal degree matrix with $\mb{D}(i,i) = \sum_j \hat{\mb{A}}(i,j)$. As noted in Section~\ref{subsubsec:graph_interdependence_function}, matrix ${\mb{A}}$ can be composed with the graph-based structural interdependence function ({\ie} Equation~(\ref{equ:graph_interdependence_function})) with optional row-normalization, which models the interdependence relationships among instances.
 
Therefore, we can rewrite the SGC operator-based graph instance batch updating equation as:

\begin{equation}
\text{SGC}(\mb{X} | G) = \left \langle \kappa_{\xi} (\mb{X}), \psi(\mb{w}) \right \rangle = \left \langle \xi(\mb{X}) \kappa (\mb{X}), \psi(\mb{w}) \right \rangle =  \sigma \left( {\mb{A}} \mb{X} \mb{W}^\top \right).
\end{equation}

where

\begin{itemize}
\item $\xi(\mb{X} | G) = \mb{A} \in \mathbbm{R}^{b \times b}$: This denotes the graph-based structural interdependence function.
\item $\kappa(\mb{X}) = \mb{X} \in \mathbbm{R}^{b \times m}$: This denotes the identity data expansion function.
\item $\psi(\mb{w}) = \text{reshape}(\mb{w}) = \mb{W} \in \mathbbm{R}^{n \times m}$: This represents the identity parameter reconciliation function.
\end{itemize}
 
For graphs with a large number of nodes and links, the normalized adjacency matrix ${\mb{A}}$ can be extremely large. In such cases, as proposed in the Graph-Bert paper \cite{Zhang2020GraphBertOA}, batches of small-sized sub-graphs covering only the neighborhood can be sampled and fed into the model for node embedding vector updating, greatly reducing memory consumption.
 

\noindent \textbf{Representing GNN with {\our}}: Based on the above description, we can represent the GNN model architecture with {\our} by selecting the following component functions to compose the model layers:

\begin{itemize}

\item \textbf{SGC Layer}: The {\our} layer with one single head and one channel, comprising: (1) graph-based structural interdependence function with optional row normalizations as the post-processing function, (2) identity data transformation function, (3) identity parameter reconciliation function, and (4) zero remainder function. The layer may also use an optional activation function for output processing. For models with skip-layer residual connections (similar to Graph-Bert \cite{Zhang2020GraphBertOA}), a linear remainder function can be used instead of the zero remainder function.

\end{itemize}

An example of the GNN's representation using {\our} is shown in Plot (a) of Figure~\ref{fig:backbone_representation_2}, which includes a single graph convolutional layer. To construct deeper GNN architectures, additional SGC-based data batch updating layers can be incorporated by adding corresponding layers to {\our}.


\subsection{Unifying Transformer with {\our}}

Since being processed in 2017, Transformer \cite{Vaswani2017AttentionIA} has been the dominant backbone model used in building many AI models. In recent years, Transformer has demonstrated its effectiveness in processing the inputs in different modalities, including but not limited to images \cite{Dosovitskiy2020AnII}, point cloud \cite{Zhao2020PointT} and graphs \cite{Zhang2020GraphBertOA}. Meanwhile, we have also witnessed some criticisms about Transformer in terms of its extremely high time and space costs, which lead to the current huge demands of both computational facilities and energy consumptions. In this section, we will introduce the detailed components used in Transformer, and investigate to unify Transformer within the canonical representation of {\our}, which may illustrate potential opportunities to address such weakness.


\subsubsection{Transformer}

In this part, we will first delve into the architecture of the Transformer model, with a particular focus on its pivotal component: the scaled dot-product attention mechanism. This mechanism forms the cornerstone of the Transformer's ability to process sequential data effectively.

\begin{minipage}{\linewidth}

\begin{wrapfigure}{r}{0.32\textwidth}
    \centering
    \vspace{-15pt}
    \includegraphics[width=0.3\textwidth]{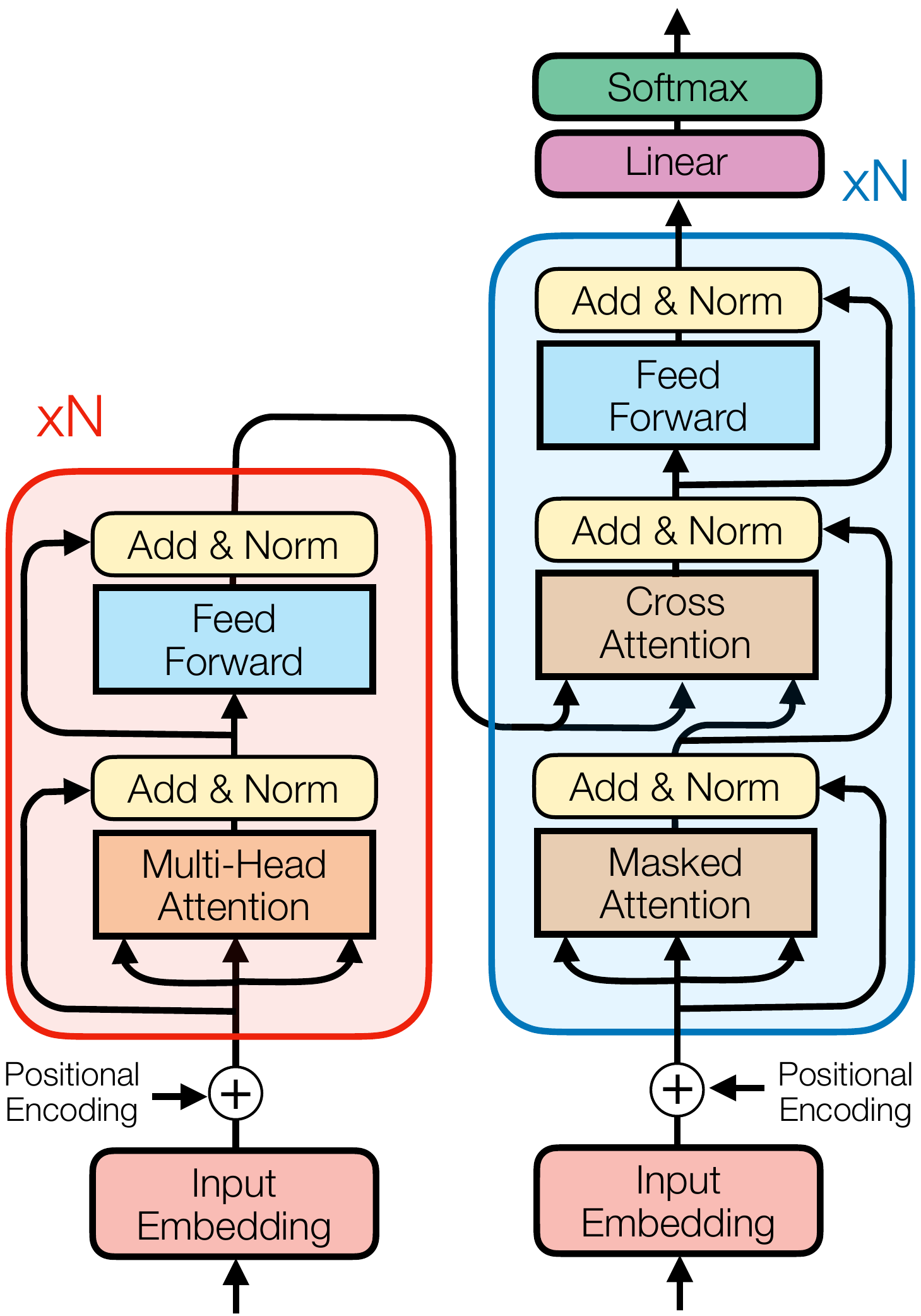}
    \caption{An illustration of the transformer model architecture.}
    \label{fig:transformer_architecture}
    \vspace{-10pt}
\end{wrapfigure}


\noindent \textbf{Transformer Model Architecture}: The right plot illustrates the Transformer model architecture, comprising an encoder (depicted in the left red block) and a decoder (shown in the right blue block). Both the encoder and decoder blocks incorporate several key functional components, including Multi-Head Attention, Feed-Forward layers, and Add \& Norm (normalization) layers. The encoder's output serves as input to the decoder, facilitating the generation of the final output. In the decoder block, the Multi-Head Attention component undergoes slight modifications, resulting in Masked Attention and Cross Attention components. The Masked Attention involves adding masks to the attention mechanism, while the Cross Attention accepts inputs from multiple sources. A crucial feature of both encoder and decoder blocks is their incorporation of positional information through positional encoding. This positional data, combined with the input embedding, is fed into the attention component to enhance the learning process. The decoder's output undergoes further processing through Linear and Softmax layers to produce the final output. This sophisticated architecture enables the Transformer to effectively handle sequential data while capturing long-range dependencies, making it particularly effective for tasks such as machine translation and text generation.

\end{minipage}


\noindent \textbf{Scaled Dot-Product Attention in Transformer}: In our previous {\old} paper \cite{zhang2024rpnreconciledpolynomialnetwork}, we have already demonstrated how to represent the MLP model within the {\old} framework. Building upon that foundation, we will now focus on examining the attention mechanism integral to the Transformer model.

Formally, consider an input data batch $\mb{X} \in \mathbbm{R}^{b \times m}$ containing $b$ instances. The Transformer model processes and embeds this input into latent representations, taking into account the element-wise relationships within the input through its attention mechanism. In the case of single-head attention, we can represent the calculated pairwise attention matrix among the input elements as follows:

\begin{equation}\label{equ:transformer_attention}
\mb{A} = \text{softmax}\left(\frac{\mb{Q} \mb{K}^\top}{\sqrt{r}} \right) \in \mathbbm{R}^{b \times b} \text{, where } \mb{Q} = \mb{X} \mb{W}_q \in \mathbbm{R}^{b \times r} \text{, and } \mb{K} = \mb{X} \mb{W}_k \in \mathbbm{R}^{b \times r}.
\end{equation}

In the attention matrix $\mb{A}$, each element $\mb{A}(i, j)$ represents the calculated attention score between the $i_{th}$ and $j_{th}$ instances of the input batch. The model employs learnable parameter matrices $\mb{W}_q, \mb{W}_k \in \mathbbm{R}^{m \times r}$ to compress the input batch $\mb{X}$ into query and key matrices $\mb{Q}, \mb{K} \in \mathbbm{R}^{b \times r}$, respectively. This transformation allows the model to project the input into a space where relevant similarities can be more easily computed.

Building upon the attention matrix, we can formally express the learned representations of the input batch as follows:

\begin{equation}\label{equation:transformer_updating_equation}
\mb{H} = \text{softmax}\left(\frac{\mb{Q} \mb{K}^\top}{\sqrt{r}} \right) \mb{V} \in \mathbbm{R}^{b \times n}\text{, where } \mb{V} = \mb{X} \mb{W}_v \in \mathbbm{R}^{b \times n}.
\end{equation}

In the above formula, matrix $\mb{W}_v \in \mathbbm{R}^{m \times n}$ denotes the learnable parameters involved in defining the value matrix $\mb{V} \in \mathbbm{R}^{b \times n}$ of the input batch.


\subsubsection{Representing Transformer with {\our}}

In this part, we will explore how to represent the Transformer architecture using {\our} by defining its internal component functions. We will particularly focus on the scaled dot-product attention module, a key component of the Transformer model.


\noindent \textbf{Representing Attention with Parameter Efficient Bilinear Interdependence Function}: We can rewrite the attention matrix from Equation~(\ref{equ:transformer_attention}) by substituting $\mb{Q}$ and $\mb{K}$ with their detailed representations $\mb{X} \mb{W}_q$ and $\mb{X} \mb{W}_k$:

\begin{equation}
\mb{A} = \text{softmax}\left(\frac{\mb{X} \mb{W}_q (\mb{X} \mb{W}_k)^\top}{\sqrt{r}} \right) = \text{softmax}\left(\frac{\mb{X} \mb{W}_q \mb{W}_k^\top \mb{X}^\top }{\sqrt{r}} \right).
\end{equation}

The term $\mb{A} = \mb{X} \mb{W}_q \mb{W}_k^\top \mb{X}^\top$ can be represented as a low-rank parameterized bilinear interdependence function on the input data batch $\mb{X}$, as introduced in Section~\ref{subsubsec:parameterized_bilinear_interdependence_function}. Here, $\mb{W}_q, \mb{W}_k \in \mathbbm{R}^{m \times r}$ denote the low-rank parameter matrices, whose product composes the bilinear parameter matrix $\mb{W} = \mb{W}_q \mb{W}_k^\top \in \mathbbm{R}^{m \times m}$.

To handle the division by $\sqrt{r}$, we introduce a special normalization called scaled-softmax normalization:

\begin{equation}
\text{scaled-softmax}(\mb{A} | r) =  \text{softmax} \left(\frac{\mb{A}}{\sqrt{r}} \right).
\end{equation}

Thus, we can represent the input data batch updating in Transformer, {\ie} Equation~(\ref{equation:transformer_updating_equation}), as:

\begin{equation}
\left \langle \kappa_{\xi}(\mb{X}), \psi(\mb{w}) \right \rangle = \left \langle \xi(\mb{X}) \kappa(\mb{X}), \psi(\mb{w}) \right \rangle = \mb{A} \mb{X} \mb{W}^\top,
\end{equation}

where

\begin{itemize}

\item $\xi(\mb{X})  = \mb{A} = \mb{X} \mb{W}_q \mb{W}_k^\top \mb{X}^\top \in \mathbbm{R}^{b \times b}$: This denotes the low-rank parameterized bilinear interdependence function with scalaed-softmax row normalization.

\item $\kappa(\mb{X}) = \mb{X} \in \mathbbm{R}^{b \times m}$: This denotes the identity data transformation function.

\item $\psi(\mb{w}) = \mb{W} \in \mathbbm{R}^{n \times m}$: This denotes the identity parameter reconciliation function.

\end{itemize}

Residual connections in Transformer can be effectively implemented with remainder functions. Other normalization functions in the module can be implemented with pre-, post-, and output-processing functions in {\our}, as introduced in \cite{zhang2024rpnreconciledpolynomialnetwork}.


\noindent \textbf{Representing Transformer with {\our}}: Based on this analysis, we can represent the Transformer model within {\our} by selecting the following component functions to compose the layers:

\begin{itemize}

\item \textbf{Attention Layer}: The {\our} layer with multi-head, comprising: (1) low-rank parameterized bilinear interdependence function with scalaed-softmax row normalization, (2) identity data transformation function, (3) identity parameter reconciliation function, and (4) linear remainder function. The layer may also use an optional normalization function for output processing.

\end{itemize}

An example of the Transformer's representation using {\our} is shown in Plot (b) of Figure~\ref{fig:backbone_representation_2}, which includes a single transformer block. Additional attention layers in the Transformer can be represented by adding corresponding layers to {\our}. The feed-forward layer in the Transformer is similar to that in an MLP and will not be discussed in detail here.

\section{Empirical Evaluations of {\our}}\label{sec:experiments}

To evaluate the effectiveness of the proposed {\our} model for function learning tasks on different interdependent datasets, this section presents empirical studies conducted on real-world benchmark datasets across various modalities, including images, language, time-series and graphs. The organization of this section is as follows: Section~\ref{subsec:vision_classification} explores the {\our} model's performance on discrete classification tasks involving vision and language benchmark datasets, which exhibit grid- and chain-structured interdependence relationships, respectively. Moving beyond discrete data, Section~\ref{subsec:time_sereies_prediction} focuses on the {\our} model's application to continuous time-series prediction and graph-structured data classification tasks, both characterized by inherent chain- and graph-structured interdependence.

Beyond the main experimental results, we provide a comprehensive analysis of the method’s performance, including learning convergence, parameter sensitivity, ablation studies, interpretability, and visualizations. These investigations offer deeper insights into the method’s advantages and robustness when addressing function learning tasks across datasets with diverse interdependence structures.


\subsection{Discrete Vision and Language Data Classification}\label{subsec:vision_classification}

Equipped with the grid and chain-based structural interdependence functions, the proposed {\our} model can effectively capture local interdependence relationships among image pixels and sequential interdependence relationships among words or tokens in documents. These capabilities are expected to enhance {\our}'s learning performance for vision and language classification tasks. To validate the effectiveness of the proposed {\our} model and its interdependence functions, extensive experiments were conducted on several benchmark image and language datasets.

This subsection is organized as follows. We begin by introducing the datasets and experimental setups. A detailed analysis of the grid-based structural interdependence function is presented, exploring factors such as local patch shapes, sizes, and packing strategies. For the chain-based structural interdependence function, we examine the impact of various chain-based interdependence matrix representations on language classification tasks. In addition to analyzing the interdependence functions, we investigate the influence of other model components and architectural variations on learning performance for vision and language classification. Finally, we report the main results of {\our}, comparing its performance to several baseline methods, including MLP, the earlier version {\old} (without interdependence functions), CNN, and RNN.

\subsubsection{Dataset Description and Experiment Setups}


\begin{table}[t]
\centering
\caption{The table summarizes key statistical information for the discrete image and language datasets. These datasets are pre-partitioned into training and testing sets. For each image and document instance, the table also specifies their input and output sizes.}
\label{tab:vision_dataset_statistics}
\begin{tabular}{|c|c|c|c|c|c|}
\hline
\multirow{2}{*}{} & \multicolumn{2}{c|}{\textbf{Image Datasets}} & \multicolumn{3}{c|}{\textbf{Language Classification Datasets}} \\ \cline{2-6}
                          & \textbf{MNIST} & \textbf{CIFAR-10} & \textbf{IMDB} & \textbf{AGNews} & \textbf{SST2} \\ \hline
Input Image Size                     & 28 $\times$ 28  & 32 $\times$ 32 $\times$ 3 & $512 \times 300$  & $64 \times 300$  & $32 \times 300$ \\ \hline
Output Class \#                     & 10  & 10 & 2  & 4  & 2 \\ \hline
Train Instance \#                     & 60,000  & 50,000 & 25,000  & 120,000  & 67,349 \\ \hline
Test Instance \#                     & 10,000  & 10,000 & 25,000  & 7,600  & 872  \\ \hline
\end{tabular}
\end{table}


\noindent \textbf{Dataset Descriptions}: Following our previous work \cite{zhang2024rpnreconciledpolynomialnetwork}, we evaluate the proposed method on two image benchmark datasets (MNIST and CIFAR-10) and three language datasets (IMDB, AGNews, and SST2). Comparisons are made with MLP, the previous {\our} method, CNN, and RNN. The basic statistical details of these datasets are summarized in Table~\ref{tab:vision_dataset_statistics}.

\begin{itemize}

\item \textbf{Image Datasets}: For the MNIST and CIFAR-10 datasets, the images are flattened into vectors and normalized using mean-std normalization. No image augmentations ({\eg} horizontal/vertical flipping, rotation, or noise addition) were applied in these experiments.

\item \textbf{Language Datasets}: For IMDB, AGNews, and SST2, we utilize the pre-trained GloVe 6B encoder to convert each token into a $300$-dimensional embedding vector. To manage variable-length documents, dataset-specific maximum lengths are employed for truncation or padding: $512$ for IMDB, $64$ for AGNews, and $32$ for SST2.

\end{itemize}

\noindent \textbf{Experiment Setups}: All image and language datasets used in the experiments come with pre-defined training and testing splits. For consistency and fairness, all baseline methods use the same data partitions. Model performance is evaluated using accuracy as the default metric.

\subsubsection{Grid Interdependence Investigation: Patch Shapes, Sizes and Packing Strategies}

In the {\our} model learning framework, images are represented as grid structures, where each pixel is uniquely identified by a coordinate tuple. The model captures local interdependence relationships between pixels through patches of distinct geometric shapes—including cuboids, cylinders, and spheres, which are defined as an ordered set of pixel coordinates in the underlying grid structure. While cuboid and cylindrical patches are particularly effective for image data analysis, spherical patches are better suited for 3D point-cloud data—a topic beyond the scope of this section but slated for future investigation. Here, we explore how the choice of patch shape (cuboid or cylinder), size, and packing strategy impacts the performance of the {\our} model in image classification tasks.

To systematically evaluate these effects, we conduct extensive experiments on the CIFAR-10 dataset using the {\our} model with a consistent architecture comprising three main components: (1) \textit{Attribute Interdependence Layers}: Two layers implementing grid-based structural attribute interdependence functions, each utilizing identity data transformation, identity parameter reconciliation, and zero remainder functions, with both layers maintaining 128 output channels; (2) \textit{Compression Layer}: A single layer employing grid-based data compression function, constant eye parameter reconciliation, and zero remainder functions, outputting 128 channels; and (3) \textit{Perceptron Layers}: Three successive layers utilizing identity data transformation, identity parameter reconciliation, and zero remainder functions, with output dimensions progressively decreasing from 1024 to 512, and finally to 10, respectively.

\begin{figure}[t]
    \centering
    \begin{subfigure}{\textwidth}
        \centering
        \includegraphics[width=\textwidth]{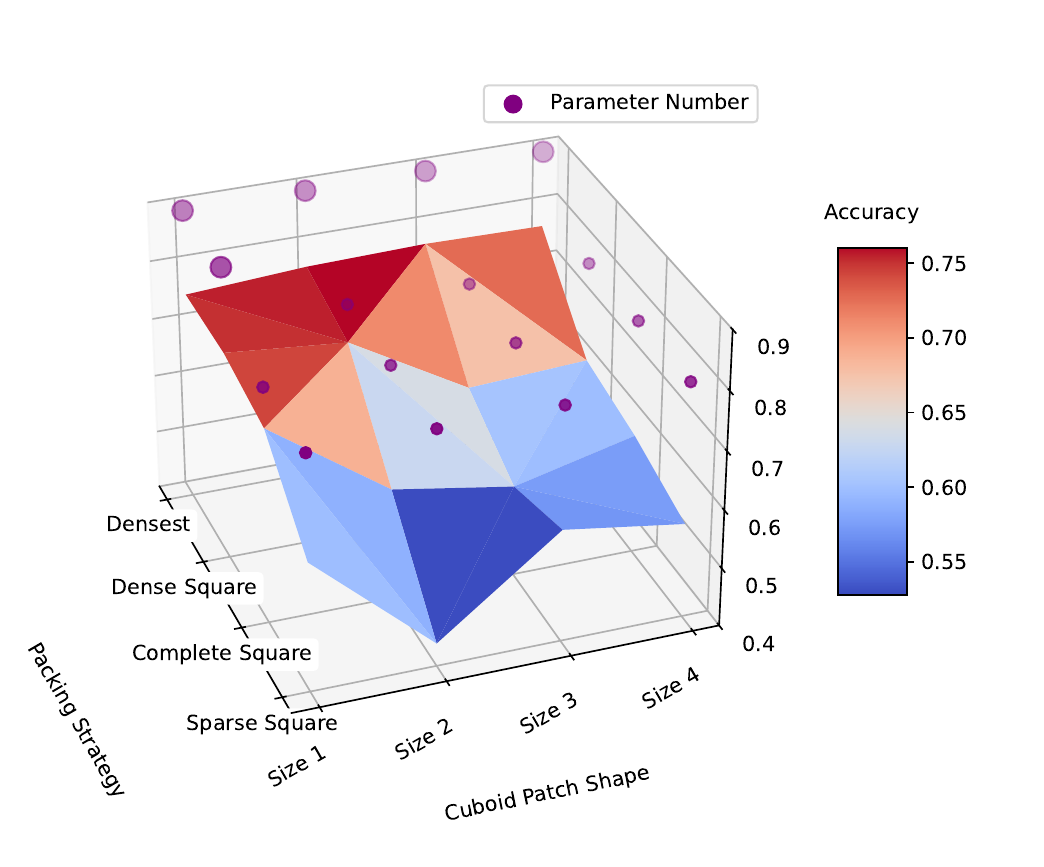}
    \end{subfigure}
        \caption{Analysis of the size and packing strategies of the cuboid patch shape. Size 1: $p_h=p_h'=1$, $p_w=p_w'=1$; Size 2: $p_h=p_h'=2$, $p_w=p_w'=2$; Size 3: $p_h=p_h'=3$, $p_w=p_w'=3$; and Size 4: $p_h=p_h'=4$, $p_w=p_w'=4$. For all these shapes, we have $p_d=p_d'=0$ by default.}
    \label{fig:cuboid_patch_results}
\end{figure}

\begin{figure}[t]
    \centering
    \begin{subfigure}{\textwidth}
        \centering
        \includegraphics[width=\textwidth]{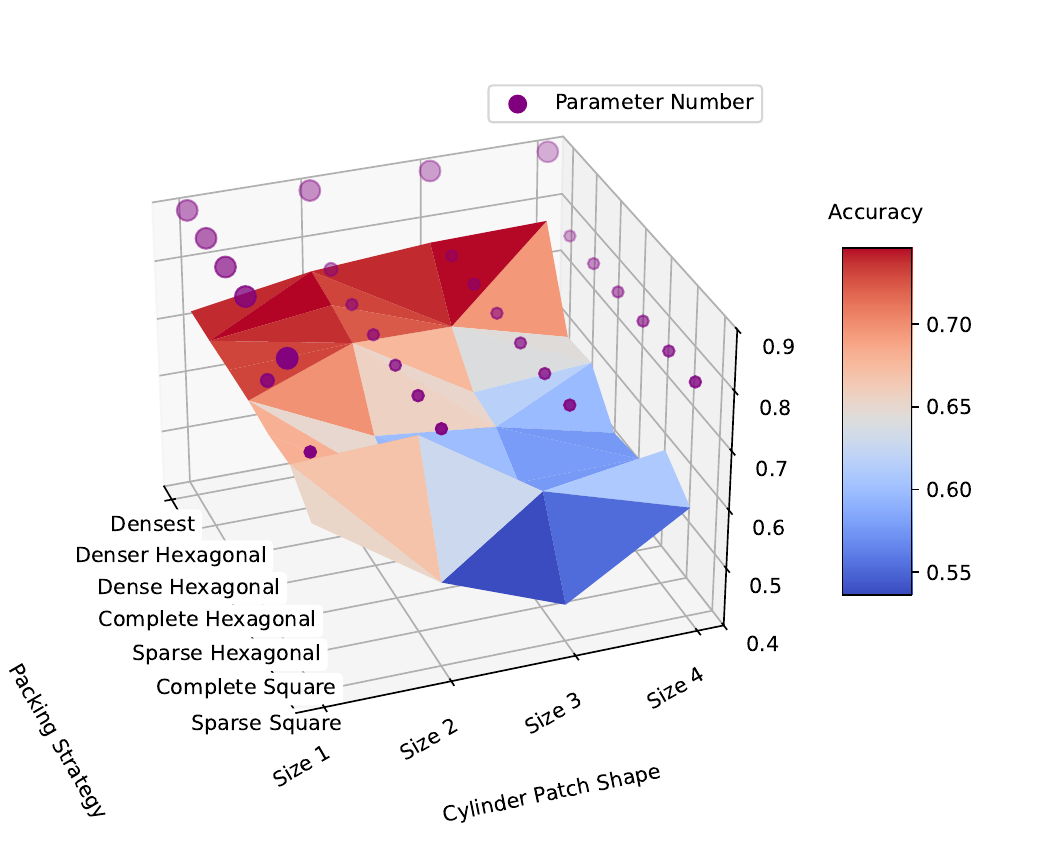}
    \end{subfigure}
        \caption{Analysis of the size and packing strategies of the cylinder patch shape. Size 1: $p_r=1$; Size 2: $p_r=2$; Size 3: $p_r=3$; and Size 4: $p_r=4$. For all these shapes, we have $p_d=p_d'=0$ by default.}
    \label{fig:cylinder_patch_results}
\end{figure}

Our comprehensive evaluation of {\our}'s performance examines variations in both patch shapes and their packing strategies within the underlying grid structure. This analysis focuses particularly on the impact of patch shape modifications in two critical components: the grid-based structural attribute interdependence function and the grid-based data compression function. The comparative performance results for cuboid and cylinder patch shapes are presented in Figures~\ref{fig:cuboid_patch_results} and \ref{fig:cylinder_patch_results}, respectively.

\noindent \textbf{Patch Shapes}: Our analysis reveals that patch shape significantly influences both learning performance and the number of learnable parameters. Specifically, a cuboid patch with dimensions $p_h=p_h'=4, p_w=p_w'=4$ (and 128 channels) covers $10,368$ feature map elements, while a cylinder patch with radius $p_r=4$ (and 128 channels) encompasses $6,272$ patch elements. Using the \textit{sparse square} packing strategy for projections between $128$ input and output channels, the interdependence function requires $2,021,514$ learnable parameters for cuboid patches versus $1,485,834$ for cylinder patches. Under the \textit{densest packing} strategy, {\our} achieves an accuracy of $0.771$ with cuboid patches and $0.780$ with cylinder patches.

\noindent \textbf{Patch Sizes}: The relationship between patch size and {\our}'s learning performance is multifaceted, as it influences not only the patch dimensions but also the center distances in packing and compression functions. For the interdependence function with cuboid patches, we find that minimal dimensions ($p_h=p_h'=1$ and $p_w=p_w'=1$) yield superior performance across all packing strategies except \textit{densest packing}. This pattern holds true for cylindrical patches as well. The enhanced performance likely stems from the smaller packing parameters, which minimize information loss during processing.

\noindent \textbf{Packing Strategies}: Among the investigated patch-related parameters, packing strategies demonstrate the most substantial impact on model performance. Our experiments show that the ``\textit{densest packing}'' and ``\textit{dense/denser packing}'' strategies consistently outperform sparse or complete packing approaches. These results suggest that maintaining some degree of redundancy is crucial for effectively capturing interdependence relationships and extracting relevant information from image data. The \textit{densest packing} strategy achieves the highest accuracy scores, particularly with larger patch configurations—such as cuboid patches with dimensions $p_h=p_h'=3$ and $p_w=p_w'=3$, or cylinder patches with radius $p_r = 4$. However, this superior performance comes at a computational cost, as these configurations require over $135$M learnable parameters, significantly more than other packing strategies.

\subsubsection{Grid Interdependence Layer Depth and Width Investigation}

\begin{figure}[t]
    \centering
    \begin{subfigure}{\textwidth}
        \centering
        \includegraphics[width=\textwidth]{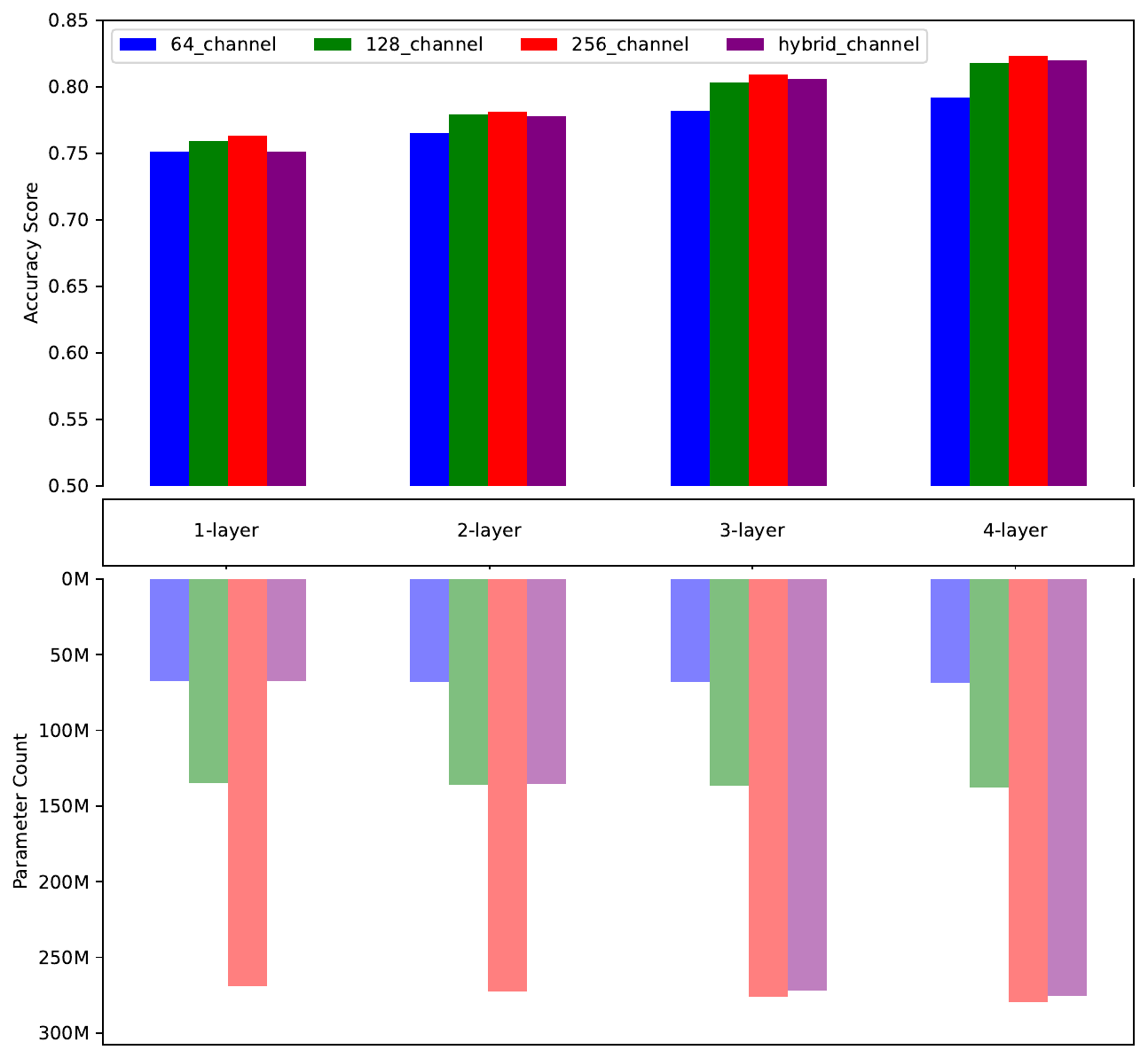}
    \end{subfigure}
        \caption{Analysis of {\our}'s performance as a function of model depth (number of interdependence layers) and width (number of channels). The hybrid channel configurations implement progressively increasing channel numbers across layers: (1) one-layer model: $[3, 64]$; (2) two-layer model: $[3, 64, 128]$; (3) three-layer model: $[3, 128, 128, 256]$; and (4) four-layer model: $[3, 128, 128, 256, 256]$.}
    \label{fig:depth_width_results}
\end{figure}

We further investigate the effects of model depth ({\ie} number of interdependence layers) and width ({\ie} number of channels) using {\our} with a cylinder patch of size $d_r=4$ and the \textit{densest} packing strategy. Our investigation encompasses architectures with $1$, $2$, $3$, and $4$ grid-based interdependence layers, featuring channel configurations of $64$, $128$, $256$, and hybrid combinations. Throughout these variations, we maintain consistent compression and perceptron layers with identical output dimensions. Figure~\ref{fig:depth_width_results} presents both the accuracy scores achieved by these architectural variations and their corresponding numbers of learnable parameters.

The experimental results demonstrate that increasing both depth and width enhances {\our}'s learning capacity, resulting in steady improvements in accuracy scores. Among all tested configurations, the architecture with $4$ layers and $256$ channels achieves superior performance compared to other variants. However, we observe that increasing model width leads to a dramatic expansion in the number of learnable parameters, with the majority of these additional parameters concentrated in the subsequent perceptron layers.


\subsubsection{Grid based Compression Layer Analysis}

\begin{figure}[t]
    \centering
    \begin{subfigure}{\textwidth}
        \centering
        \includegraphics[width=\textwidth]{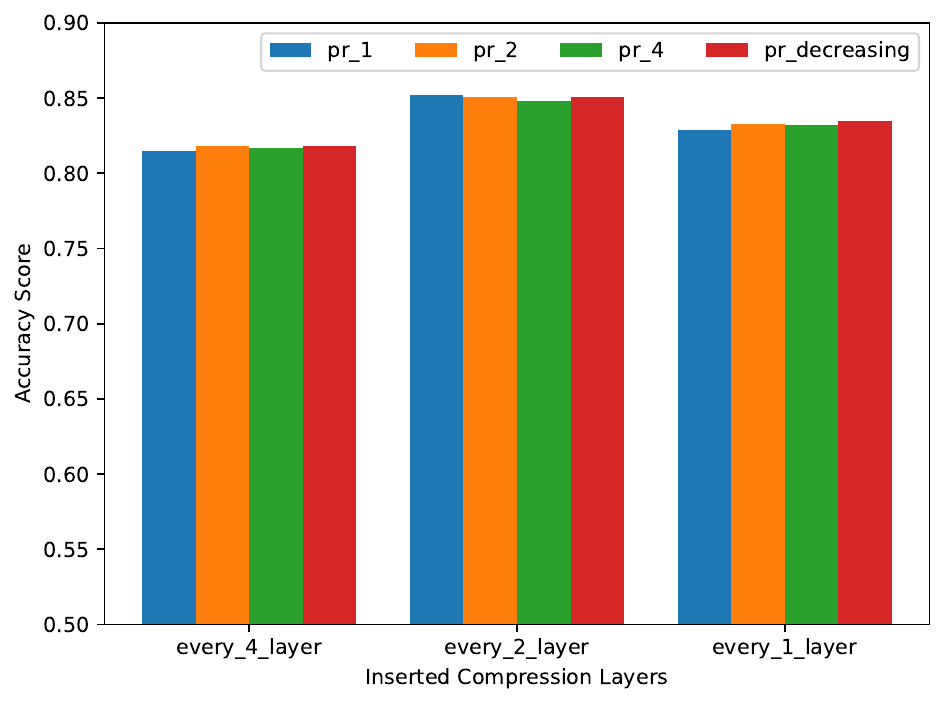}
    \end{subfigure}
        \caption{Analysis of compression layer configurations in {\our}, where compression layers are strategically inserted between interdependence layers for feature map compression. The insertion frequency follows three patterns: ``{Every 4 layer}'', ``{Every 2 layer}'', or ``{Every 1 layer}'', indicating a compression layer follows every 4, 2, or 1 interdependence layers, respectively. The parameter $p_r$ denotes the cylinder patch radius used in compression functions, while pr\_{\text{decreasing}}'' represents an adaptive configuration starting with $p_r = 4$ and halving at each successive compression layer. All compression layers maintain consistent patch packing strategy parameters with $cd_h=cd_w=2$.}
    \label{fig:compression_function_analysis}
\end{figure}

Our previous analyses employed a single compression layer to process feature maps learned by grid-based interdependence layers before flattening and feeding them to subsequent perceptron layers. Here, we extend our investigation to examine the effects of incorporating multiple compression layers between interdependence layers, enabling progressive compression of learned feature maps throughout the network.

The analysis provided in this part focuses on the {\our} model with 4 layers, where we systematically evaluate compression layer insertion frequencies of every 1, 2, or 4 interdependence layer(s). Additionally, we investigate the impact of varying cylinder patch sizes in compression layers, departing from our previous assumption of identical patch sizes between compression and interdependence layers. Our investigation encompasses compression layers with fixed patch radii ($p_r = 1$, $p_r = 2$, $p_r = 4$) and an adaptive configuration where $p_r$ begins at 4 and halves after each compression layer. For example, in a 4-layer {\our} model with compression layers after each interdependence layer, the adaptive configuration yields four compression layers with progressively decreasing patch radii of 4, 2, 1, and 1.

Results shown in Figure~\ref{fig:compression_function_analysis} reveal three key findings: (1) inserting compression layers every two interdependence layers yields superior performance compared to other frequencies; (2) when compression layers are added every two interdependence layers, cylinder patches with $p_r=1$ achieve optimal performance; and (3) while large patch sizes ($p_r=4$) initially degrade model performance, implementing progressive size reduction across compression layers leads to improved results.

\subsubsection{Chain Interdependence Investigation: Uni-directional, Bi-directional, Single-Hop, and Multi-Hop}

\begin{figure}[t]
    \centering
    \begin{subfigure}{\textwidth}
        \centering
        \includegraphics[width=\textwidth]{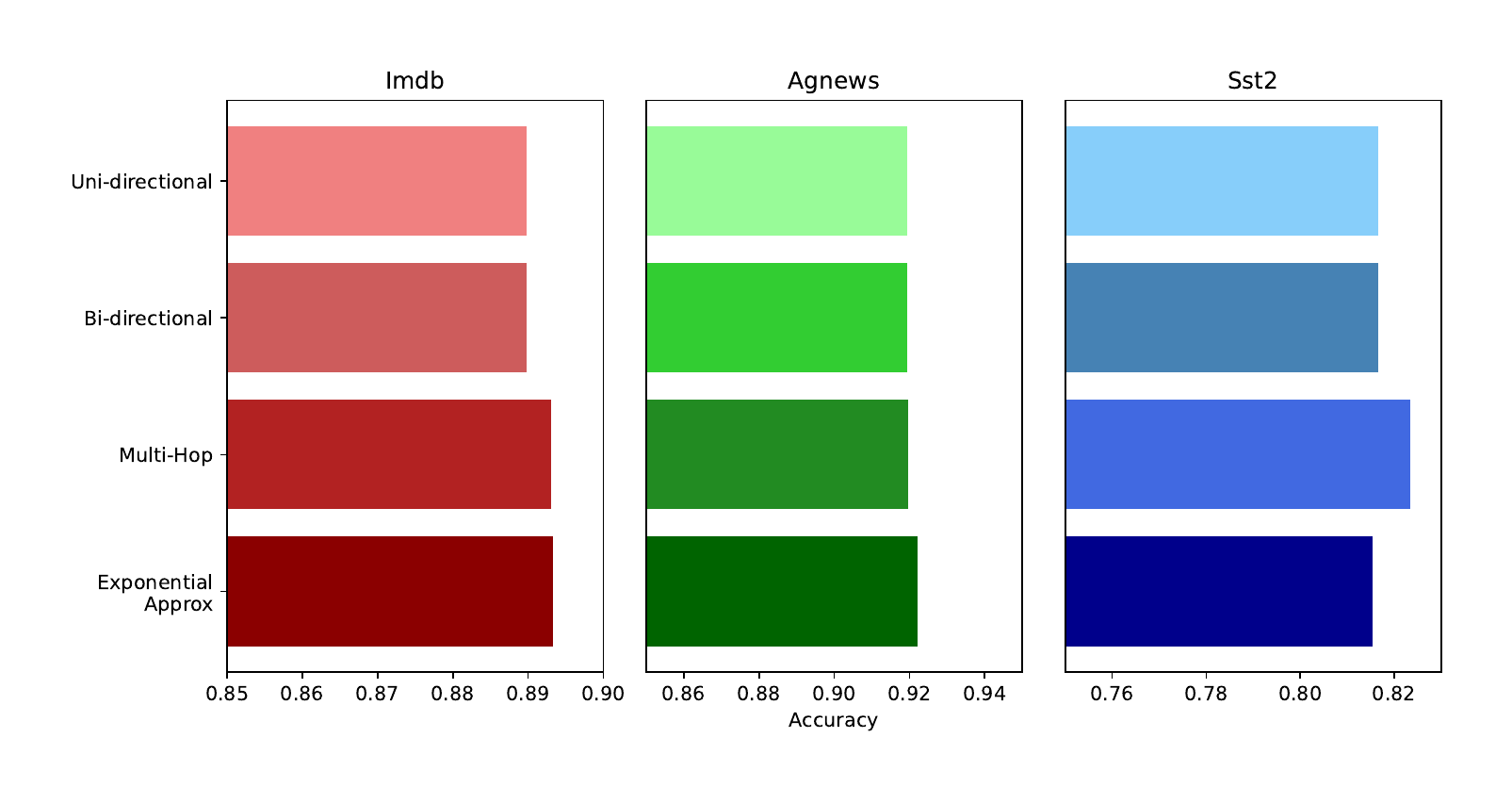}
    \end{subfigure}
        \caption{Analysis of chain-based structural interdependence functions' impact on {\our}'s classification performance across IMDB, AGNews, and SST2 datasets. Performance comparison encompasses four distinct interdependence modes: ``Uni-directional'', ``Bi-directional'', Multi-Hop'', and ``Exponential Approx''.}
    \label{fig:chain_interdependence_function}
\end{figure}

Expanding beyond our analysis of grid-based structural interdependence functions on image data, we evaluate the effectiveness of chain-based structural interdependence functions on language data. Our investigation employs a five-layer {\our} architecture comprising: an input perceptron layer, two chain-based structural interdependence function layers with an intervening perceptron layer, and an output perceptron layer. The model processes classification by aggregating learned token representations from the second interdependence layer via summation before feeding them to the output perceptron layer. Figure~\ref{fig:chain_interdependence_function} presents {\our}'s performance results using chain structural interdependence functions across three datasets: IMDB, AGNews, and SST2.

Our analysis reveals several key insights about chain-based structural interdependence functions when their outputs are aggregated through summation. Notably, uni-directional and bi-directional chain structural interdependence functions achieve comparable performance levels. Multi-hop interdependence functions (with hop counts $h=3$ or $h=5$) demonstrate significant performance improvements over their single-hop counterparts, with particularly pronounced gains on the SST2 dataset. Furthermore, the exponential approximation of multi-hop interdependence functions achieves comparable—and in some cases slightly superior—performance on IMDB and AGNews datasets, validating the effectiveness of our approximation approach.

\begin{table}[t]
    \caption{The learning performance of the comparison methods on discrete image and language datasets is evaluated using Accuracy as the default metric. For each dataset, the best-performing method is highlighted in bold math font.}
    \label{tab:discrete_data_classification_results}
    \centering
    \setlength{\tabcolsep}{2pt}
    \begin{tabular}{|c|c|c|c|c|c|}
        \hline
        \multirow{2}{*}{\textbf{Models}} & \multicolumn{2}{c|}{\textbf{Image Datasets}} & \multicolumn{3}{c|}{\textbf{Language Datasets}}  \\
        \cline{2-6}
         & \textbf{MNIST} & \textbf{CIFAR10} & \textbf{IMDB} & \textbf{AGNews} & \textbf{SST2}  \\
        \hline
         \hline
	\makecell{MLP} 
	& \makecell{ $9.82 \times 10^{-1}$}
	& \makecell{ $5.63 \times 10^{-1}$}
	& \makecell{ $8.85 \times 10^{-1}$} 
        & \makecell{ $9.21 \times 10^{-1}$} 
        & \makecell{ $8.05 \times 10^{-1}$} \\
        \hline
        \hline
        \makecell{Previous {\old}\\ \cite{zhang2024rpnreconciledpolynomialnetwork}} 
        & \makecell{ $9.86 \times 10^{-1}$} 
        & \makecell{ $5.61 \times 10^{-1}$}
	& \makecell{$8.86 \times 10^{-1}$} 
        & \makecell{$9.19 \times 10^{-1}$} 
        & \makecell{$8.07 \times 10^{-1}$}\\
        \hline
        \hline
        \makecell{CNN (Image) \& \\ LSTM (Language)} 
        & \makecell{ $9.96 \times 10^{-1}$} 
	& \makecell{ $8.47 \times 10^{-1}$}
	& \makecell{\boldmath $8.93 \times 10^{-1}$} 
	& \makecell{$9.14 \times 10^{-1}$} 
	& \makecell{\boldmath $8.27 \times 10^{-1}$} \\
        \hline
        \hline
        \makecell{{\our}} 
        & \makecell{\boldmath $9.98 \times 10^{-1}$} 
        & \makecell{\boldmath $8.52 \times 10^{-1}$}
	& \makecell{\boldmath $8.93 \times 10^{-1}$} 
        & \makecell{\boldmath $9.22 \times 10^{-1}$} 
        & \makecell{$8.23 \times 10^{-1}$} \\
        \hline
    \end{tabular}
\end{table}


\subsubsection{The Main Results of {\our} on Vision Data Classification}

Table~\ref{tab:discrete_data_classification_results} presents a comprehensive performance comparison between {\our} and several baseline methods. For comparison, we include the standard MLP, our previous {\old} model \cite{zhang2024rpnreconciledpolynomialnetwork}, CNN (for image classification tasks), and LSTM (for language classification tasks). While we also evaluated vanilla RNN for language classification, its performance was significantly inferior to LSTM, hence we report LSTM results as the representative sequential model baseline.

\begin{figure}[t]
    \centering
    \begin{subfigure}{0.22\textwidth}
        \centering
        \includegraphics[width=\textwidth]{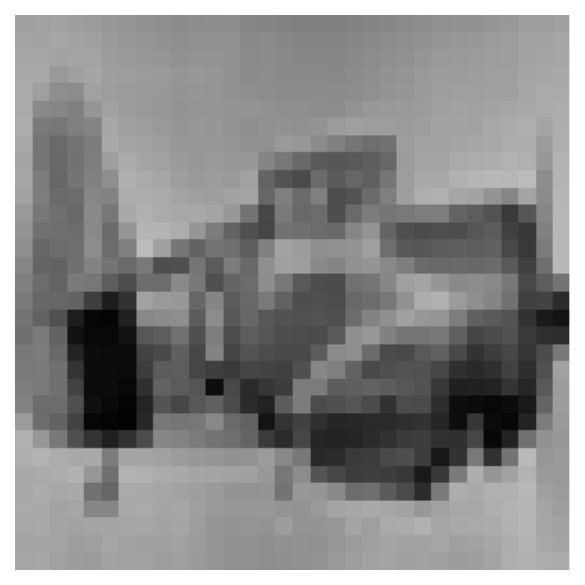}
        \label{fig:grid_1}
    \end{subfigure}
    \hfill
    \begin{subfigure}{0.22\textwidth}
        \centering
        \includegraphics[width=\textwidth]{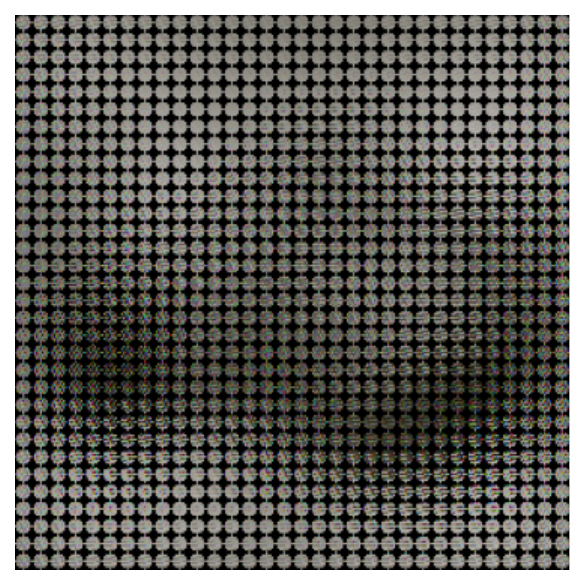}
        \label{fig:grid_2}
    \end{subfigure}
    \hfill
    \begin{subfigure}{0.22\textwidth}
        \centering
        \includegraphics[width=\textwidth]{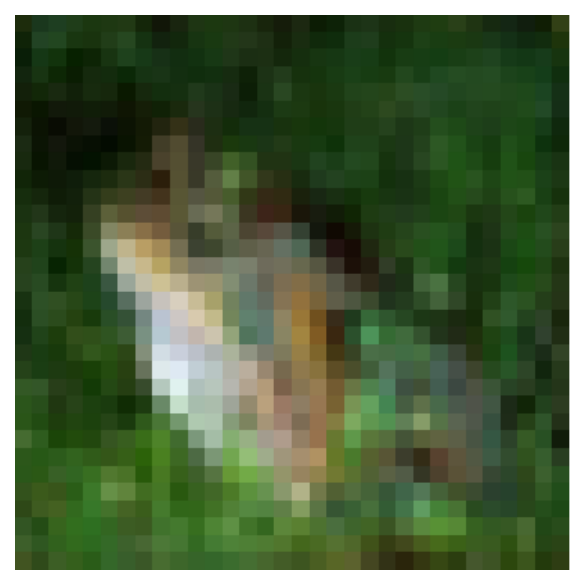}
        \label{fig:grid_3}
    \end{subfigure}
    \hfill
    \begin{subfigure}{0.22\textwidth}
        \centering
        \includegraphics[width=\textwidth]{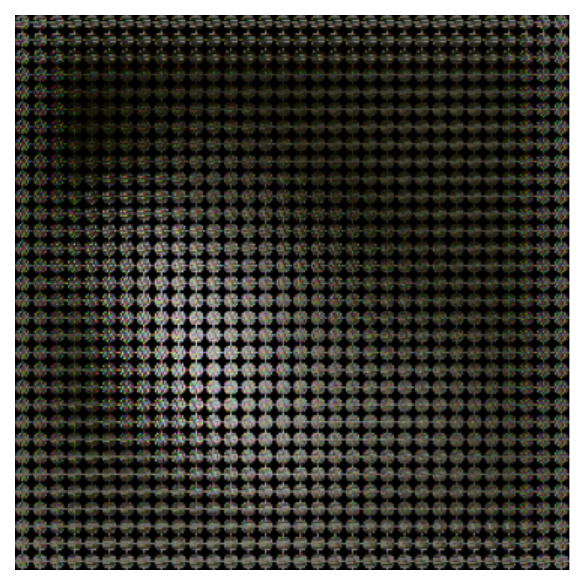}
        \label{fig:grid_4}
    \end{subfigure}
    
    \begin{subfigure}{0.22\textwidth}
        \centering
        \includegraphics[width=\textwidth]{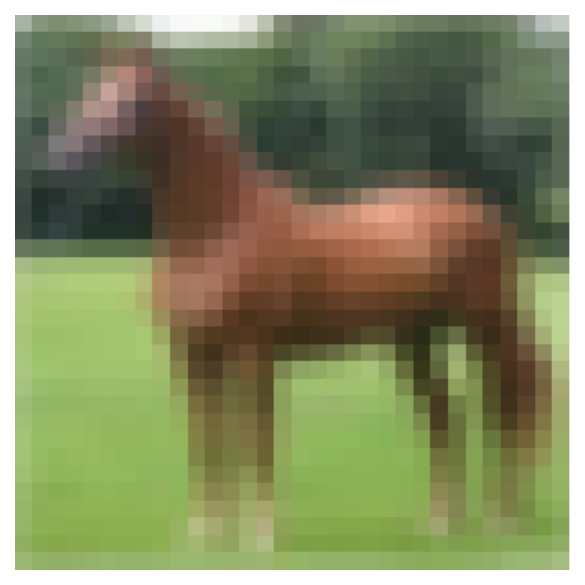}
        \label{fig:grid_3}
    \end{subfigure}
    \hfill
    \begin{subfigure}{0.22\textwidth}
        \centering
        \includegraphics[width=\textwidth]{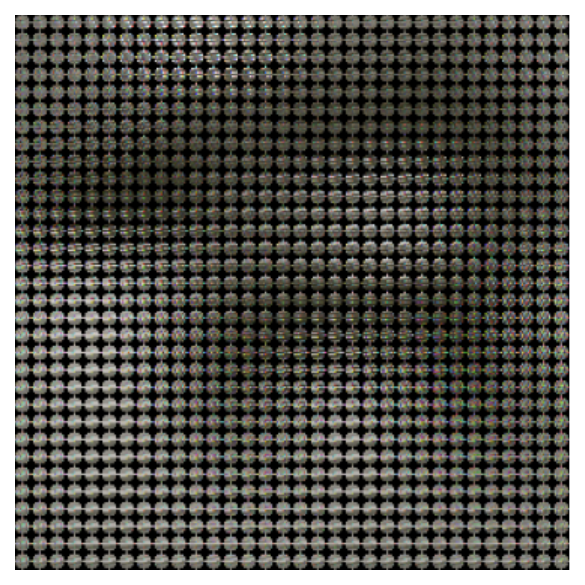}
        \label{fig:grid_4}
    \end{subfigure}
    \hfill
    \begin{subfigure}{0.22\textwidth}
        \centering
        \includegraphics[width=\textwidth]{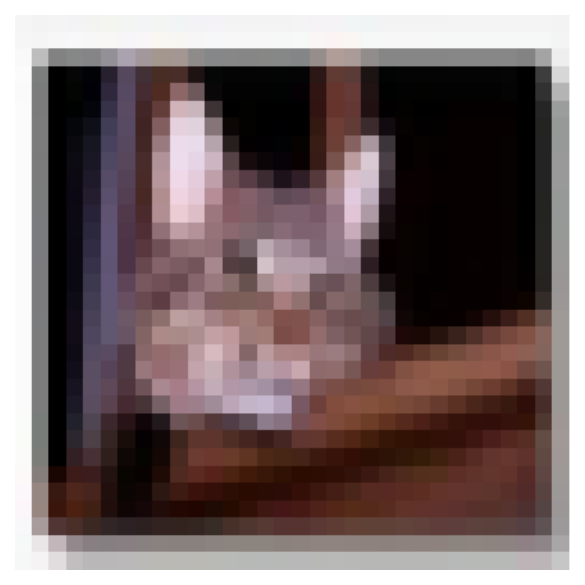}
        \label{fig:grid_3}
    \end{subfigure}
    \hfill
    \begin{subfigure}{0.22\textwidth}
        \centering
        \includegraphics[width=\textwidth]{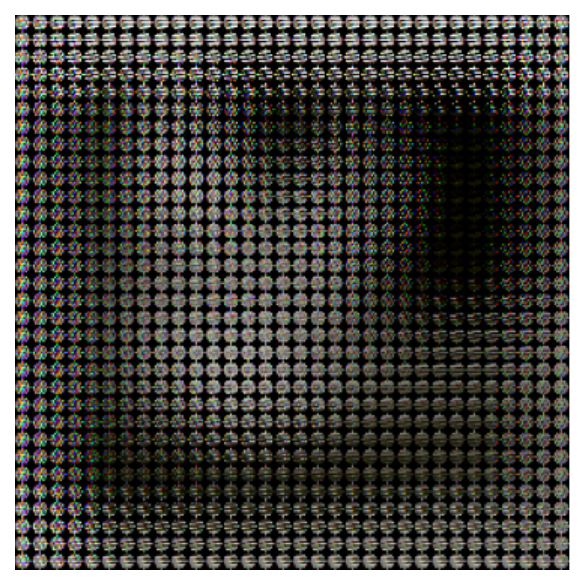}
        \label{fig:grid_4}
    \end{subfigure}
    
    \begin{subfigure}{0.22\textwidth}
        \centering
        \includegraphics[width=\textwidth]{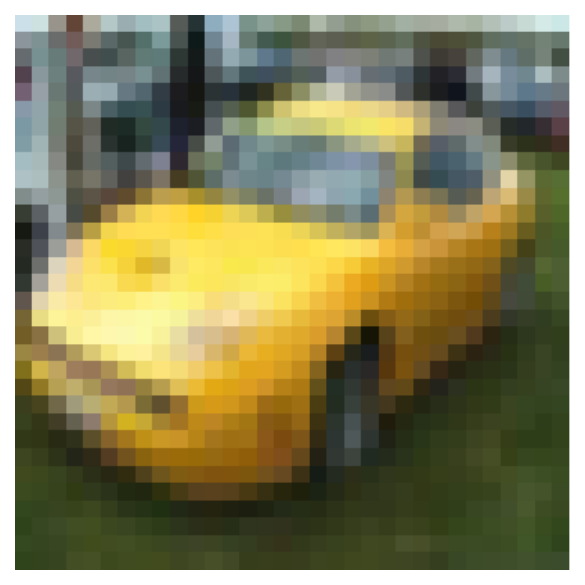}
        \label{fig:grid_3}
    \end{subfigure}
    \hfill
    \begin{subfigure}{0.22\textwidth}
        \centering
        \includegraphics[width=\textwidth]{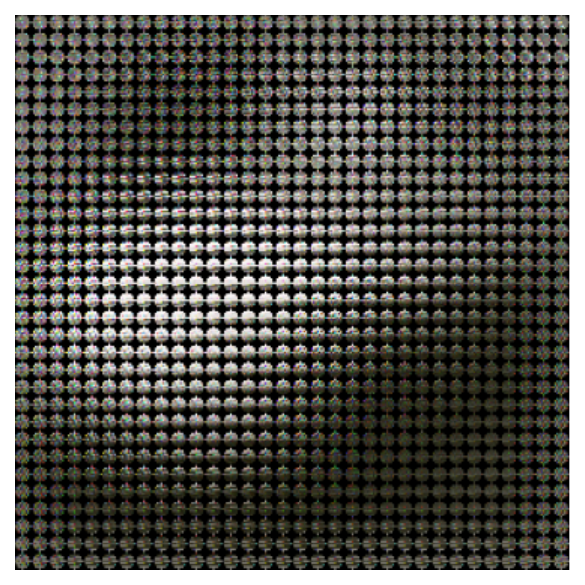}
        \label{fig:grid_4}
    \end{subfigure}
    \hfill
    \begin{subfigure}{0.22\textwidth}
        \centering
        \includegraphics[width=\textwidth]{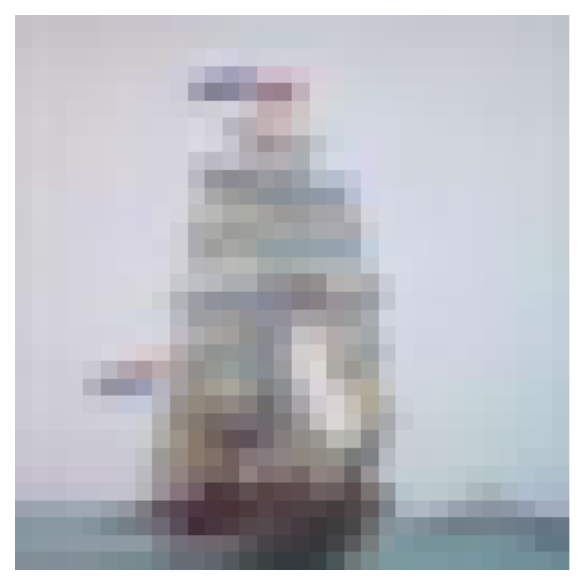}
        \label{fig:grid_3}
    \end{subfigure}
    \hfill
    \begin{subfigure}{0.22\textwidth}
        \centering
        \includegraphics[width=\textwidth]{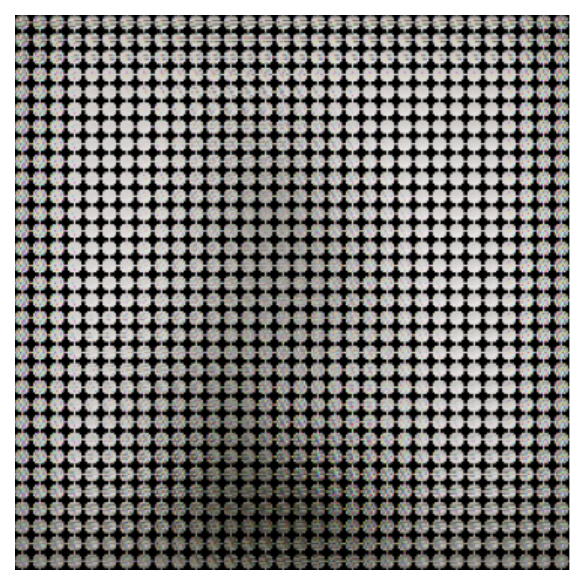}
        \label{fig:grid_4}
    \end{subfigure}
    
    \begin{subfigure}{0.22\textwidth}
        \centering
        \includegraphics[width=\textwidth]{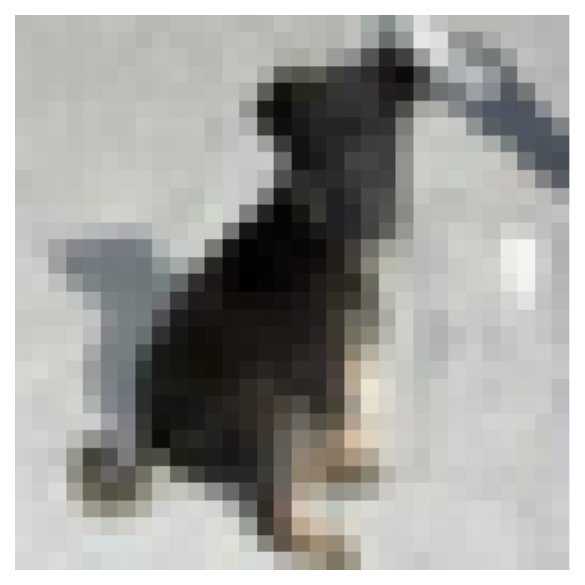}
        \label{fig:grid_3}
    \end{subfigure}
    \hfill
    \begin{subfigure}{0.22\textwidth}
        \centering
        \includegraphics[width=\textwidth]{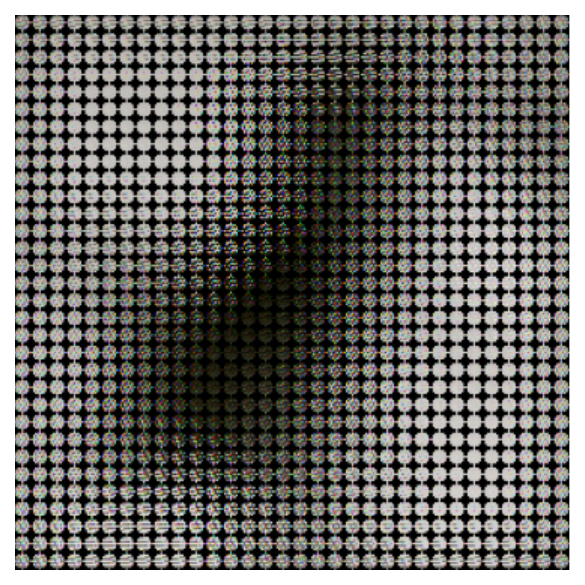}
        \label{fig:grid_4}
    \end{subfigure}
    \hfill
    \begin{subfigure}{0.22\textwidth}
        \centering
        \includegraphics[width=\textwidth]{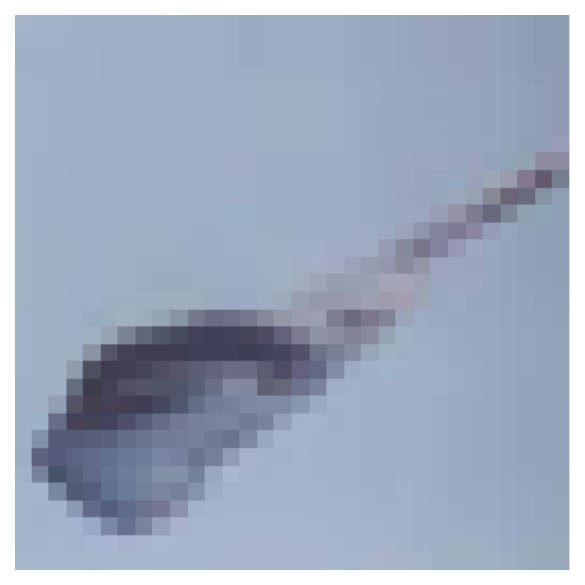}
        \label{fig:grid_3}
    \end{subfigure}
    \hfill
    \begin{subfigure}{0.22\textwidth}
        \centering
        \includegraphics[width=\textwidth]{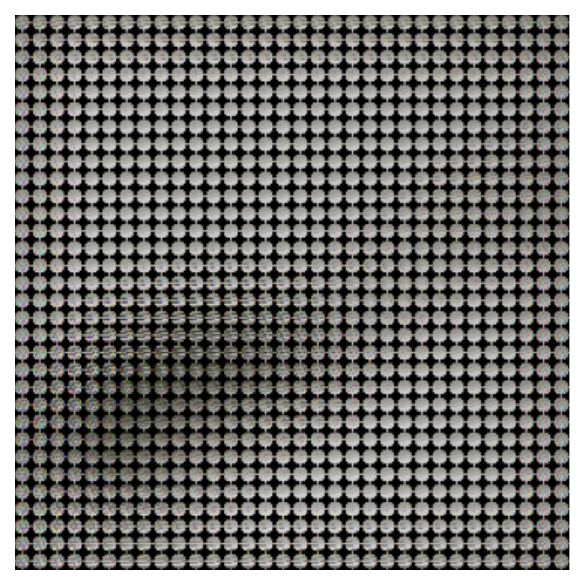}
        \label{fig:grid_4}
    \end{subfigure}
    
    \begin{subfigure}{0.22\textwidth}
        \centering
        \includegraphics[width=\textwidth]{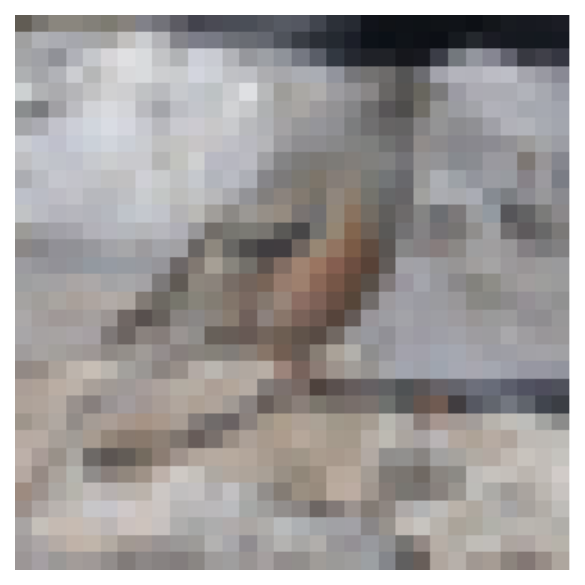}
        \label{fig:grid_3}
    \end{subfigure}
    \hfill
    \begin{subfigure}{0.22\textwidth}
        \centering
        \includegraphics[width=\textwidth]{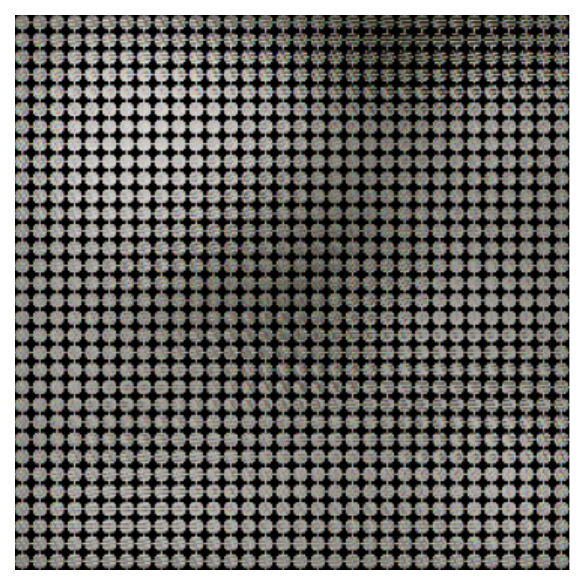}
        \label{fig:grid_4}
    \end{subfigure}
    \hfill
    \begin{subfigure}{0.22\textwidth}
        \centering
        \includegraphics[width=\textwidth]{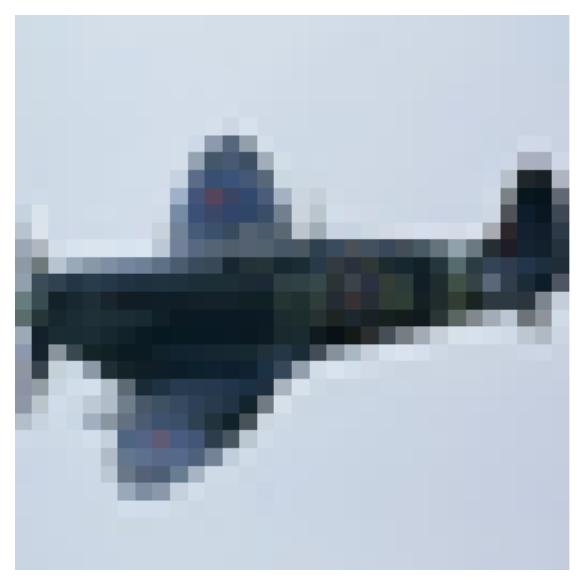}
        \label{fig:grid_3}
    \end{subfigure}
    \hfill
    \begin{subfigure}{0.22\textwidth}
        \centering
        \includegraphics[width=\textwidth]{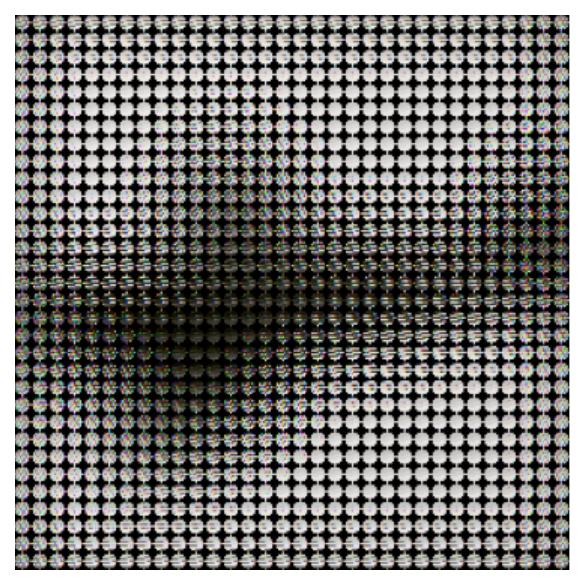}
        \label{fig:grid_4}
    \end{subfigure}
    \caption{An illustration of raw images (shown in the first and third columns) from the CIFAR-10 dataset and their expanded counterparts (shown in the second and fourth columns), generated using the grid-based structural interdependence functions with a cylindrical patch of radius $p_r=4$.}
    \label{fig:grid_interdependence_visualization}
\end{figure}

As shown in Table~\ref{tab:discrete_data_classification_results}, the proposed {\our}, equipped with interdependence functions, significantly outperforms the earlier {\old} model introduced in \cite{zhang2024rpnreconciledpolynomialnetwork}, particularly on the CIFAR-10 dataset. These results highlight the effectiveness and importance of the interdependence functions in capturing relationships among attributes and instances. Furthermore, {\our} achieves comparable—and occasionally superior—performance to CNN and LSTM models. This demonstrates {\our}'s potential not only to theoretically unify CNN and RNN architectures within its canonical representation but also to empirically deliver performance on par with these established backbone models in practical applications.


\subsubsection{Visualization of Cylinder Patches of Images}

In Figure~\ref{fig:grid_interdependence_visualization}, we present several examples of the raw images from the CIFAR-10 dataset and their corresponding expansions using the grid-based interdependence function. The first and third columns contain $10$ raw images sampled from the CIFAR-10 dataset, while the second and fourth columns display their respective expanded images.

For each pixel in a raw image, the grid-based structural interdependence function pads the pixels within the cylindrical patch of radius $p_r=4$ (with a diameter of $9$). To visualize these patches, we include some pixels with dummy values to enclose each cylindrical patch within a cuboid of size $9 \times 9 \times 3$. Compared to the raw image pixels, the expanded images still retain the outlines of the objects. Moreover, for each pixel in the raw images, their expansions provide richer contextual information about their local learning neighborhood. This additional information enables the {\our} model to utilize a more detailed learning context for classifying these images, leading to improved performance compared to the previous {\old} model \cite{zhang2024rpnreconciledpolynomialnetwork}.

\subsubsection{Visualization of Chain-based Multi-Hop Interdependence Functions}

\begin{figure}[t]
    \centering
    \begin{subfigure}{0.48\textwidth}
        \centering
        \includegraphics[width=\textwidth]{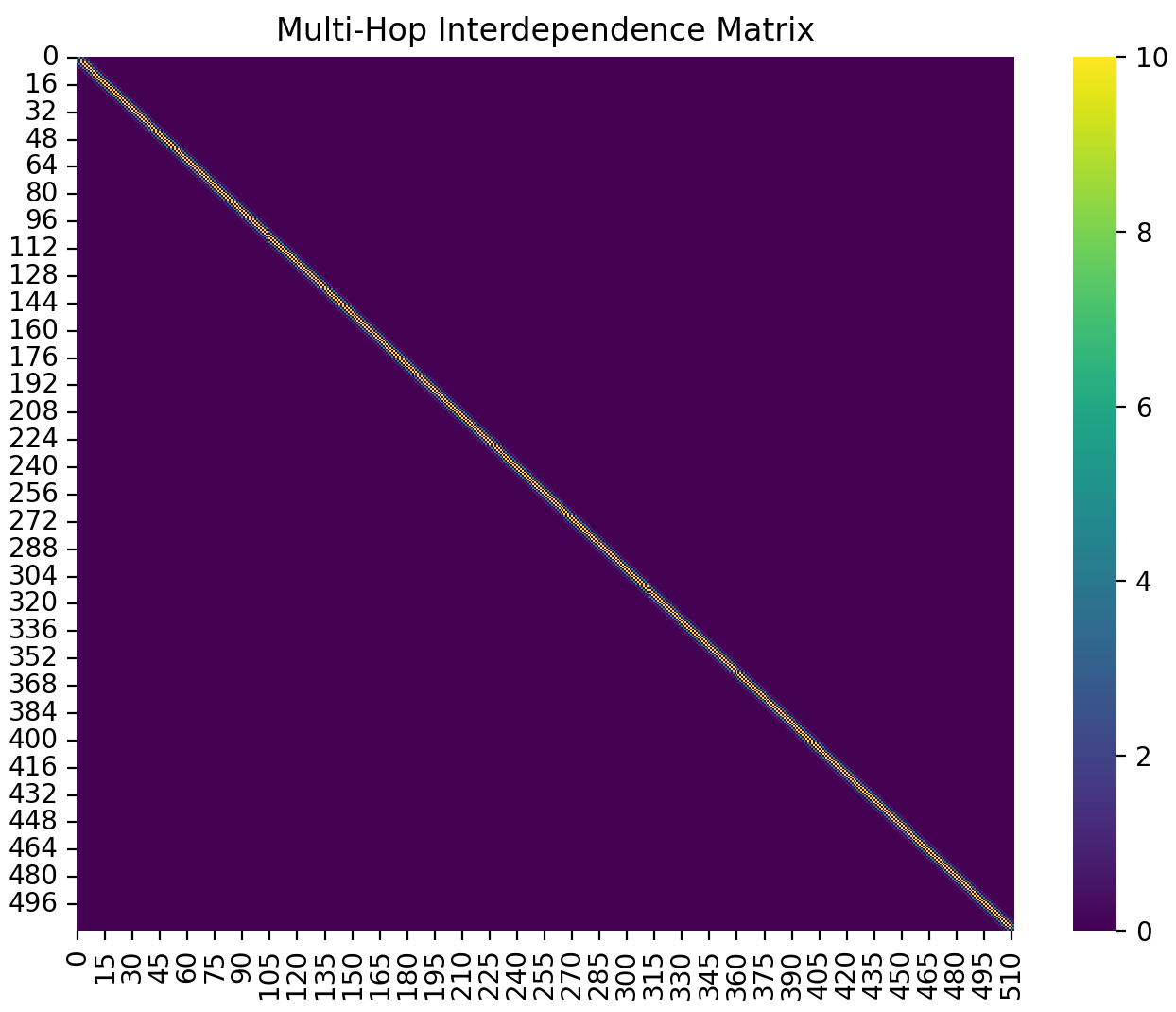}
        \caption{Multi-Hop Interdependence Matrix}
        \label{fig:chain_multi_hop}
    \end{subfigure}
    \hfill
    \begin{subfigure}{0.48\textwidth}
        \centering
        \includegraphics[width=\textwidth]{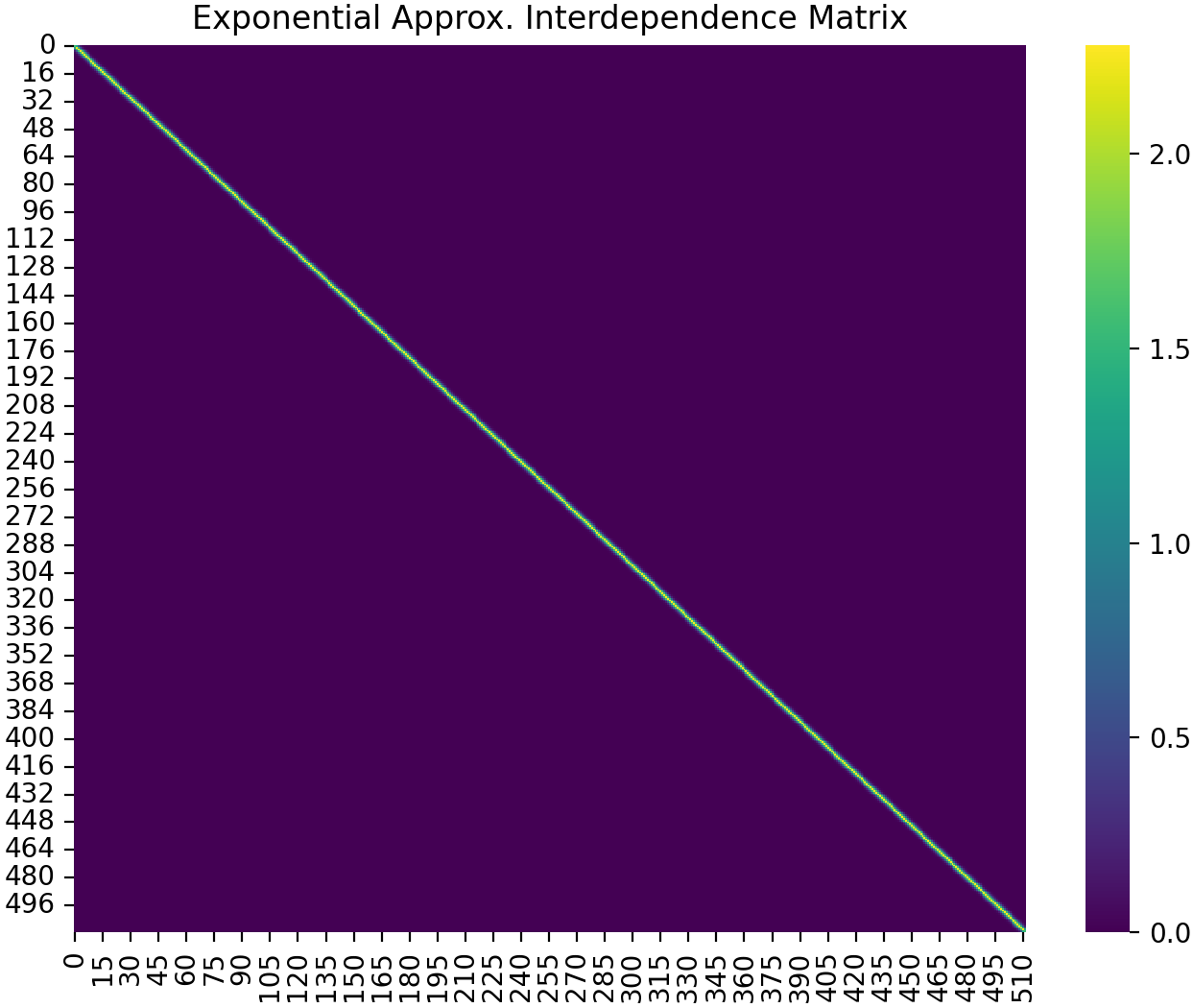}
        \caption{Exponential Approx. Interdependence Matrix}
        \label{fig:chain_exponential_approx}
    \end{subfigure}
    \caption{An illustration of the chain-based structural interdependence matrices for the IMDB language dataset classification. The percentages of non-zero elements in the two matrices are as follows: (1) multi-hop ($h=5$) interdependence matrix: $1.165\%$, and (2) exponential approximation interdependence matrix: $7.096\%$.}
    \label{fig:chain_interdependence_visualization}
\end{figure}

In Figure~\ref{fig:chain_interdependence_visualization}, we illustrate the chain-based structural interdependence matrices used in {\our} for classifying the IMDB dataset. The left plot represents the multi-hop interdependence matrix with hop $h=5$, while the right plot shows its exponential approximation.

From the visualization, both matrices exhibit non-zero entries concentrated along their diagonal regions. However, as indicated by the diagonal colors, the non-zero entries in the multi-hop interdependence matrix have relatively larger values compared to those in the exponential approximation matrix, which penalizes higher-order matrix powers using constant factors. Additionally, we calculate the percentages of non-zero entries in these matrices. The exponential approximation interdependence matrix contains approximately $7.096\%$ non-zero entries, significantly higher than the sparse multi-hop interdependence matrix, which has only $1.165\%$ non-zero entries.

\subsection{Time-Series Data Prediction and Graph Data Classification}\label{subsec:time_sereies_prediction}

In addition to discrete image and language data, the {\our} model has also demonstrated effectiveness in learning from continuous data and graph-structured data. In this subsection, we present the experimental evaluation of {\our} and several comparison methods on continuous time-series benchmark datasets related to finance and traffic over time, as well as on graph-structured benchmark datasets pertaining to citation networks.

We begin by providing a brief overview of these time-series and graph benchmark datasets. Next, we present the experimental results obtained by the comparison methods. Finally, we illustrate some of the learned interdependence matrices specifically for the graph data.

\subsubsection{Time-Series Dataset Description and Experiment Setups}

\begin{table}[t]
\centering
\caption{The table summarizes the key statistics of the raw time-series benchmark datasets. For the Stock and ETF datasets, instances vary in length, with the mean and standard deviation (mean$\pm$std) provided. In contrast, instances in the LA and Bay Area traffic datasets are aligned and have identical lengths. Additionally, the table includes the number of attributes recorded for each data instance at each timestamp.}
\label{tab:time_series_dataset_statistics}
\begin{tabular}{|c|c|c|c|c|}
\hline
\multirow{2}{*}{} & \multicolumn{4}{c|}{\textbf{Time-Series Datasets}} \\ \cline{2-5}
                          & \textbf{Stock} & \textbf{ETF} & \textbf{Traffic-LA} & \textbf{Traffic-Bay} \\ \hline\hline
Instance \#                     &7,163 & 1,344 & 207 & 325  \\ \hline
Timestamp \#                     &$2,078 \pm 833$ &$1,908 \pm 893$ & 34,272 & 52,116  \\ \hline
Attribute \#	& 6 & 6 & 1 & 1 \\ \hline
\end{tabular}
\end{table}


\noindent \textbf{Dataset Descriptions}: The time-series data studied in the experiments include stock market and urban traffic datasets. Specifically, the stock market datasets comprise Nasdaq-traded stocks and ETFs, while the urban traffic datasets include data collected from the Los Angeles region and the San Francisco Bay Area. Basic statistical information about these datasets is summarized in Table~\ref{tab:time_series_dataset_statistics}. For the stock and ETF datasets, each instance has $7$ attributes at each timestamp, corresponding to ``\textit{Open}'', ``\textit{Close}'', ``\textit{High}'', ``\textit{Low}'', ``\textit{Volume}'', and ``\textit{OpenInt}''. In this project, we specifically predict the ``\textit{Open}'' attribute, which represents the opening price of the stocks and ETFs for each trading day. Meanwhile, for the traffic data, there is a single attribute representing the accumulated traffic counts within a specific time period, which is used as the target for prediction in the experiments.

\noindent \textbf{Experimental Setups}: In the experiments, these datasets were cleaned, pre-processed, and partitioned according to their temporal granularities: per day for stock and ETF data, and per minute for traffic data. For each instance in the finance datasets ({\ie} the stock and ETF datasets), we extracted the recent one-year trading records and partitioned them into input-output pairs. The input spans $10$ prior trading days, while the output represents the next day’s opening price, with a time gap of $1$. Similarly, for the traffic data ({\ie} the LA and Bay Area datasets), all available records were used to create input-output pairs with the same setup as the finance data. The performance of all comparison methods is evaluated using MSE (Mean Squared Error) as the default metric.

\subsubsection{The Main Results of {\our} on Time-Series Prediction}

\begin{table}[t]
    \caption{The learning results of the comparison methods on the time-series benchmark datasets are presented in the table. The scores represent the MSE achieved by each method. For the best performance on each dataset, the corresponding scores are highlighted in bold math font.}
    \label{tab:time_series_prediction_benchmark_results}
    \centering
    \setlength{\tabcolsep}{2pt}
    \begin{tabular}{|c|c|c|c|c|}
        \hline
        \multirow{2}{*}{\textbf{Models}} & \multicolumn{4}{c|}{\textbf{Time-Series Prediction Datasets}}  \\
        \cline{2-5}
         & \textbf{Stock} & \textbf{ETF} & \textbf{Traffic-LA} & \textbf{Traffic-Bay}  \\
        \hline
        \hline
        \makecell{MLP} 
        & $1.96 \times 10^{-4}$
        & $3.36 \times 10^{-4}$
        & {\boldmath $1.70 \times 10^{-1}$}
        & $8.13 \times 10^{-2}$\\
        \hline
        \hline
        \makecell{Previous {\old}\\ \cite{zhang2024rpnreconciledpolynomialnetwork}} 
        & $1.94 \times 10^{-4}$
        & {\boldmath $3.31 \times 10^{-4}$}
        & $1.71 \times 10^{-1}$
        & $8.12 \times 10^{-2}$\\
        \hline
        \hline
        \makecell{RNN} 
        & {\boldmath $1.85 \times 10^{-4}$}
        & $3.47 \times 10^{-4}$
        & $1.74 \times 10^{-1}$
        & $8.13 \times 10^{-2}$\\
        \hline
        \hline
        \makecell{{\our}}
        & {\boldmath $1.85 \times 10^{-4}$}
        & $3.35 \times 10^{-4}$
        & $1.74 \times 10^{-1}$
        & {\boldmath $7.99 \times 10^{-2}$}\\
        \hline
    \end{tabular}
\end{table}


In Table~\ref{tab:time_series_prediction_benchmark_results}, we present the forecasting results of {\our} on the time-series datasets. For comparison, the table also includes the performance of RNN, the previous {\old}, and MLP, where MLP and {\old} project the input sequence of length $10$ to the desired output. The scores in the table represent the MSE achieved by these methods on the four time-series benchmark datasets.

From the results, we observe that {\our} achieves comparable performance to RNN on the Stock, ETF, and Traffic-LA datasets. However, compared to MLP and the previous {\old} model, the advantages of RNN and {\our} in sequence data modeling are less pronounced. This is likely because MLP and {\old}, built with fully connected layers, can effectively utilize the historical record information. Notably, on the Traffic-Bay dataset, {\our} outperforms the other methods. This improvement can be attributed to the relatively longer sequences in the Traffic-Bay dataset, which provide more training data samples to enhance the learning process of {\our}.

\subsubsection{Graph Dataset Description and Experiment Setups}

\noindent \textbf{Dataset Descriptions}: We studied three citation graph benchmark datasets: Cora, Citeseer, and Pubmed. Basic statistical information about these datasets is provided in Table~\ref{tab:graph_dataset_statistics}. Each dataset consists of a set of nodes representing papers, connected by citation links among them. Specifically, the Cora, Citeseer, and Pubmed datasets contain $2,708$, $3,327$, and $19,717$ nodes, connected by $10,556$, $9,104$, and $88,648$ links, respectively. In addition to the graph structure, each node is associated with a feature vector (a bag-of-words representation of the paper's textual description, weighted by TF-IDF) and a label denoting the paper's topic. Detailed statistical information is also included in Table~\ref{tab:graph_dataset_statistics}.

\noindent \textbf{Experimental Setups}: In the experiments, subsets of nodes were sampled for training and testing the graph learning and node classification models. Specifically, for each class in the three graph datasets, $20$ instances were sampled for training, and a set of $1,000$ nodes was randomly sampled for testing. For the graph neural networks and {\our}, training was conducted using semi-supervised and transductive learning settings, effectively incorporating unlabeled instances into the model training. Conversely, the MLP and the previous {\old} model were trained using classic supervised and inductive learning settings, relying solely on the labeled training set.

\subsubsection{The Main Results of {\our} on Graph Classification}

\begin{table}[t]
\centering
\caption{The table provides an overview of the graph benchmark datasets, including the number of nodes and links, as well as the input attributes and output label classes. For the experiments, we randomly sampled 20 node instances per class as the training set for each graph dataset, and 1,000 random nodes as the testing set.}
\label{tab:graph_dataset_statistics}
\begin{tabular}{|c|c|c|c|}
\hline
\multirow{2}{*}{} & \multicolumn{3}{c|}{\textbf{Graph Datasets}} \\ \cline{2-4}
                          & \textbf{Cora} & \textbf{Citeseer} & \textbf{Pubmed} \\ \hline\hline
Node \#	& 2,708 & 3,327 & 19,717 \\ \hline
Link \#	& 10,556 & 9,104 & 88,648 \\ \hline
Input Dim.	& 1,433 & 3,703 & 500   \\ \hline
Output Dim.& 7  & 6 & 3  \\ \hline
Train \#	& 140 & 120 & 60  \\ \hline
Test \#	& 1,000 & 1,000 & 1,000  \\ \hline
\end{tabular}
\end{table}

\begin{table}[t]
    \caption{The classification performance of the comparison methods on nodes in the graph benchmark datasets is reported. All comparison methods share the same input, hidden, and output layer dimensions, with detailed architectural information provided below the graph datasets.}
    \label{tab:graph_classification_benchmark_results}
    \centering
    \setlength{\tabcolsep}{2pt}
    \begin{tabular}{|c|c|c|c|}
        \hline
        \multirow{5}{*}{\textbf{Models}} & \multicolumn{3}{c|}{\textbf{Graph Datasets}}  \\
        \cline{2-4}
         & \makecell{ \textbf{Cora}  \\ {- - - - - - - - - - - -} \\ $[$1433, 16, 7$]$} 
         & \makecell{ \textbf{Citeseer}  \\ {- - - - - - - - - - - -} \\ $[$3703, 16, 6$]$} 
         & \makecell{ \textbf{Pubmed}  \\ {- - - - - - - - - - - -} \\ $[$500, 16, 3$]$} \\
        \hline
        \hline
        \makecell{MLP} 
        & \makecell{ $6.61 \times 10^{-1}$ } 
        & \makecell{ $3.05 \times 10^{-1}$ }
        & \makecell{ $3.83 \times 10^{-1}$ } \\
        \hline
        \hline
        \makecell{Previous {\old}\\ \cite{zhang2024rpnreconciledpolynomialnetwork}} 
        & \makecell{ $6.82 \times 10^{-1}$ } 
        & \makecell{ $3.17 \times 10^{-1}$ }
        & \makecell{ $4.06 \times 10^{-1}$ } \\
        \hline
        \hline
        \makecell{GCN} 
        & \makecell{ $8.15 \times 10^{-1}$ } 
        & \makecell{ $7.03 \times 10^{-1}$ } 
        & \makecell{ $7.90 \times 10^{-1}$ } \\
        \hline
        \hline
        \makecell{{\our}\\ (graph interdependence)}
        & \makecell{ $8.22 \times 10^{-1}$ } 
        & \makecell{\boldmath $7.09 \times 10^{-1}$ } 
        & \makecell{ $7.96 \times 10^{-1}$ } \\
        \hline
        \hline
        \makecell{{\our}\\ (hybrid interdependence)}
        & \makecell{\boldmath $8.30 \times 10^{-1}$ } 
        & \makecell{ $7.05 \times 10^{-1}$ }
        &  \makecell{\boldmath $8.02 \times 10^{-1}$ } \\
        \hline
    \end{tabular}
\end{table}


In Table~\ref{tab:graph_classification_benchmark_results}, we present the learning performance of {\our} and the baseline methods on the three graph benchmark datasets. For each dataset, the table also includes the model architectures, with the numbers inside brackets representing the input, hidden, and output dimensions, respectively.

The reported results show that the {\our} model, equipped with graph-based interdependence functions, achieves a significant performance improvement compared to both the previous {\old} model and the MLP model. Additionally, {\our} achieves comparable performance to the GCN model across these datasets. Furthermore, the table includes the learning performance of {\our} using a hybrid interdependence function, which combines the graph interdependence function with a bilinear interdependence function, using the product as the fusion function. The results indicate that the hybrid interdependence function slightly enhances the performance of {\our} on the Cora and Pubmed datasets, demonstrating the effectiveness of bilinear functions in capturing interdependence relationships within input data batches.

\subsubsection{Graph Interdependence Matrix Visualization}

\begin{figure}[t]
    \centering
    \begin{subfigure}{0.49\textwidth}
        \centering
        \includegraphics[width=\textwidth]{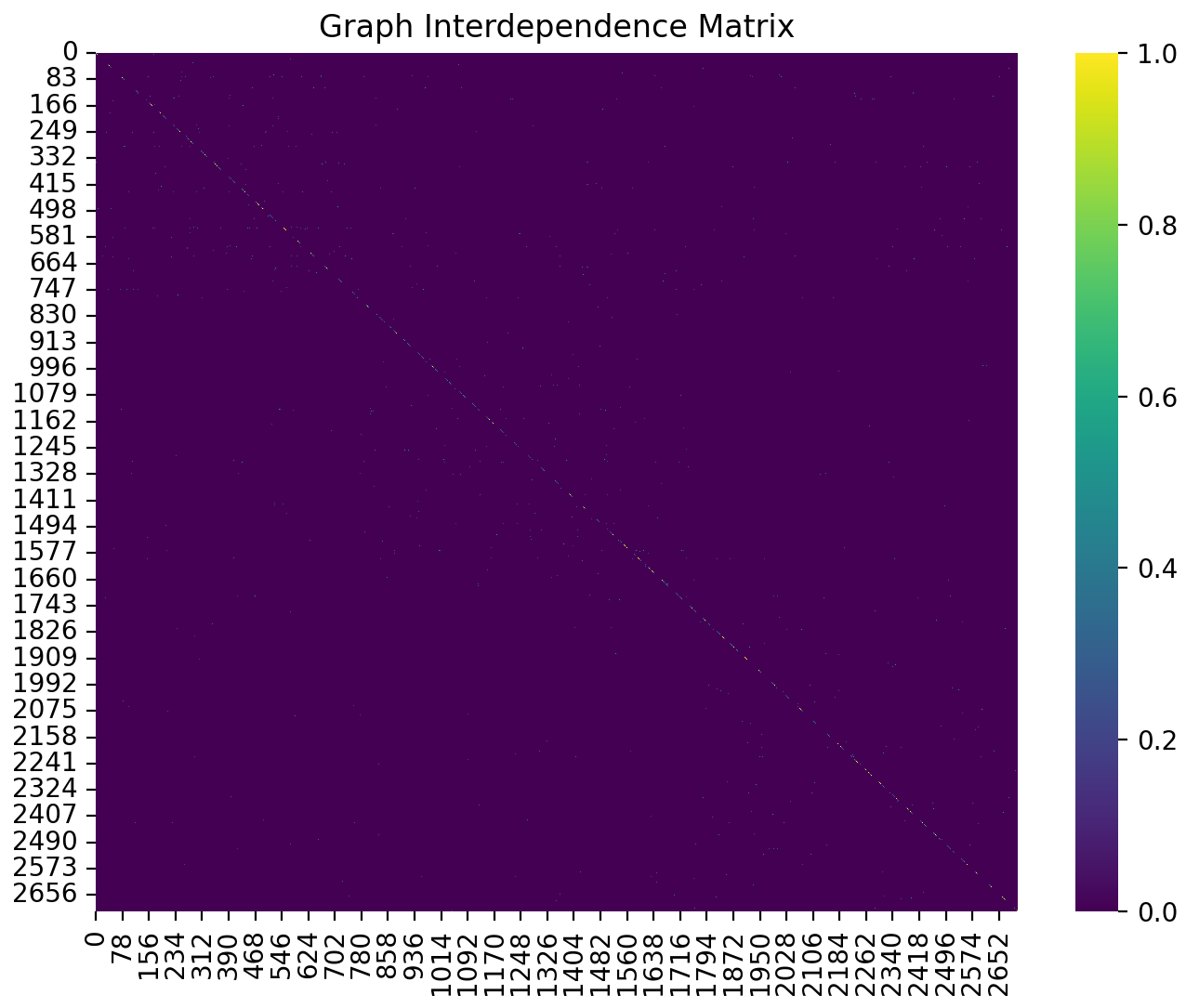}
        \caption{Graph Interdependence}
        \label{fig:graph_bilinear_interdependence_visualization_1}
    \end{subfigure}
    \begin{subfigure}{0.49\textwidth}
        \centering
        \includegraphics[width=\textwidth]{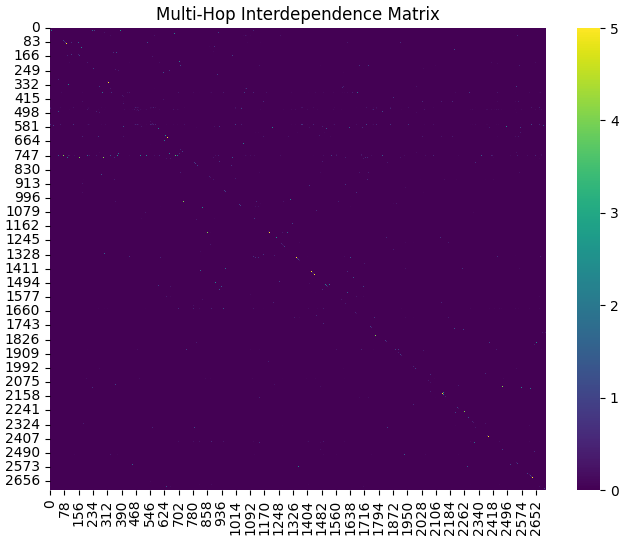}
        \caption{Multihop Interdependence}
        \label{fig:graph_bilinear_interdependence_visualization_2}
    \end{subfigure}
    \begin{subfigure}{0.49\textwidth}
        \centering
        \includegraphics[width=\textwidth]{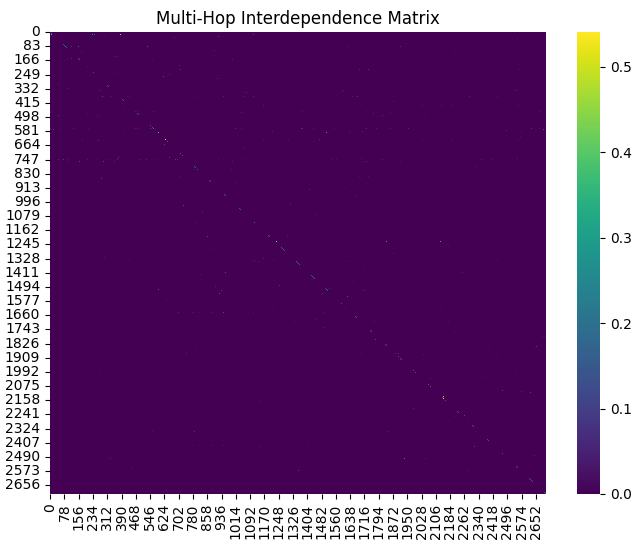}
        \caption{Pagerank Interdependence}
        \label{fig:graph_bilinear_interdependence_visualization_3}
    \end{subfigure}
    \begin{subfigure}{0.49\textwidth}
        \centering
        \includegraphics[width=\textwidth]{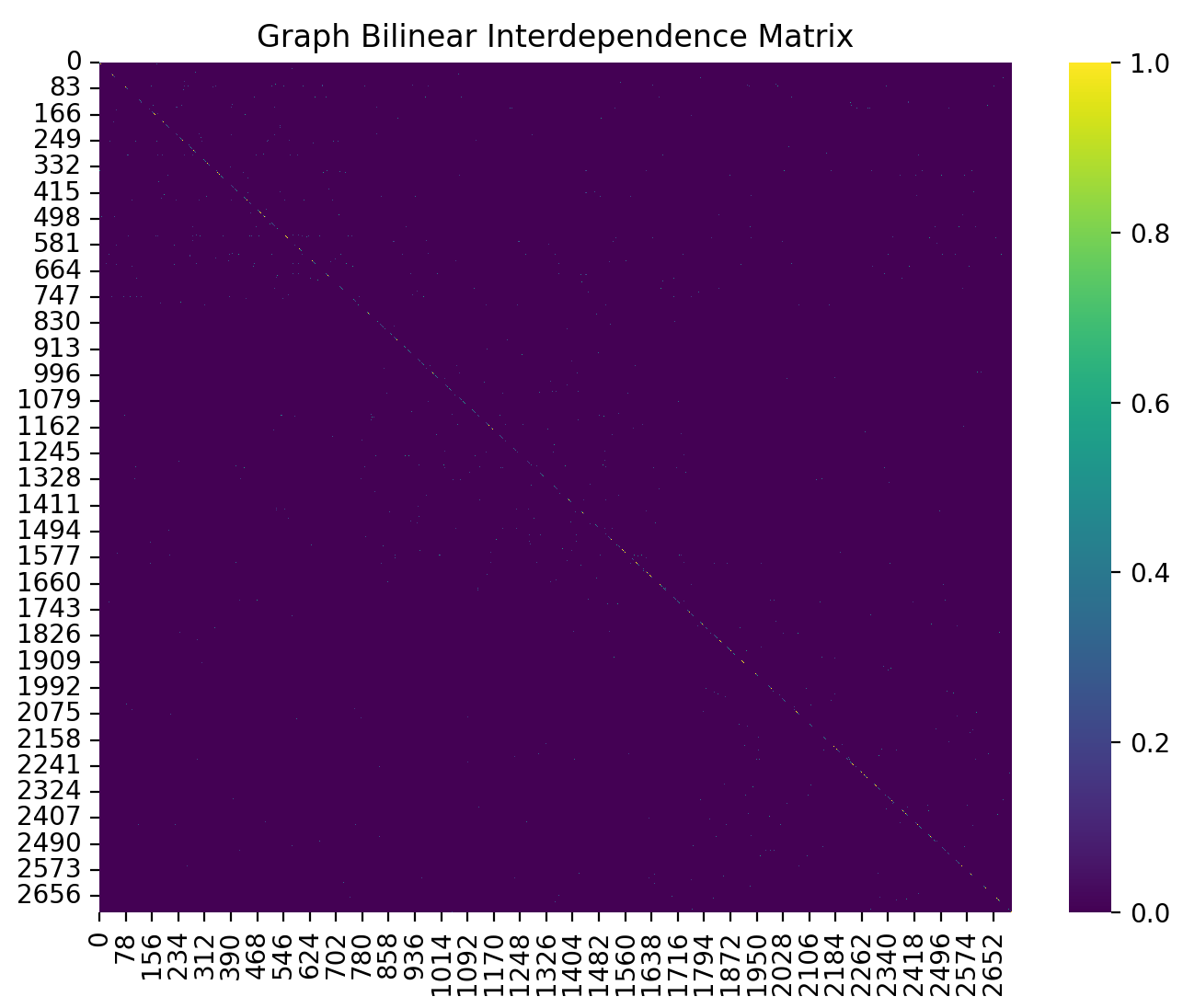}
        \caption{Hybrid Interdependence}
        \label{fig:graph_bilinear_interdependence_visualization_5}
    \end{subfigure}
    \caption{An illustration of the graph, multi-hop graph, PageRank, and hybrid interdependence matrices (combining graph and bilinear methods) computed on the Cora dataset is provided. The non-zero entry ratios of these matrices are $0.110\%$, $0.454\%$, $0.783\%$, and $0.110\%$, respectively. (For better visibility of the non-zero matrix entry values, please zoom in on the figure.)}
    \label{fig:graph_bilinear_interdependence_visualization}
\end{figure}

In Figure~\ref{fig:graph_bilinear_interdependence_visualization}, we display the graph-based interdependence matrices calculated using various graph structural interdependence functions. Specifically, Plot (a) is based on the graph adjacency matrix, Plots (b) and (c) represent the multi-hop and PageRank graph interdependence matrices, and Plots (d) and (e) correspond to the bilinear and hybrid graph interdependence matrices.

From the visualizations, the value scales of the entries in the multi-hop interdependence matrix are relatively larger compared to the other matrices. In terms of sparsity, the multi-hop and PageRank interdependence matrices have more non-zero entries than the one-hop and hybrid interdependence matrices. The hybrid interdependence matrix is constructed by masking the bilinear interdependence matrix with a binary graph interdependence matrix, where non-zero entries are column-normalized. Some of these matrices are highly sparse. For better visibility of the matrix entry values, it is recommended to zoom in on the figure to examine the sparsely distributed non-zero entries.


\section{Interpretation of {\our} with Interdependence Functions}\label{sec:interpretation}

Empirical evaluations provided in the previous section show that incorporating data interdependence functions significantly enhances the learning performance of {\our}, especially when compared to the previous {\old} model introduced in \cite{zhang2024rpnreconciledpolynomialnetwork}. In this section, we discuss interpretations of {\our} with data and structural interdependence functions from both theoretical machine learning and biological neuroscience perspectives. These interpretations also highlight the motivations and advantages of integrating interdependence functions within the {\our} model architecture.


\subsection{Theoretic Machine Learning Interpretations}

The {\our} model introduced in previous sections effectively captures interdependent relationships among both instances and attributes, making it adaptable to both \textit{inductive} and \textit{transductive} learning settings. In the following, we use the \textit{transductive} learning setting as an example to examine the theoretical learning performance of {\our} with instance interdependence functions; similar results also apply to {\our} with attribute interdependence functions in other learning settings.

Formally, in the transductive learning setting, let the dataset be $\mc{D} = \{\mb{x}_1, \mb{x}_2, \cdots, \mb{x}_n\}$, consisting of $n$ data instances, with a subset of these instances having known labels for model training, denoted as $\mc{T} = \{\mb{x}_1, \mb{x}_2, \cdots, \mb{x}_m\} \subset \mc{D}$. For each instance $\mb{x}_i \in \mc{T}$, its known label vector is represented by $f(\mb{x}_i)$. As introduced in \cite{zhang2024rpnreconciledpolynomialnetwork}, the learning loss incurred by the model $g(\cdot | \mb{w})$ on the dataset $\mc{D}$ can be formally represented as follows:

\begin{equation}
{\scriptsize 
\underbrace{\int_{\mb{x} \in \mc{D}} p(\mb{x}) \left\| g(\mb{x} | \mb{w}) - f(\mb{x}) \right\| \mathrm{d} \mb{x} }_{\text{overall error } \mc{L}}
=
\underbrace{\int_{\mb{x} \in \mc{T}} p(\mb{x}) \left\| g(\mb{x} | \mb{w}) - f(\mb{x}) \right\| \mathrm{d} \mb{x} }_{\text{empirical error } \mc{L}_{em}}
+
\underbrace{\int_{\mb{x} \in \mc{D} \setminus \mc{T}} p(\mb{x}) \left\| g(\mb{x} | \mb{w}) - f(\mb{x}) \right\| \mathrm{d} \mb{x} }_{\text{expected error } \mc{L}_{exp}},
}
\end{equation}

where $p(\mb{x})$ denoting the sampling probability for instance $\mb{x} \in \mc{D}$.

Motivated by recent work \cite{Esser2021LearningTC}, we aim to derive the generalization error bound of {\our} with interdependence functions to interpret its learning performance as follows:

\begin{equation}\label{equ:theoretic_interpretation_objective}
\mc{L}_{exp} \le \mc{L}_{em} + \text{generalization error bound}.
\end{equation}

The term ``\textit{generalization error bound}'' in the above equation can be derived based on various learning theories, such as VC Dimension \cite{doi:10.1137/1116025, 10.1145/76359.76371} and Rademacher Complexity \cite{10.5555/944919.944944}. These theories consider different factors—such as model architecture, component modules, and input data—in defining generalization error bounds, which will be discussed in detail below.


\subsubsection{Generalization Error-Bound based on VC Dimension}

To streamline the analysis, we consider a binary classification transductive learning problem using the {\our} model, which includes identity expansion, instance interdependence, identity reconciliation, zero remainder, and $K$ layers as an example.

Formally, given a data batch $\mb{X} \in \mathbbm{R}^{b \times d_0}$, where $d_0 = m$ represents the input dimension, we can represent the {\our} model as follows:

\begin{equation}
g_K \circ g_{K-1} \circ \cdots \circ g_1: \mathbbm{R}^{b \times d_0} \to \{+1, -1\}^b,
\end{equation}

where the $k_{th}$-layer defines a mapping $g_k: \mathbbm{R}^{b \times d_{k-1}} \to \mathbbm{R}^{b \times d_{k}}$, {\ie}

\begin{equation}
\begin{aligned}
g_k(\mb{X} | \mb{w}_{\psi, k}) &= \left \langle \kappa_{\xi, k} (\mb{X}), \psi_{k}(\mb{w}_{\psi, k}) \right \rangle + \pi_{k}(\mb{X}).
\end{aligned}
\end{equation}

The corresponding hypothesis class of the above model can be represented as
\begin{equation}
\mc{H} = \left\{ g(\mb{X} | \mb{w}, K) = g_K \circ g_{K-1} \circ \cdots \circ g_1: \mathbbm{R}^{b \times m} \to \{+1, -1\}^b \right\},
\end{equation}
which together with the output label ``sign'' function actually define the VC-dimension of the model:

\begin{equation}
\begin{aligned}
\text{VC-dimension}(\text{sign} \circ \mc{H}) &= \min \left\{m, \text{rank}(\mb{A}_{\xi_i}), \{d_k\}_{k \in \{1, \cdots, K\}} \right\} \\
&\le \min \left\{ \text{rank}(\mb{A}_{\xi_i}), d_{K-1} \right\}.
\end{aligned}
\end{equation}

Based on the above bound, as indicated in \cite{Esser2021LearningTC}, for any $\delta \in (0, 1)$, with probability $1-\delta$, the expected generalization error for any model $g \in  \text{sign} \circ \mc{H}$ satisfies

\begin{equation}
\mc{L}_{exp}(g) - \mc{L}_{em} (g) \le \sqrt{  \frac{8}{m} \left( \min \{ \text{rank}(\mb{A}_{\xi_i}), d_{K-1} \} \cdot \ln(em) + \ln \left(\frac{4}{\delta}\right) \right)}.
\end{equation}

The error bound derived above offers valuable insights into the {\our} model:

\begin{itemize}
\item On $\text{rank}(\mb{A}_{\xi_i}) \le b$: The previous {\old} model \cite{zhang2024rpnreconciledpolynomialnetwork}, which lacks the capability to model interdependence relationships, can be viewed as a special case of {\our} with an identity interdependence matrix, {\ie} $\mb{A}_{\xi_i} = \mb{I} \in \mathbbm{R}^{b \times b}$, yielding a default rank of $b$. This suggests that the new {\our} model—and all the backbone models discussed in previous sections that can be represented as {\our}—achieves a tighter error bound by incorporating interdependence than the previous {\old} model, along with all models previously discussed in \cite{zhang2024rpnreconciledpolynomialnetwork} that can be represented by {\old}.

\item On $\text{rank}(\mb{A}_{\xi_i}) \text{ v.s. } d_{K-1}$: If $d_{K-1} < \text{rank}(\mb{A}_{\xi_i})$, the introduced interdependence function becomes redundant; however, if $d_{K-1} \ge \text{rank}(\mb{A}_{\xi_i})$, the interdependence function reduces the model's error bound. In other words, the model architecture itself influences the effectiveness of the interdependence function.
\end{itemize}

\subsubsection{Generalization Error-Bound based on Rademacher Complexity}

Compared to the classic VC dimension theory, Rademacher complexity analysis provides a data-dependent bound for broader hypothesis classes. Building on the {\our} model discussed above, we derive its generalization error bound based on Rademacher complexity.

Formally, consider the constrained hypothesis classes $\mc{H}^{\beta, \omega} \subset \mc{H}$, where model parameters satisfy $\left\| \mb{w}k \right\| {\infty} \le \omega$ and the optional bias term satisfies $\left\| \mb{b}_k \right\|_1 \le \beta$ for each layer $k \in {1, 2, \cdots, K}$. As analyzed in \cite{Esser2021LearningTC}, the Rademacher complexity of this restricted hypothesis class can be represented as follows:

\begin{equation}\label{equ:rademacher_complexity}
\mc{R}_{m, n} ( \mc{H}^{\beta, \omega} ) \le \frac{c_1 n^2}{m(n-m)} \left( \sum_{k=0}^{K-1} c_2^k \left\| \mb{A}_{\xi_i} \right\|_{\infty}^k \right) + c_3 c_2^K \left\| \mb{A}_{\xi_i} \right\|_{\infty}^K \left\| \mb{A}_{\xi_i} \mb{X} \right\|_{2 \to \infty} \sqrt{\log(n)},
\end{equation}

where $c_1 = 2 L \beta$, $c_2 = 2 L \omega$, $c_3 = L \omega \sqrt{\frac{2}{d}}$, and $L$ is the Lipschitz constant of the layer's optional activation function. Furthermore, following the derivations in \cite{ElYaniv2007TransductiveRC}, for any $\delta \in (0, 1)$, with probability $1 - \delta$, the generalization error for any model $g \in \mc{H}^{\beta, \omega}$ satisfies the following:

\begin{equation}
\mc{L}_{exp}(g) - \mc{L}_{em}(g) \le \mc{R}_{m, n} ( \mc{H}^{\beta, \omega} ) + c_4 \frac{n \sqrt{ \min \{ m, n-m \}} }{m(n-m)} + c_5 \sqrt{ \frac{n}{m(n-m)} \ln \left( \frac{1}{\delta} \right) },
\end{equation}

where $c_4$ and $c_5$ are constants with the value bounds: $c_4 < 5.05$ and $c_5 < 0.8$.

Based on the generalization error bound above, we gain insights, particularly from Equation~(\ref{equ:rademacher_complexity}), where both $\left\| \mb{A}_{\xi_i} \right\|_{\infty}$ and $\left\| \mb{A}_{\xi_i} \mb{X} \right\|_{2 \to \infty}$ influence the tightness of the error bound:

\begin{itemize}

\item On the norm term $\left\| \mb{A}_{\xi_i} \right\|_{\infty}$: This norm of the interdependence matrix $\mb{A}_{\xi_i}$ reflects the largest absolute row sums of the matrix. Therefore, to tighten the error bound, normalizing the matrix $\mb{A}_{\xi_i}$ may be necessary in practical applications.

\item On the norm term $\left\| \mb{A}_{\xi_i} \mb{X} \right\|_{2 \to \infty}$: This norm of the interdependence matrix represents $\sup_{\left\| \mb{z} \right\|_2 = 1} \left\| \mb{A}_{\xi_i} \mb{X} \mb{z} \right\|_{\infty}$, which can be relaxed as follows:

\begin{equation}
\left\| \mb{A}_{\xi_i} \mb{X} \right\|_{2 \to \infty} = \max_j \left\| (\mb{A}_{\xi_i} \mb{X})_{\cdot, j} \right\|_{2} \le \left\| \mb{A}_{\xi_i} \right\|_{2} \max_j  \left\| \mb{X}_{\cdot, j} \right\|_{2} \le \left\| \mb{A}_{\xi_i} \right\|_{2} \left\| \mb{X} \right\|_{2 \to \infty}.
\end{equation}

In other words, both the matrix $\mb{A}_{\xi_i}$ and the input data batch $\mb{X}$ affect the error bound. Normalizing $\mb{A}_{\xi_i}$ can effectively reduce its $2$-norm term $\left\| \mb{A}_{\xi_i} \right\|_{2}$. As for $\left\| \mb{X} \right\|_{2 \to \infty}$, data batches with smaller row norms, balanced row magnitudes, or properties such as sparsity, low rank, and orthogonality result in smaller norm values.

\end{itemize}

\subsection{Biological Neuroscience Interpretations}

In addition to theoretical machine learning interpretations, the interdependence function components in the {\our} model architecture may also simulate certain compensatory functions of biological nervous systems, including (but not limited to) sensory integration in the brain cortex, relational memory in the hippocampus, and working attention mechanisms.

\subsubsection{Understanding Interdependence from Sensory Integration at Brain Cortex}

The brain integrates interdependent information from multiple sensory inputs (such as vision, hearing, and touch) in multimodal areas of the cortex, including the parietal and prefrontal cortices. These regions serve as hubs where sensory inputs from diverse sources converge, enabling the brain to form unified representations of objects and events. During sensory information integration in the cortex, the brain naturally captures relationships among these multimodal sensory sources.

\begin{wrapfigure}{r}{0.5\textwidth}
  \begin{center}
  \vspace{-20pt}
    \includegraphics[width=0.5\textwidth]{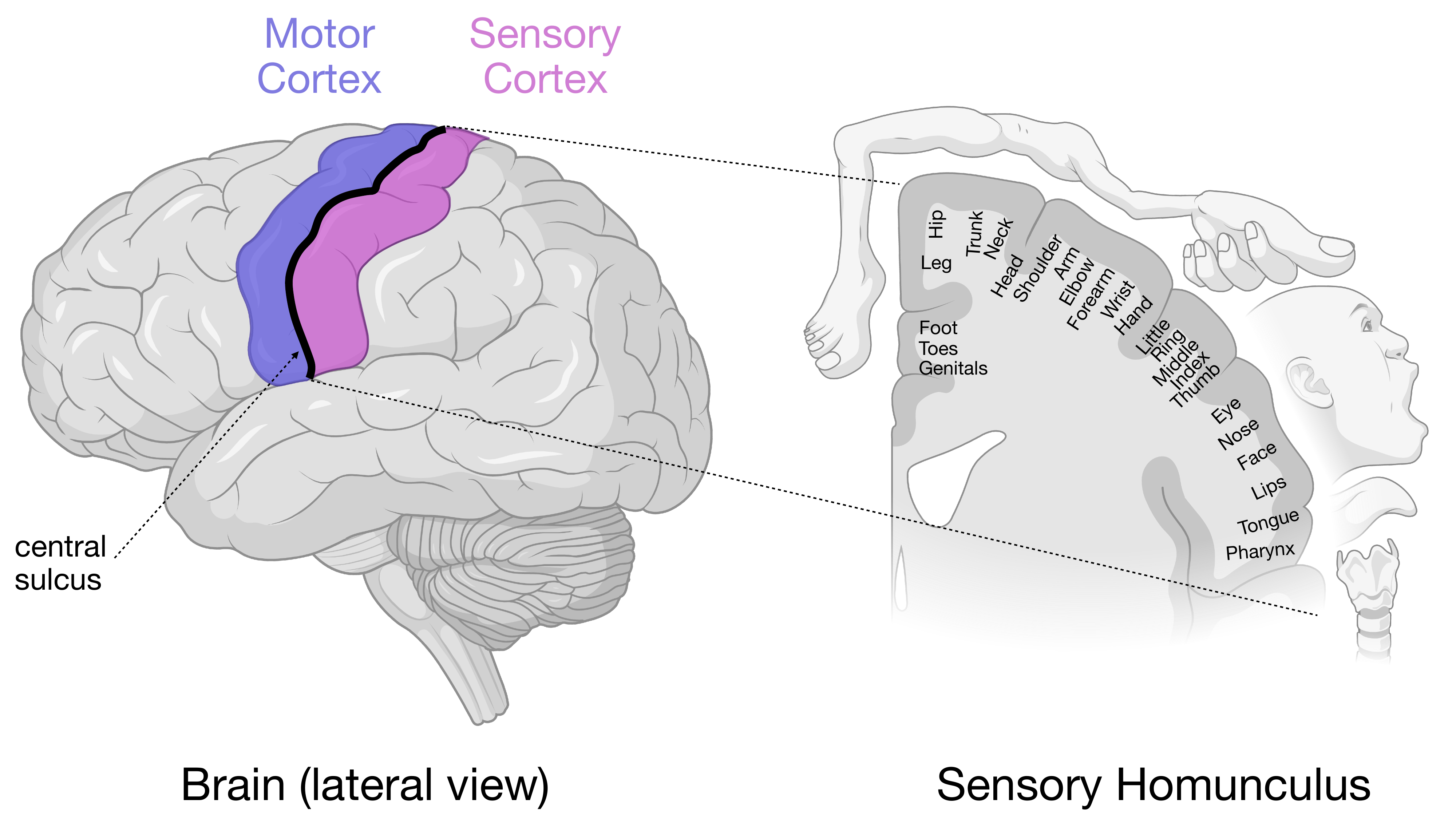}
  \end{center}
  \vspace{-7pt}
  \caption{Brain Sensory Cortex.}\label{fig:sensory_cortex}
  \vspace{-8pt}
\end{wrapfigure}

As shown in Figure~\ref{fig:sensory_cortex}, we illustrate a lateral view of the human brain, highlighting its four main lobes: the frontal, parietal, temporal, and occipital lobes (also illustrated in Figure~\ref{fig:cortex_thalamus}). The frontal and parietal lobes are separated by the central sulcus, a groove between tissues, where the sensory cortex (in pink) and motor cortex (in purple) are located adjacent to each other on either side of the sulcus. In the right panel of Figure~\ref{fig:sensory_cortex}, we see the primary sensory cortex with an orderly (inverted) tactile representation, extending from the toe (at the top left of the cerebral hemisphere) to the mouth (at the bottom right). Each cerebral hemisphere in the primary somatosensory cortex represents tactile sensations from the opposite (contralateral) side of the body. 

The amount of primary somatosensory cortex dedicated to a body part is not proportional to the size of the body surface but, rather, to the relative density of tactile receptors. Body parts with a higher density of tactile receptors, such as the lips and hands, occupy larger areas in the somatosensory cortex, reflecting their heightened sensitivity to tactile stimulation.

With this diverse and extensive sensory information, the brain rarely processes inputs independently. Instead, it fuses them, captures interdependent relationships, and generates an optimal response for the motor systems. This integration and processing of multimodal sensory inputs mirror how structured models in machine learning capture interdependencies across different data attributes and instances.

\subsubsection{Understanding Interdependence from from Hippocampus and Relational Memory}

The hippocampus plays a crucial role in relational memory, enabling the brain to encode relationships between objects, contexts, and spatial locations. This function is especially important for integrating interdependencies across time and space, allowing the brain to learn sequences, temporal patterns, and associations between events.

\begin{wrapfigure}{r}{0.5\textwidth}
  \begin{center}
    \vspace{-20pt}
    \includegraphics[width=0.5\textwidth]{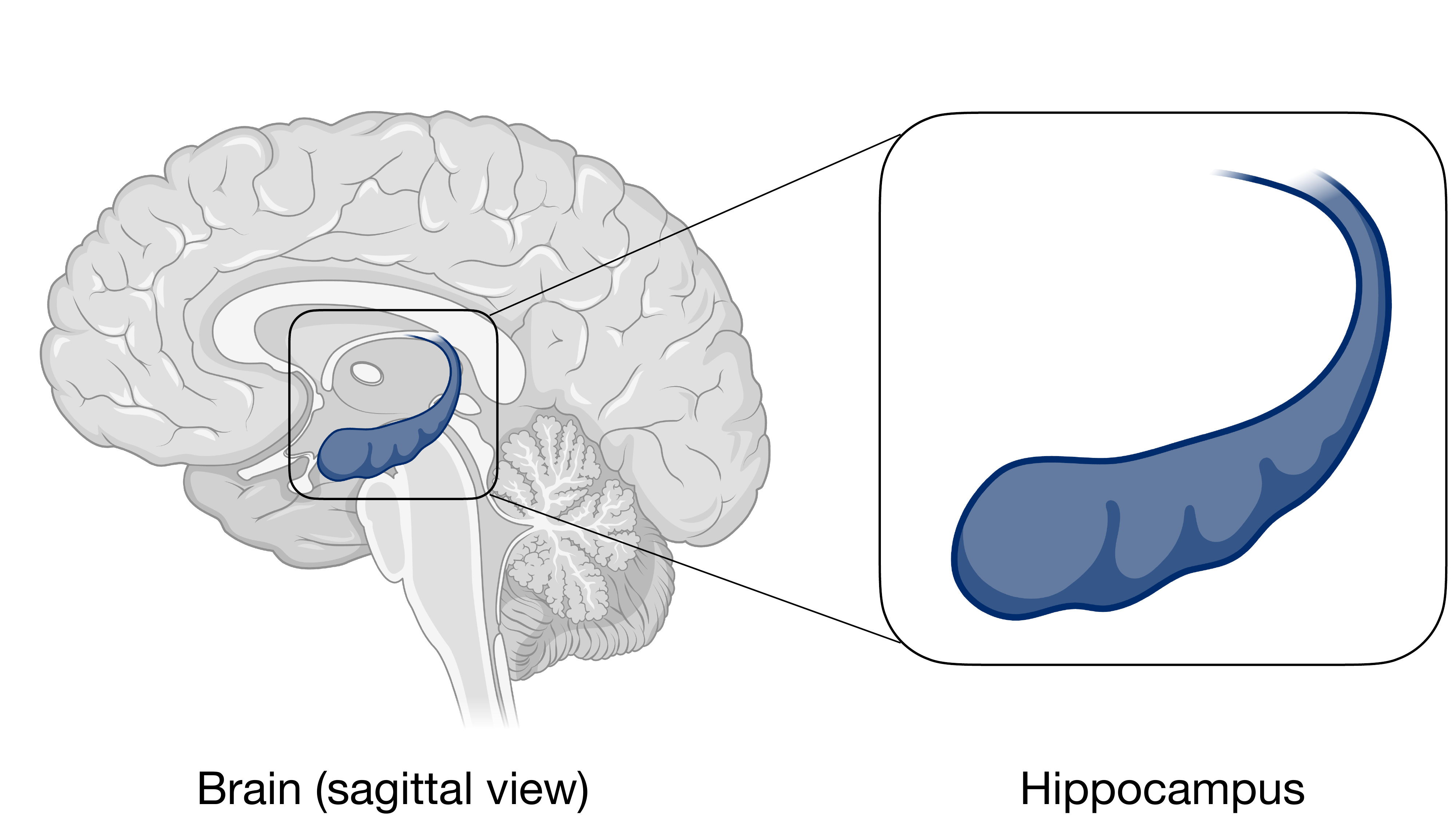}
  \end{center}
    \vspace{-7pt}
  \caption{Brain Hippocampus.}\label{fig:hippocampus}
    \vspace{-8pt}
\end{wrapfigure}

As illustrated in Figure~\ref{fig:hippocampus}, the hippocampus is located within the temporal lobe (see also Figure~\ref{fig:cortex_thalamus}). Key structures within the temporal lobe that support long-term memory include the hippocampus and surrounding regions, such as the perirhinal, parahippocampal, and entorhinal cortices. The hippocampus, along with the dentate gyrus, has a curved shape reminiscent of a seahorse. Humans and other mammals possess two hippocampi, one on each side of the brain. As part of the limbic system, the hippocampus plays a central role in consolidating information from short-term to long-term memory and in spatial memory, which supports navigation. In addition to the hippocampus, regions like the prefrontal and visual cortices also contribute to explicit memory; however, these will not be discussed in this paper.

The memory mechanism, supported by the hippocampus and other cortical areas, enables the brain to encode, store, retain, and retrieve information and past experiences, capturing interdependencies both explicitly and implicitly among instances and attributes. For example, during spatial navigation, the hippocampus integrates environmental information (attributes) with specific locations or landmarks (instances), allowing for memory retrieval based on complex interdependencies between these elements. This is analogous to how the {\our} model captures interdependence in structured data through the data and structural interdependence functions introduced in this paper.

\subsubsection{Understanding Interdependence from Attention Mechanisms}

In addition to sensory information integration and memory, attention mechanisms in the brain—particularly involving the prefrontal cortex and posterior parietal cortex—enable selective processing of interdependent information by dynamically modulating neural activity. By focusing on specific attributes or instances while filtering out irrelevant details, the brain can prioritize important interdependent relationships, such as the correlation between auditory and visual stimuli within a given context.

\begin{wrapfigure}{r}{0.5\textwidth}
  \begin{center}
    \vspace{-20pt}
    \includegraphics[width=0.5\textwidth]{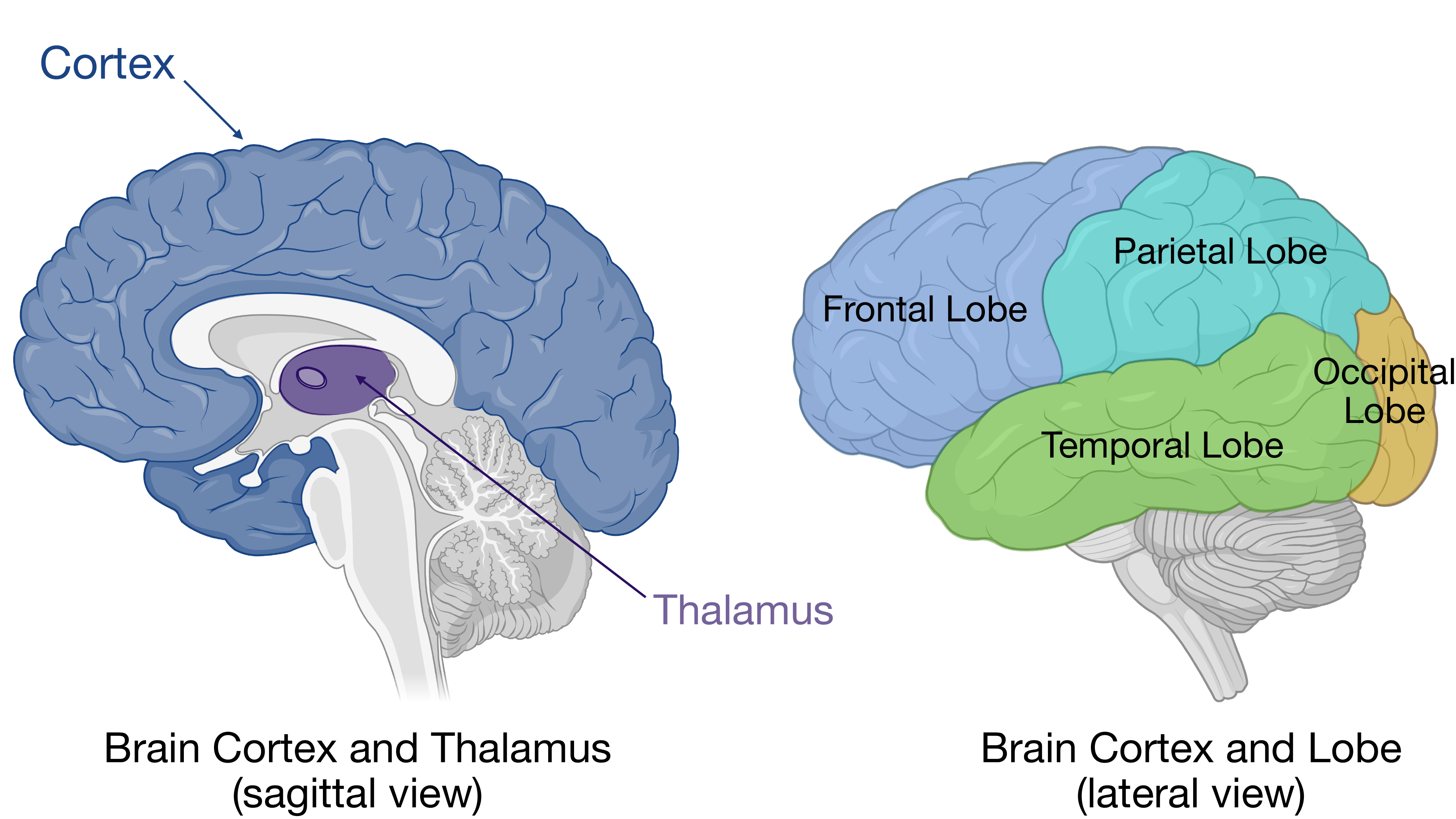}
  \end{center}
    \vspace{-7pt}
  \caption{Brain Cortex and Thalamus.}\label{fig:cortex_thalamus}
    \vspace{-8pt}
\end{wrapfigure}

Attention is the behavioral and cognitive process of selectively focusing on specific aspects of information—whether subjective or objective—while disregarding other perceivable information. This process involves a complex network of brain regions that work together to regulate and sustain focus. Key areas include the prefrontal cortex, parietal cortex, anterior cingulate cortex, thalamus, and basal ganglia, some of which are shown in Figure~\ref{fig:cortex_thalamus}. In particular, the thalamus, the purple region depicted in the figure, consists of two oval collections of nuclei that make up most of the diencephalon’s mass. Often described as a relay station, the thalamus directs nearly all sensory information (except for olfactory signals) to the cortex, making an initial stop in the thalamus before reaching its cortical destination. The thalamus is subdivided into specialized nuclei, each handling specific types of sensory information, which it routes to the appropriate area in the cortex for further processing.

Supported by the thalamus and other cortical regions, the brain’s attention mechanisms are essential for capturing interdependencies among sensory information and sequential events. Attention enables the brain to prioritize relevant information, filter out distractions, and integrate sensory inputs and events over time—key to understanding relationships between different stimuli.


\section{Intellectual Merits, Limitations and Future Work Timeline}\label{sec:discussion}

In this section, we discuss the intellectual merits of the {\our} model equipped with interdependence functions. Additionally, we address the current model's limitations, highlighting potential directions for future development and the next phase of this project.

\subsection{Intellectual Merits}

In this paper, we propose a redesign of the {\our} model architecture by incorporating data interdependence functions that model relationships among both attributes and instances. Based on empirical experiments and theoretical analyses provided in previous sections, these interdependence functions significantly enhance the learning performance of the {\our} model. Below, we summarize the intellectual merits of the newly proposed {\our} model.

\noindent \textbf{Theoretical Merits}: Unlike the previous {\old} model \cite{zhang2024rpnreconciledpolynomialnetwork}, which assumes attributes and instances are independent and identically distributed, the newly designed {\our} model effectively captures interdependent relationships among both attributes and instances through a set of interdependence functions defined using information from the input data batch. These data interdependence functions significantly expand the modeling capabilities of {\our} for complex function learning tasks on interdependent data. The theoretical analyses provided in this paper also offer insights into defining optimal interdependence functions that lead to tighter generalization error bounds based on both VC dimension and Rademacher complexity theories. From a biological neuroscience perspective, these interdependence functions emulate certain compensatory functions of the biological nervous system within the {\our} model.

\noindent \textbf{Technical Merits}: The {\our} model introduces a diverse family of interdependence functions that capture relationships among attributes and instances from various perspectives, including the input data batch, underlying topological and geometrical structures, and combinations of these. These interdependence functions enable unification of diverse backbone models, such as CNNs, RNNs, GNNs, and Transformers (and their variants), as discussed in this paper. We demonstrate that these backbone models share similar architectures, with key differences arising from how the interdependence functions are defined. These observations provide valuable insights for designing the future ``Transformer-Next'' new backbone models.

\noindent \textbf{Computational Merits}: The interdependence functions introduced here compute matrices to model relationships among instances and attributes. These computed interdependence matrices are typically small and often sparse across many learning settings, resulting in minimal storage requirements. Additionally, because interdependence functions operate on the data batch as input—applying attribute interdependence prior to data transformation—they save considerable computational time compared to operations performed on transformed data batches. As with the previous paper \cite{zhang2024rpnreconciledpolynomialnetwork}, the {\our} model architecture allows for computations of different component functions to be distributed across multiple chips, machines, or cloud platforms, enhancing learning efficiency and safeguarding data privacy and model parameter security.


\subsection{Limitations}

When implementing the {\our} architecture and its component modules, we identified several limitations, particularly in modeling capabilities for dynamic data, learning algorithms, and the potential deployment of large-scale intelligent systems.

\noindent \textbf{Learning Limitations}: One challenge encountered in the implementation of the {\our} model lies in adapting loss functions, optimizers, and error backpropagation-based learning algorithms. While conventional representation learning-based loss functions, optimizers, and algorithms are applicable to the current function learning task, they exhibit certain inconsistencies with function-oriented learning models. In both this paper and the previous work \cite{zhang2024rpnreconciledpolynomialnetwork}, our function learning models require compressing the input data batch into the output space using perceptron-based layers (involving identity data transformation, identity reconciliation, and zero remainder functions). Although such layers are necessary for traditional representation learning models, they perform minimal data transformation in function learning models, significantly reducing the data batch’s representational capabilities as dimensions are compressed.

\noindent \textbf{Modeling Limitations}: This paper successfully unifies several dominant backbone models within the canonical {\our} representation. However, challenges remain in representing RNN models with {\our}, as the current model lacks dynamic processing capabilities, requiring temporal interdependence in RNNs to be converted into spatial interdependence instead. This highlights a limitation of {\our} in modeling ``temporal dynamics'', a key factor in developing future world models with spatial intelligence. To address this, we plan to redesign modules in the current {\our} model to enable ``temporal dynamics'' modeling, which will be a primary focus of our future work.

\noindent \textbf{Large-Scale Intelligent Systems}: To demonstrate the effectiveness of our proposed techniques, we aim to build a large-scale intelligent system based on the {\our} model, showcasing its multimodal modeling capabilities, learning performance advantages, and inherent parameter efficiency. Creating such systems will require redesigning many component functions and models to ensure learning efficiency when handling large-scale data and vast numbers of parameters. Once the above limitations in learning and modeling are addressed, we will initiate the large-scale intelligent system project, which will be developed in parallel with the current {\our}-based backbone framework project.


\subsection{Future Work Timeline}

In the upcoming year of 2025, we plan to address the current limitations of the {\our} model and the {\toolkit} library. Our primary focus will be on refining the {\our} model framework and developing application projects that leverage the enhanced {\our} model and {\toolkit} toolkit.

\noindent \textbf{Framework Enhancement Projects}: Based on our current development pace, we estimate to spend another six months on addressing the learning limitations in the current {\our} model. Specifically, we aim to explore function learning-oriented loss functions, optimized objective functions, and model learning algorithms. A new version of {\our}, featuring these learning-related enhancements, is expected for release by mid-2025. Additionally, we plan to incorporate ``time'' and related functions into the {\our} model to enable dynamic learning scenarios. Modeling ``temporal dynamics'' will require a substantial redesign of the current {\our} functions and components, which will be time-intensive. Following the resolution of learning limitations, we anticipate another six months to integrate ``temporal dynamics'' into the model architecture, with a target release by the end of 2025.

\noindent \textbf{Application Projects}: In parallel with addressing the challenges in learning and modeling, we plan to undertake several system-building projects to conduct preliminary testing of {\our} on large-scale datasets with one or a few modalities. We expect to initiate a large-scale intelligent system project by the end of 2025, a project that will likely take several years due to its anticipated challenges and complexities. Technical reports on the progress of the {\our} and intelligent system projects will be released as new developments become available.

\section{Related Work}\label{sec:related_work}

In this section, we briefly discuss existing work related to this paper, including various backbone models proposed for data across different modalities and recent advancements in multi-modality foundation model learning.

\subsection{Related Backbone Models}

With the integration of data interdependence functions, the {\our} model proposed in this paper has demonstrated the capability to unify several dominant backbone models, including convolutional neural networks, recurrent neural networks, graph neural networks, and Transformers. In this section, we briefly introduce these backbone models and highlight key papers that have contributed critical technical breakthroughs.

\subsubsection{Convolutional Neural Networks}

Originally defined in the 19th century in Fourier's work \cite{Fourier1822}, convolution has become a widely used mathematical operator in both continuous signal processing and discrete image processing. In traditional computer vision, convolution is extensively applied in tasks such as image blurring, sharpening, resizing, and edge detection. Manually defined convolution kernels can provide meaningful physical interpretations but may have limited applicability for diverse vision tasks. To address this limitation, Yann LeCun introduced the convolutional neural network (CNN), known as LeNet \cite{726791}, proposing that convolution kernels be defined as learnable parameters that can be automatically learned from input data. These learnable convolution kernels significantly improve the learning performance of vision models \cite{OShea2015AnIT} and expand their applicability to various vision-related tasks \cite{cnn_model, 8308186}. Following LeNet, additional CNN variants have been developed, such as AlexNet \cite{NIPS2012_c399862d} and VGG16 \cite{Simonyan2014VeryDC}. AlexNet, with its 5 convolutional layers, achieved 84.6\% accuracy on ImageNet, while VGG16, with 13 convolutional layers, further boosted accuracy to 92.7\%. However, as model architectures became deeper, performance degradation emerged, where deeper models tended to show higher training and testing errors than shallower ones. To address this issue, Kaiming He introduced ResNet \cite{He2015DeepRL}, incorporating skip-layer residual connections into the CNN architecture, which increased performance to 96.4\% on ImageNet.

\subsubsection{Recurrent Neural Networks}

The study of recurrent neural networks (RNNs) originated from research in associative memory. Frank Rosenblatt \cite{Rosenblatt1960} introduced a three-layer perceptron neural network with recurrent connections in the middle layer. Another foundation of associative memory came from statistical mechanics with the Ising model \cite{RevModPhys.39.883}, which modeled thermal equilibrium. Later, Roy J. Glauber \cite{10.1063/1.1703954} extended the Ising model by adding a time component, allowing for temporal evolution. Building on the Ising model, Sherrington and Kirkpatrick developed the Sherrington-Kirkpatrick model as an exactly solvable model of spin glass, featuring an energy function with multiple local minima. Based on this model, John Hopfield introduced the Hopfield network with binary activation functions \cite{Hopfield1982}, later extended to continuous activation functions \cite{Hopfield1984}. In recent years, Hopfield has further investigated methods to increase memory storage capacity by modifying network dynamics and energy functions \cite{Krotov2016DenseAM, Krotov2020LargeAM, Ramsauer2020HopfieldNI}. With the resurgence of neural networks in the 1980s, new RNN architectures, such as the Jordan network \cite{JORDAN1997471} and Elman network \cite{elmannet}, emerged to study cognitive psychology. To address gradient explosion and vanishing issues, Hochreiter and Schmidhuber \cite{10.1162/neco.1997.9.8.1735} introduced the Long Short-Term Memory (LSTM) network, which has since become the dominant RNN model. To model bidirectional dependencies in input sequences, the Bidirectional RNN (Bi-RNN) \cite{650093} was developed, using two RNNs to process inputs in different directions. For a simplified LSTM structure, the Gated Recurrent Unit (GRU) \cite{Cho2014LearningPR} was introduced as an alternative RNN architecture. More recently, Gu and Dao proposed the Mamba state space model \cite{Gu2023MambaLS, Dao2024TransformersAS}, which incorporates recurrent models for sequence data by updating states over time.

\subsubsection{Graph Neural Networks}

Unlike images and language, graphs, as topological structures, lack a fixed order of nodes \cite{Garg2020GeneralizationAR}, necessitating distinct model designs. Before the advent of graph neural networks (GNNs), graph learning algorithms were primarily developed based on topological structures. Yan and Han proposed gSpan \cite{1184038}, a graph-based substructure pattern mining algorithm that extracts frequent subgraphs for feature learning. Sun introduced path-based algorithms \cite{10.14778/3402707.3402736} to extract features from graphs for various learning tasks. With the rise of neural networks, unsupervised algorithms for learning graph representations emerged, such as DeepWalk \cite{Perozzi2014DeepWalkOL} and node2vec \cite{10.1145/2939672.2939754}. To effectively leverage labeled node information, the Graph Convolutional Network (GCN) \cite{Kipf2016SemiSupervisedCW} was developed to learn node representations by aggregating information from neighboring nodes. The Graph Attention Network (GAT) \cite{Velickovic2017GraphAN} further enhanced GCN by incorporating an attention mechanism based on pairwise node representations. However, as GNN architectures deepen, issues such as the over-smoothing problem \cite{Li2018DeeperII} and the suspended animation problem \cite{Zhang2019GResNetGR} can arise. Techniques such as edge dropout \cite{Rong2019DropEdgeTD} and graph residual learning \cite{Zhang2019GResNetGR} have been proposed to mitigate these issues. Inspired by the effectiveness of Transformer models in language processing, researchers have recently explored Transformer-based GNN models, including Graph-Transformer \cite{Yun2019GraphTN}, GPT-GNN \cite{Hu2020GPTGNNGP}, and Graph-BERT \cite{Zhang2020GraphBertOA}.

\subsubsection{Attention and Transformer}

The concept of attention was first introduced by Google DeepMind in \cite{Mnih2014RecurrentMO} to model adaptive selection of high-resolution image regions. Around the same time, Bengio and collaborators applied attention mechanisms in machine translation models \cite{Bahdanau2014NeuralMT}. Following these initial explorations, attention mechanisms have proven effective across a range of deep learning tasks and are widely implemented in various deep learning models \cite{Xu2015ShowAA, yang-etal-2016-hierarchical, Wang2017ResidualAN, Bello2019AttentionAC}. Based on scaled dot-product attention, researchers at Google introduced the Transformer model \cite{Vaswani2017AttentionIA} in 2017, which represents language data using pairwise self-attention scores derived from inputs. Shortly after, Google launched the pre-trained BERT (Bidirectional Transformer) model \cite{Devlin2019BERTPO} for language understanding in 2018. Around the same time, other Transformer-based language models were proposed by teams from different organizations, including GPT from OpenAI \cite{Radford2018ImprovingLU} and BART from Meta \cite{Lewis2019BARTDS}. Today, Transformer has become the dominant backbone for large language models, with model sizes expanding significantly in recent years. For example, the initial GPT-1 model had only 117 million parameters, while the recent GPT-4 model has grown to 1.8 trillion parameters. Inspired by the success of Transformer models in natural language processing, researchers have also explored applying Transformer architectures to other fields, including ViT \cite{Dosovitskiy2020AnII} for image processing, Graph-BERT \cite{Zhang2020GraphBertOA} for graph learning, and DiT \cite{Peebles2022ScalableDM} for image and video generation.

\subsection{Multi-Modality Foundation Models}

The {\our} model introduced in this paper can also serve as a foundation model applicable to learning tasks across datasets with different modalities, aligning closely with current advancements in multi-modality foundation model learning.

\subsubsection{Multi-Modality Data Alignment}

Multi-modal data representation learning aims to integrate and interpret heterogeneous data types ({\eg} images, text, audio, video, and sensor data) through unified representations. OpenAI introduced Contrastive Language-Image Pretraining (CLIP) \cite{Radford2021LearningTV}, which correlates images and language and enables zero-shot image classification. Google Research proposed ALIGN \cite{Jia2021ScalingUV}, which aligns visual and textual features by training on large-scale image-text pairs with contrastive objectives. Another approach focuses on learning hierarchical representations that capture both intra- and inter-modal relationships. The VATT model (Video-Audio-Text Transformer) \cite{Akbari2021VATTTF} employs self-supervised learning across multiple modalities to capture high-level correlations. Similarly, recent advancements in large-scale foundation models, such as Unified-IO \cite{Lu2022UnifiedIOAU}, aim to unify inputs from multiple modalities within a common encoder-decoder framework, enhancing generalization across modalities. Additional models, such as LXMERT \cite{Tan2019LXMERTLC}, Unicoder-VL \cite{Li2019UnicoderVLAU}, Oscar \cite{Li2020OscarOA}, and ViLBERT \cite{Lu2019ViLBERTPT}, have also been proposed for multi-modal representation learning, particularly for vision-language tasks. Recently, Meta introduced MetaCLIP \cite{Xu2023DemystifyingCD}, pre-trained on a larger image-text dataset, which outperforms CLIP on zero-shot image classification tasks.

\subsubsection{Multi-Modality Tokenizer}

Tokenization is essential in multi-modal learning, as it converts raw inputs from different modalities into token representations that machine learning models can process. Extending the concept of tokenization from natural language processing (NLP) to a multi-modal context has driven innovations in representing complex, multi-source input data. A key development is the introduction of multi-modal tokenizers that handle diverse inputs, such as text, images, and videos, enabling downstream tasks like multi-modal fusion. Early models, such as VisualBERT \cite{Li2019VisualBERTAS}, combined pre-trained BERT embeddings with visual features by treating images as sequences of region tokens derived from object detection models like Faster R-CNN. More recent models, such as BEiT-3 \cite{Wang2022ImageAA} and FLAVA \cite{Singh2021FLAVAAF}, expand on this concept by introducing universal tokenizers that map images, text, and audio into a unified token space. Other advances in multi-modal tokenization, such as Pix2Seq \cite{Chen2021Pix2seqAL}, represent images as sequences of tokens, facilitating the adaptation of NLP-like transformers to the vision domain. With the rise of large language models, recent research has proposed converting video data into discrete tokens \cite{Yu2022MAGVITMG} to enable reasoning in language models. For instance, VideoPoet \cite{Kondratyuk2023VideoPoetAL} employs video tokenizers to transform multi-modal data into discrete tokens for video generation.

\subsubsection{Multi-Modality Generative Models}

The field of generative modeling has advanced rapidly, with recent models demonstrating impressive capabilities in data synthesis. The denoising diffusion probabilistic model \cite{Ho2020DenoisingDP} has proven effective in generating image data and serves as a foundational model for the current surge in generative AI. To improve efficiency, the latent diffusion model \cite{Rombach2021HighResolutionIS} performs generation within the latent space using cross-attention mechanisms. The DiT model \cite{Peebles2022ScalableDM} further introduces a scalable diffusion approach with Transformers as the backbone. Building on these diffusion models, multi-modal generative models can produce coherent outputs (e.g., text, images, or video) from multi-modal inputs by learning cross-modal relationships. Models like DALL-E \cite{Ramesh2021ZeroShotTG} and Imagen \cite{Saharia2022PhotorealisticTD} leverage text-to-image generation via diffusion, pushing the boundaries of visually rich outputs conditioned on text. GLIDE \cite{Nichol2021GLIDETP} explores text-guided image generation and editing with diffusion models. Recently, Stability AI released Stable Diffusion 3 \cite{Esser2024ScalingRF}, which incorporates scaling rectified flow Transformers. These models not only generate high-quality images but also demonstrate creativity and abstraction across modalities. Beyond text-to-image generation, video generation has gained focus in the community. Make-A-Video \cite{Singer2022MakeAVideoTG}, Dreamix \cite{Molad2023DreamixVD}, and Video Diffusion \cite{Ho2022VideoDM} extend diffusion concepts to video by leveraging pretrained text-to-image models with additional motion learning. Using the latent alignment algorithm \cite{Blattmann2023AlignYL}, Stability AI has recently expanded their image generation model for video \cite{Blattmann2023StableVD} and 3D object generation from image inputs \cite{Voleti2024SV3DNM}.


\section{Conclusion}\label{sec:conclusion}

In this paper, we redesigned the Reconciled Polynomial Network and introduced the new {\our} model, which incorporates an innovative component—the data interdependence function—into its architecture to explicitly model diverse relationships among both data instances and attributes. These data interdependence functions significantly enhance {\our}'s capabilities for complex function learning tasks on interdependent data, including but not limited to images, language, time series, and graphs. To demonstrate the efficacy of {\our}'s data interdependence modeling, we conducted extensive experiments on various multi-modal benchmark datasets, showing that {\our} consistently outperforms existing backbone models in multiple deep function learning tasks.

This enhancement also expands {\our}'s unifying potential, enabling it to encompass a broader range of prevalent backbone architectures within its canonical representation, including convolutional neural networks (CNNs), recurrent neural networks (RNNs), graph neural networks (GNNs), and Transformers. Through these unifications, we illustrate the shared architectures and unique differences of these backbone models. These insights not only open new avenues for designing superior ``Transformer-Next'' models but also position {\our} as a robust framework for advancing function learning architecture design. Beyond empirical experiments, this paper also interprets the {\our} model from theoretical machine learning and biological neuroscience perspectives.

To support the adoption, implementation, and experimentation of {\our}, we have updated our toolkit to the new {\toolkit}, which incorporates interdependence modeling capabilities in model design and learning, along with updates to head and layer modules and model architecture. This updated toolkit enables researchers to efficiently design, customize, and deploy new {\our} models across a wide range of function learning tasks on diverse interdependent datasets.


\acksection

This work is partially supported by NSF through grants IIS-1763365 and IIS-2106972.


{
\newpage
\bibliography{./sections/reference}
\bibliographystyle{plain}
}

{
\newpage
\appendix
\section{Appendix}

\subsection{Licensing Rights of Using BioRender Created Contents in This Paper}

\noindent \textbf{Descriptions}: For the biomedical image content presented in Figures~\ref{fig:sensory_cortex}-\ref{fig:cortex_thalamus} in Section~\ref{sec:interpretation}, we have obtained the necessary permissions for their use in publications. Confirmation letters granting licensing rights from BioRender for these generated contents are included in the following pages.

\newpage


\section*{Supplementary material}

# Confirmation of Publication and Licensing Rights - Open Access

**October 10th, 2024**

**Subscription Type:**    Institution - Academic
**Agreement number:**    EB27ERFE3F
**Publisher Name:**    arxiv

**Figure Title:**    sensory_cortex

**Citation to Use:**    *Created in BioRender. Zhang, J. (2024)* [BioRender.com/f39x623](BioRender.com/f39x623)

To whom this may concern,

This document ("Confirmation") hereby confirms that Science Suite Inc. dba BioRender ("BioRender") has granted the following BioRender user: Jiawei Zhang ("User") a BioRender Academic Publication License in accordance with BioRender's [Terms of Service](Terms of Service) and [Academic License Terms](Academic License Terms) ("License Terms") to permit such User to do the following on the condition that all requirements in this Confirmation are met:

1) publish their Completed Graphics created in the BioRender Services containing both User Content and BioRender Content (as both are defined in the License Terms) in publications (journals, textbooks, websites, etc.); and
2) sublicense such Completed Graphics under "open access" publication sublicensing models such as CC-BY 4.0 and more restrictive models, so long as the conditions set forth herein are fully met.

Requirements of User:

1) All Completed Graphics to be published in any publication (journals, textbooks, websites, etc.) must be accompanied by the following citation either as a caption, footnote or reference for each figure that includes a Completed Graphic:
   "Created in BioRender. Zhang, J. (2024) [BioRender.com/f39x623](BioRender.com/f39x623) ".
2) All terms of the License Terms including all Prohibited Uses are fully complied with.  E.g. For Academic License Users, no commercial uses (beyond publication in journals, textbooks or websites) are permitted without obtaining or switching to a BioRender Industry Plan.
3) A Reader (defined below) may request that the User allow their figure to be a public template for Readers to view, copy, and modify the figure.  It is up to the User to determine what level of access to grant.

Open-Access Journal Readers:

Open-Access journal readers ("Reader") who wish to view and/or re-use a particular Completed Graphic in an Open-Access journal subject to CC-BY sublicensing may do so by clicking on the URL link in the applicable citation for the subject Completed Graphic.

The re-use/modification options below are available after the Reader requests the User to adapt their

figure as a BioRender template and the User has granted such access.
1) <u>View-Only/Free Plan Use:</u> A Reader who wishes to only view the Completed Graphic may do so in the BioRender Services as either a BioRender Free Plan user or simply as a viewer. By becoming a BioRender Free Plan user, the Reader may view, modify and re-use the Completed Graphic as permitted under BioRender's [Basic License Terms](#) (e.g. personal use only, no publishing or commercial use permitted).
2) <u>Re-Use/Publish with No Modifications:</u> For any re-use and re-publication of a Completed Graphic with no modification(s) to the Completed Graphic made by the Reader, a Reader may do so by citing the original author using the citation noted above with the Completed Graphic. The Reader must also comply with the underlying License Terms which apply to the Completed Graphic as noted above (e.g. no commercial use for Academic License).
3) <u>Re-Use/Publish with Modifications:</u> For any re-use and re-publication of a Completed Graphic with a modification(s) made by the Reader, the Reader may do so by becoming a BioRender user themselves under either an Academic or Industry Plan, citing the original author using the citation noted above with the Completed Graphic and complying with the applicable License Terms.

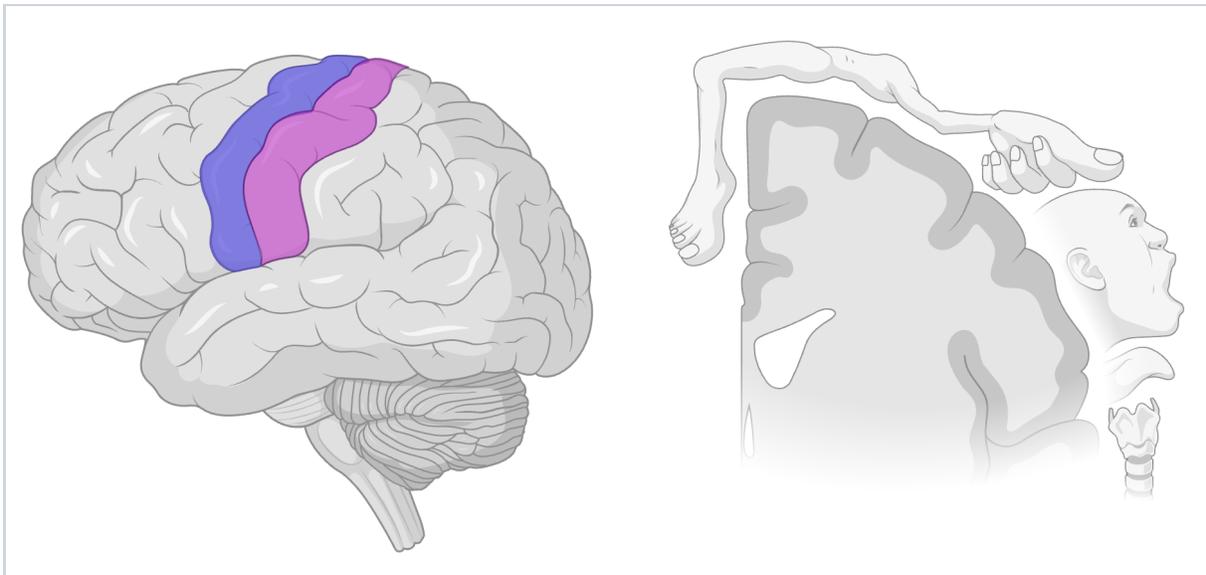

*For any questions regarding this document, or other questions about publishing with BioRender, please refer to our [BioRender Publication Guide](#), or contact BioRender Support at [support@biorender.com](mailto:support@biorender.com).*

# Confirmation of Publication and Licensing Rights - Open Access

**October 10th, 2024**

| | |
|---|---|
| **Subscription Type:** | *Institution - Academic* |
| **Agreement number:** | *JT27ER7T81* |
| **Publisher Name:** | *arxiv* |
| **Figure Title:** | *hippocampus* |
| **Citation to Use:** | *Created in BioRender. Zhang, J. (2024)* [BioRender.com/j80y259](BioRender.com/j80y259) |

To whom this may concern,

This document ("Confirmation") hereby confirms that Science Suite Inc. dba BioRender ("BioRender") has granted the following BioRender user: Jiawei Zhang ("User") a BioRender Academic Publication License in accordance with BioRender's [Terms of Service](Terms of Service) and [Academic License Terms](Academic License Terms) ("License Terms") to permit such User to do the following on the condition that all requirements in this Confirmation are met:

1) publish their Completed Graphics created in the BioRender Services containing both User Content and BioRender Content (as both are defined in the License Terms) in publications (journals, textbooks, websites, etc.); and
2) sublicense such Completed Graphics under "open access" publication sublicensing models such as CC-BY 4.0 and more restrictive models, so long as the conditions set forth herein are fully met.

Requirements of User:

1) All Completed Graphics to be published in any publication (journals, textbooks, websites, etc.) must be accompanied by the following citation either as a caption, footnote or reference for each figure that includes a Completed Graphic:
   "Created in BioRender. Zhang, J. (2024) [BioRender.com/j80y259](BioRender.com/j80y259) ".
2) All terms of the License Terms including all Prohibited Uses are fully complied with. E.g. For Academic License Users, no commercial uses (beyond publication in journals, textbooks or websites) are permitted without obtaining or switching to a BioRender Industry Plan.
3) A Reader (defined below) may request that the User allow their figure to be a public template for Readers to view, copy, and modify the figure. It is up to the User to determine what level of access to grant.

Open-Access Journal Readers:

Open-Access journal readers ("Reader") who wish to view and/or re-use a particular Completed Graphic in an Open-Access journal subject to CC-BY sublicensing may do so by clicking on the URL link in the applicable citation for the subject Completed Graphic.

The re-use/modification options below are available after the Reader requests the User to adapt their

figure as a BioRender template and the User has granted such access.
1) <u>View-Only/Free Plan Use:</u> A Reader who wishes to only view the Completed Graphic may do so in the BioRender Services as either a BioRender Free Plan user or simply as a viewer. By becoming a BioRender Free Plan user, the Reader may view, modify and re-use the Completed Graphic as permitted under BioRender's [Basic License Terms](#) (e.g. personal use only, no publishing or commercial use permitted).
2) <u>Re-Use/Publish with No Modifications:</u> For any re-use and re-publication of a Completed Graphic with no modification(s) to the Completed Graphic made by the Reader, a Reader may do so by citing the original author using the citation noted above with the Completed Graphic. The Reader must also comply with the underlying License Terms which apply to the Completed Graphic as noted above (e.g. no commercial use for Academic License).
3) <u>Re-Use/Publish with Modifications:</u> For any re-use and re-publication of a Completed Graphic with a modification(s) made by the Reader, the Reader may do so by becoming a BioRender user themselves under either an Academic or Industry Plan, citing the original author using the citation noted above with the Completed Graphic and complying with the applicable License Terms.

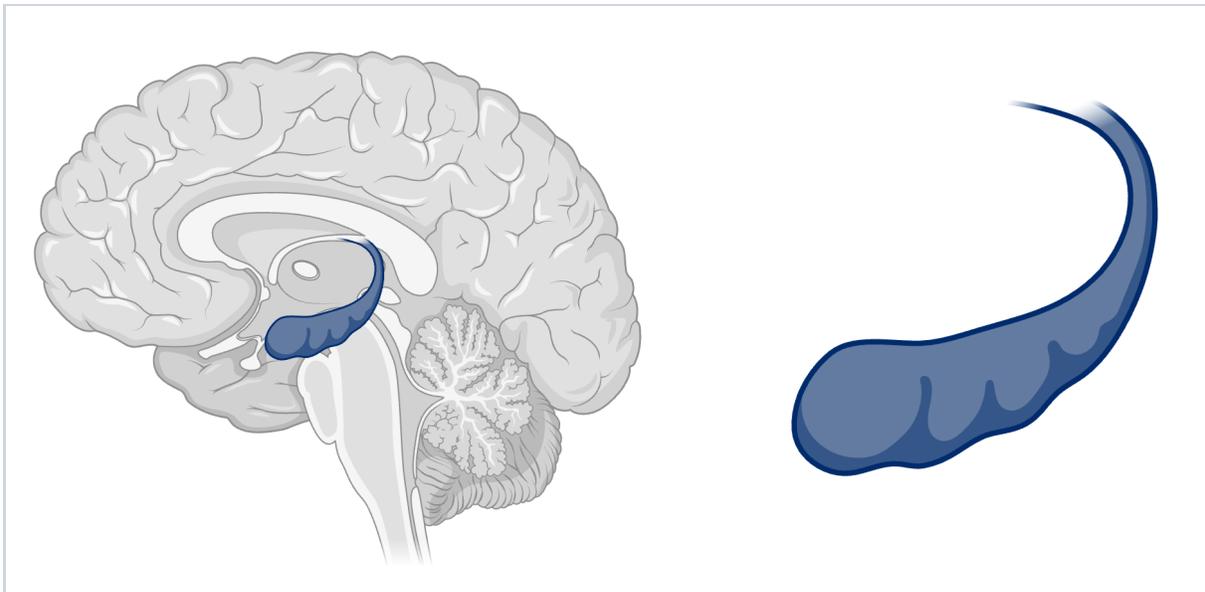

*For any questions regarding this document, or other questions about publishing with BioRender, please refer to our [BioRender Publication Guide](#), or contact BioRender Support at [support@biorender.com](mailto:support@biorender.com).*

# Confirmation of Publication and Licensing Rights - Open Access

**October 10th, 2024**

***Subscription Type:***   *Institution - Academic*
***Agreement number:***   *AQ27ERC06B*
***Publisher Name:***   *arxiv*

***Figure Title:***   *cortex and thalamus*

***Citation to Use:***   *Created in BioRender. Zhang, J. (2024) [BioRender.com/v74r180](BioRender.com/v74r180)*

To whom this may concern,

This document ("Confirmation") hereby confirms that Science Suite Inc. dba BioRender ("BioRender") has granted the following BioRender user: Jiawei Zhang ("User") a BioRender Academic Publication License in accordance with BioRender's [Terms of Service](#) and [Academic License Terms](#) ("License Terms") to permit such User to do the following on the condition that all requirements in this Confirmation are met:

1)    publish their Completed Graphics created in the BioRender Services containing both User Content and BioRender Content (as both are defined in the License Terms) in publications (journals, textbooks, websites, etc.); and
2)    sublicense such Completed Graphics under "open access" publication sublicensing models such as CC-BY 4.0 and more restrictive models, so long as the conditions set forth herein are fully met.

Requirements of User:

1)    All Completed Graphics to be published in any publication (journals, textbooks, websites, etc.) must be accompanied by the following citation either as a caption, footnote or reference for each figure that includes a Completed Graphic:
    "Created in BioRender. Zhang, J. (2024) [BioRender.com/v74r180](BioRender.com/v74r180) ".
2)    All terms of the License Terms including all Prohibited Uses are fully complied with.  E.g. For Academic License Users, no commercial uses (beyond publication in journals, textbooks or websites) are permitted without obtaining or switching to a BioRender Industry Plan.
3)    A Reader (defined below) may request that the User allow their figure to be a public template for Readers to view, copy, and modify the figure.  It is up to the User to determine what level of access to grant.

Open-Access Journal Readers:

Open-Access journal readers ("Reader") who wish to view and/or re-use a particular Completed Graphic in an Open-Access journal subject to CC-BY sublicensing may do so by clicking on the URL link in the applicable citation for the subject Completed Graphic.

The re-use/modification options below are available after the Reader requests the User to adapt their

figure as a BioRender template and the User has granted such access.
1) <u>View-Only/Free Plan Use:</u> A Reader who wishes to only view the Completed Graphic may do so in the BioRender Services as either a BioRender Free Plan user or simply as a viewer. By becoming a BioRender Free Plan user, the Reader may view, modify and re-use the Completed Graphic as permitted under BioRender's [Basic License Terms](#) (e.g. personal use only, no publishing or commercial use permitted).
2) <u>Re-Use/Publish with No Modifications:</u> For any re-use and re-publication of a Completed Graphic with no modification(s) to the Completed Graphic made by the Reader, a Reader may do so by citing the original author using the citation noted above with the Completed Graphic. The Reader must also comply with the underlying License Terms which apply to the Completed Graphic as noted above (e.g. no commercial use for Academic License).
3) <u>Re-Use/Publish with Modifications:</u> For any re-use and re-publication of a Completed Graphic with a modification(s) made by the Reader, the Reader may do so by becoming a BioRender user themselves under either an Academic or Industry Plan, citing the original author using the citation noted above with the Completed Graphic and complying with the applicable License Terms.

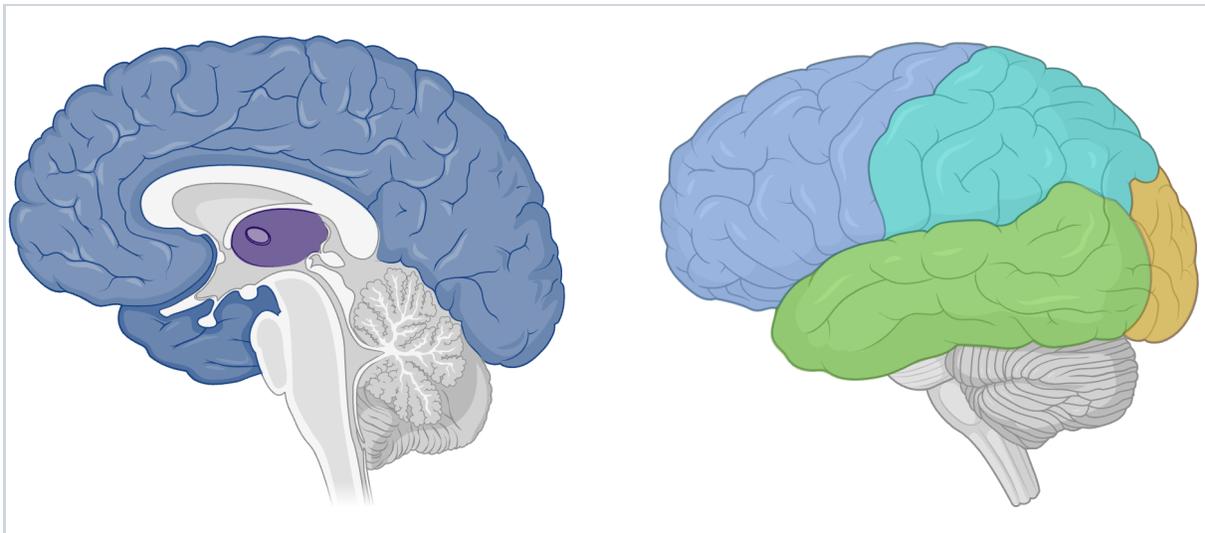

*For any questions regarding this document, or other questions about publishing with BioRender, please refer to our [BioRender Publication Guide](#), or contact BioRender Support at [support@biorender.com](mailto:support@biorender.com).*


}


\end{document}